\documentclass[letterpaper]{article} 
\usepackage{arxiv_aaai}  
\usepackage{times}  
\usepackage{helvet}  
\usepackage{courier}  
\usepackage[hyphens]{url}  
\usepackage{color}
\usepackage{graphicx} 
\usepackage{natbib}  
\usepackage{caption} 
\usepackage{algorithm}
\usepackage{algorithmic}

\usepackage{newfloat}
\usepackage{stfloats}

\usepackage{listings}
\DeclareCaptionStyle{ruled}{labelfont=normalfont,labelsep=colon,strut=off} 
\floatstyle{ruled}
\newfloat{listing}{tb}{lst}{}
\floatname{listing}{Listing}
\usepackage{booktabs}
\usepackage{subfigure}
\usepackage{amsfonts}       
\usepackage{tabularx}

\title{Details Enhancement in Unsigned Distance Field Learning for High-fidelity 3D Surface Reconstruction}
\author{
    Cheng Xu\textsuperscript{\rm 1, \rm 2, \rm 3}, 
    Fei Hou\textsuperscript{\rm 1, \rm 2}\thanks{Corresponding author}, 
    Wencheng Wang\textsuperscript{\rm 1, \rm 2},
    Hong Qin\textsuperscript{\rm 4}, 
    Zhebin Zhang\textsuperscript{\rm 5}, 
    Ying He\textsuperscript{\rm 6}
}
\affiliations{
   \textsuperscript{\rm 1}Key Laboratory of System Software (CAS), State Key Laboratory of Computer Science, Institute of Software, Chinese Academy of Sciences, China\\
    \textsuperscript{\rm 2}University of Chinese Academy of Sciences, China\\
    \textsuperscript{\rm 3}School of Advanced Interdisciplinary Sciences, University of Chinese Academy of Sciences, China\\
    \textsuperscript{\rm 4}Department of Computer Science, Stony Brook University, USA.\\
    \textsuperscript{\rm 5}InnoPeak Technology, USA\\
    \textsuperscript{\rm 6}College of Computing and Data Science, Nanyang Technological University, Singapore

    \{xucheng, houfei, whn\}@ios.ac.cn, qin@cs.stonybrook.edu, zhebin.zhang@oppo.com, yhe@ntu.edu.sg

}

\usepackage{bibentry}

\begin{document}

\maketitle
\begin{abstract}
\label{abstract}

While Signed Distance Fields (SDF) are well-established for modeling watertight surfaces, Unsigned Distance Fields (UDF) broaden the scope to include open surfaces and models with complex inner structures. Despite their flexibility, UDFs encounter significant challenges in high-fidelity 3D reconstruction, such as non-differentiability at the zero level set, difficulty in achieving the exact zero value, numerous local minima, vanishing gradients, and oscillating gradient directions near the zero level set. To address these challenges, we propose Details Enhanced UDF (DEUDF) learning that integrates normal alignment and the SIREN network for capturing fine geometric details, adaptively weighted Eikonal constraints to address vanishing gradients near the target surface, unconditioned MLP-based UDF representation to relax non-negativity constraints, and DCUDF for extracting the local minimal average distance surface. These strategies collectively stabilize the learning process from unoriented point clouds and enhance the accuracy of UDFs. Our computational results demonstrate that DEUDF outperforms existing UDF learning methods in both accuracy and the quality of reconstructed surfaces. Our source code is at https://github.com/GiliAI/DEUDF.
\end{abstract}

\section{Introduction}
\label{sec:introduction}
While signed distance fields (SDF) are favored for their capability to represent watertight surfaces, unsigned distance fields (UDFs) provide a means to model both open surfaces and objects with complex inner structures. However, achieving high-quality UDFs that accurately reconstruct 3D surfaces with fine geometric details is challenging for several reasons. Firstly, UDFs struggle to precisely achieve a zero value, making it difficult to generate open boundaries. Secondly, UDFs are theoretically non-differentiable at the zero level set, resulting in vanishing gradients near the target surface. This issue leads to numerous undesired local minima, complicating the extraction of the zero level set. Thirdly, the gradient directions of UDFs tend to oscillate near the surface, causing the reconstructed surfaces to be  fragmented~\cite{Guillard2022MeshUDF}. 

Due to the inherently low accuracy of learned UDFs, the extracted zero level sets are typically over-smoothed and lack crucial geometric details. Several studies have aimed to enhance the precision of UDF learning. For instance, NDF~\cite{Chibane2020NDF} trains a shape encoder and a decoder from 3D surfaces of various types, including point clouds, meshes and mathematical functions. As a supervised method, its performance heavily relies on the quality and diversity of the training dataset. Unsupervised approaches, such as CAP-UDF~\cite{Zhou2022CAPUDF} and LevelSetUDF~\cite{Zhou2023levelset}, offer greater flexibility in handling a wider range of 3D models. Despite advancements in UDF learning techniques, all existing  methods still suffer from  relatively low accuracy in the learned distance fields compared to SDFs. This limitation significantly diminishes their practical usage in real-world applications. 

This paper introduces a new method, called Details Enhanced UDF (DEUDF) learning, aimed at enhancing the accuracy of UDF learning from unoriented point clouds to ensure that learned UDFs can capture the fine geometric details of target surfaces. A key observation is the significant role normal directions play in learning fine details. Although obtaining globally consistent orientations~\cite{Rui2023GCNO} is challenging due to its combinatorial and global optimization nature, acquiring normal directions locally, for instance, through principal component analysis~\cite{Hoppe1992}, is feasible. Consequently, we constrain the UDF gradients to align with normal directions to enhance detail capture, while disregarding normal orientations. 

\begin{figure*}[!htbp]
    \centering
    \subfigure[Existing UDF learning architectures]{
		\includegraphics[height=0.095\textheight]{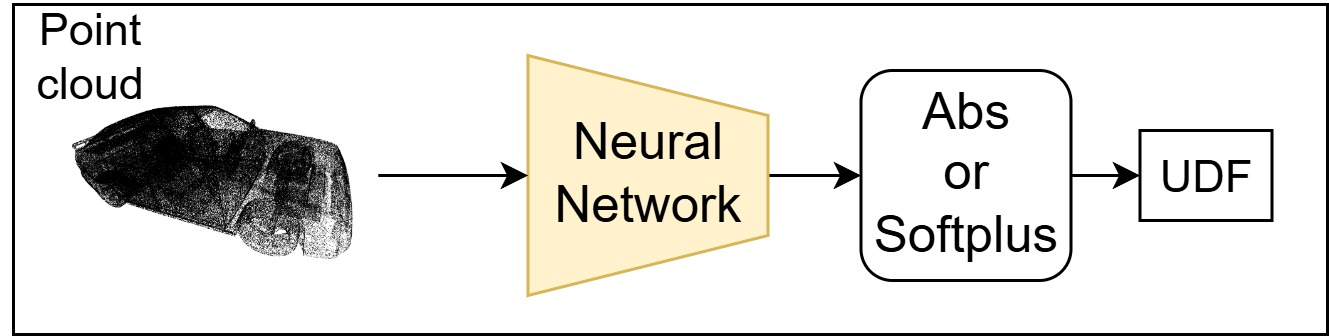}}
    \subfigure[Our architecture]{
		\includegraphics[height=0.095\textheight]{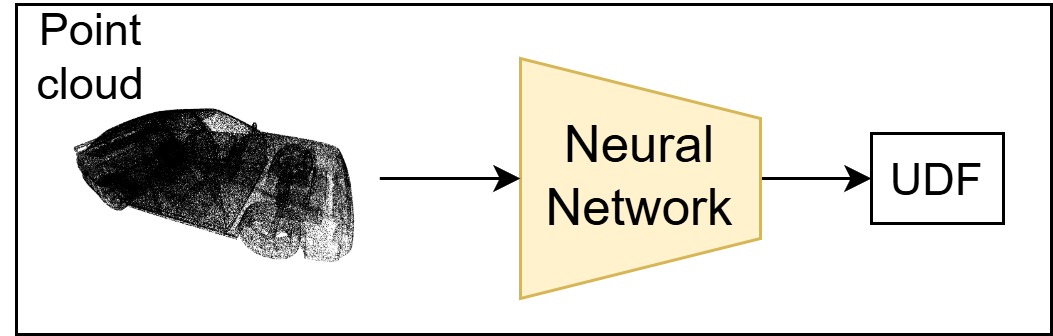}}\\
    \subfigure[3D setup]{
		\includegraphics[width=0.21\textwidth]{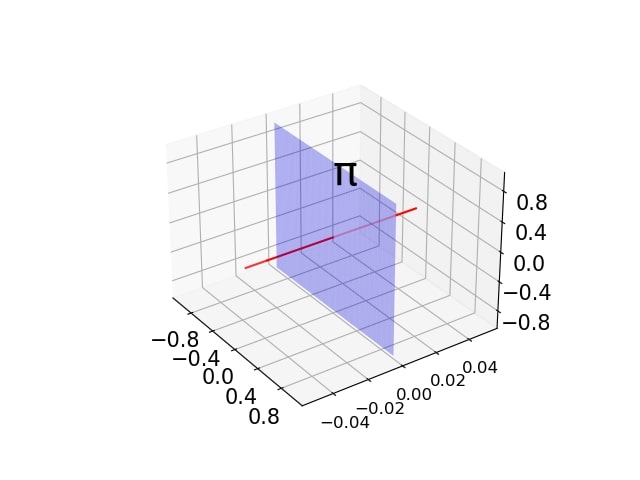}}
  \subfigure[Ground truth]{
		\includegraphics[width=0.17\textwidth]{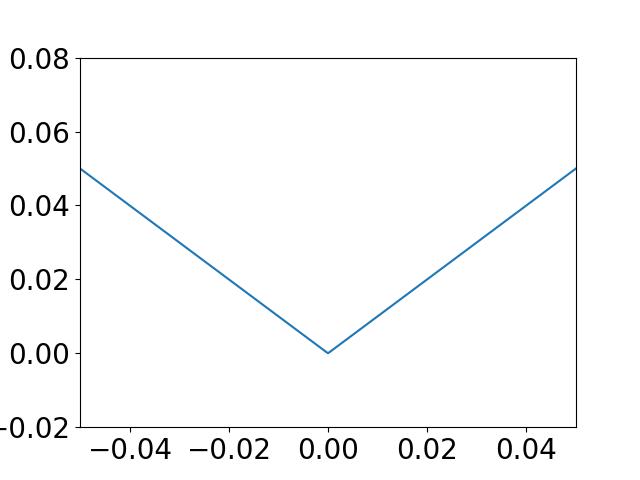}}  
    \subfigure[MLP+Abs.]{
        \includegraphics[width=0.17\textwidth]{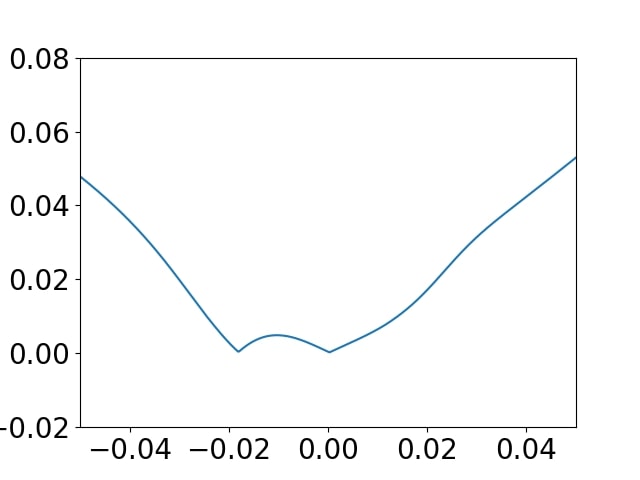}}
    \subfigure[MLP+Softplus]{
		\includegraphics[width=0.17\textwidth]{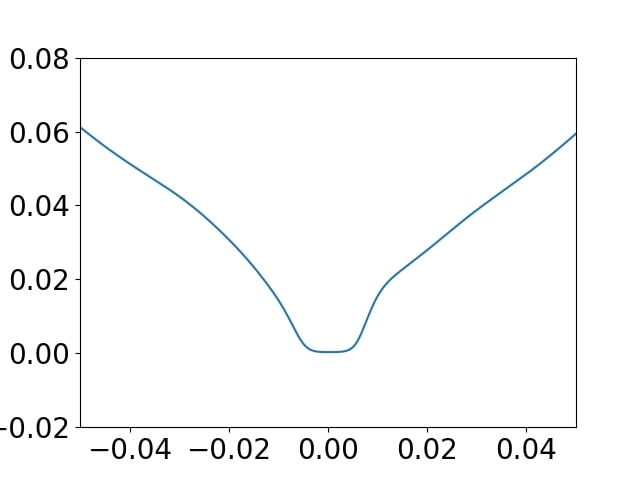}}
    \subfigure[Ours]{
		\includegraphics[width=0.17\textwidth]{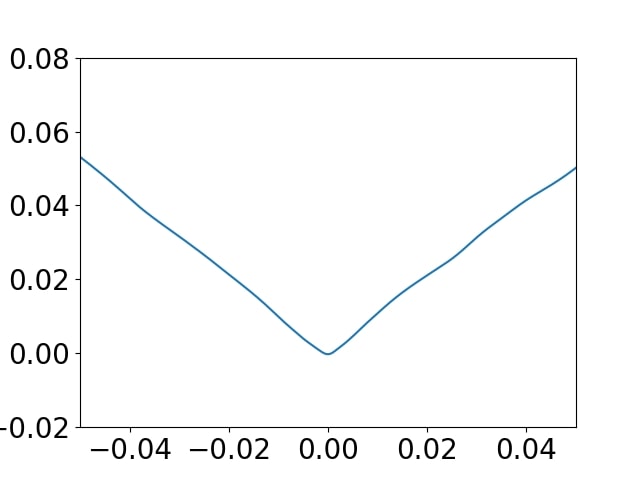}}\\
    \caption{Illustration of UDF learning with various neural representations. (a) Existing neural network architectures often use an absolute value or softplus function to prevent negative distances. (b) In contrast, our method relaxes the non-negative condition and employs an unconditioned MLP with the SIREN activation function for predicting the distances. (c) To show the differences between existing representations and ours, we consider a plane $\pi$ and a line perpendicular to $\pi$. We plot the unsigned distance to $\pi$ on this line learned by our method, but with different MLP output layers. (d) The horizontal axis represents a signed distance range from -0.05 to 0.05, while the vertical axis measures the learned unsigned distance. Ideally, the unsigned distance should exhibit a perfect ``V'' shape relative to the signed distance. (e) However, UDFs parameterized by conditioned MLPs can present defects. For example, learning a negative value followed by taking the absolute value results in a ``W''-shaped distance field around the zero level set. (f) Employing the softplus activation function to eliminate negative values yields learned UDFs with vanishing gradients across a relatively large distance range near the zero value. (g) In contrast, our method, which employs unconditioned MLPs, significantly narrows this range of vanishing gradients.
    }
    \label{fig:UDF_comp}
\end{figure*}

Existing methods strictly constrain UDFs to non-negative values, so the UDFs are hard to achieve exact zero values. Meanwhile, such strict constraint has brought some other problems as illustrated in Figure~\ref{fig:UDF_comp}. To overcome this limitation, we relax the strict requirements that UDFs must be non-negative and that the surface must precisely correspond to the zero iso-surfaces. Though the distances are not strictly non-negative, we still call it UDF. This adaptation enables the use of an unconditioned multilayer perceptron (MLP), meaning an MLP that outputs its value directly without additional operations to make the output positive. Unlike traditional methods that generate UDFs by taking the absolute value of a learned SDF~\cite{Zhou2023levelset} -- prone to inducing oscillating gradients -- or by using the $\mathrm{softplus}$ activation function in MLPs to eliminate negative values~\cite{Liu2023} -- leading to vanishing gradients -- our relaxation not only addresses the vanishing gradients but also stabilizes the oscillation of gradient directions near the surface. 

While SDFs maintain well-behaved gradients with consistent unit length throughout 3D space, UDFs often experience vanishing gradients at the zero level set, diminishing the effectiveness of uniformly applied Eikonal constraints for UDF learning. To address this issue, we propose an adaptively weighted Eikonal constraint, specifically tailored to align with the unique properties of UDFs. Moreover, we incorporate the SIREN network~\cite{Sitzmann2020SIREN} to represent high-frequency details in UDFs, thereby enhancing the encoding capabilities of our model. We consider the intended surface should be around zero and the distance values---both positive and negative---should be as small as possible. To achieve this, we adopt DCUDF~\cite{Hou2023}, an optimization-based surface extraction algorithm.

By integrating normal alignment, unconditioned MLPs with SIREN activation functions, adaptively weighted Eikonal constraints, and UDF-tailored surface extraction techniques, DEUDF significantly improves the accuracy of UDF learning. Evaluations on benchmark datasets demonstrate our method outperforms baseline methods in terms of UDF accuracy and quality of reconstructed surfaces.

\section{Related work}
\label{sec:related-works}

Surface reconstruction from point clouds has been studied extensively for the last three decades. The field has seen significant evolution, from computational geometry methods~\cite{Amenta1998,Dey2003} to implicit function techniques~\cite{Hoppe1992,Ohtake2003,Kazhdan2006,Kazhdan2013,Hou2022,DWG}, and more recently to deep learning approaches~\cite{Park2019,Chibane2020NDF,Zhou2023levelset,Ren2023,AlignGraidentandHessian,fainstein2024dudf}. Due to space constraints, this section primarily focuses on deep learning-based 3D reconstruction techniques.

Both signed distance fields and occupancy fields effectively represent closed surfaces. An occupancy field defines whether each point in  space is inside or outside a given shape. ONet~\cite{Mescheder2019} employs a deep neural network classifier to implicitly represent 3D surfaces as a continuous decision boundary, while IF-Net~\cite{Chibane2020IFNET} and CONet~\cite{Peng2020} use encoders to capture shape. Compared to occupancy fields, SDFs provide additional information about the distance of a point form the surface of the object, making them favored for applications that require accurate shape representation, such as reconstruction, shape interpolation and completion.  DeepSDF~\cite{Park2019} introduces an innovative implicit encoder that defines the boundary of a 3D shape as the zero level set of a learned implicit function. Following this, numerous neural SDF-based works have been developed. For example, DeepLS~\cite{Chabra2020} utilizes a grid structure to store latent codes for local shape features, SIREN~\cite{Sitzmann2020SIREN} introduces a novel activation function for increasing the network's capability to capture high-frequency signals, and  IDF~\cite{Wang2022IDF} employs displacement maps to enhance the representation of fine details. Additionally, SDFs have been utilized to represent geometric shapes for neural rendering tasks, such as  NeuS~\cite{wang2021neus} and VolSDF~\cite{VolSDF}, which leverage SDFs for 3D reconstruction from multi-view images.

To model general non-watertight surface, Chibane et al.~\cite{Chibane2020NDF} introduced neural UDFs, which predict the unsigned distance from a query point to the nearest surface point. GIFS~\cite{Ye2022} models the relationship between points rather than between points and surfaces, while NVF~\cite{Yang_2023_CVPR} learns a vector as an alternative to calculating gradient from UDFs, representing the direction from query points to the target surface. Unlike these methods, which utilize separate neural networks to extract supplementary information that aids UDF learning, CAP-UDF~\cite{Zhou2022CAPUDF} and GeoUDF~\cite{Ren2023} focus on enhancing the density of the input point clouds via adopting upsampling techniques. Despite these advancements, the challenge of ambiguous gradients near the zero level set remains.
To address the non-differentiability issue of UDFs at zero, LevelSetUDF~\cite{Zhou2023levelset} projects the properties of the non-zero level set to the zero level set to learn a continuous and smooth UDF, while DUDF~\cite{fainstein2024dudf} adopts a new representation to maintain differentiability at points close to the target surface. Although LevelSetUDF and DUDF tackle the non-differentiable issues, they still struggle to match the quality of reconstruction--particularly for surfaces with fine details-achieved SDF learning methods.
Additionally, similar to SDFs, UDFs are also used to implicitly represent 3D shapes in neural rendering tasks, such as 3D reconstruction from multi-view images (e.g., NeuralUDF~\cite{Long2023}, NeUDF~\cite{Liu2023}, 2S-UDF~\cite{2SUDF}) and 3D generation~\cite{yu2025surfd,udiff}. See Table~\ref{tab:methods-comp} for a qualitative comparison of existing UDF learning methods.

Extracting the zero level set from UDFs is technically non-trivial, as it is rare for the learned UDFs to precisely reach zero values. There are several research efforts aiming at addressing this issue. Gradient-based methods such as CAP-UDF~\cite{Zhou2022CAPUDF}, MeshUDF~\cite{Guillard2022MeshUDF} and GeoUDF~\cite{Ren2023} use both gradient directions and UDF values to detect zero crossings, while optimization-based techniques, such as DCUDF~\cite{Hou2023}, optimize the surface locally to minimize the average distance value on the surface.

\begin{table}[]
    \centering

    \renewcommand\arraystretch{0.5}
    \setlength{\tabcolsep}{1pt}
    \scriptsize
    \begin{tabular}{cccccc}
        \toprule
         Method & Input & MLP & Eikonal &  Non-negativity & Learning\\
        \midrule
         NeUDF & multi-view images & softplus+PE  & uniform &  softplus & unsupervised\\
         NeuralUDF & multi-view images  & softplus+PE  & uniform &  ABS & unsupervised\\
         2S-UDF & multi-view images &  softplus+PE & uniform & softplus & unsupervised\\
         NDF & sparse point clouds &  ReLU  & - & ABS & supervised\\
         GIFS & sparse point clouds &  ReLU  & - & ABS & supervised\\
         GeoUDF & sparse point clouds & LeakyReLU & - &  ABS & supervised \\
         DUDF & dense point clouds & SIREN  & uniform & ABS+HS & supervised\\
         CAP-UDF & sparse point clouds & ReLU+PE & - & ABS & unsupervised\\
         LevelSetUDF & dense point clouds & ReLU+PE & - & ABS & unsupervised\\
        \midrule
         Ours & dense point clouds &  SIREN & adaptive & no & unsupervised\\
        \bottomrule
    \end{tabular}
    \caption{Qualitative comparison of existing UDF learning methods. HS: hyperbolic scaling; PE: positional encoding; ABS: absolute value.}
    \label{tab:methods-comp}
\end{table}

\section{Method}
\label{sec:method}

Let $\mathcal{P}=\left\{\mathbf{p}_i\in \mathbb{R}^3\right\}_{i=1}^n$ represent the input raw point cloud, which has been uniformly scaled to fit within the cube domain $\Omega=[-1,1]^3$. We employ an MLP to parameterize the UDF for $\mathcal{P}$, denoted by $f$. Our objective is to accurately learn $f$ in order to extract a high-fidelity mesh that represents the geometric structure of $\mathcal{P}$.

\subsection{Relaxation of non-negative constraints}

Traditional methods for leaning UDFs generally ensure non-negative distance values by adopting specific strategies, such as taking the absolute value or using the $\mathrm{softplus}$ in the last layer. However, as illustrated in Figure~\ref{fig:UDF_comp}, these approaches have significant drawbacks in accurately representing distances near zero. For example, using the absolute value results in UDFs exhibiting a ``W'' shape, leading to changes in gradient directions and the presence of multiple minimum values. Moreover, when the absolute value is applied to an SDF, the resulting UDF exhibits characteristics similar to those of an SDF. This leads to the unintended consequence of gap filling even in point clouds that represent open surfaces. See Figure~\ref{fig:open-surfcomparison} for an example. On the other hand, employing the $\mathrm{softplus}$ activation function helps avoid the W-shaped artifacts associated with the absolute value approach. Nonetheless, this method tends to generate a U-shaped distance field, characterized by a relatively width bandwidth around the zero value, approximately between 0 and 0.04. This occurs because for $x\in (-\infty,0)$, $\mathrm{softplus}(x)$ yields a small positive value with almost zero derivatives. Consequently, this results in vanishing gradients for query points near the target surface, which can significantly hinder the effectiveness of network training that relies on gradient-based optimization techniques.

Observing both vanishing gradients and oscillating gradient directions stem from the strict non-negative constraint on distance values, we propose relaxing the conditions that require UDFs to be non-negative and the surface to coincide precisely with zero iso-surface. We use unconditioned MLPs to represent UDFs and consider the local minimal distance surface of the UDF value around zero, which may be either positive or negative, as the intended surface. As illustrated in Figure~\ref{fig:UDF_comp} (e), this relaxation results in a distance function with a significantly narrower bandwidth compared to using the softplus activation function, thereby providing a high-quality approximation to the ground truth distance filed, which exhibits a V-shaped profile.

With UDFs parameterized by unconditioned MLPs, we define the following loss functions for learning UDFs without ground-truth supervision:
\begin{equation}
    \mathcal{L}_{\mathrm{dist}}=\sum_{\mathbf{p}_i\in \mathcal{P}}|f(\mathbf{p}_i)|,
\end{equation}
and
\begin{equation}    \mathcal{L}_{\mathrm{positive}}=\sum_{\mathbf{x}\in \Omega}\exp\left(-100f(\mathbf{x})\right).
\end{equation}
The distance term $\mathcal{L}_\mathrm{dist}$ encourages the zero level set of the learned UDFs to pass through the input points $\mathbf{p}_i$. The positivity enforcement term 
$\mathcal{L}_{\mathrm{positive}}$ is designed to ensure that values of $f(\mathbf{x})$ for off-surface points $\mathbf{x}$ are large and positive. This term encourages the majority of sample points are assigned positive values, effectively preventing the generation of negative distance values and ensuring the function behaves like a true UDF. Additionally, it helps to maintain a clear distinction between surface and non-surface regions, cruicial for accurate surface reconstruction. 

\paragraph{Remark.} In NeuralUDF~\cite{Long2023}, a similar loss term in the form  $\exp(-100|f|)$ was used. It is important to note that our loss term does not include the absolute value. This subtle difference significantly impacts the behavior of the learned distance field $f$. With the absolute value, their loss encourages $|f|$ being a large positive value, which consequently reduces the occurrence of points with zero distance values. This reduction minimizes the presence of small disconnected components in the reconstructed surfaces~\cite{Wang2022IDF,Long2023}. Therefore, the $\exp$ in their loss functions acts as a regularizer to smooth the learned distance fields. As mentioned above, the use of the absolute value $|f|$ in the loss function can lead to undesired side effects, such as a W-shaped profile in the learned UDFs, which may consequently result in watertight models. In sharp contrast, our loss term, which omits the absolute value, serves as a soft non-negative constraint. This encourages $f$ to remain positive as much as possible, thus differentiating it from an SDF, and enabling $f$ to mimic a true UDF. Even though DCUDF~\cite{Hou2023} used an unconditional MLP to represent UDF, it needs the ground truth UDF to supervise and keep the distances almost positive. While, DEUDF is unsupervised learning.

\subsection{Normal alignment}
\label{subsec:normalalignment}

Normal directions are critical for enhancing surface details in the reconstruction process. Let $\mathcal{N}=\{\mathbf{n}_i\}_{i=1}^{n}$ represent the set of unit normals for the point set $\mathcal{P}$. Following~\cite{Hoppe1992}, we apply principal component analysis to each point $\mathbf{p}_i$ to determine its normal direction $\mathbf{n}_i$. Since UDF gradients typically vanish on the surface, it is impractical to directly constrain the gradients of $\mathcal{P}$.

To address this issue, we generate a set of sample point pairs $\mathcal{Q}=\{(\mathbf{q}^1_i,\mathbf{q}^2_i)\}_{i=1}^n$ in each training epoch, where each point $\mathbf{q}_i$ is strategically displaced from the surface. Specifically, $\mathbf{q}^1_i=\mathbf{p}_i+\lambda_i\mathbf{n}_i$ and $\mathbf{q}^2_i=\mathbf{p}_i-\lambda_i\mathbf{n}_i$, with the displacement $\lambda_i$ being randomly chosen in the range $(0,0.003]$. This ensures that $\mathcal{Q}$ contains samples on both sides of the surface, enabling a balanced evaluation on both sides of the geometric structure of interest. We then impose constraints on the UDF gradient directions at points in $\mathcal{Q}$, using the following normal alignment loss term:
\begin{equation}
\mathcal{L}_{\mathrm{normal}}=\sum_{\stackrel{\mathbf{n}_i \in \mathcal{N},} {k \in \{1,2\}}}\left(1+(-1)^k\frac{\nabla f(\mathbf{q}^k_i)\cdot \mathbf{n}_i}{\|\nabla f(\mathbf{q}^k_i)\|_2\cdot  \|\mathbf{n}_i\|_2}\right).
\end{equation}

\subsection{Adaptively weighted Eikonal constraints}

The Eikonal constraint, expressed as $\|\nabla f\|=1$, is extensively utilized in the learning processes for SDFs. However, when applied to UDFs, this approach faces challenges due to the diminished gradient magnitudes near the zero level set. Direct application of Eikonal constraints to regularize UDFs may cause the actual surface to deviate from the input point cloud $\mathcal{P}$ and may also increase the minima of the learned UDF, as illustrated in Table~\ref{tab:ablation} and Figure~\ref{fig:ablation}. To address this issue, we propose a formulation for an adaptively weighted Eikonal loss term:
\begin{equation}
\mathcal{L}_{\mathrm{eikonal}}=\sum_{\mathbf{x}\in \mathcal{Q}\bigcup\Omega}\delta(f(\mathbf{x})) \big|\|\nabla f(\mathbf{x})\|_2-1\big|,
\end{equation}
where the weight function, $\delta(\cdot)$, is designed to reduce the contribution from points close to the target surface. A U-shaped function with controllable bandwidth serves this purpose effectively. In our implementation, we employ the attenuation function used in IDF~\cite{Wang2022IDF} as our weight as
$\delta(d)=\left(1+(\frac{\xi}{d})^4\right)^{-1}$, 
where $\xi$ represents the threshold beyond which the influence of the Eikonal constraint begins to diminish significantly. In our experiments, we initially set $\xi$ to 0.01 and gradually decrease $\xi$ to 0.002 over the course of the learning process, following the learning rate. This adjustment is made to enhance the attenuation effect. We evaluate the Eikonal loss $\mathcal{L}_{\mathrm{eikonal}}$ for points in the set $\mathcal{Q}$, which serves as a proxy of the target geometry, as well as for randomly sampled points throughout the entire domain $\Omega$.

\subsection{UDF learning}

For the network architecture, we employ a 5-layer SIREN network~\cite{Sitzmann2020SIREN}, which consists of 256 units per layer. The network utilizes a sinusoidal activation function  $\sin(\omega x)$, with a default frequency\footnote{For noisy point clouds, using a lower frequency enhances resilience  against noise.} $\omega=60$ for clean point clouds in our implementation, effectively encoding fine geometric details. Our network takes spatial coordinates $(x, y, z)$ as inputs and outputs the predicted unsigned distance. The training process aims to minimize the following loss function:
\begin{equation}    \mathcal{L}=\lambda_1\mathcal{L}_{dist}+\lambda_2\mathcal{L}_{\mathrm{positive}}+\lambda_3\mathcal{L}_{\mathrm{normal}}+\lambda_4\mathcal{L}_{\mathrm{eikonal}}\label{equ-weikonal}
\end{equation}
where $\lambda_i$s are weights assigned to balance the contributions from the four loss terms. We empirically set $\lambda_1=400$, $\lambda_2=50$, $\lambda_3=40$ and $\lambda_4=10$ in our implementation.
We train the neural network using the Adam~\cite{2014Adam} optimizer, starting with a learning rate of $5\times 10^{-5}$. The learning rate decays to zero following a cosine annealing schedule~\cite{2017SGDR}.

\begin{table*}[!htbp]
    \centering
    \renewcommand\arraystretch{0.9}
    \setlength{\tabcolsep}{1pt} 
    \begin{tabular}{cccccccccccccc}
        \toprule
          & & \multicolumn{4}{c}{Stanford 3D Scene}& \multicolumn{4}{c}{Stanford 3D Scan}& \multicolumn{4}{c}{ShapeNet-Cars}\\
         \cmidrule(r){1-14}
         & & \multicolumn{2}{c}{Chamfer-L1 $(\downarrow)$} &\multicolumn{2}{c}{F-score $(\uparrow)$} & \multicolumn{2}{c}{Chamfer-L1 $(\downarrow)$} &\multicolumn{2}{c}{F-score $(\uparrow)$} & \multicolumn{2}{c}{Chamfer-L1 $(\downarrow)$} &\multicolumn{2}{c}{F-score $(\uparrow)$}\\
        \cmidrule(r){3-4}
        \cmidrule(r){5-6}
        \cmidrule(r){7-8}
        \cmidrule(r){9-10}        
        \cmidrule(r){11-12}
        \cmidrule(r){13-14}
         Method & Distance &  Mean & Median & $F1^{0.005}$& $F1^{0.0025}$ &  Mean & Median & $F1^{0.005}$& $F1^{0.0025}$ &  Mean & Median & $F1^{0.005}$& $F1^{0.0025}$\\
        \midrule

         CAP-UDF & Unsigned  & 3.37 & 3.33& 98.96 & 84.51 &
         4.12 & 3.87& 99.12 & 69.02 &
         4.97 & 4.63& 95.37 & 56.42\\
         DUDF & Unsigned  & 3.79 & 3.26 & 97.33 & 79.43 &
         4.20 & 3.95 & 99.07 & 68.10 &
         6.05 & 5.51& 89.02 & 44.18\\

         LeverSetUDF & Unsigned  & 3.16 & 2.90& 99.17 & 85.92 &
         4.12 & 3.87& 99.04 & 68.83 &
         5.03 & 4.63& 95.01 & 55.57\\
         NSH & Signed  & - & -& - & - &
         4.21 & 3.96& 99.12 & 68.18 &
         - & -& - & -\\
         IDF & Signed & -&-& -&-&
         \textbf{4.07} & \textbf{3.83}& \textbf{99.14} & \textbf{69.66}& -&-& -&-\\
        \midrule
         Ours & Unsigned & \textbf{3.09} &\textbf{2.85}& \textbf{99.41}&\textbf{86.38} &
         4.08 &\textbf{3.83}& \textbf{99.14}&69.59 & \textbf{4.91} &\textbf{4.58}& \textbf{95.53}&\textbf{56.98} \\
        \bottomrule
    \end{tabular}
    \caption{Quantitative results on Stanford 3D Scene, Stanford 3D Scans (watertight) and ShapeNet-Cars. Chamfer distances are measured in the unit of $\times10^{-3}$. 
    For CAP-UDF and LevelSetUDF, their results often include small isolated components. To compute the Chamfer distance, we cleaned their meshes by removing components that were distant from the ground truth point cloud after mesh extraction using DCUDF.}
       
    \label{tab:surf-comp}
\end{table*}

\subsection{Surface extraction}
After obtaining the UDFs, we proceed to extract the surface from the learned UDFs. Due to the relaxation of the non-negative constraints, the target geometry does not align precisely with the zero level set. Instead, we identify the target surface as the local minimal distance surface (which can be either positive or negative) near the zero values. 

One possible method for extracting the target geometry from UDFs involves explicitly using the UDF gradient, such as MeshUDF~\cite{Guillard2022MeshUDF} and GeoUDF~\cite{Ren2023}, both of which are variants of the standard Marching Cubes algorithm. On each cube edge, if the gradient directions of the UDF at the two endpoints are opposite and their UDF values are below a specified threshold, a zero crossing is marked on that edge. However, this approach is not suitable for our purpose because our learned UDFs do not ensure the positivity of the distance value, and the distance values on the surface may exceed the specified UDF threshold. If this occurs, the iso-surfacing method would exclude these cubes, leading to the creation of extracted surfaces with undesired holes.

To tackle this challenge, we adopt the optimization-based method DCUDF~\cite{Hou2023}, which initiates by extracting a double cover using the Marching Cubes algorithm at a small positive iso-value on the UDF. Subsequently, it shrinks the double cover to the local minimal distance surface. This method does not require the UDF to be strictly positive nor does it depend on a threshold to select candidate cubes. As a result, it effectively identifies the local minimal distance surface, yielding a high-quality triangle mesh that accurately represents the target surface.

\section{Experiments}
\label{sec:experiments}

\begin{figure*}[!htbp]
    \centering
    \subfigure[DUDF]{
        \label{fig:detail.dudf}
        \begin{minipage}[b]{0.18\textwidth}
            \includegraphics[width=1\textwidth]{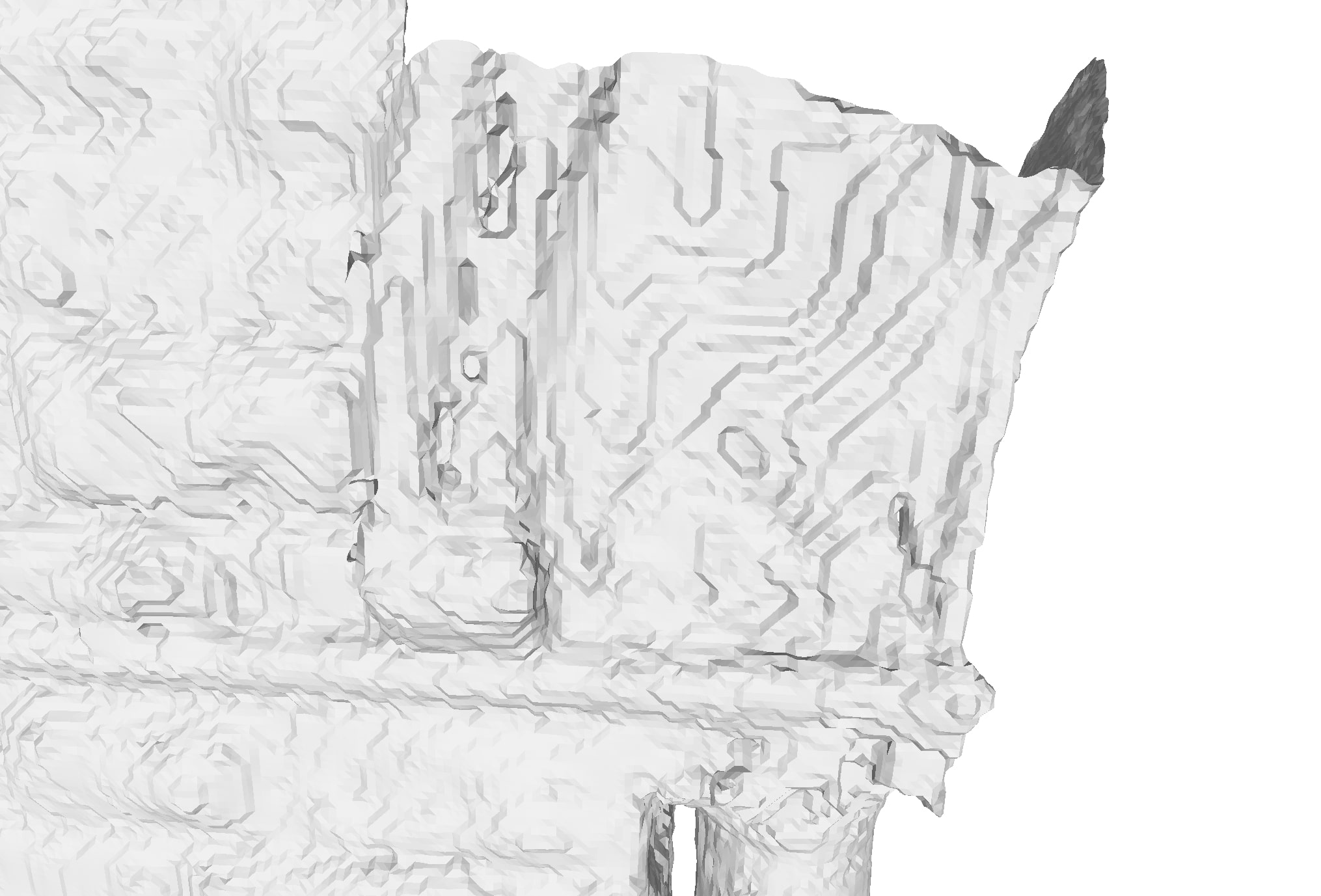}\\
            \includegraphics[width=1\textwidth]{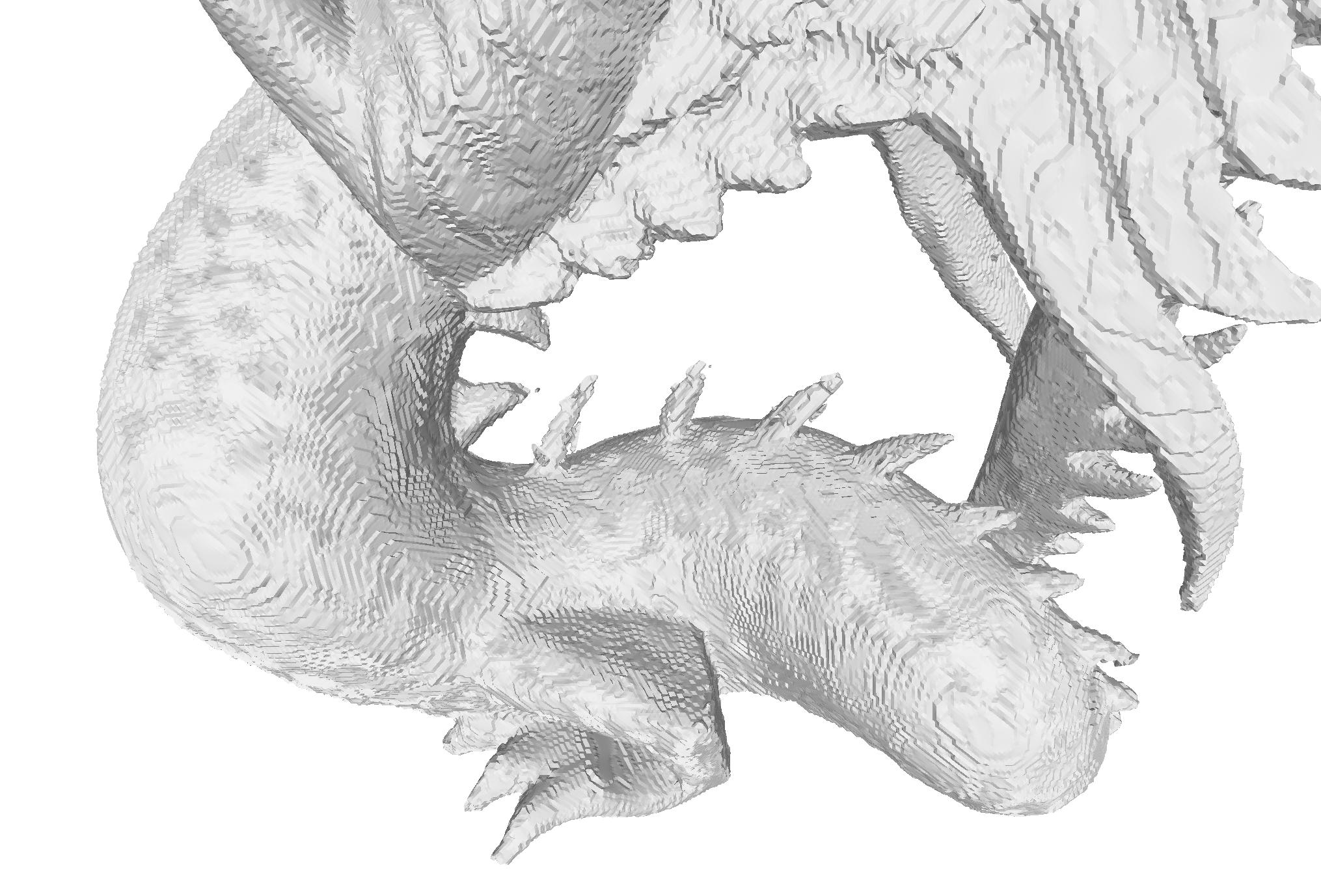} \\
            \includegraphics[width=1\textwidth]{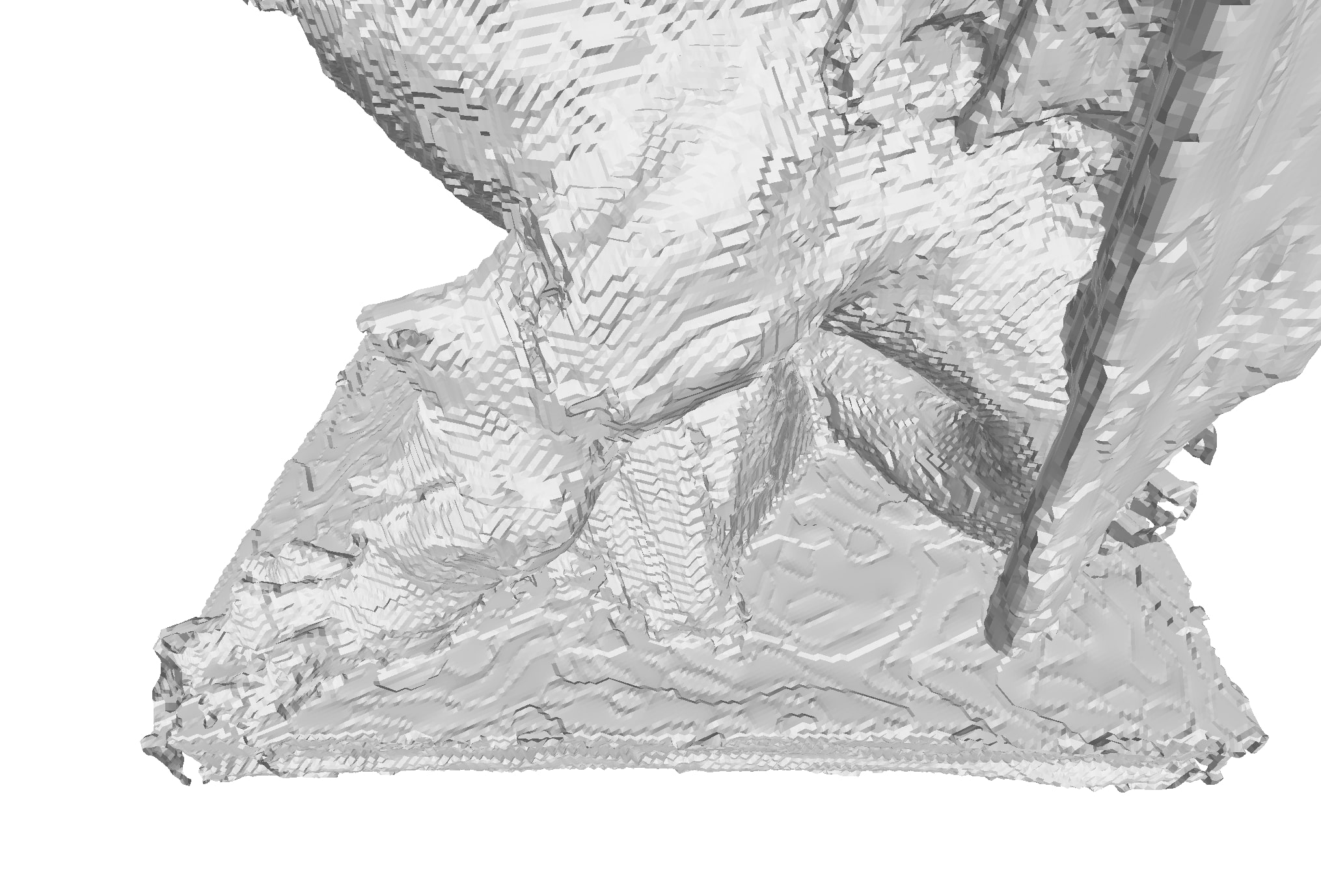}\\
            \includegraphics[width=1\textwidth]{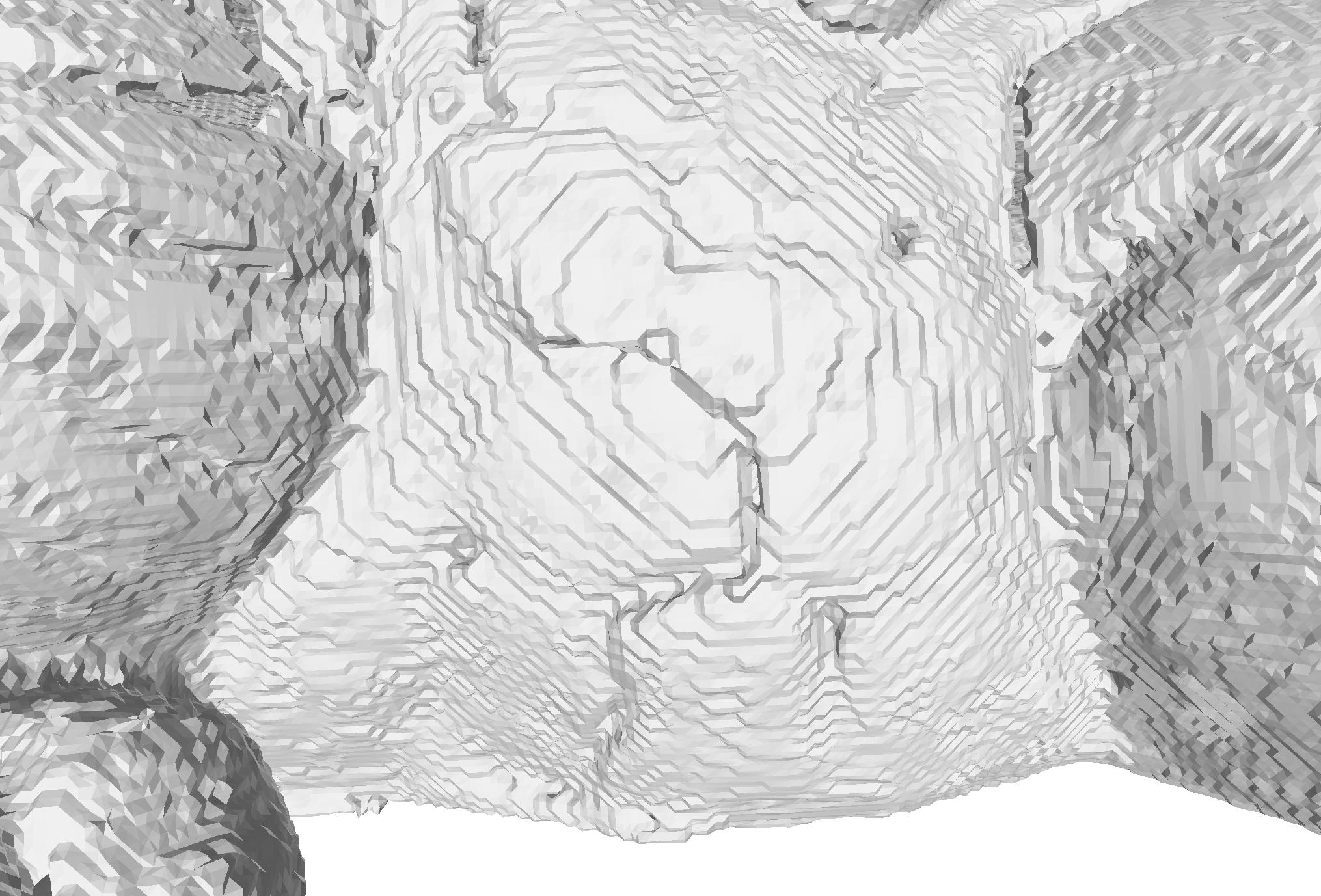}
        \end{minipage}
    }
    \subfigure[LevelSetUDF]{
        \label{fig:detail.lsudf}
        \begin{minipage}[b]{0.18\textwidth}
		    \includegraphics[width=1\textwidth]{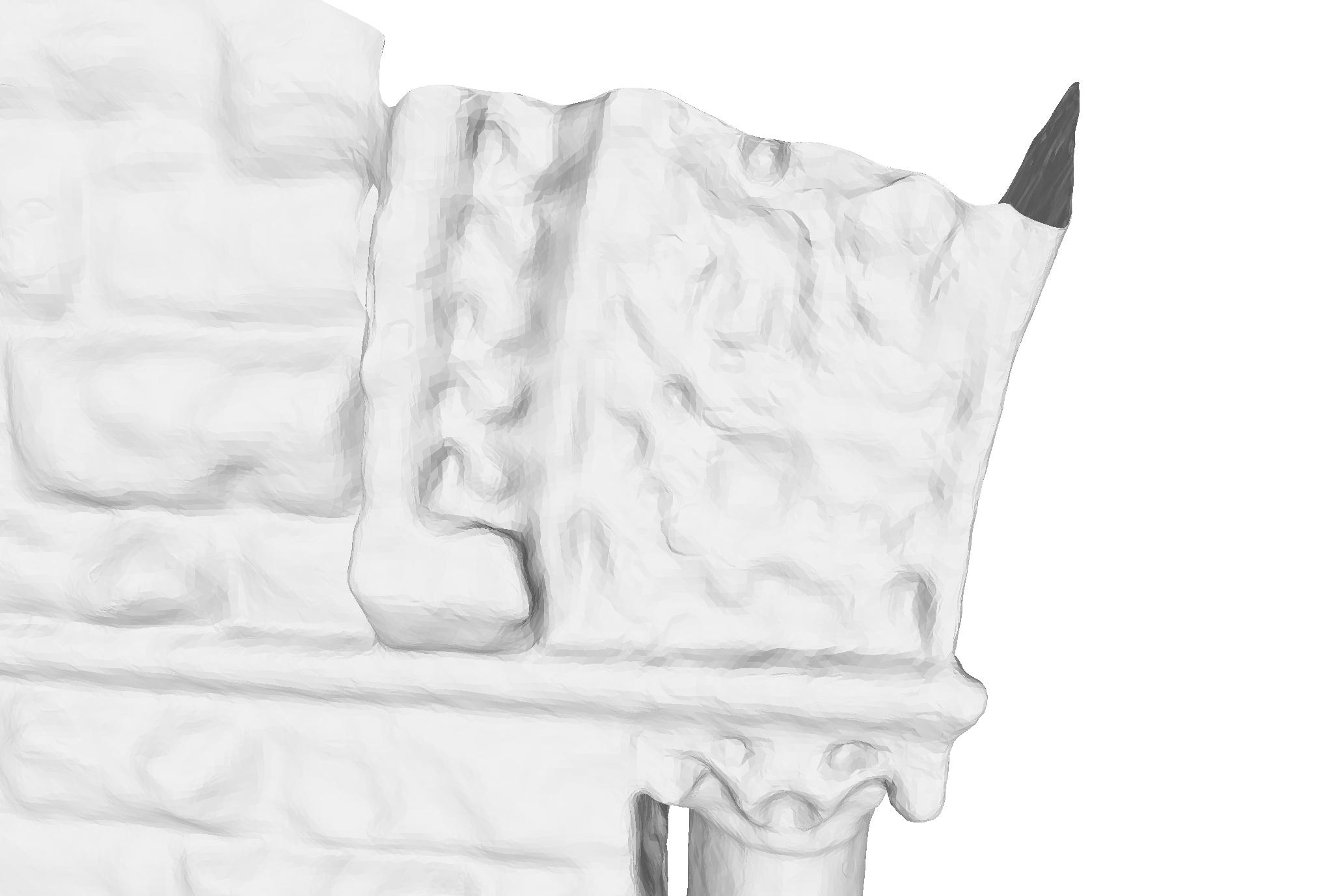}\\
            \includegraphics[width=1\textwidth]{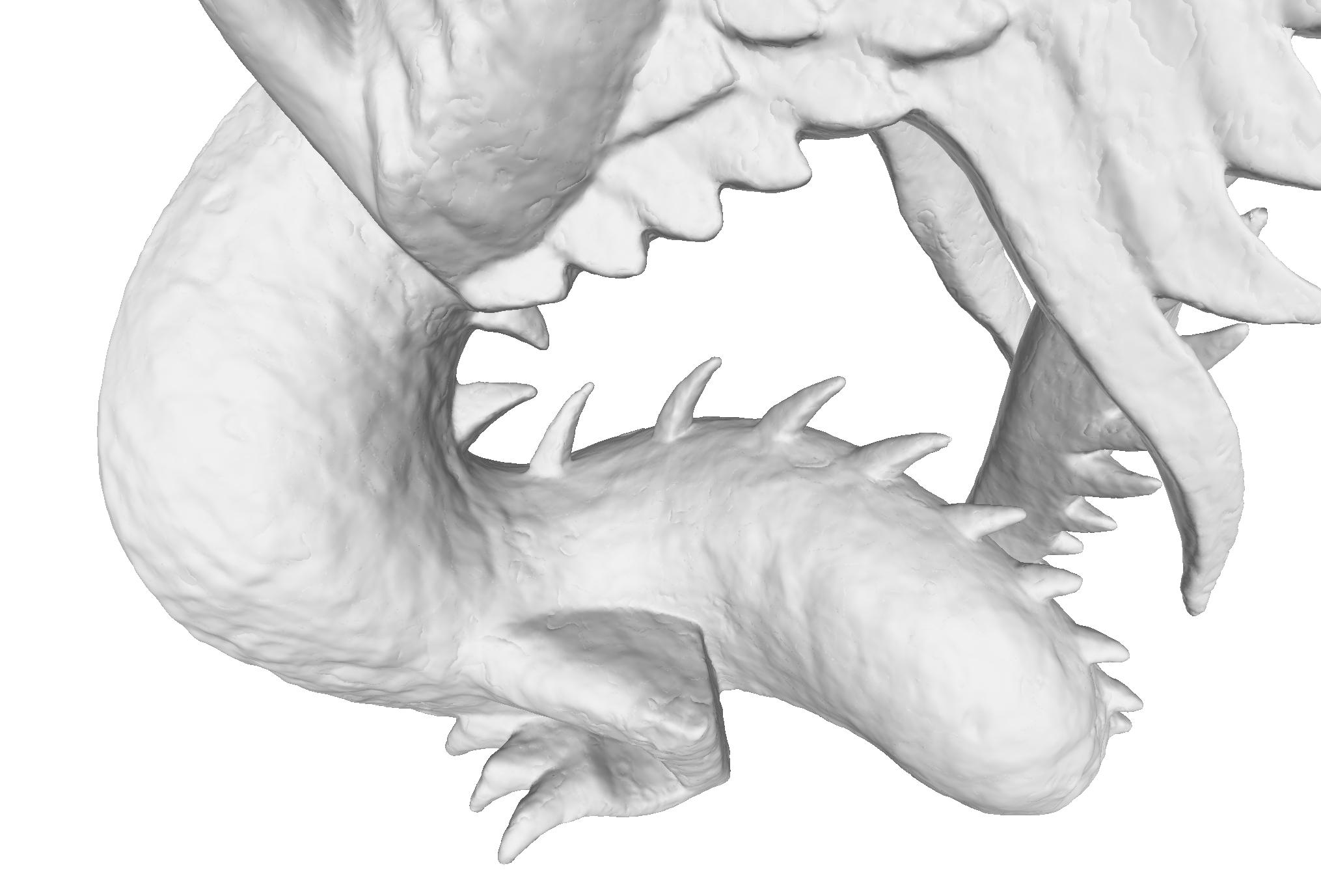} \\
  		    \includegraphics[width=1\textwidth]{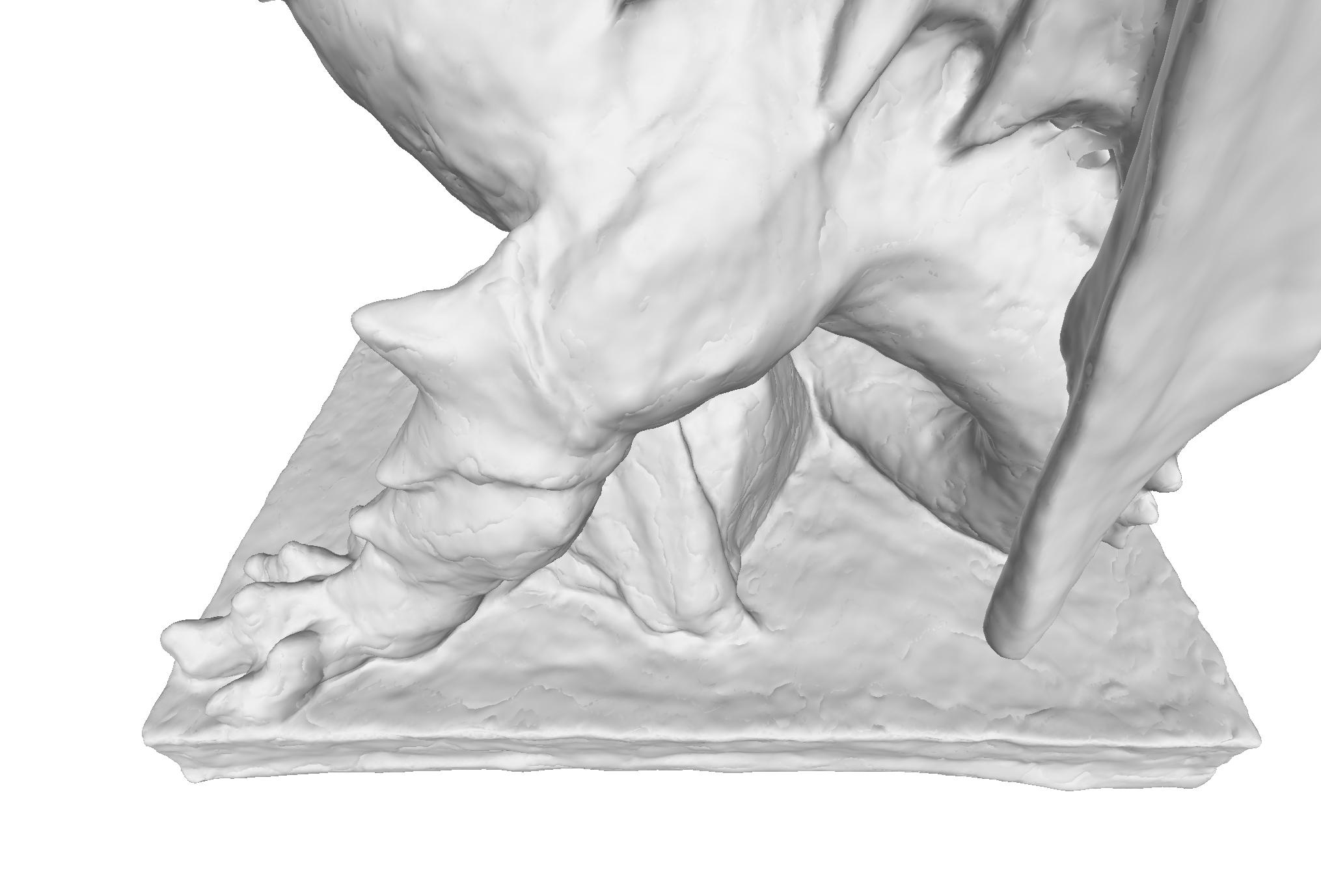}\\
  		    \includegraphics[width=1\textwidth]{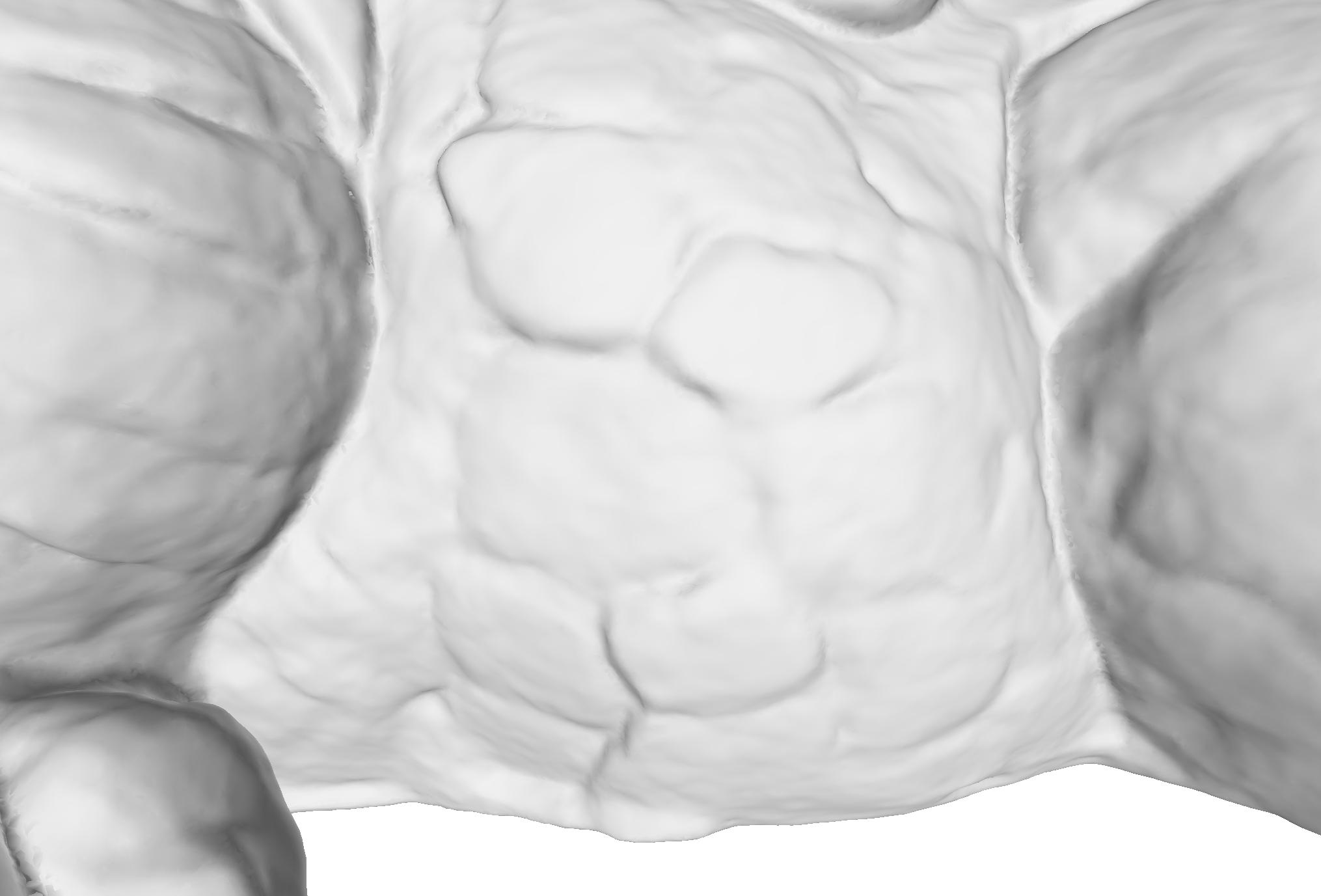}
        \end{minipage}
    }
    \subfigure[NSH]{
        \label{fig:detail.nsh}
        \begin{minipage}[b]{0.18\textwidth}
		    \includegraphics[width=1\textwidth]{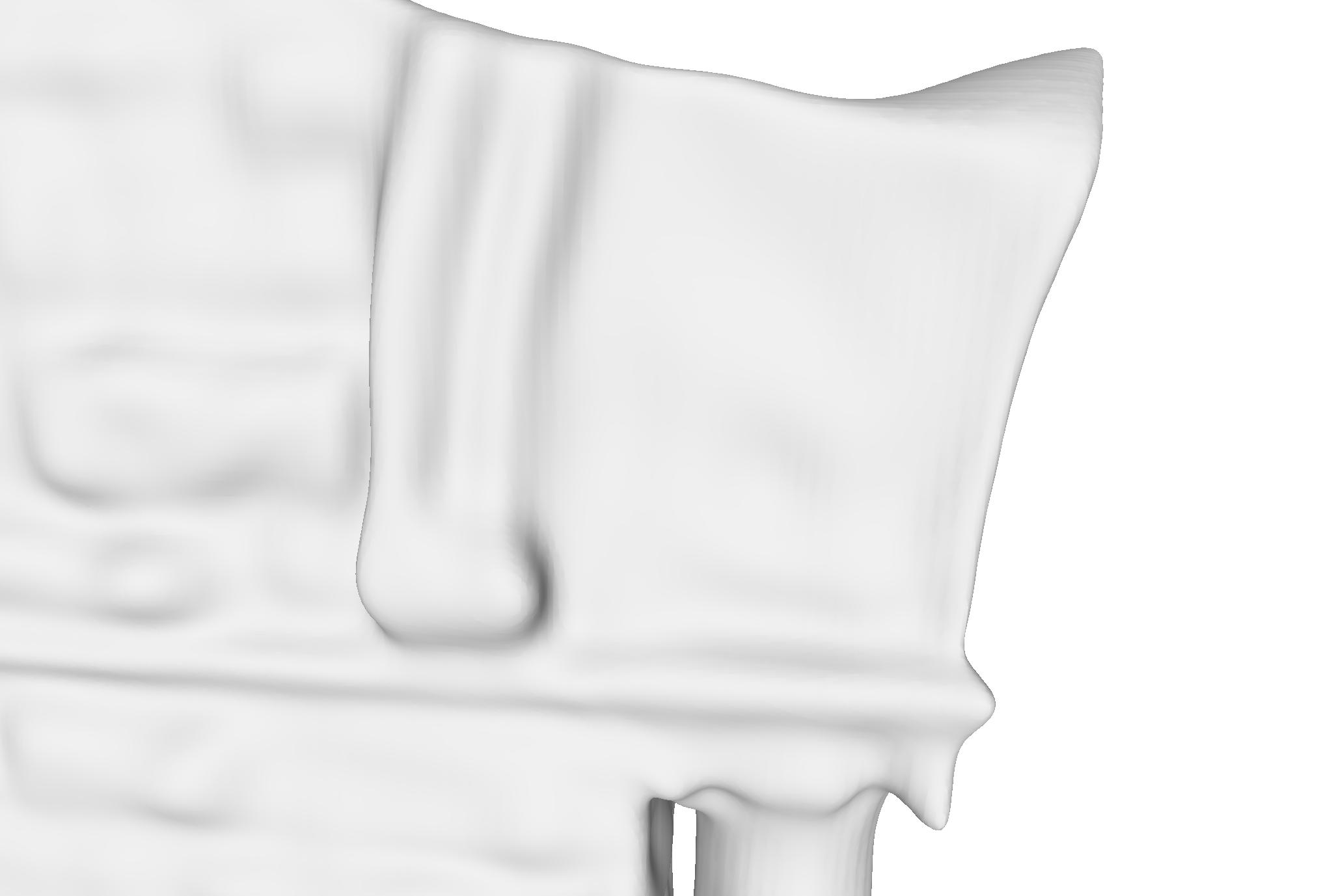}\\
            \includegraphics[width=1\textwidth]{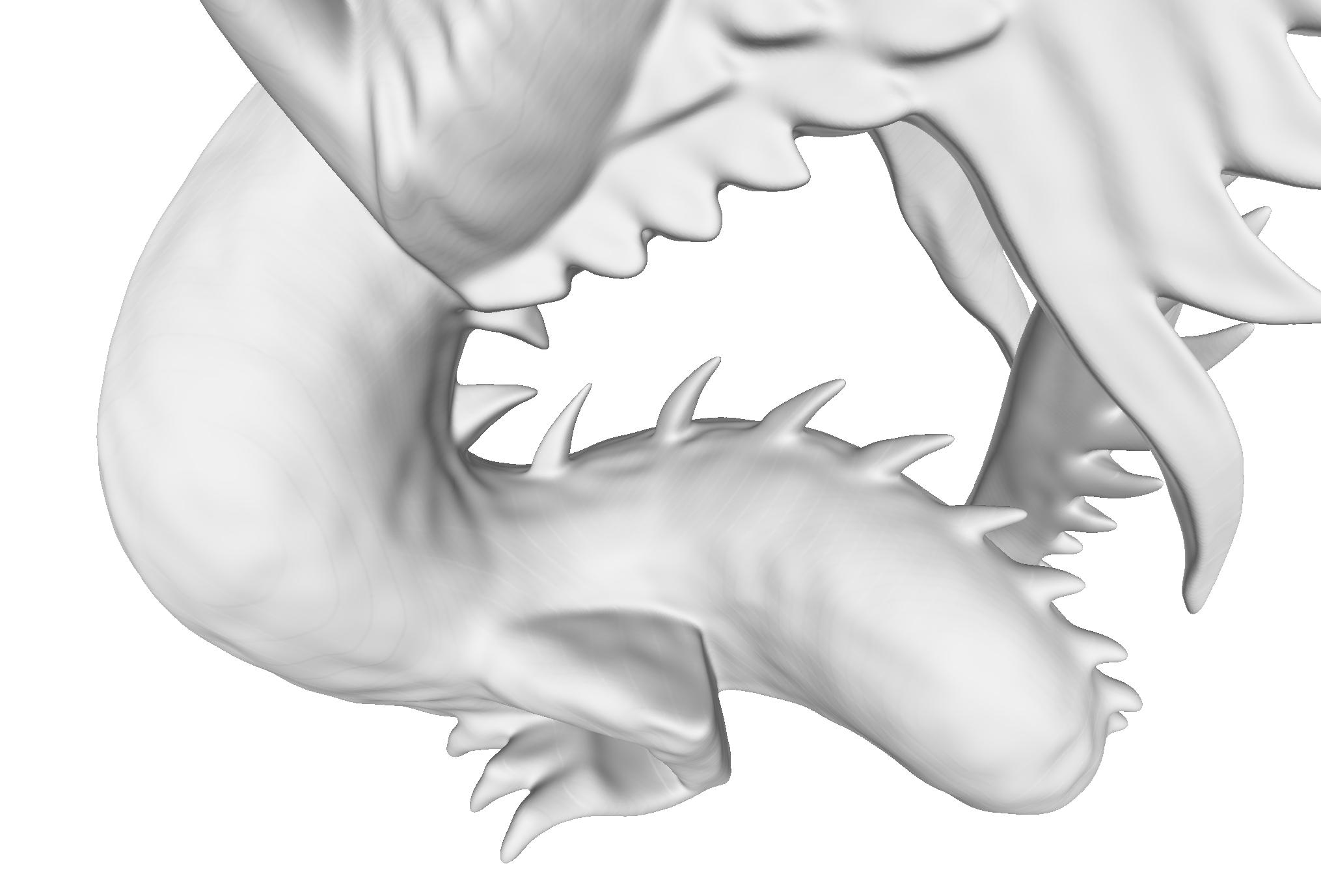} \\
            \includegraphics[width=1\textwidth]{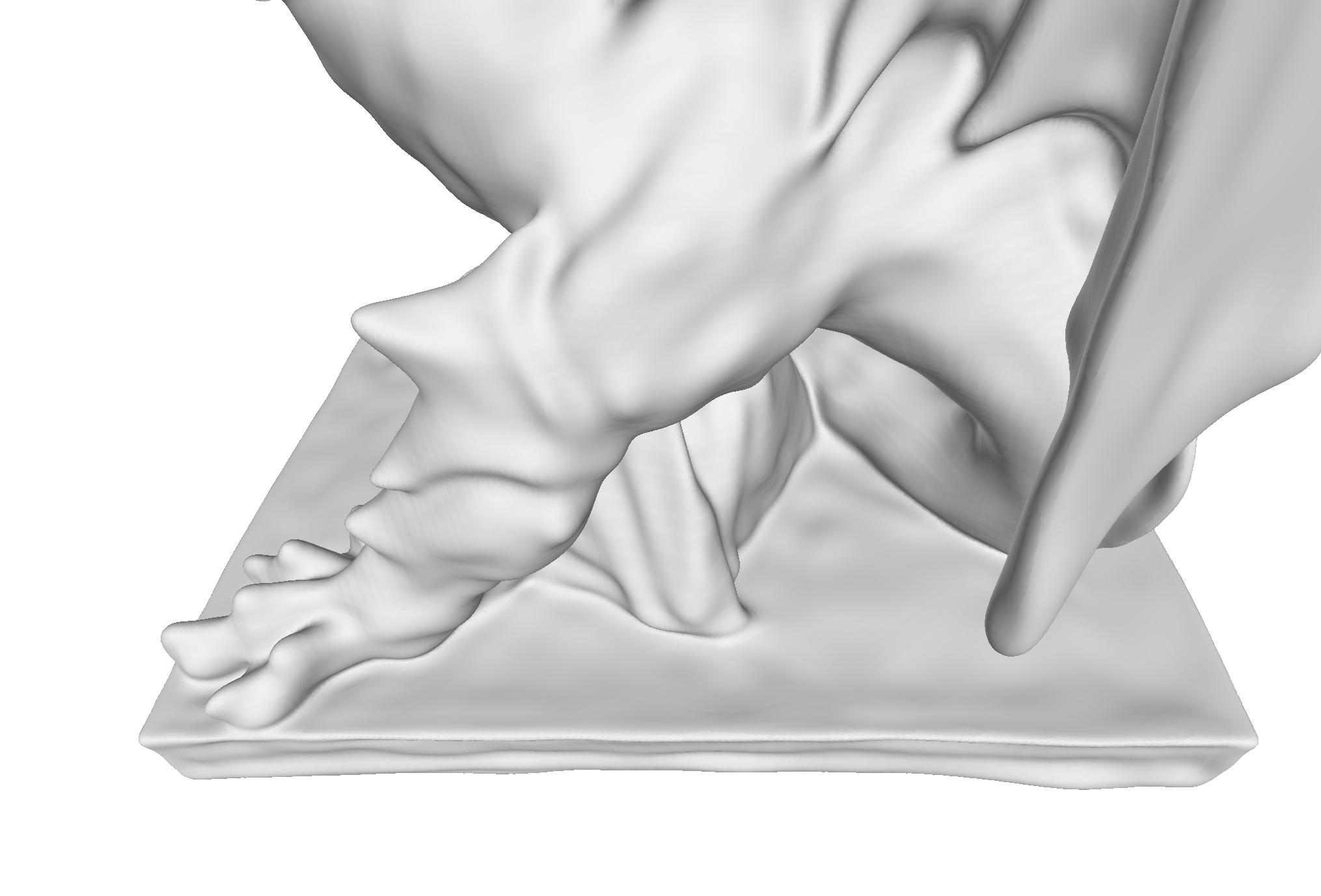}\\
            \includegraphics[width=1\textwidth]{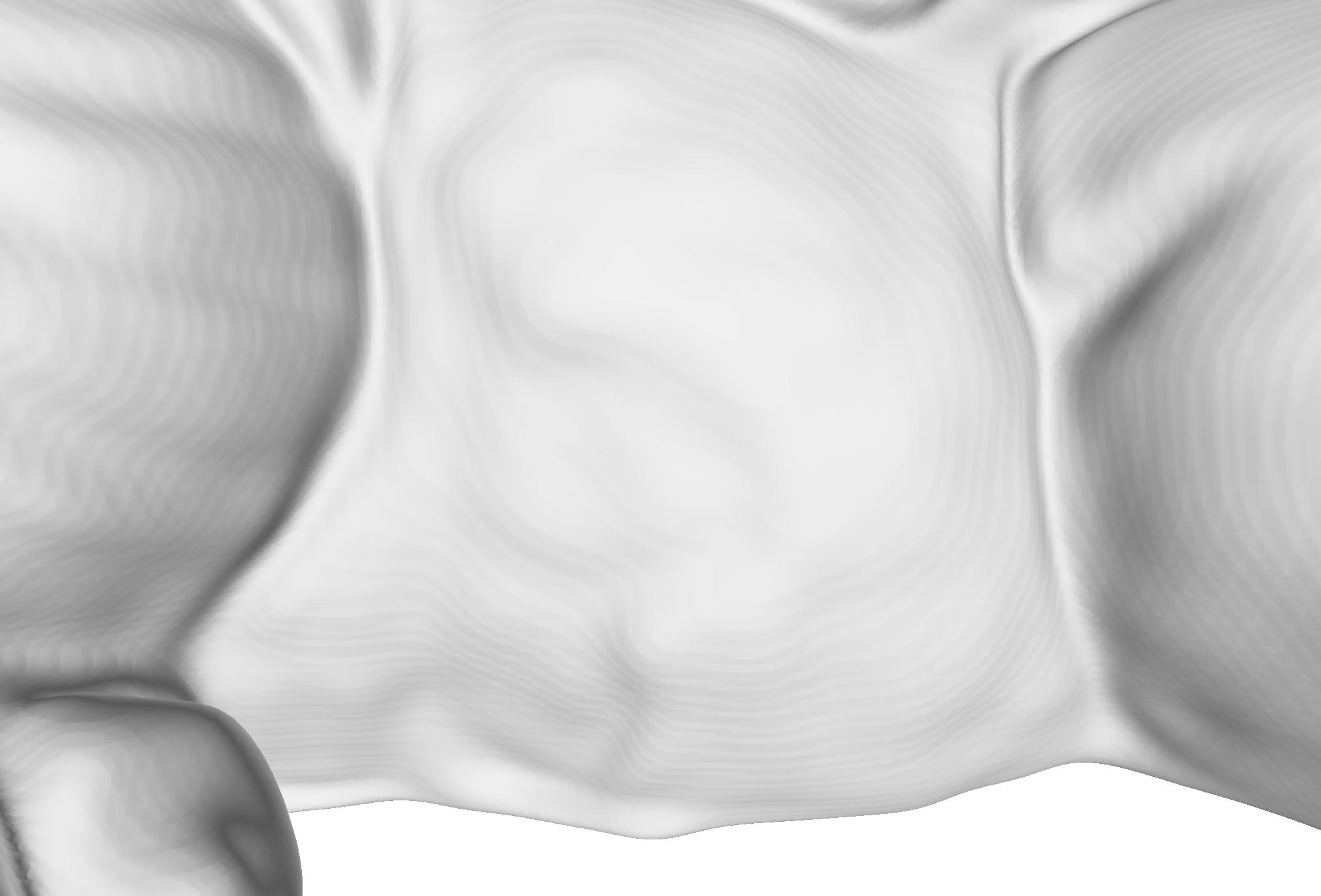}
        \end{minipage}
    }
    \subfigure[Ours]{
        \label{fig:detail.ours}
        \begin{minipage}[b]{0.18\textwidth}
            \includegraphics[width=1\textwidth]{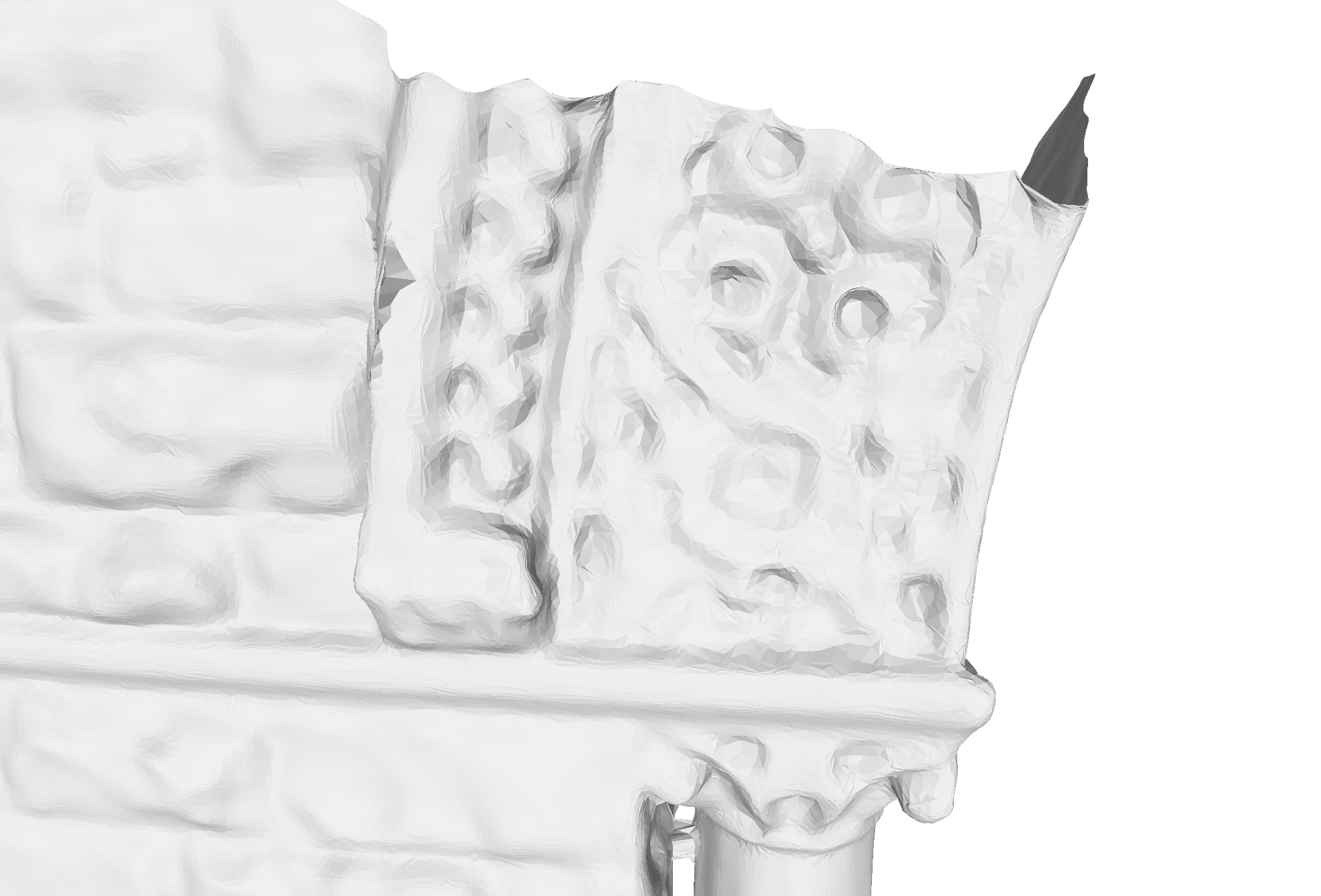}\\
            \includegraphics[width=1\textwidth]{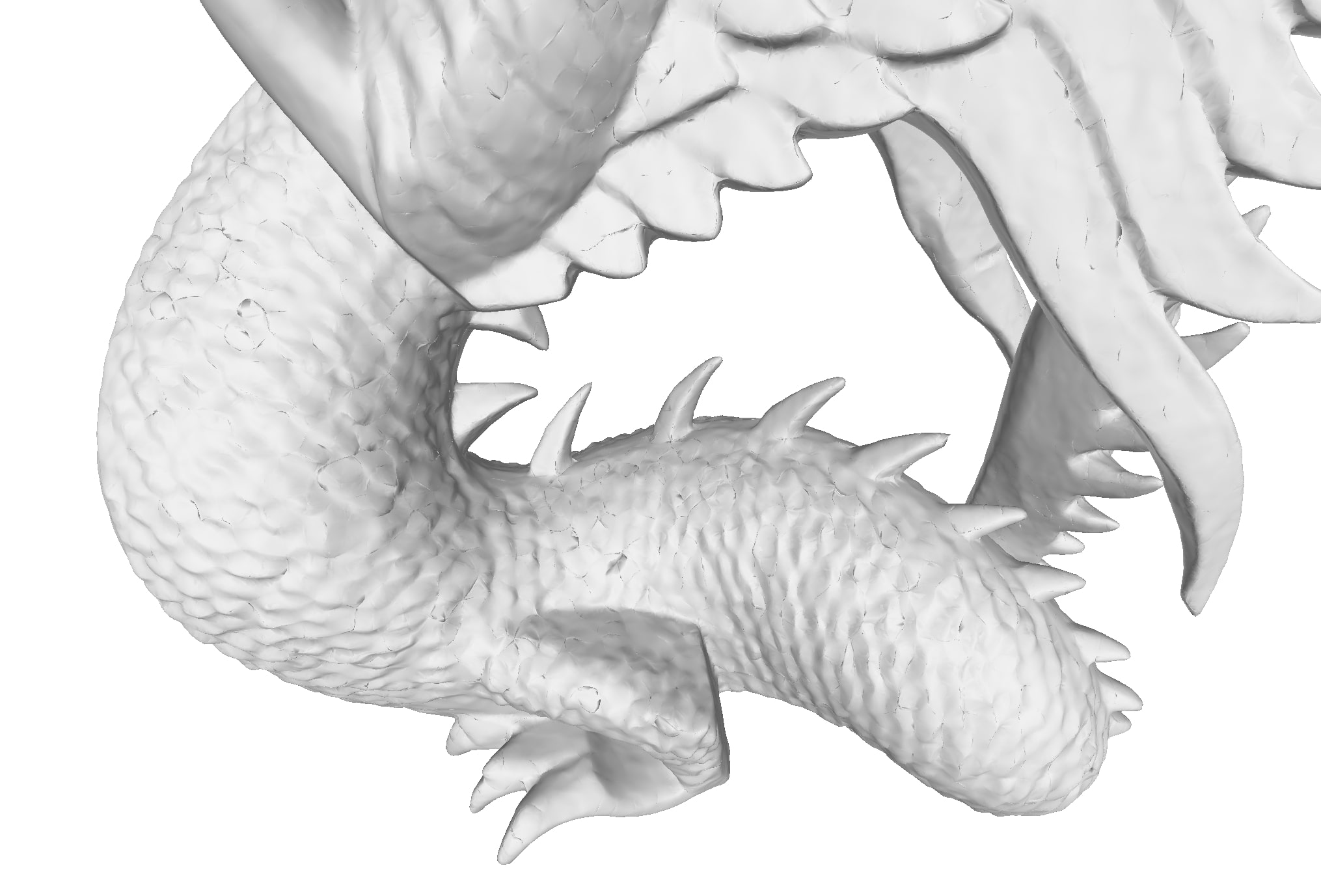} \\
  		    \includegraphics[width=1\textwidth]{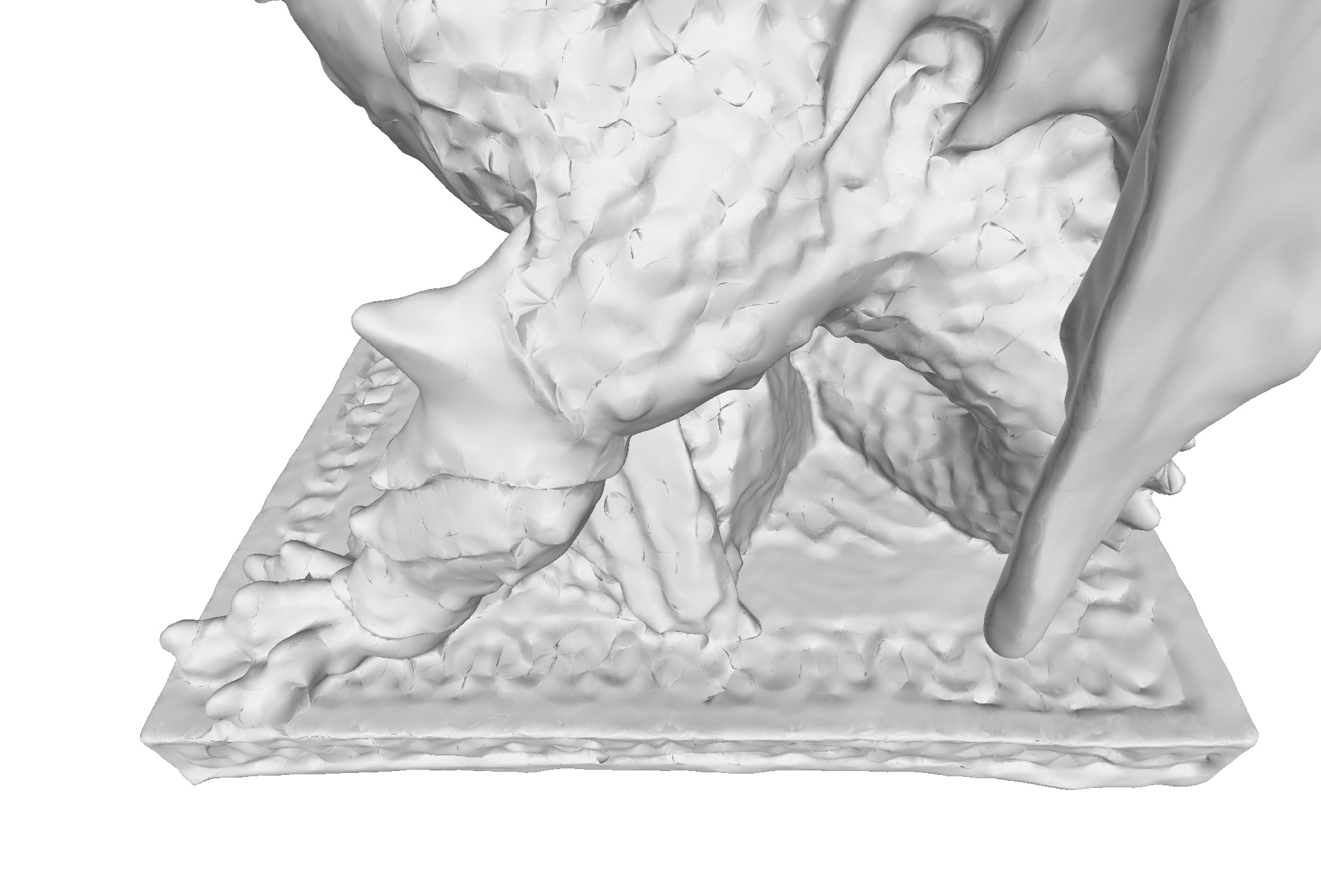}\\
            \includegraphics[width=1\textwidth]{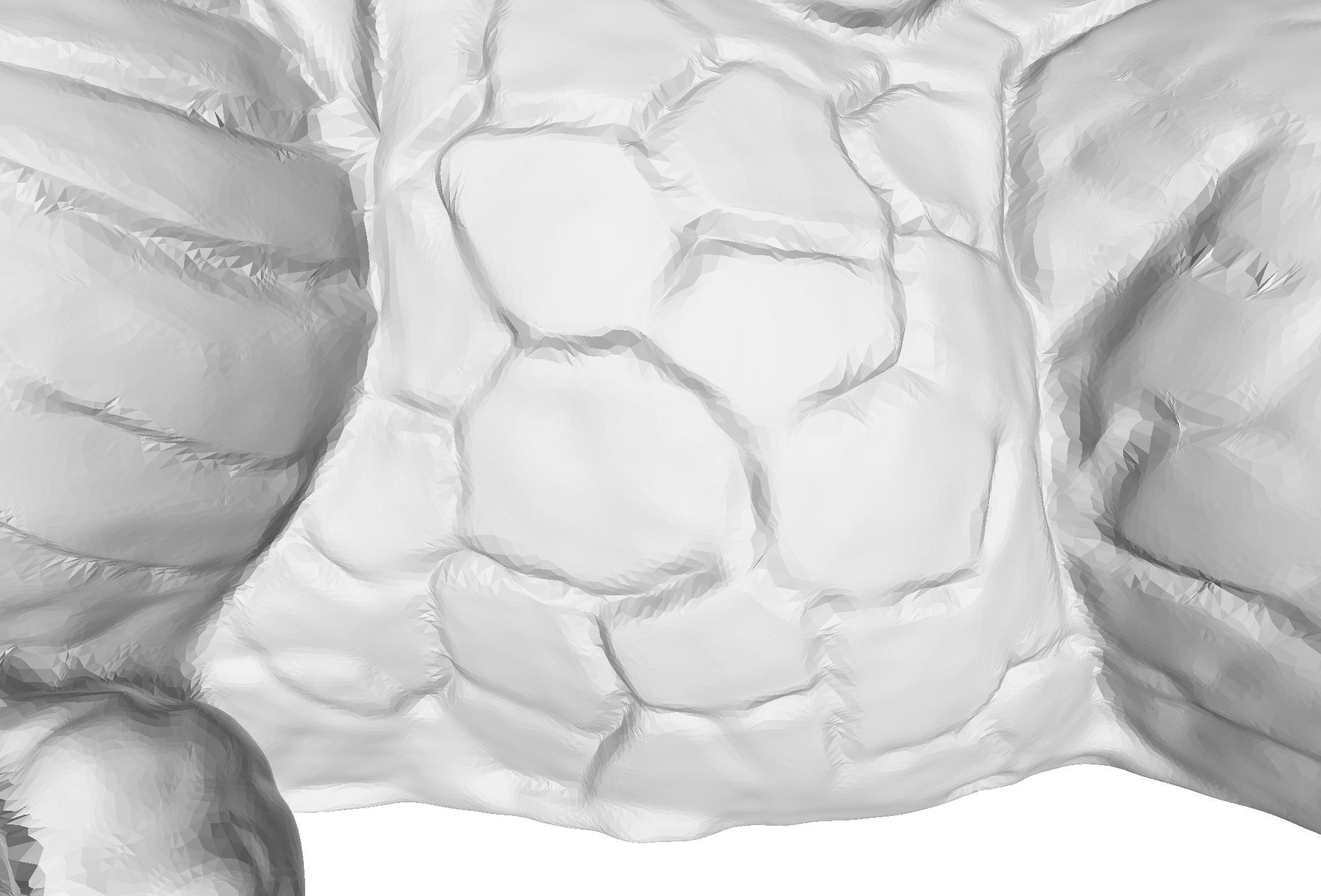}
        \end{minipage}
    }
    \subfigure[GT]{
        \label{fig:detail.gt}
        \begin{minipage}[b]{0.18\textwidth}
  		    \includegraphics[width=1\textwidth]{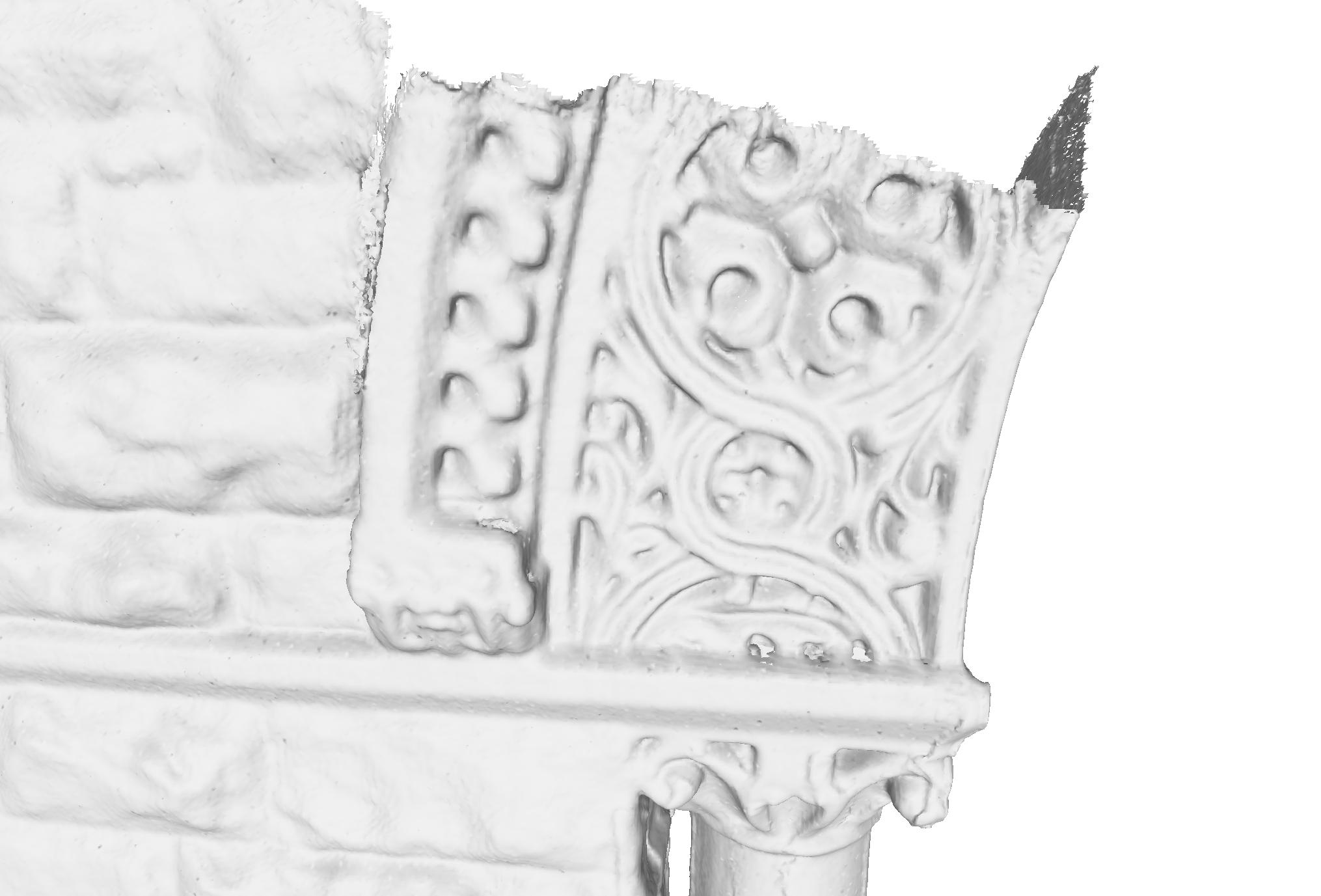}\\
            \includegraphics[width=1\textwidth]{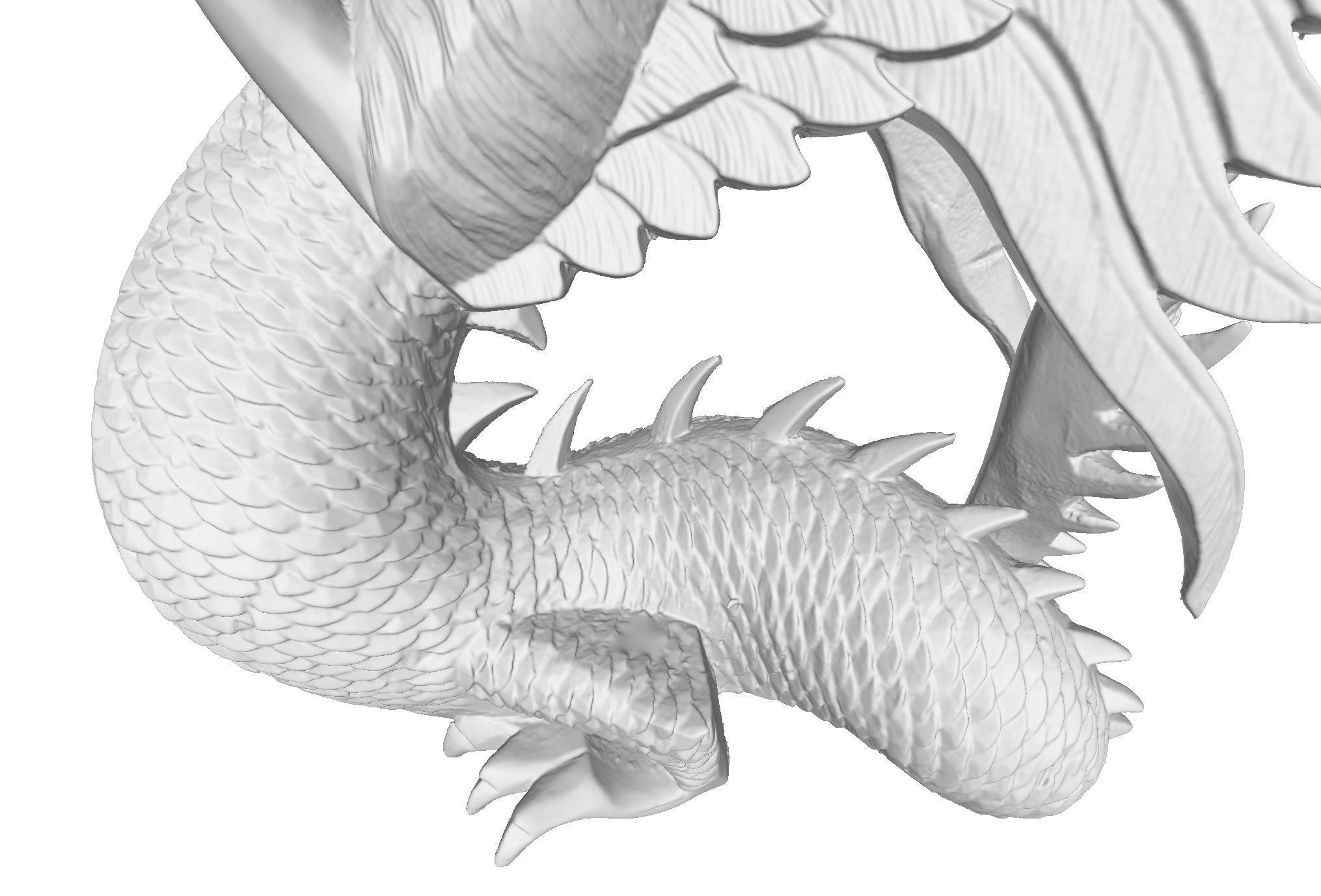} \\
            \includegraphics[width=1\textwidth]{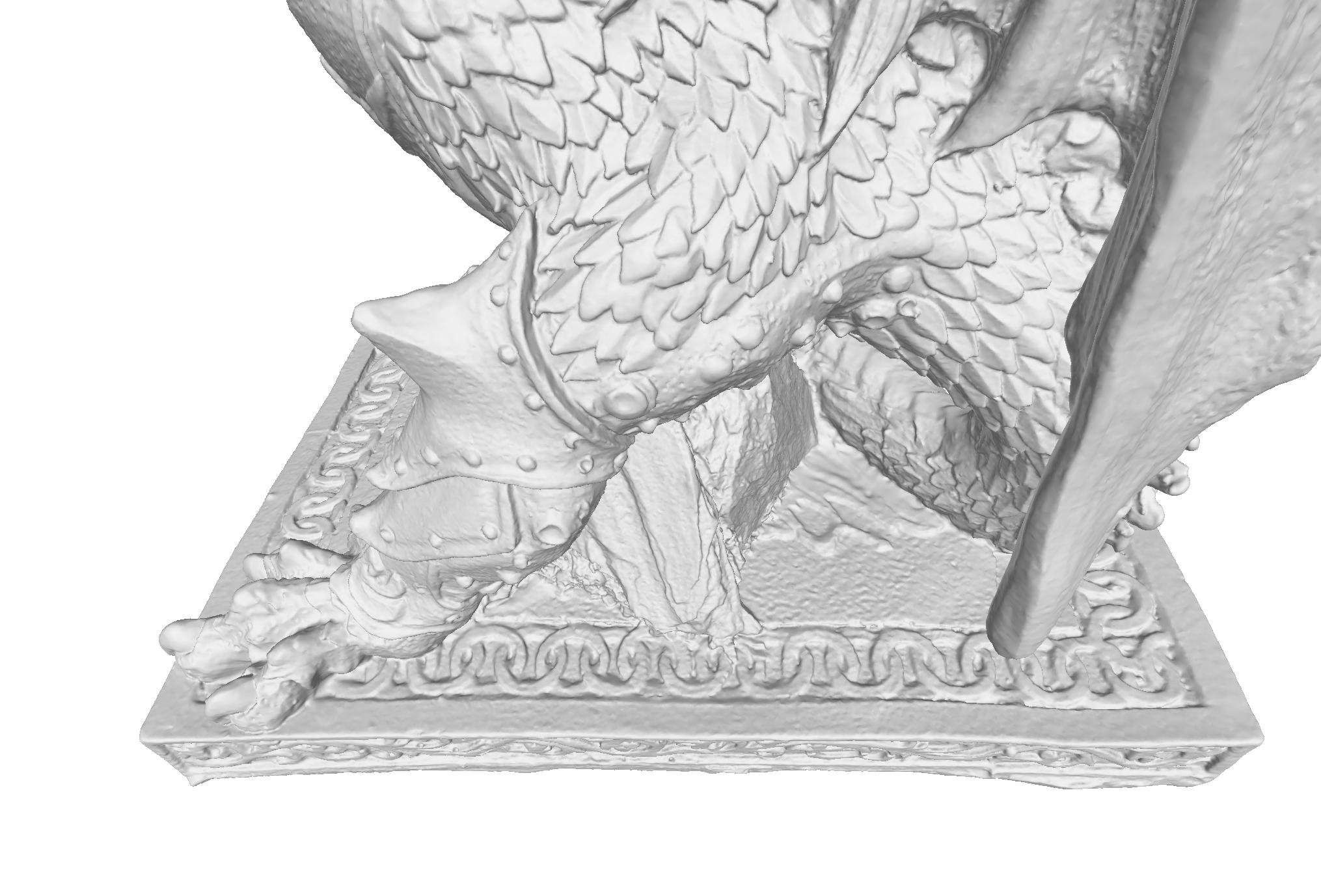}\\
            \includegraphics[width=1\textwidth]{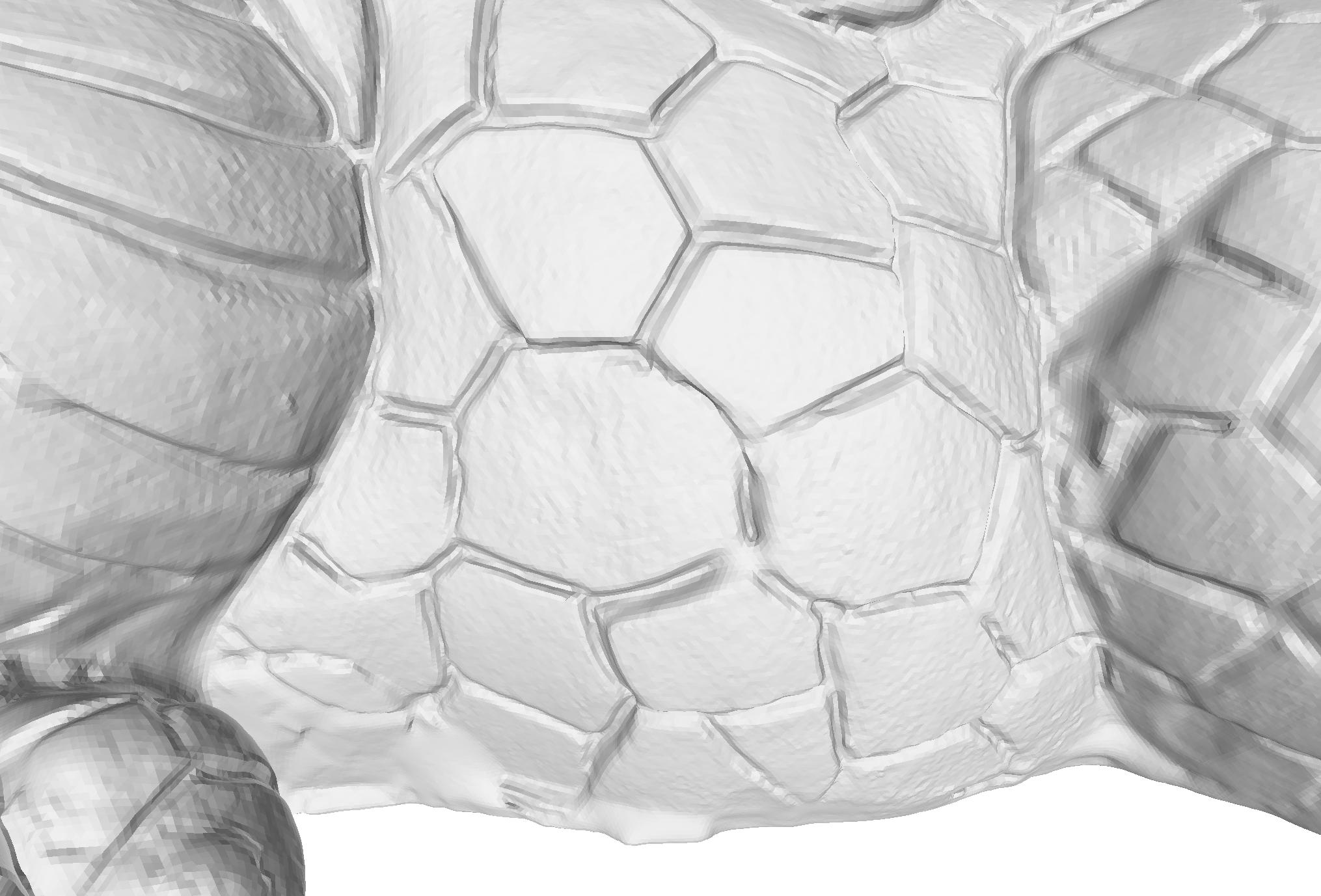}
        \end{minipage}
    }

    \caption{Visual comparison with two recent UDF learning approaches, DUDF~\cite{fainstein2024dudf} and LevelSetUDF~\cite{Zhou2023levelset}, and one recent SDF learning method, NSH~\cite{NSH}, on surfaces with fine geometric details. Our method yields visually pleasing results, reconstructing significantly more details than the other methods. }
    \label{fig:detail-surfcomparison}
\end{figure*}

\subsection{Datasets}
\label{subsec:datasets}

We evaluate our method using three datasets: ShapeNet-Cars~\cite{Chang2015} with 108 models\footnote{We select all models whose names start with ``1''.}, the Stanford 3D Scene Dataset~\cite{3DSceneDataset} with 5 models and the Stanford 3D Scan Repository \footnote{\url{https://graphics.stanford.edu/data/3Dscanrep/}} with 8 models. For each shape, we randomly sample 300K points as input. After learning the UDFs, we employ DCUDF with a resolution of $512^3$ to extract the target surface. To evaluate the accuracy, We use Chamfer distance (CD) and F-score as quantitative measures. For F-score, we set the thresholds to $0.5\%$ and $0.25\%$. Following previous methods~\cite{Zhou2022CAPUDF,Zhou2023levelset}, we randomly sample 100K points from both the reconstructed surfaces and the ground truth meshes for computing CD and F-score. All points and meshes are normalized to bounding boxes whose longest edges are 2. We test on an NVIDIA Tesla V100 GPU with 32GB memory (about 5GB used for a UDF learning). It takes about 30 minutes to learn a UDF.

\subsection{Results \& comparisons}

We compare our method with three state-of-the-art UDF learning methods: LevelSetUDF~\cite{Zhou2023levelset}, CAP-UDF~\cite{Zhou2022CAPUDF} and DUDF~\cite{fainstein2024dudf}. Since we adopt DCUDF~\cite{Hou2023} for surface extraction, we also test DCUDF for the three baselines to ensure fairness of comparisons. For CAP-UDF and LevelSetUDF, we observe that DCUDF could produce better results than their original implementations in terms of Chamfer distances and visual effects. But for DUDF, the results of DCUDF are not as good as the original results. Therefore, to report the best results of the baseline methods, we choose to adopt DCUDF for extracting the zero level set from the UDF outputs from both CAP-UDF and LevelSetUDF. While DUDF uses their original results for comparisons. For the ShapeNet-Cars and Stanford 3D Scene datasets, the outputs of DCUDF remain as double-layered meshes, bypassing the min-cut based double-layer segmentation post-processing. In addition, we assess our approach against IDF~\cite{Wang2022IDF} and NSH~\cite{NSH}, two state-of-the-art SDF learning methods, on the watertight surfaces with fine geometric details from the Stanford 3D Scan Repository. The results are illustrated in Table~\ref{tab:surf-comp} and Figure~\ref{fig:detail-surfcomparison}.

\paragraph{3D objects with fine geometric details}
To explore the ability of our method for representing 3D objects with fine geometric details, we evaluate our method on Stanford 3D Scene dataset and Stanford 3D Scan dataset. As shown in Table~\ref{tab:surf-comp} and Figure~\ref{fig:detail-surfcomparison},our method achieves the best performance in UDF-based methods, and performs close to SDF-base methods.

\begin{figure}[!htbp]
    \centering
    \subfigure[DUDF]{
        \begin{minipage}[b]{0.1\textwidth}
  		\includegraphics[width=1\textwidth]{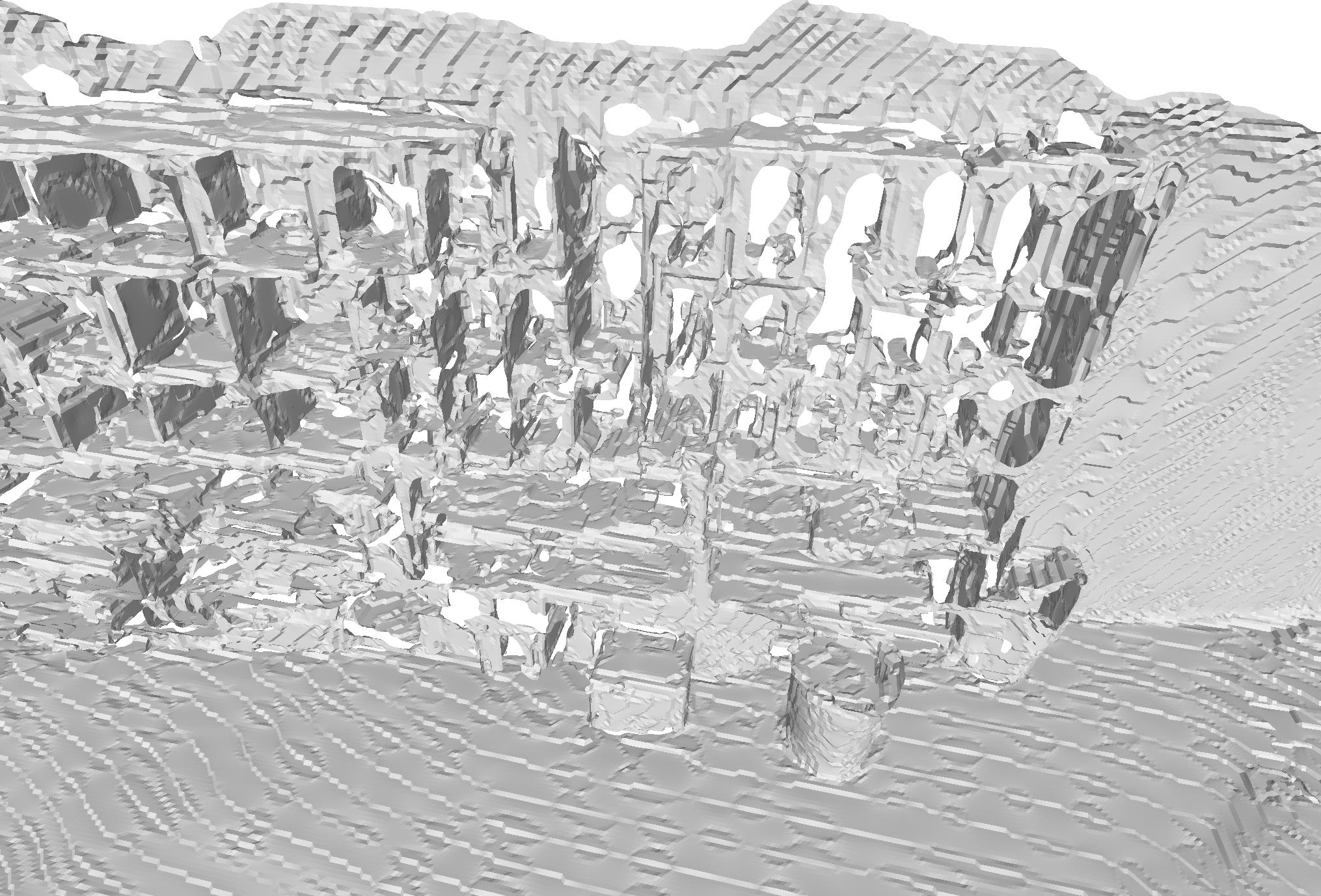}
        \end{minipage}
        \begin{minipage}[b]{0.1\textwidth}
            \includegraphics[width=1\textwidth]{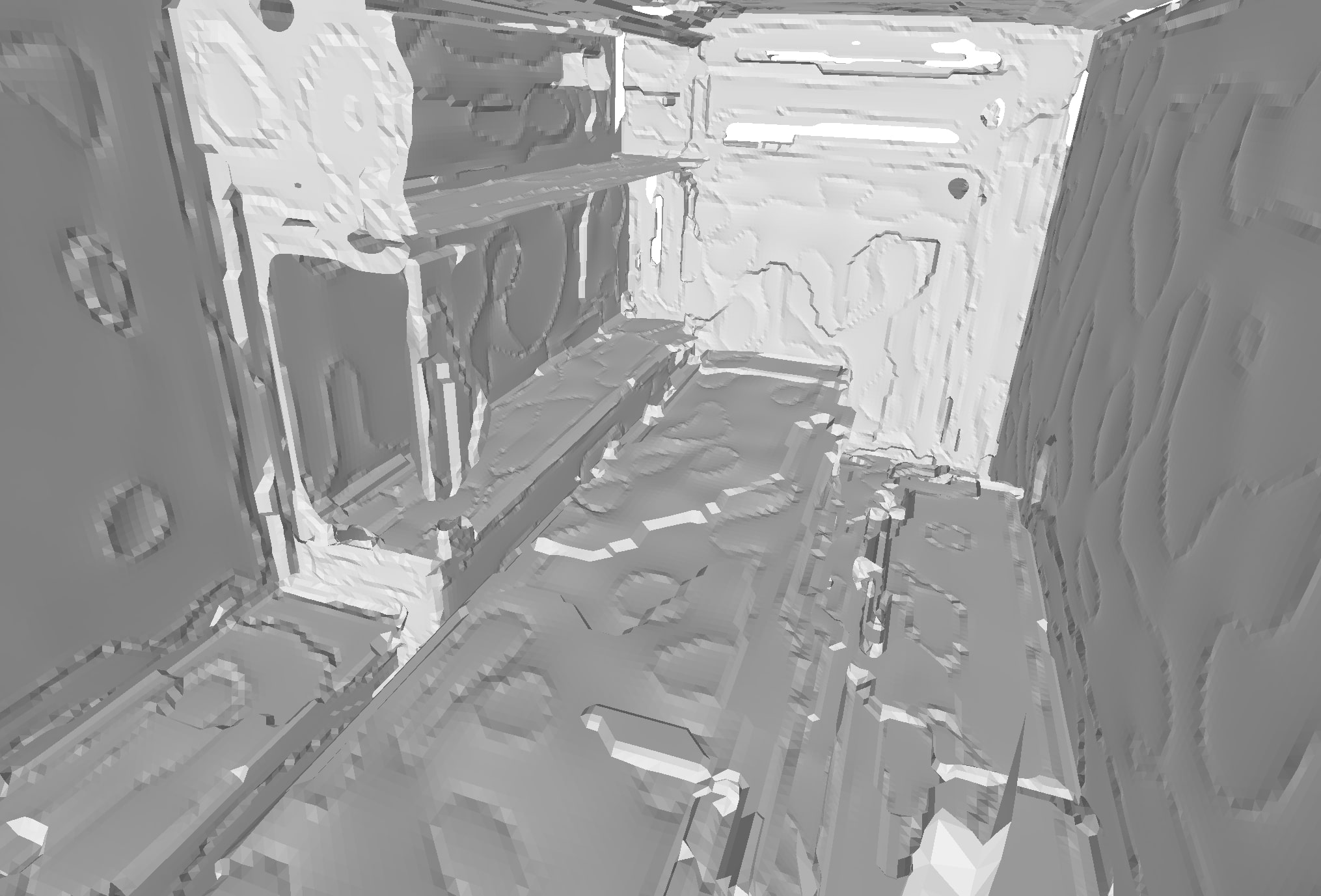}
        \end{minipage}
        \begin{minipage}[b]{0.1\textwidth}
  		\includegraphics[width=1\textwidth]{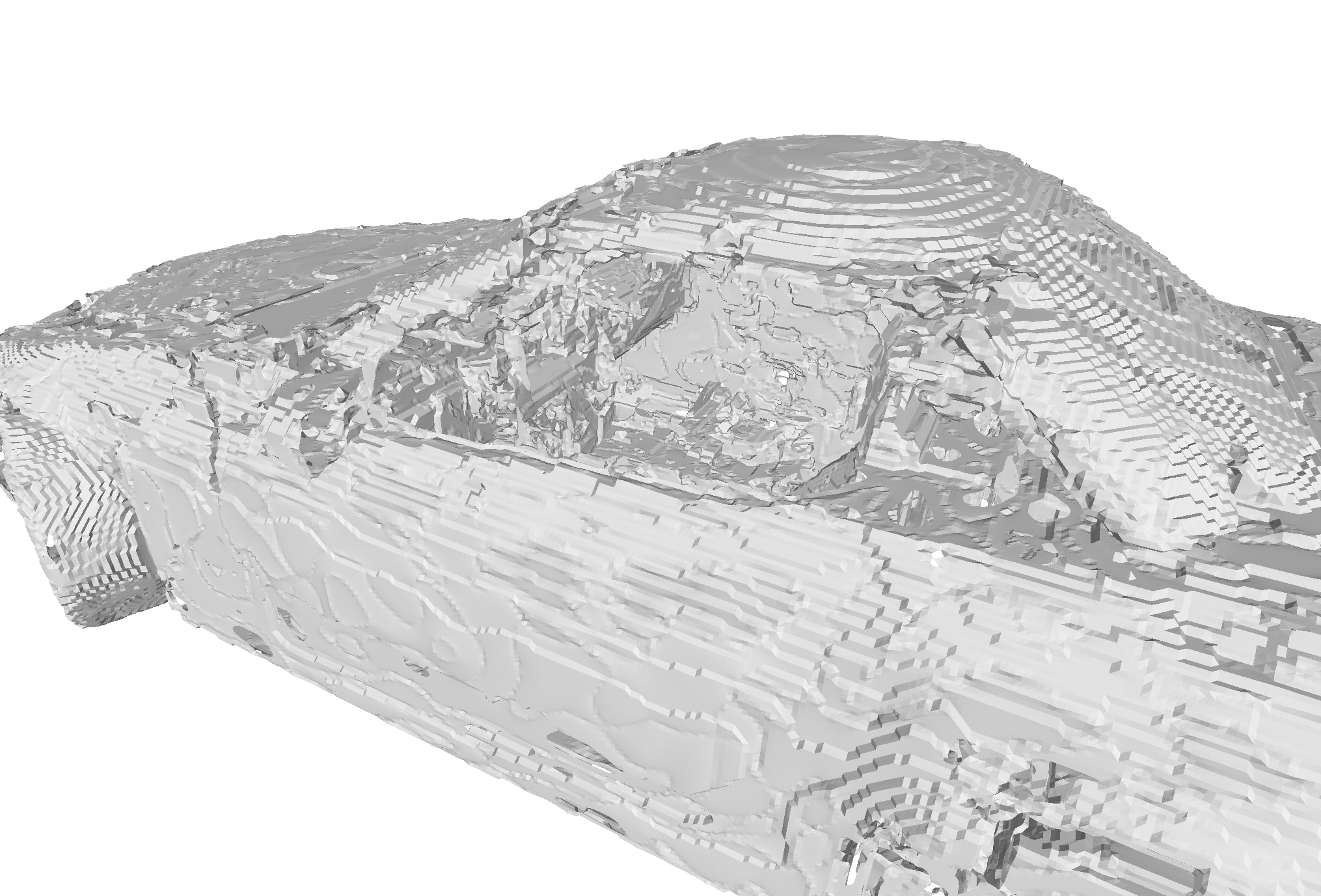}
        \end{minipage}
        \begin{minipage}[b]{0.1\textwidth}
            \includegraphics[width=1\textwidth]{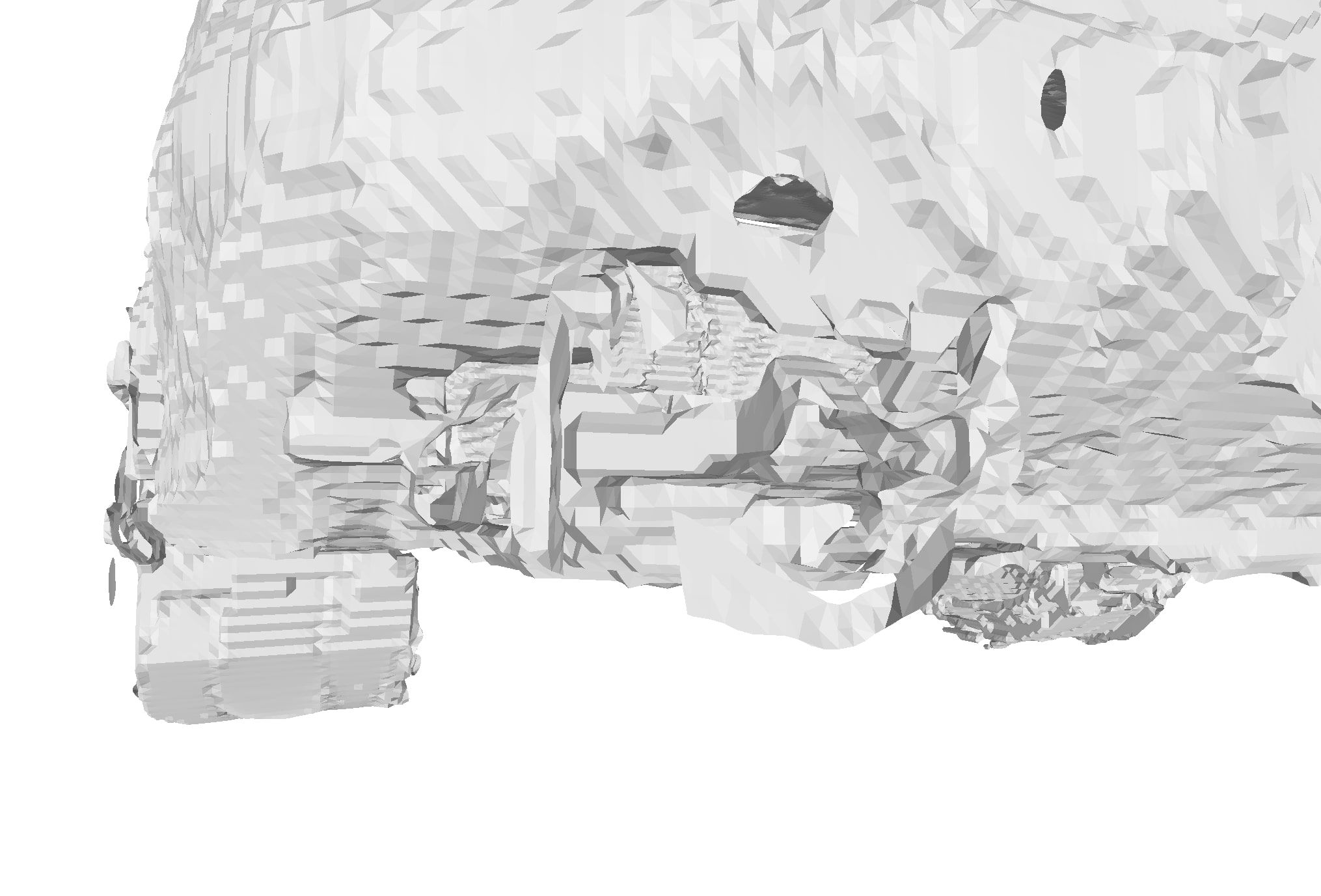} 
        \end{minipage}
    }
    \subfigure[CAP-UDF]{
        \begin{minipage}[b]{0.1\textwidth}
  		\includegraphics[width=1\textwidth]{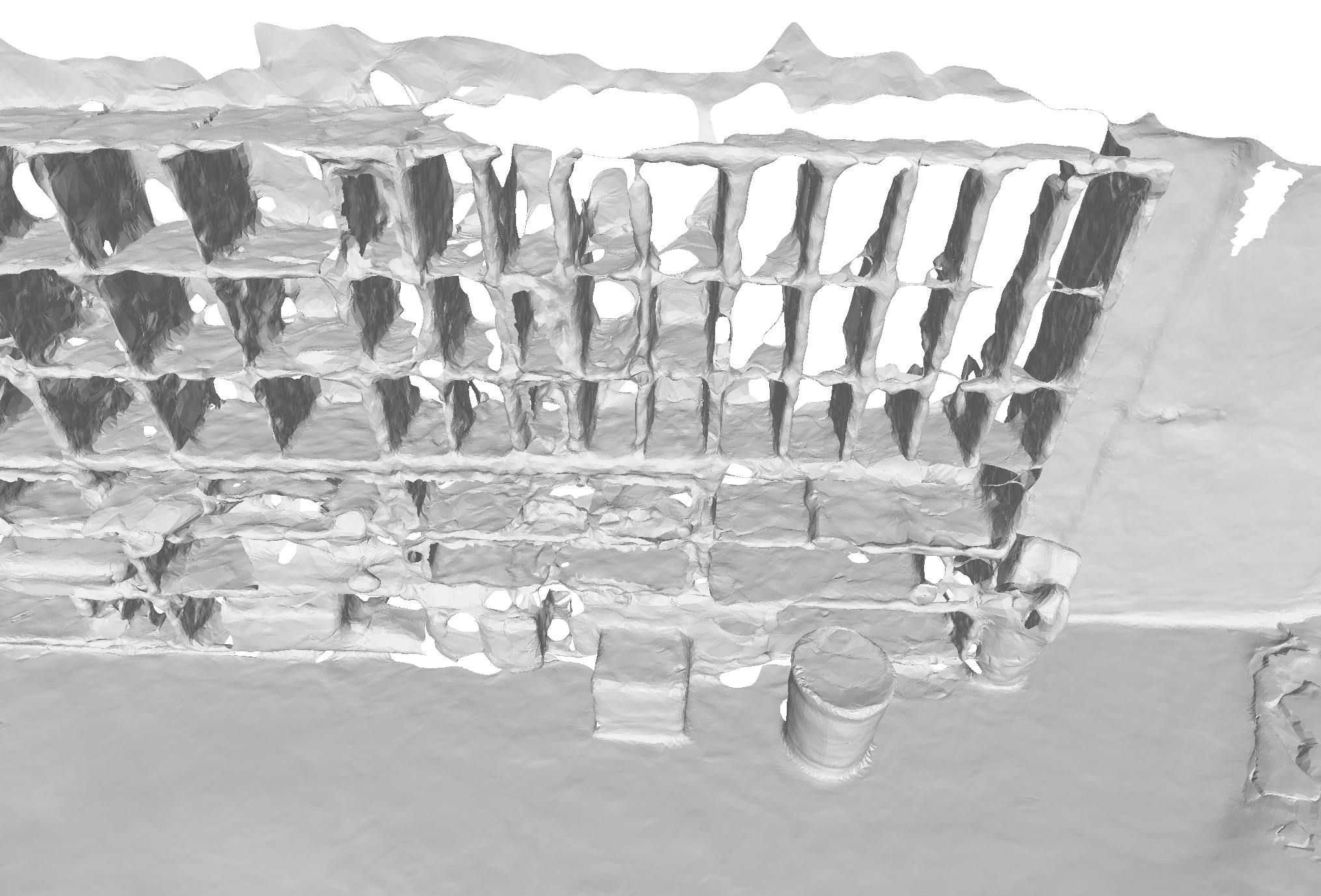}
        \end{minipage}
        \begin{minipage}[b]{0.1\textwidth}
            \includegraphics[width=1\textwidth]{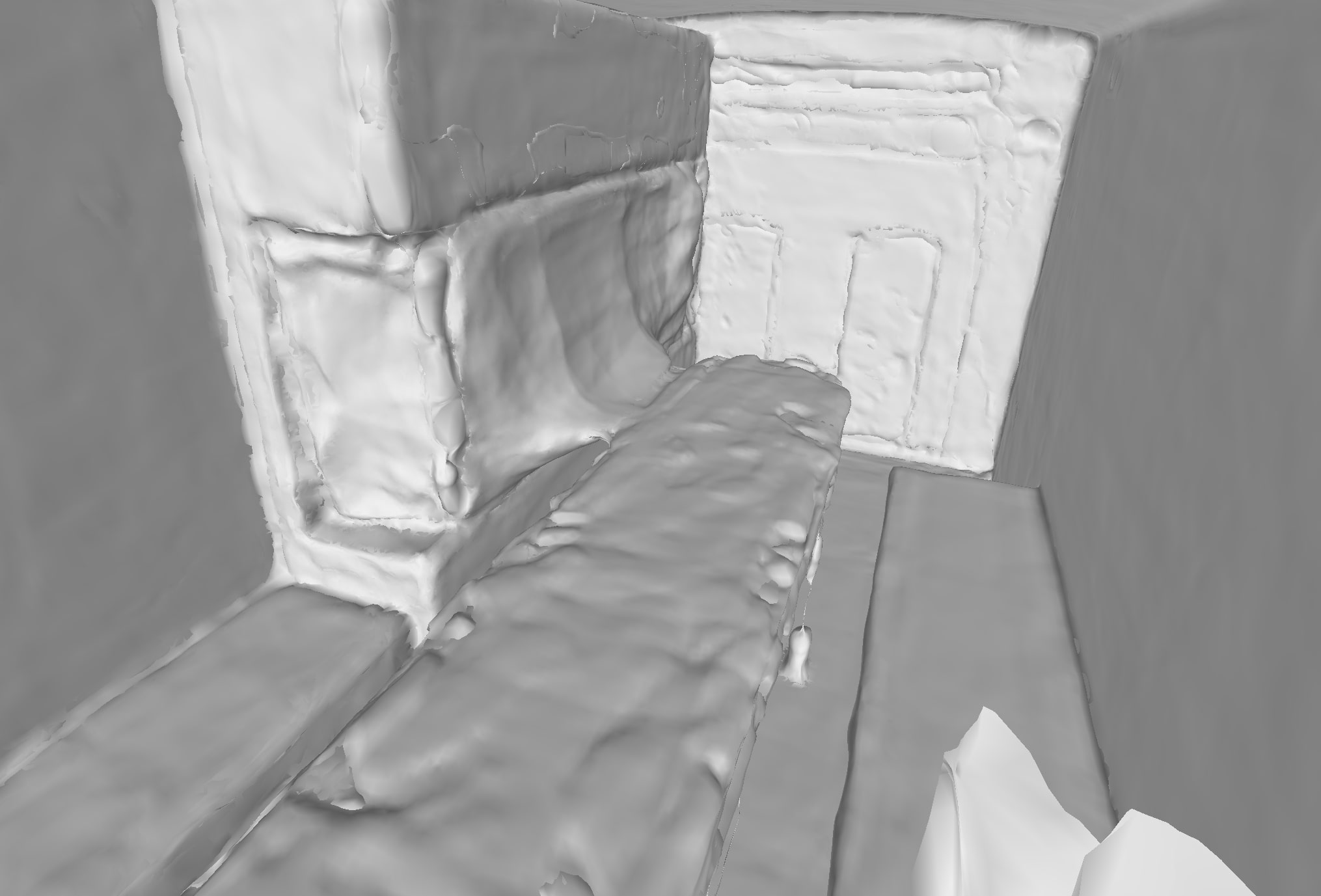} 
        \end{minipage}
        \begin{minipage}[b]{0.1\textwidth}
  		\includegraphics[width=1\textwidth]{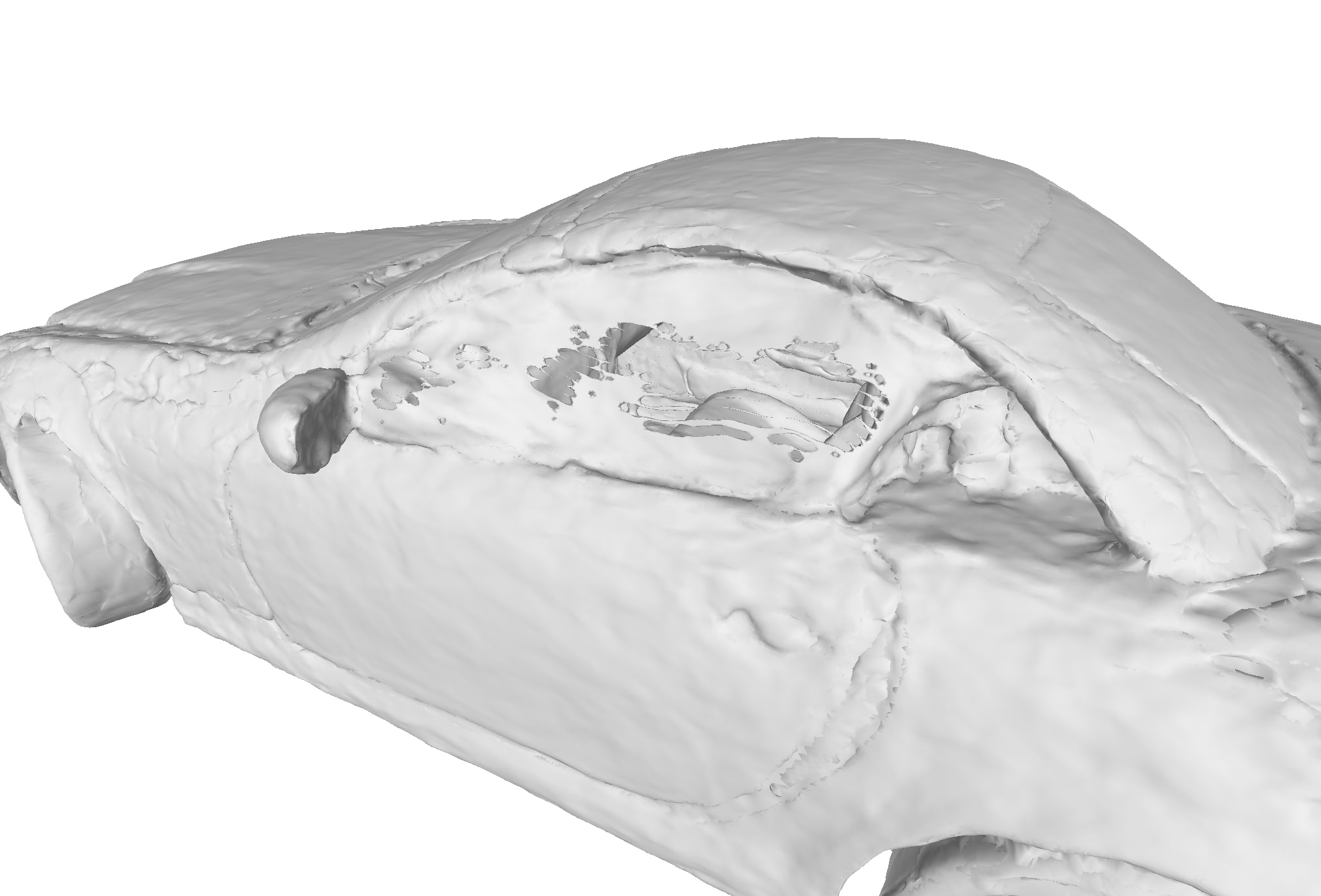}
        \end{minipage}
        \begin{minipage}[b]{0.1\textwidth}
            \includegraphics[width=1\textwidth]{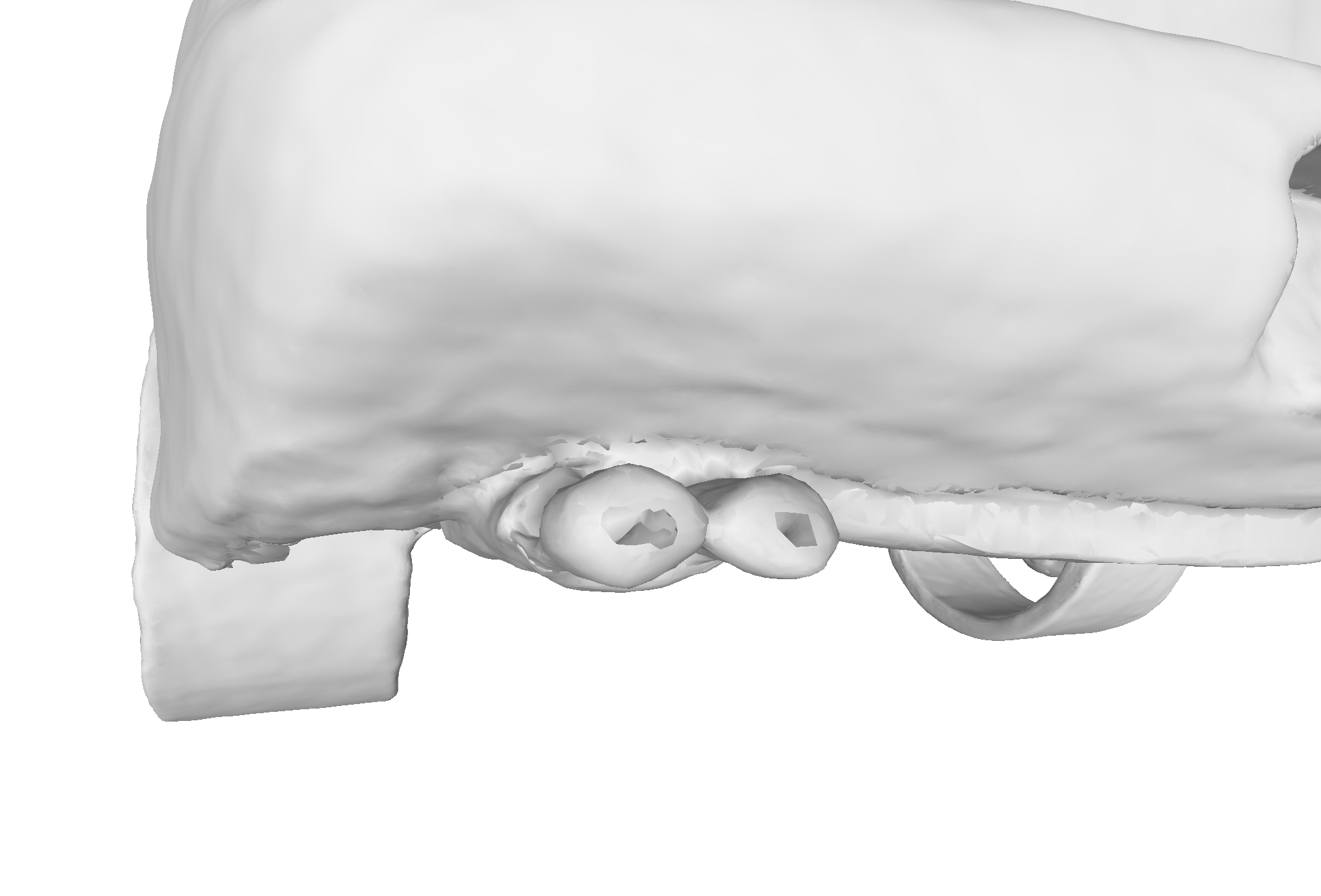} 
        \end{minipage}

    }
    \subfigure[LevelSetUDF]{
        \begin{minipage}[b]{0.1\textwidth}
		  \includegraphics[width=1\textwidth]{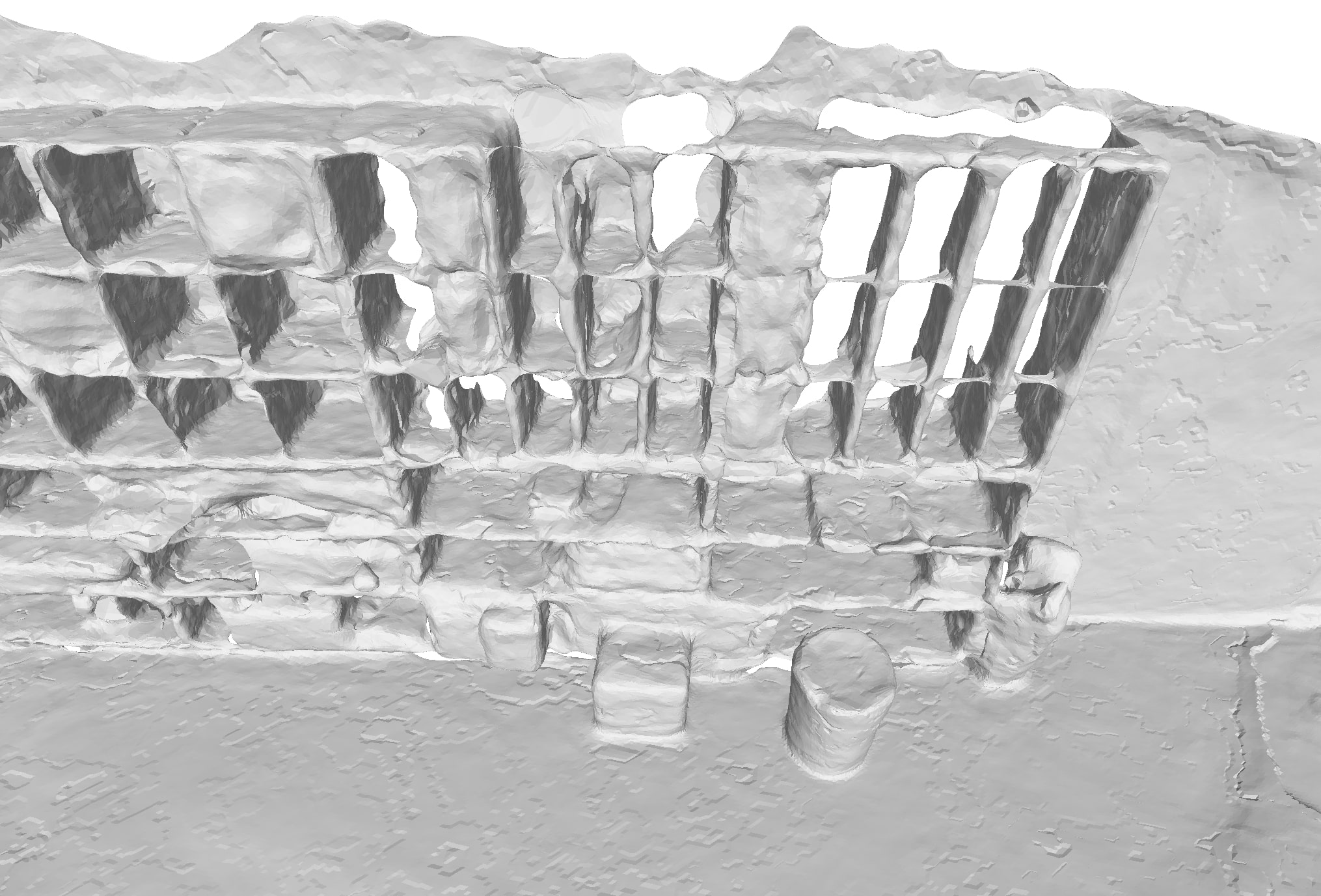} 
        \end{minipage}
        \begin{minipage}[b]{0.1\textwidth}
            \includegraphics[width=1\textwidth]{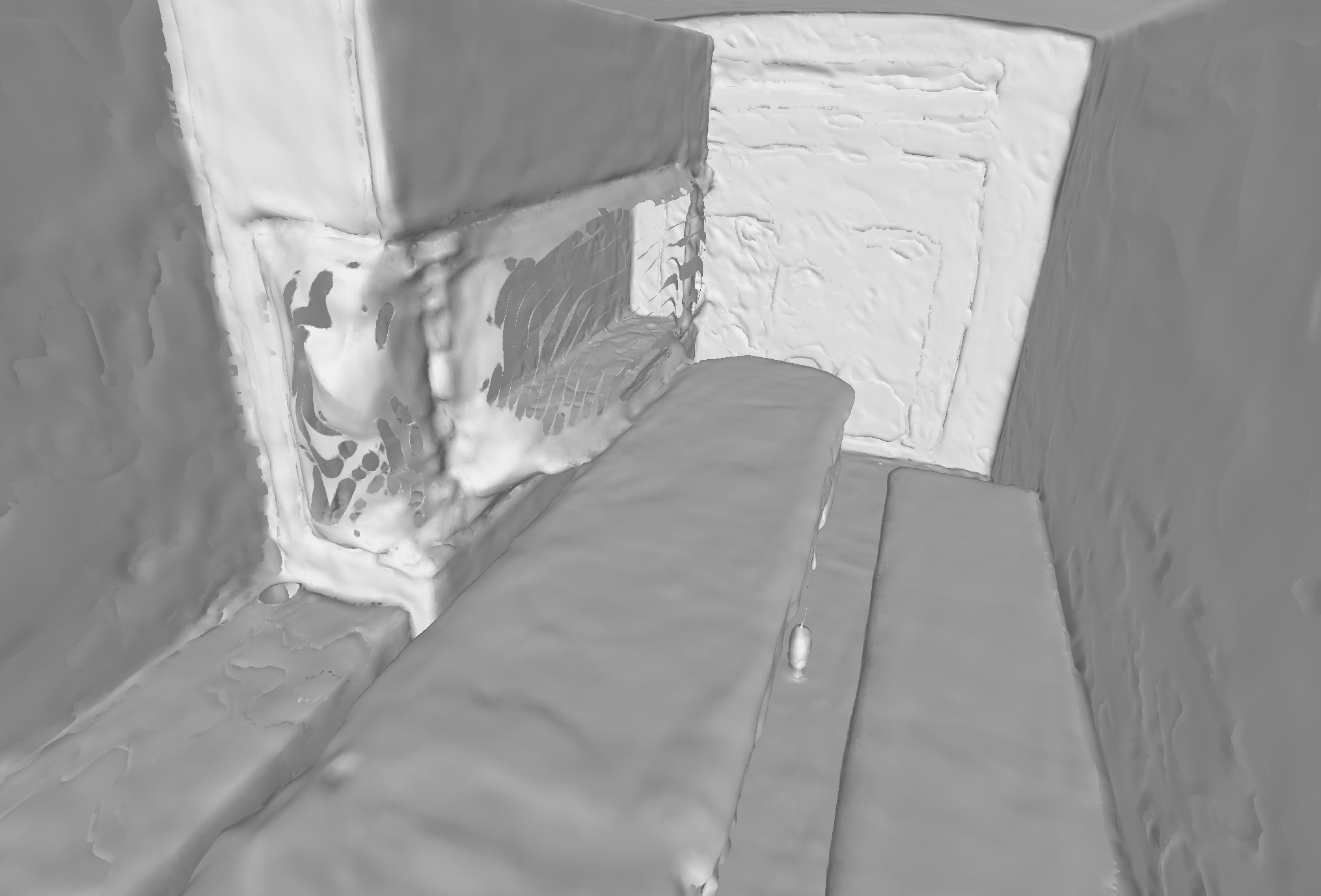} 
        \end{minipage}
        \begin{minipage}[b]{0.1\textwidth}
 	      \includegraphics[width=1\textwidth]{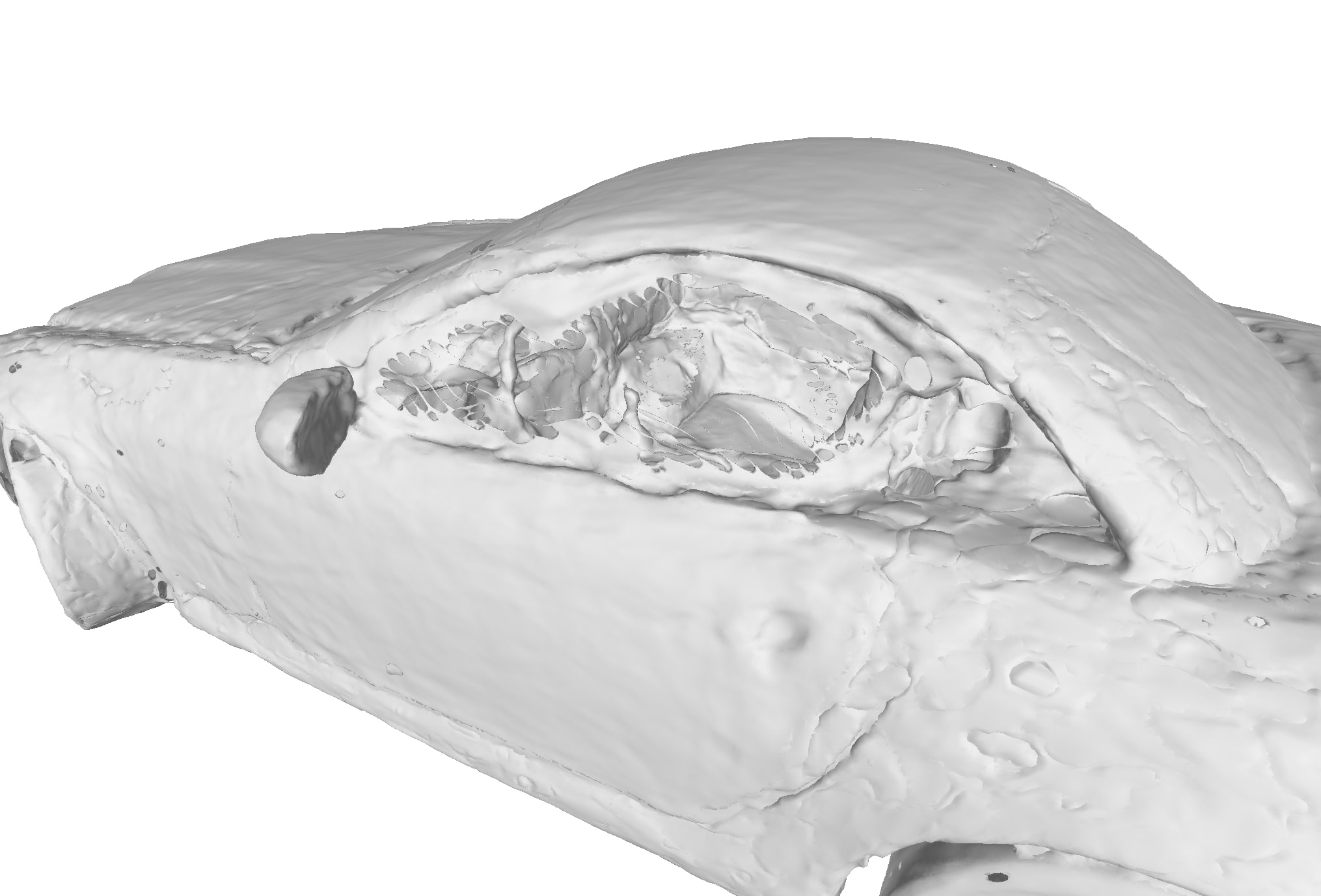}
        \end{minipage}
        \begin{minipage}[b]{0.1\textwidth}
            \includegraphics[width=1\textwidth]{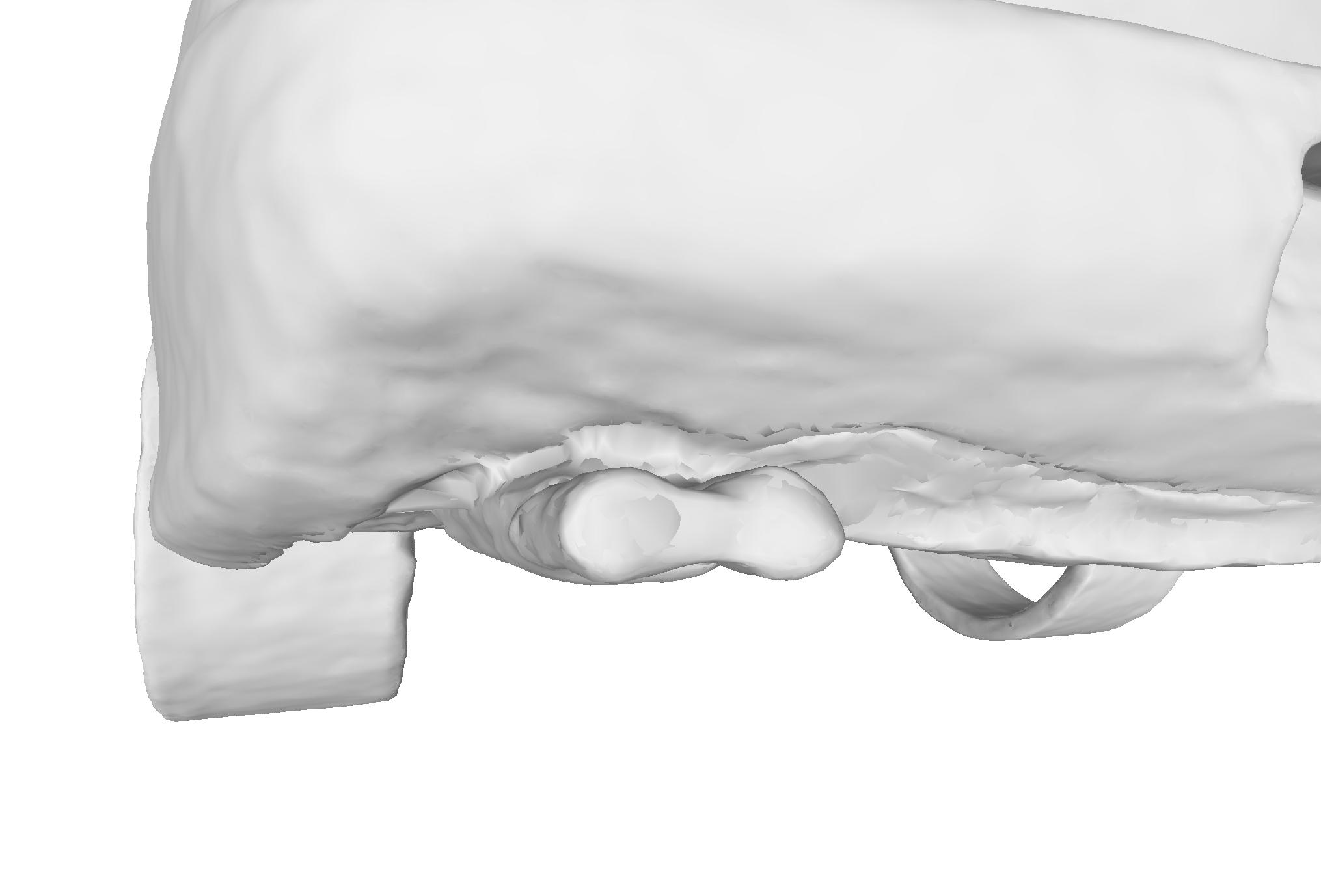} 
        \end{minipage}
    }
    \subfigure[Ours]{
        \begin{minipage}[b]{0.1\textwidth}
            \includegraphics[width=1\textwidth]{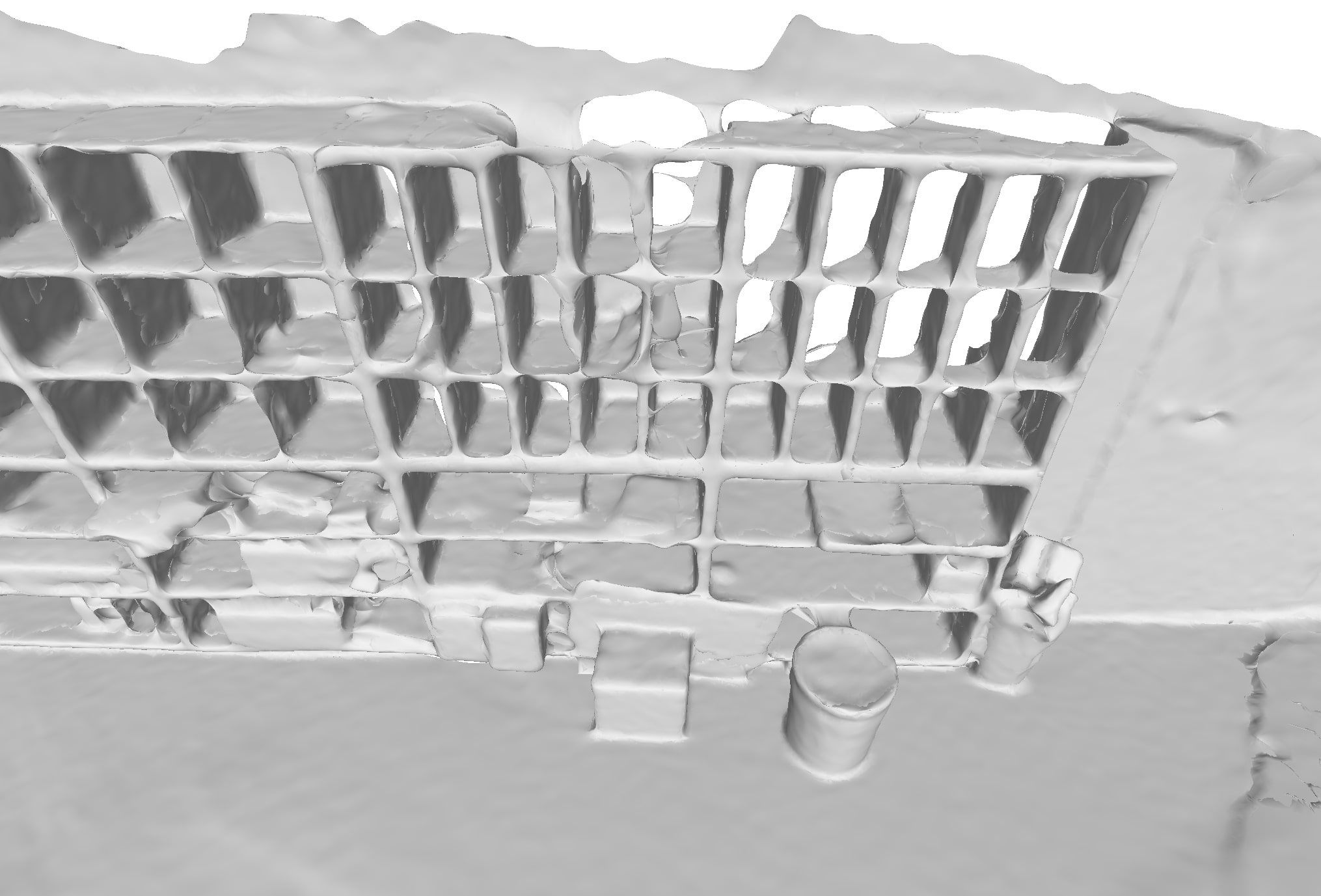}
        \end{minipage}
        \begin{minipage}[b]{0.1\textwidth}
            \includegraphics[width=1\textwidth]{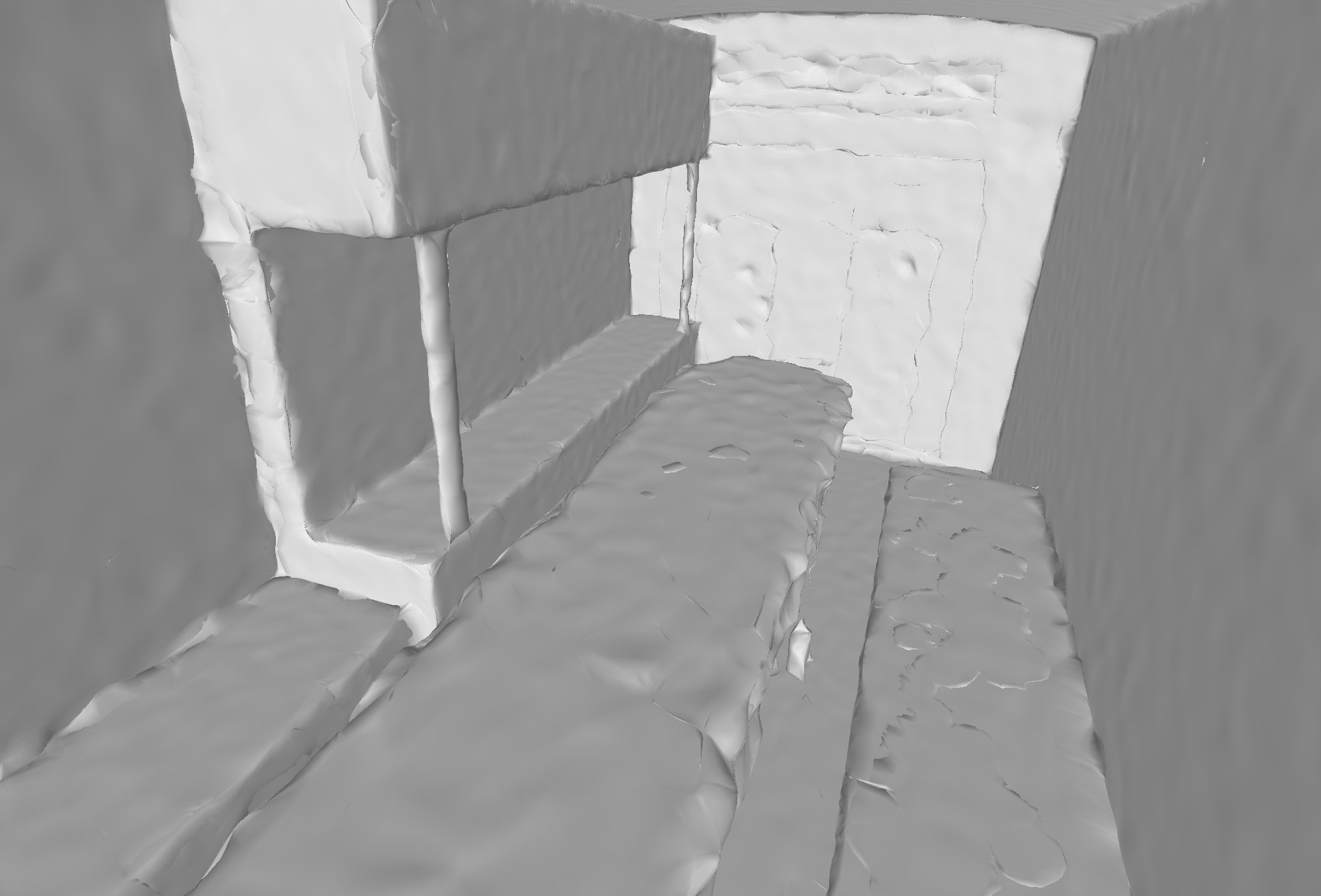} 
        \end{minipage}
        \begin{minipage}[b]{0.1\textwidth}
            \includegraphics[width=1\textwidth]{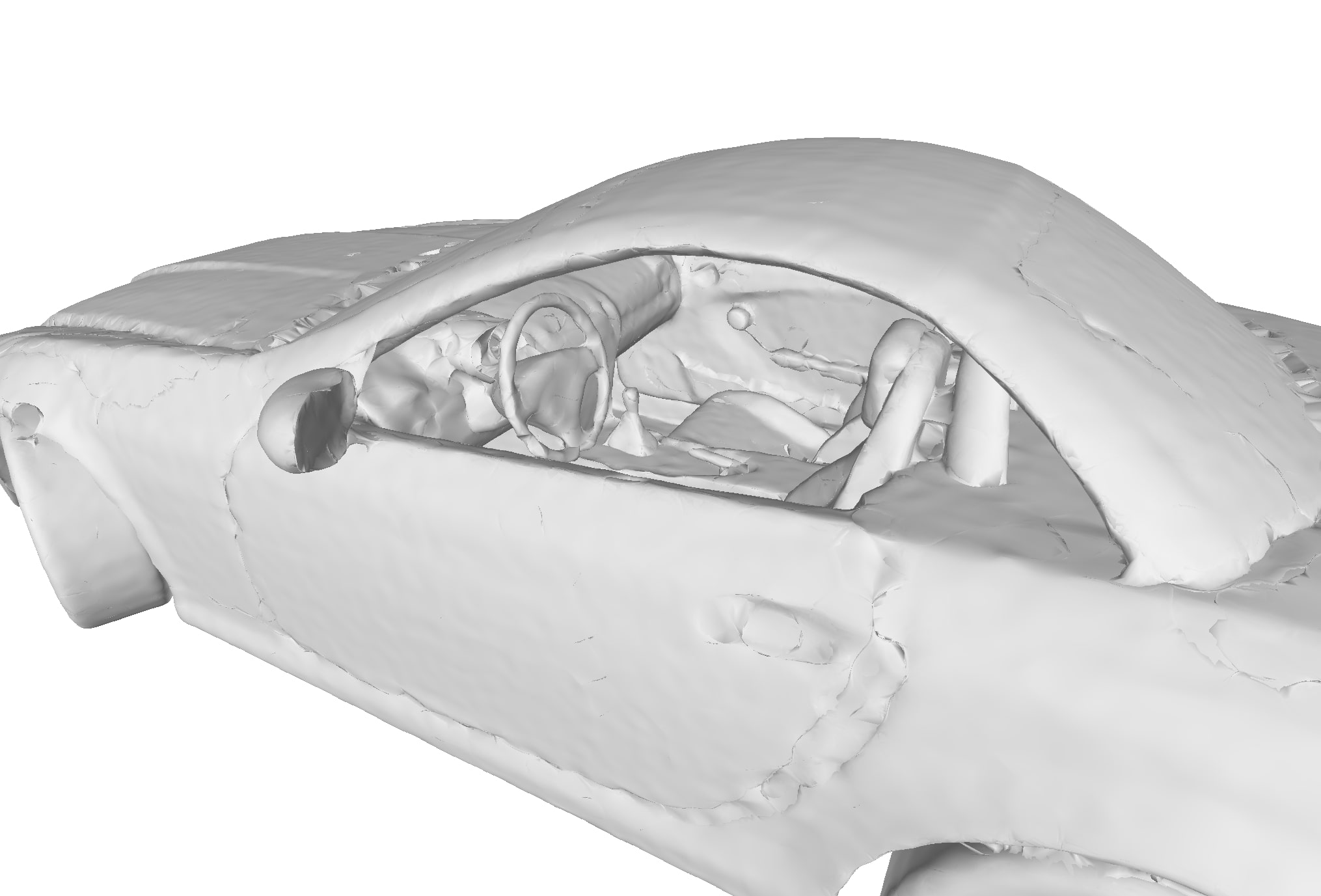}
        \end{minipage}
        \begin{minipage}[b]{0.1\textwidth}
            \includegraphics[width=1\textwidth]{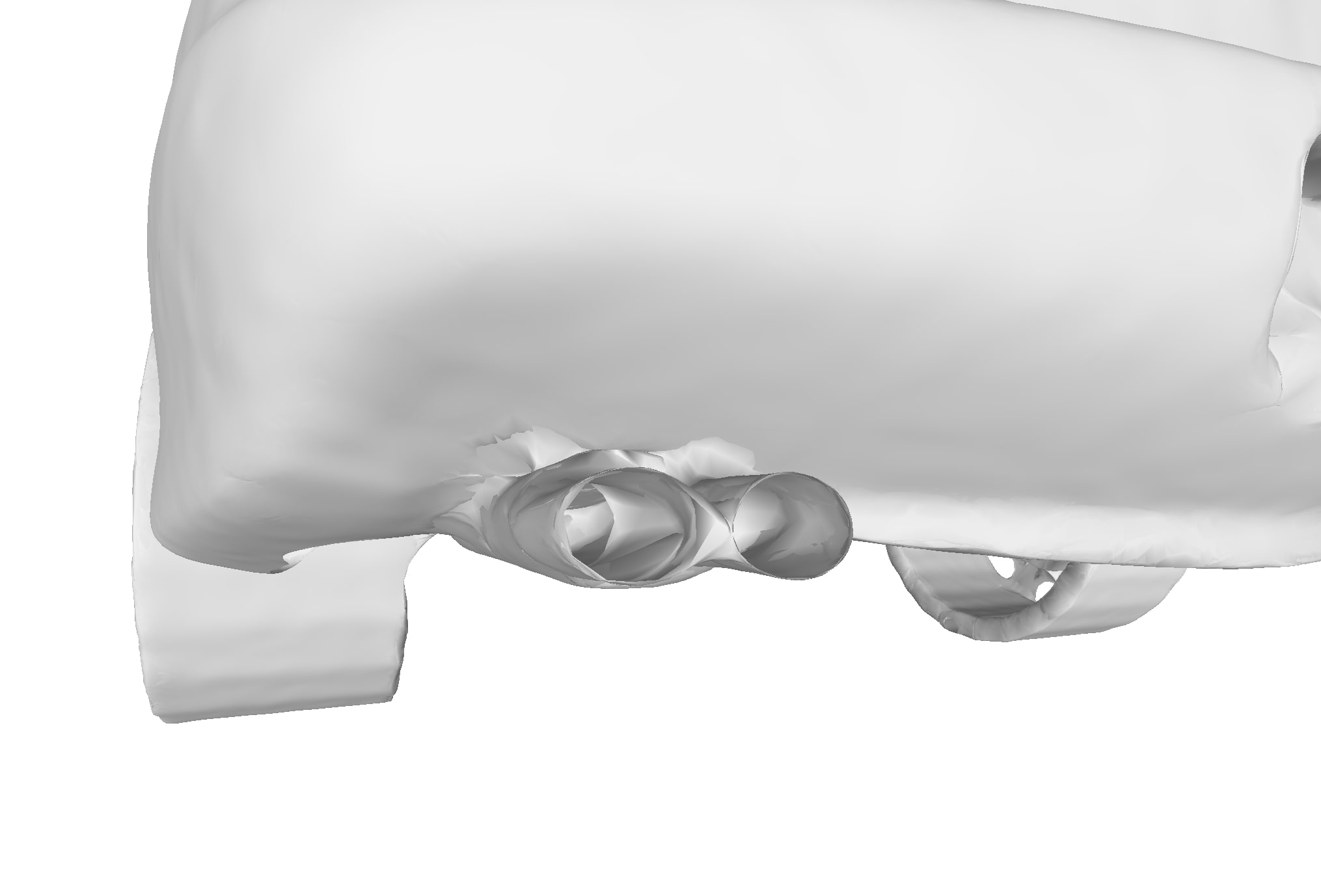} 
        \end{minipage}
    }
    \subfigure[GT]{
        \begin{minipage}[b]{0.1\textwidth}
  		\includegraphics[width=1\textwidth]{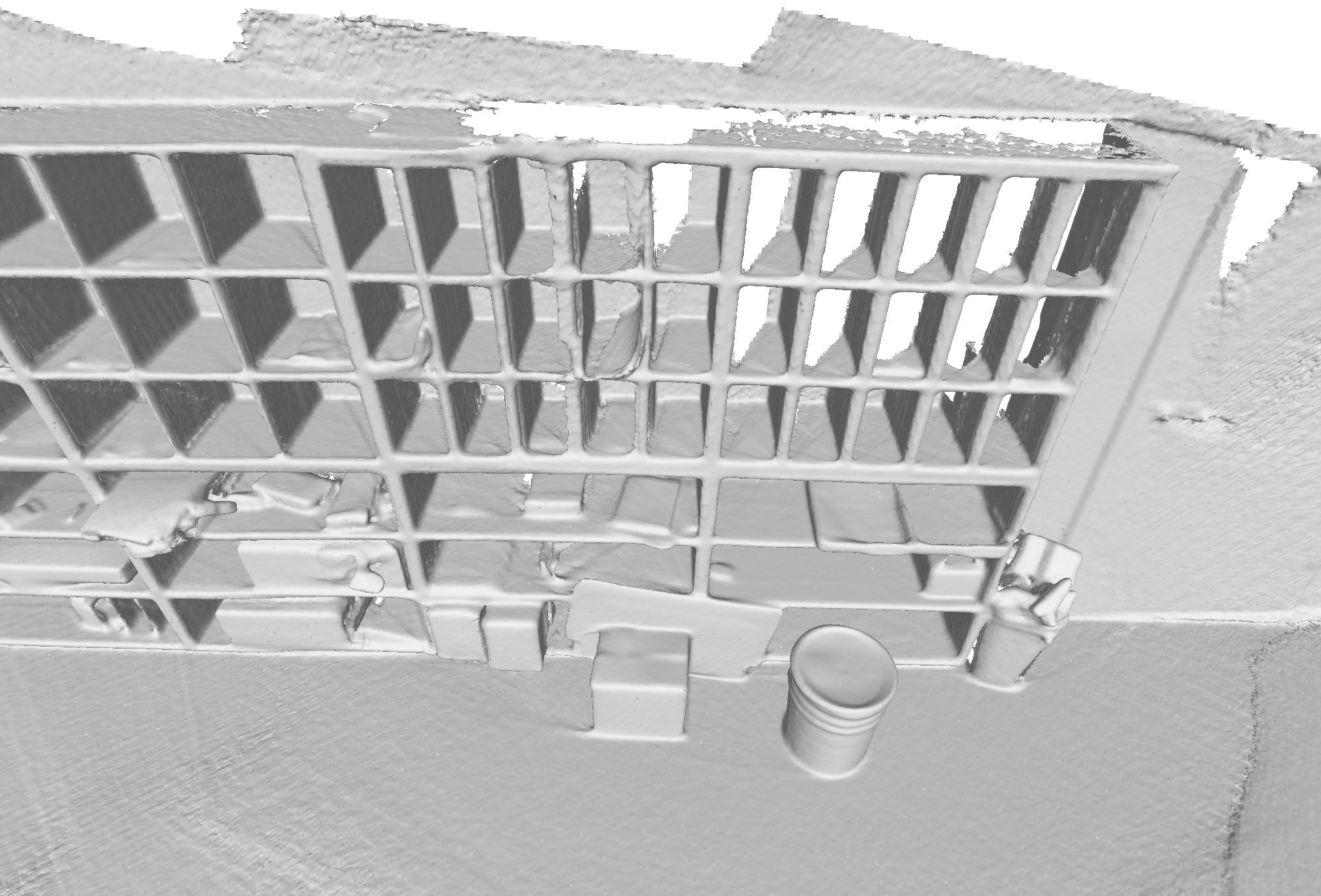}
        \end{minipage}
        \begin{minipage}[b]{0.1\textwidth}
            \includegraphics[width=1\textwidth]{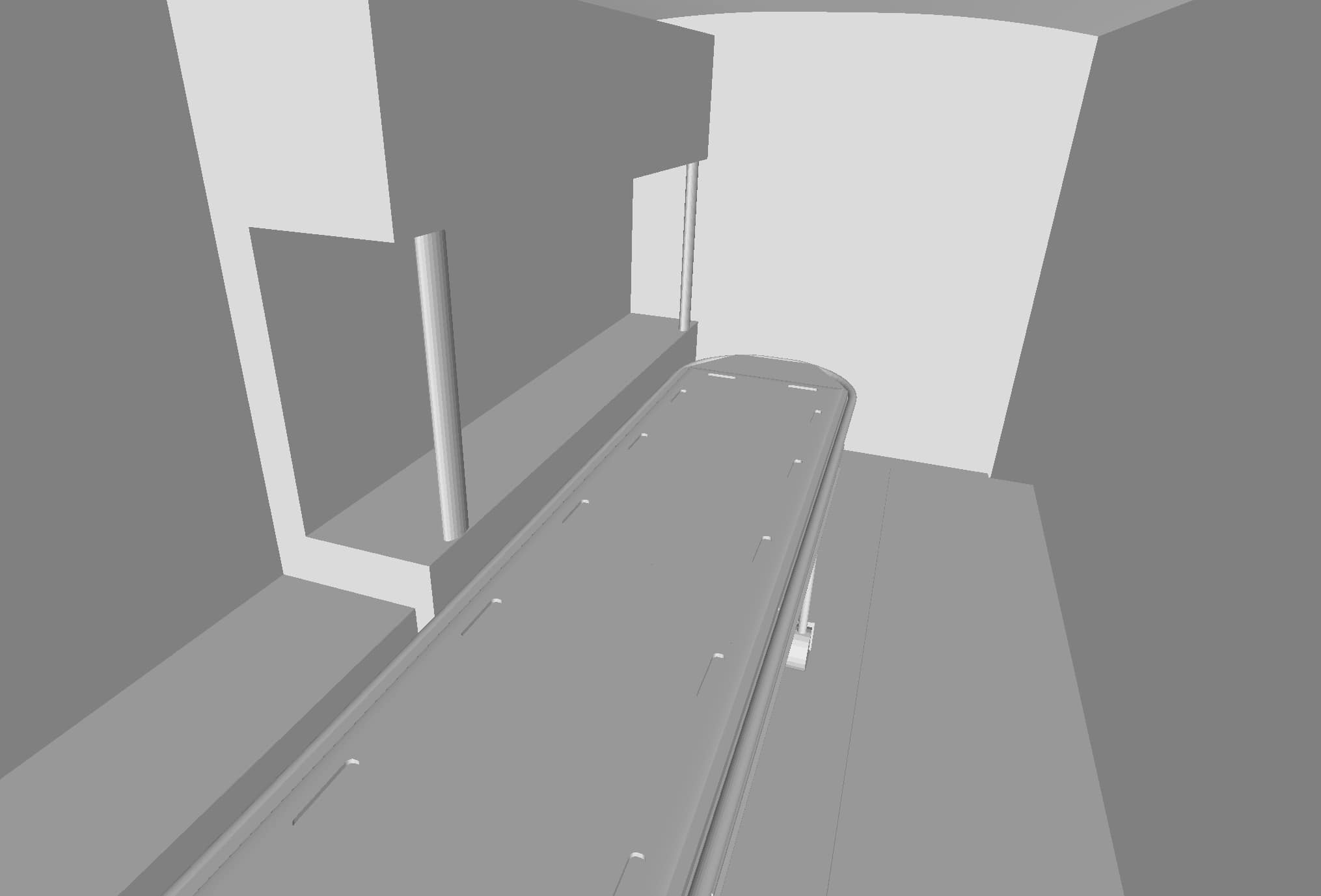} 
        \end{minipage}
        \begin{minipage}[b]{0.1\textwidth}
  		\includegraphics[width=1\textwidth]{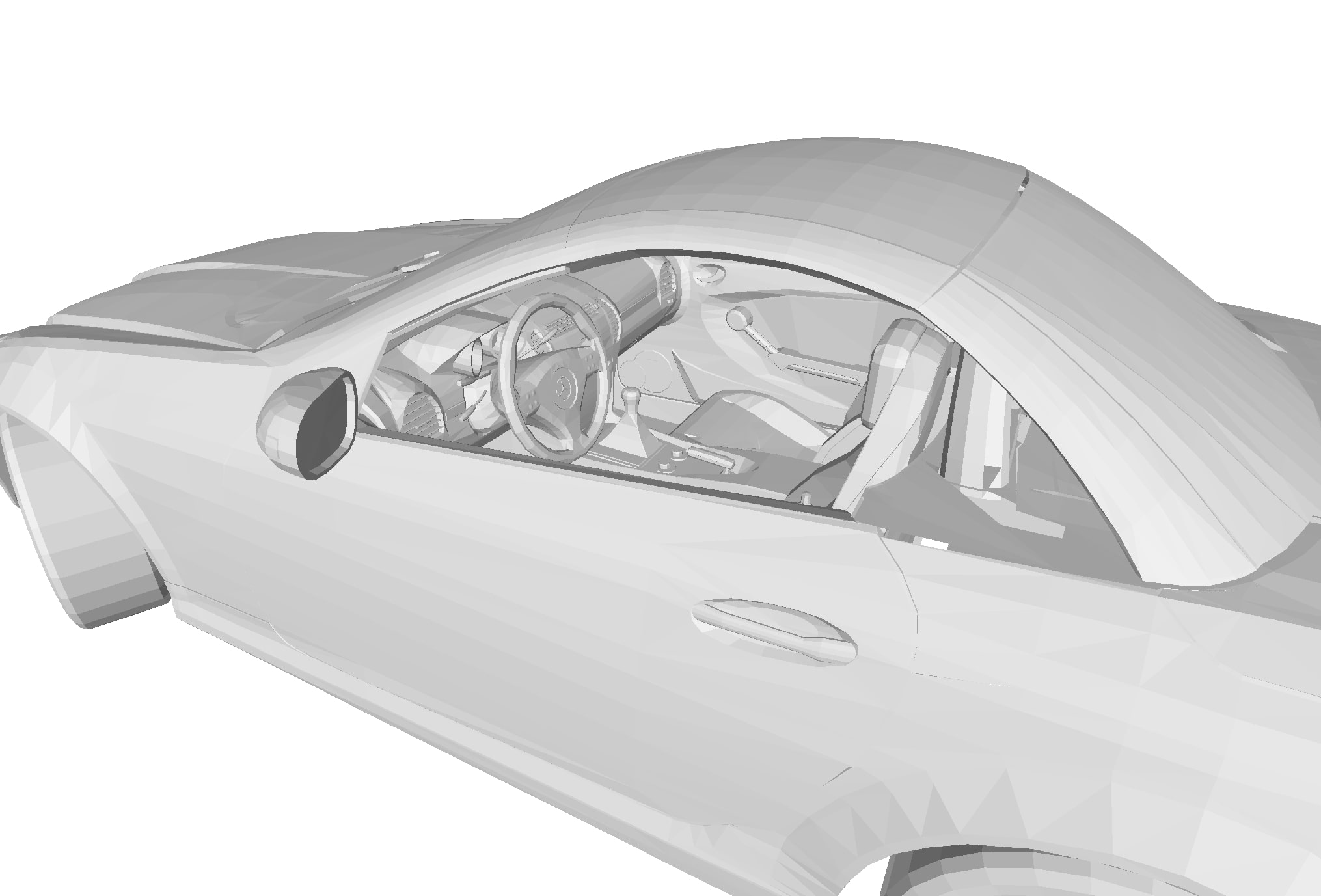}
        \end{minipage}
        \begin{minipage}[b]{0.1\textwidth}
            \includegraphics[width=1\textwidth]{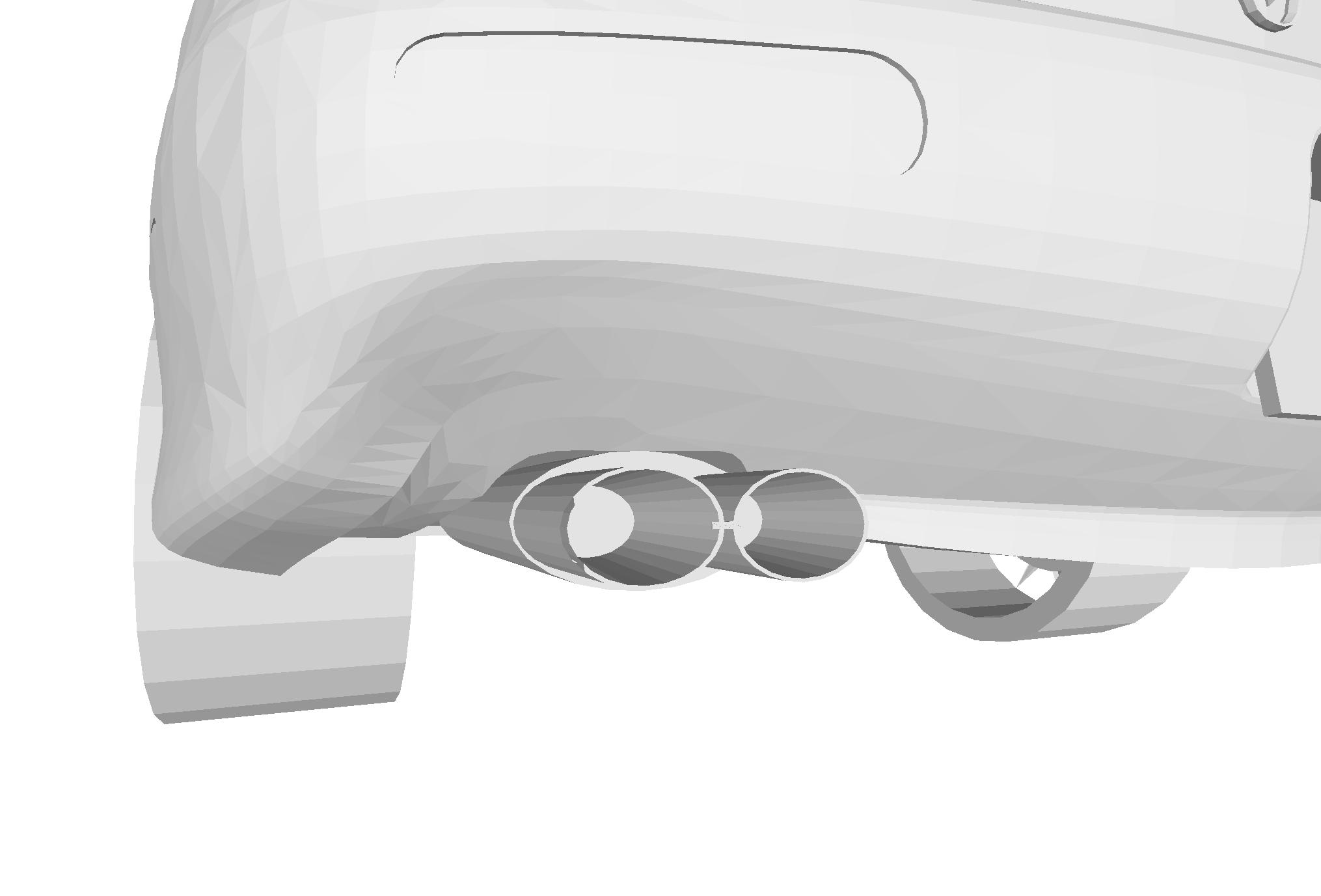} 
        \end{minipage}

    }
    \caption{Visual comparisons with DUDF, CAP-UDF and LevelSetUDF on an indoor scene of the Stanford 3D Scene dataset featuring noise, an imperfect scan, and two car models of the ShapeNet-Cars dataset showcasing complex structures. Our method remains the open boundaries, such as, the bookshelves, inner structures of the vehicle, car window and car exhaust vent.}
    \label{fig:open-surfcomparison}
\end{figure}

\paragraph{3D objects with complex inner structures}
We further explore our method for representing 3D objects and scenes with complex inner structures. We evaluate our method on Stanford 3D Scene and ShapeNet-Cars datasets. As shown in Table~\ref{tab:surf-comp} and Figure~\ref{fig:open-surfcomparison}, our method is more stable on complex structures and performs optimally for keeping open boundaries.

\begin{table}[!htbp]
    \centering
    \renewcommand\arraystretch{0.7}
    \setlength{\tabcolsep}{0pt}
    \scriptsize
    \begin{tabular}{ccccccccc}
        \toprule
          &  \multicolumn{4}{c}{Gaussian Noise-N(0,0.0025)}& \multicolumn{4}{c}{Gaussian Noise-N(0,0.005)}\\
         \cmidrule(r){1-9}
         &  \multicolumn{2}{c}{Chamfer-L1 $(\downarrow)$} &\multicolumn{2}{c}{F-score $(\uparrow)$} & \multicolumn{2}{c}{Chamfer-L1 $(\downarrow)$} &\multicolumn{2}{c}{F-score $(\uparrow)$}\\
        \cmidrule(r){2-3}
        \cmidrule(r){4-5}
        \cmidrule(r){6-7}
        \cmidrule(r){8-9}        
         Method  &  Mean & Median & $F1^{0.005}$ & $F1^{0.0025}$ &  Mean & Median & $F1^{0.005}$ & $F1^{0.0025}$ \\
        \midrule
         CAP-UDF      & 5.76 & 5.49 & 92.68 & 43.39 & 6.26 & 5.99 & 87.82 & 37.54 \\
         LeverSetUDF  & 5.63 & 5.37 & 93.39 & 45.60 & 6.42 & 6.18 & 86.45 & 36.05 \\
         DUDF         & 4.27 & 3.83 & 97.91 & 70.57 & \textbf{4.61} & \textbf{4.11} & \textbf{96.25} & \textbf{65.04} \\
        \midrule
         Ours         & \textbf{4.07} & \textbf{3.79} & \textbf{98.74} & \textbf{71.89} & 5.64 & 5.35 & 92.74 & 46.18 \\
        \bottomrule
    \end{tabular}
    \caption{
    Quantitative results for reconstruction on noisy point clouds from Stanford 3D Scenes and 3D Scans. We add Gaussian noise $N(0, 0.0025)$ and $N(0, 0.005)$ to the input point clouds and use the SIREN with a frequency of 30 to resist noises.
    }   
    \label{tab:noise-comp}
\end{table}

\paragraph{Noisy point clouds}
As illustrated in Table~\ref{tab:noise-comp}, we add $N(0, 0.0025)$ and $N(0, 0.005)$ Gaussian noise to the normalized point clouds. To improve the noise resistance, we reduce the SIREN frequency to 30. Our method is best at resisting small noises. DUDF is the best for large noise because it always oversmoothes the surface. However, for the same reason, DUDF is the worst at reconstructing details. Apart from DUDF, our DEUDF is better than others for large noisy data. Meanwhile, DEUDF is the best at reconstructing details.

\paragraph{Deviation from zero}
It is important to note that the distances of the learned UDF of a surface are unlikely to be exactly zero due to inherent learning errors. Interestingly, even though we relax the non-negativity constraint in our DEUDF, the deviation from zero in our results is actually smaller than that of UDFs learned by other methods enforcing strict non-negativity. We evaluate the average absolute distance values of the reconstructed mesh points in the learned UDF to evaluate the deviation. We test on the Stanford 3D Scene and 3D Scan Repository datasets. The average deviations are as follows: DUDF: $2.60 \times 10^{-3}$, CAP-UDF: $2.63 \times 10^{-4}$, LevelSetUDF: $1.39 \times 10^{-4}$, and our DEUDF: $1.34 \times 10^{-4}$.

\subsection{Ablation studies}
We conduct ablation studies to demonstrate the effectiveness of each component within our method.

\begin{figure}[!htbp]
    \centering
    \subfigure[]{
        \begin{minipage}[b]{0.065\textwidth}
		  \includegraphics[width=1\textwidth]{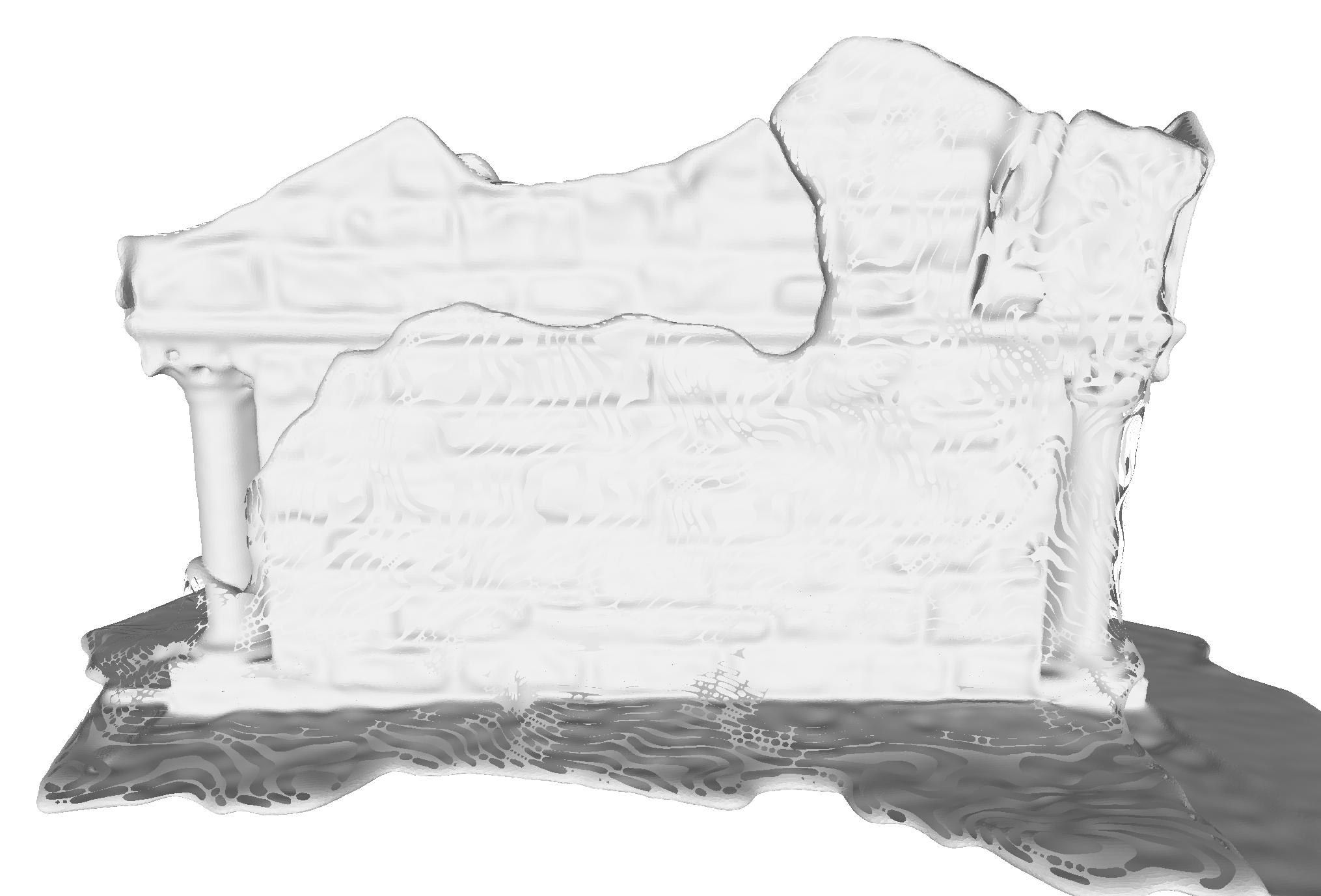} \\
  		\includegraphics[width=1\textwidth]{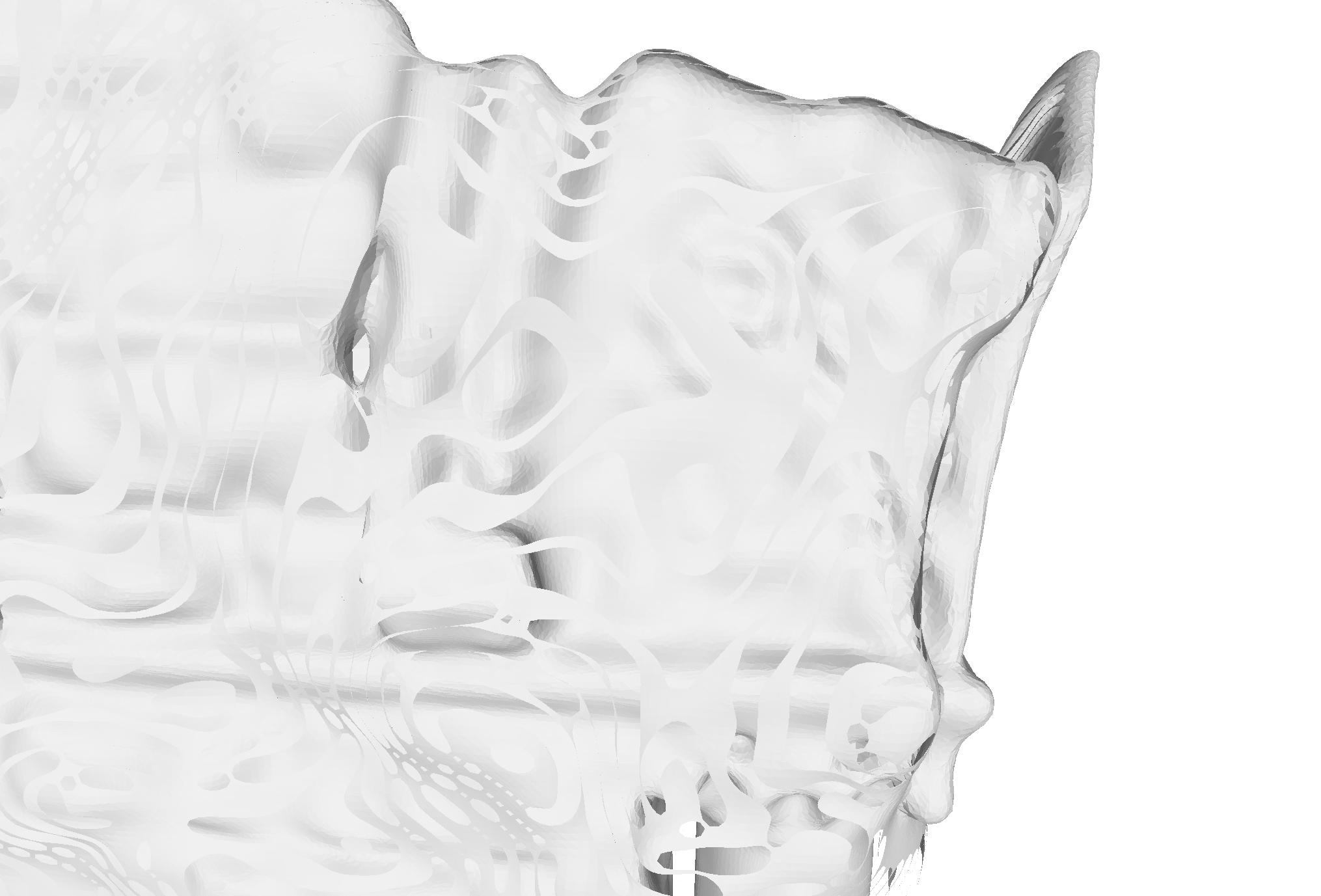}
        \end{minipage}
    }
    \subfigure[]{
        \begin{minipage}[b]{0.065\textwidth}
		  \includegraphics[width=1\textwidth]{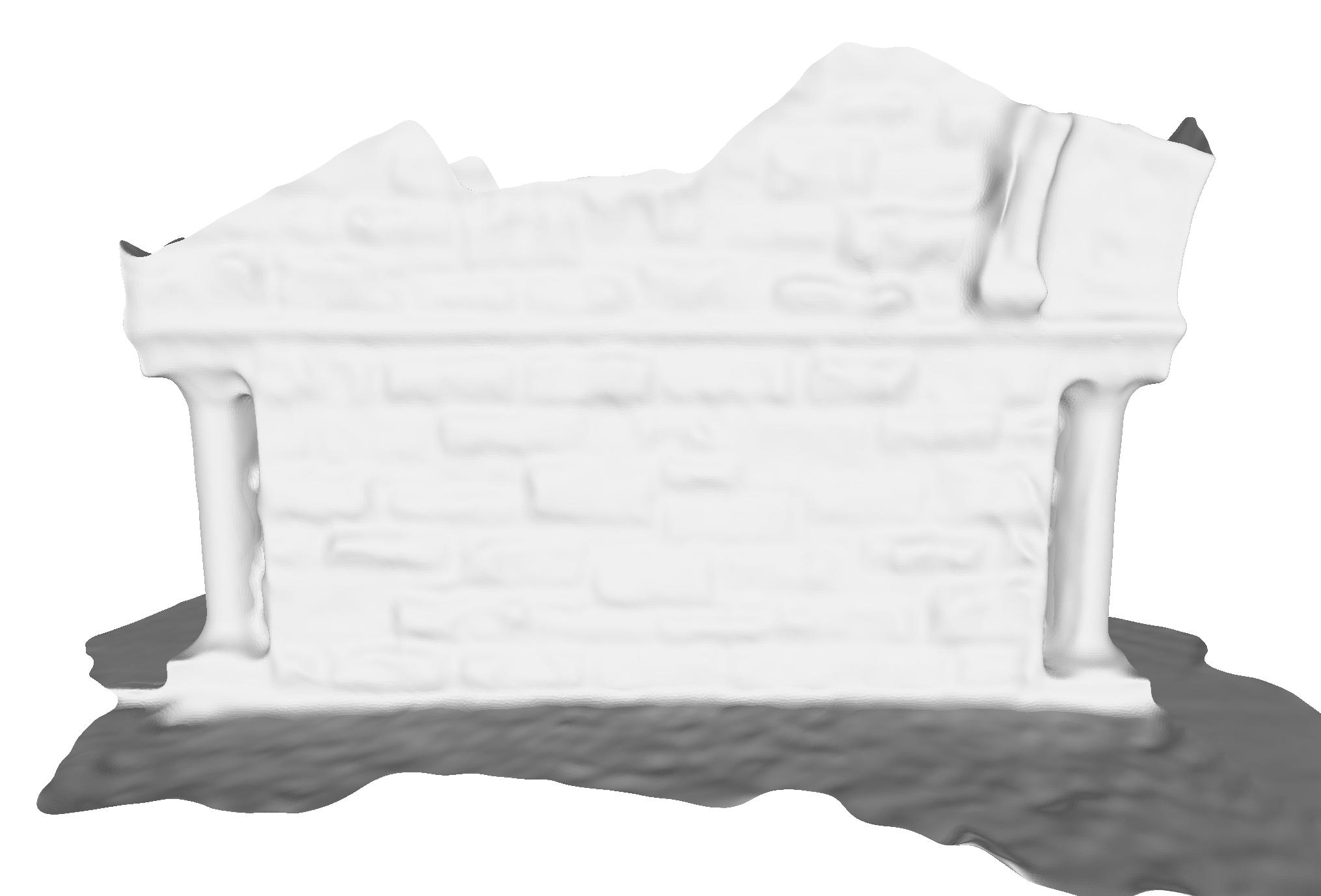}\\
		  \includegraphics[width=1\textwidth]{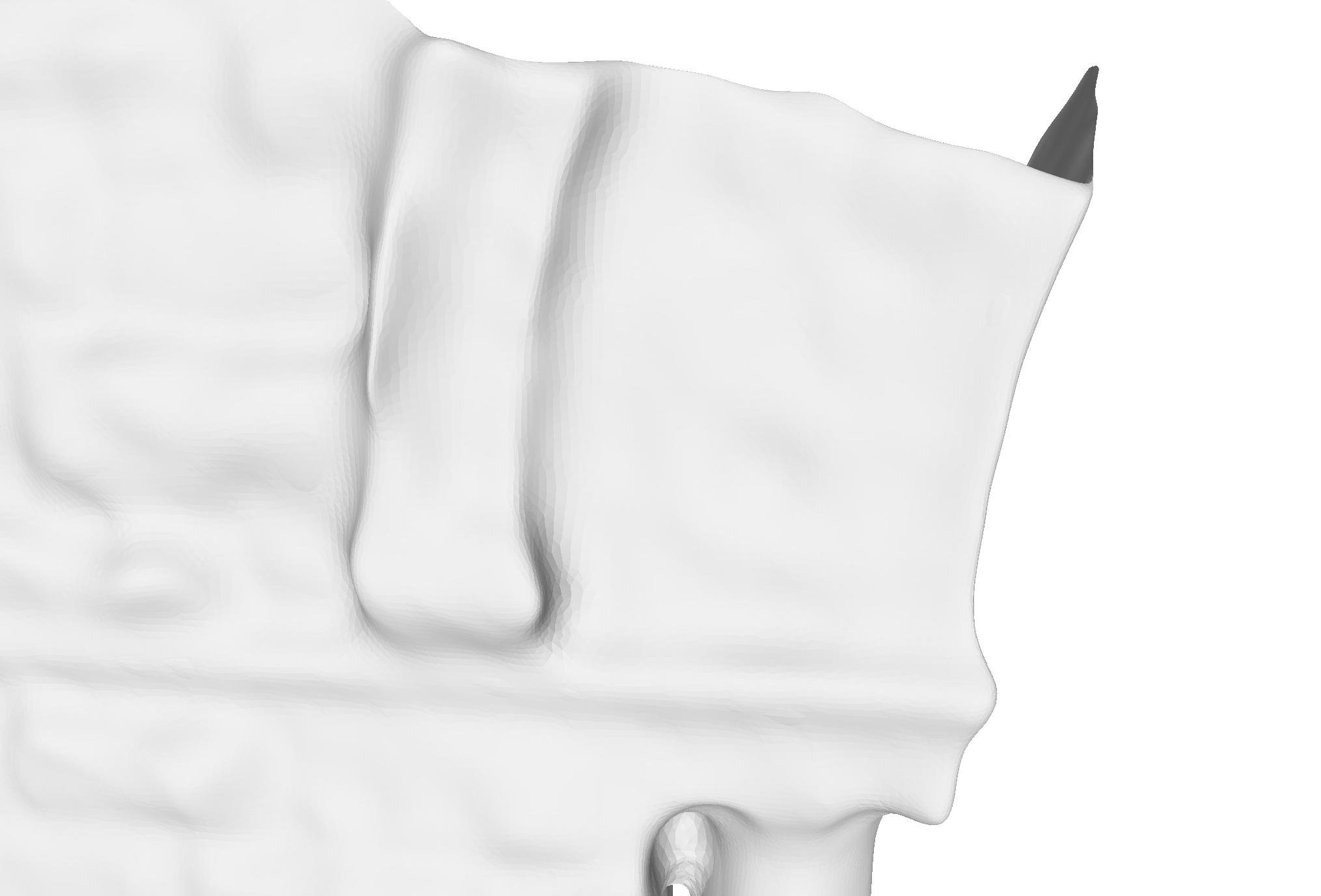}
        \end{minipage}
    }
    \subfigure[]
    {
        \begin{minipage}[b]{0.065\textwidth}
		  \includegraphics[width=1\textwidth]{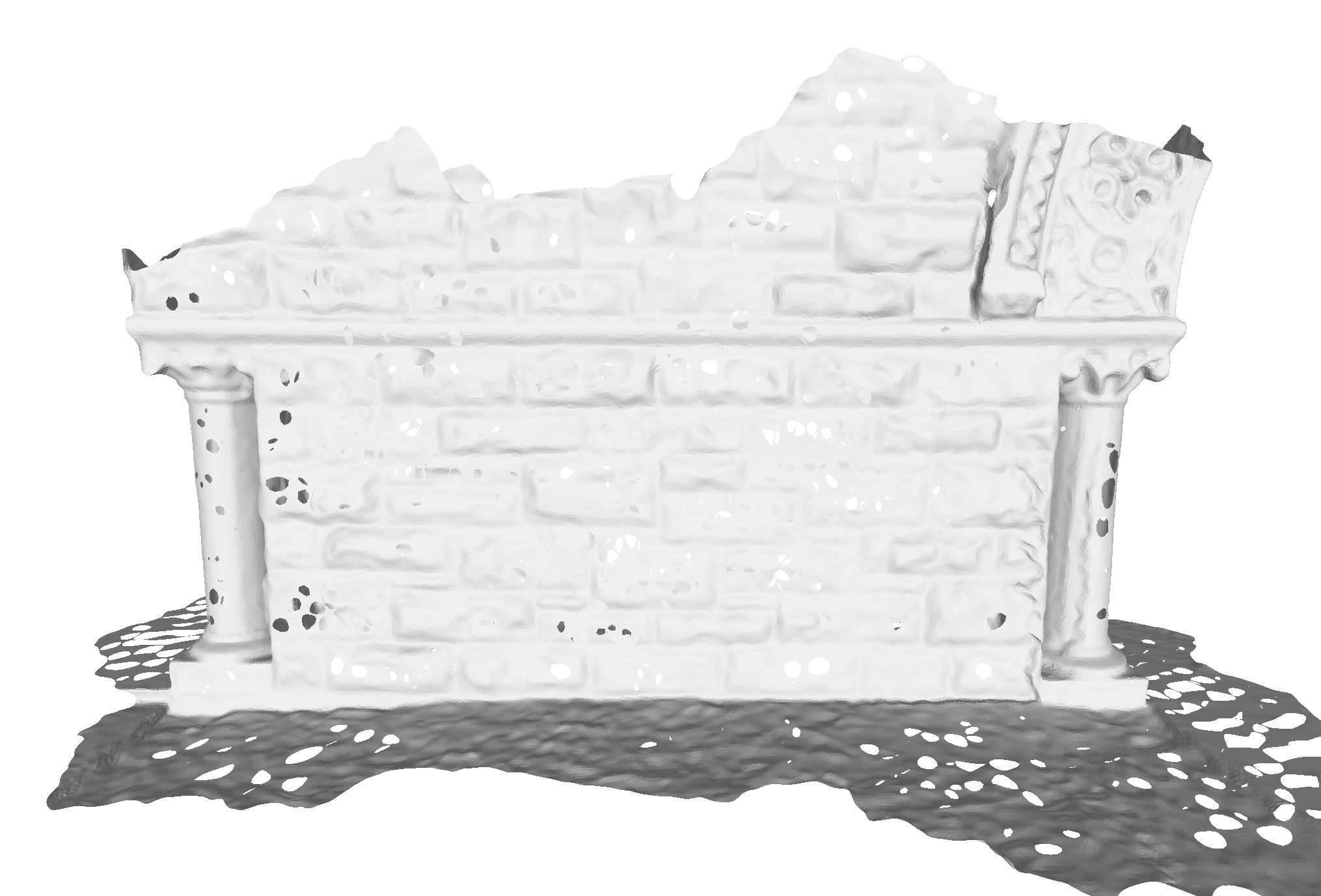}\\
		  \includegraphics[width=1\textwidth]{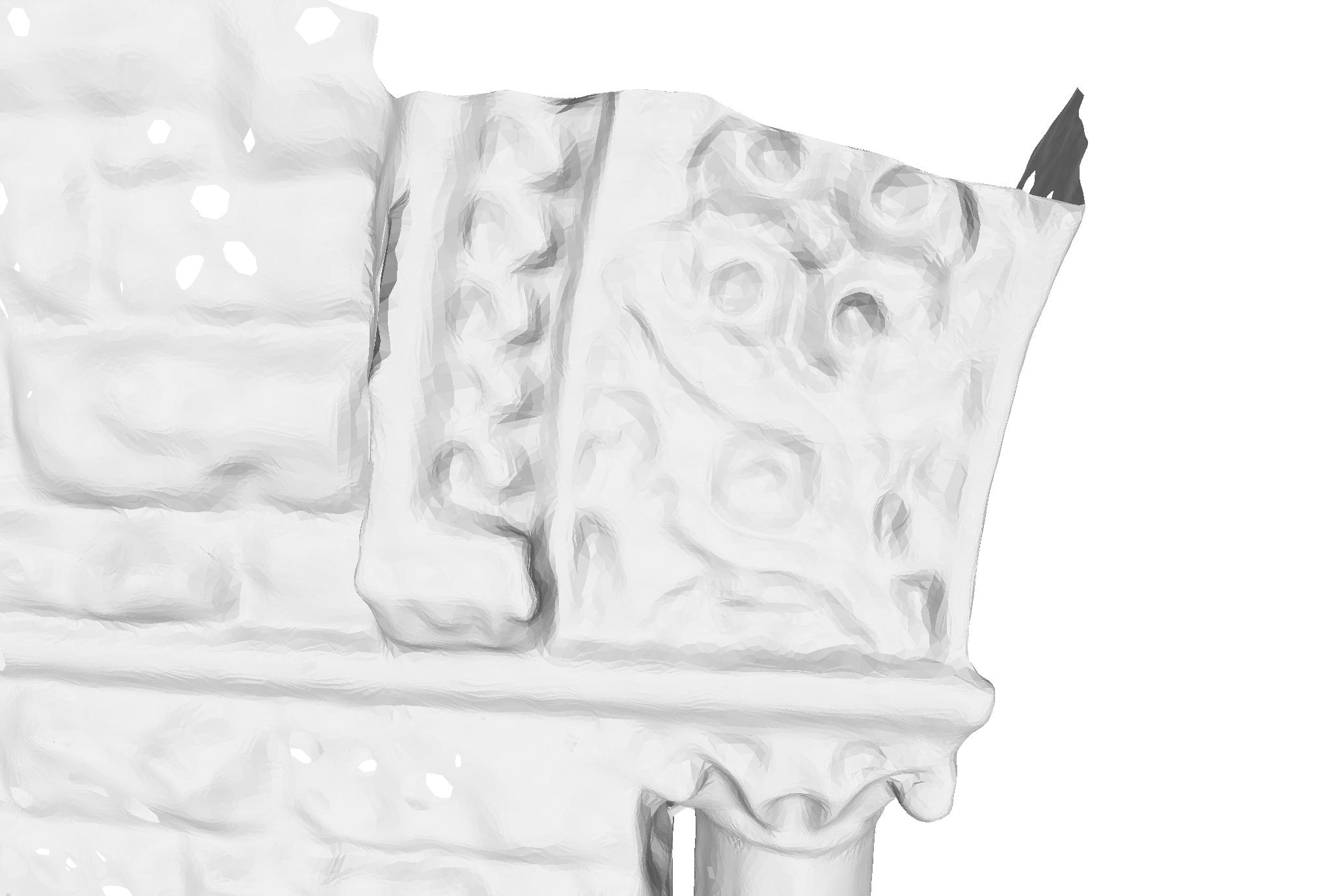}
        \end{minipage}
    }
    \subfigure[]
    {
        \begin{minipage}[b]{0.065\textwidth}
		  \includegraphics[width=1\textwidth]{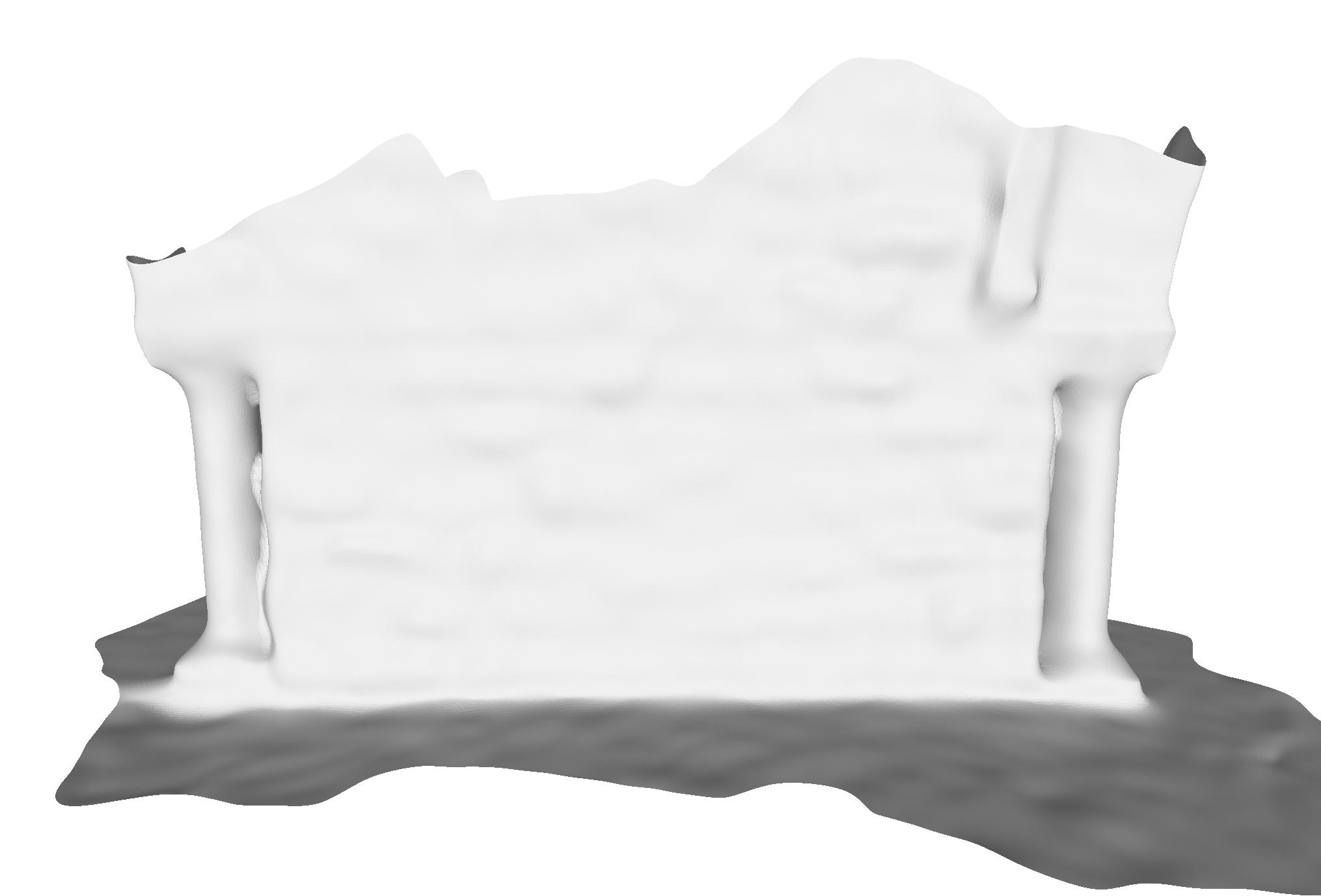}\\
		  \includegraphics[width=1\textwidth]{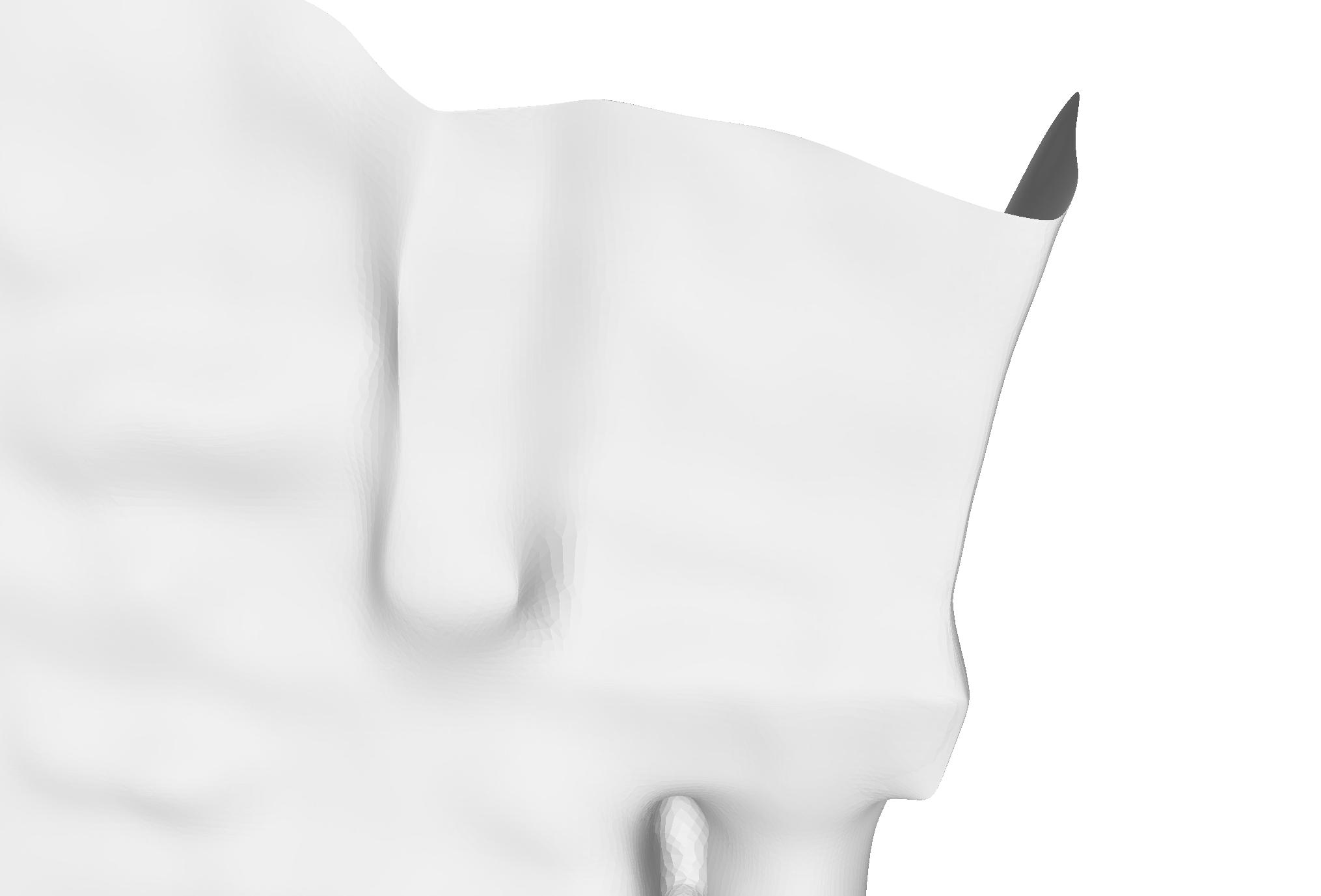}
        \end{minipage}
    }
    \subfigure[]{
        \begin{minipage}[b]{0.065\textwidth}

            \includegraphics[width=1\textwidth]{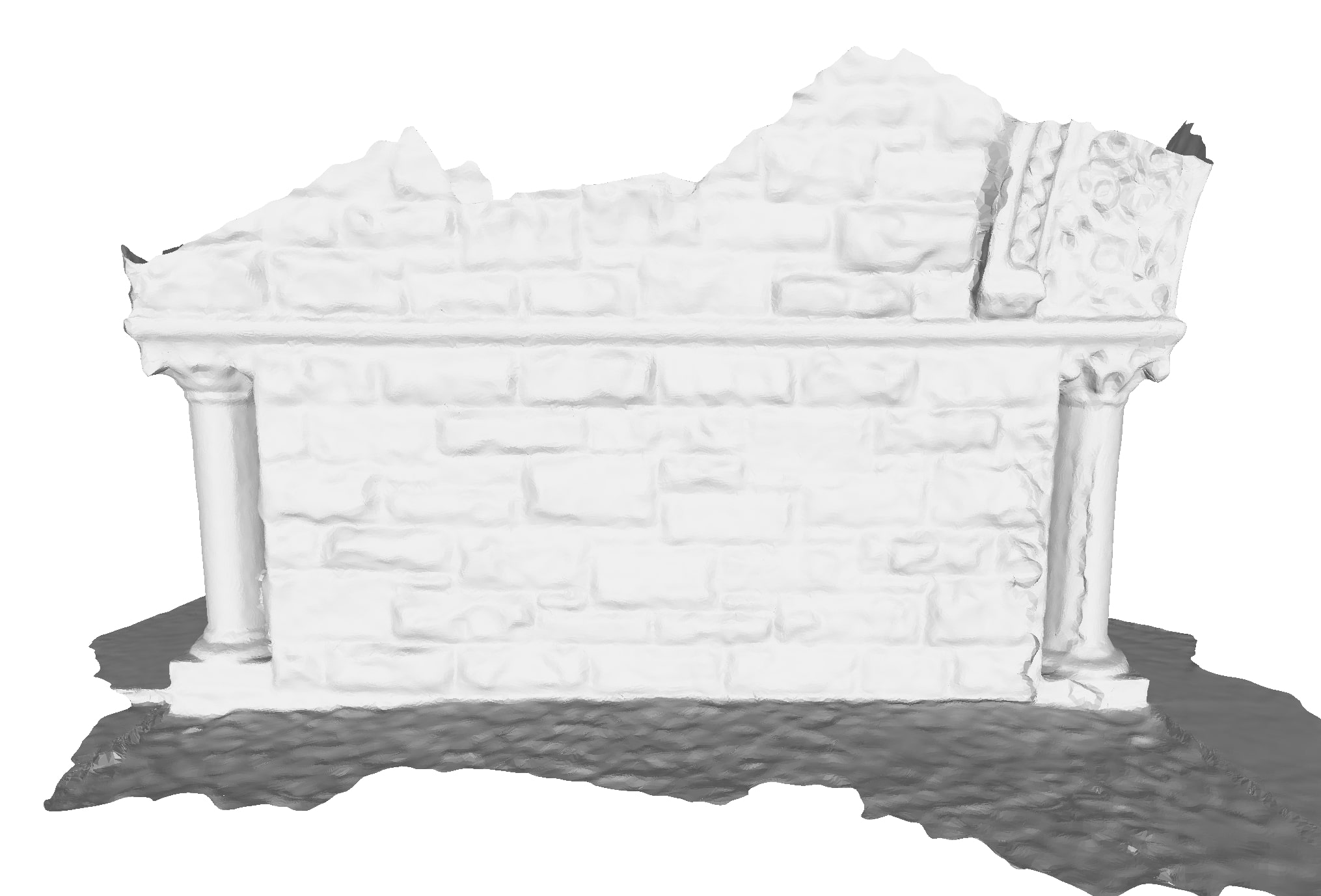} \\
            \includegraphics[width=1\textwidth]{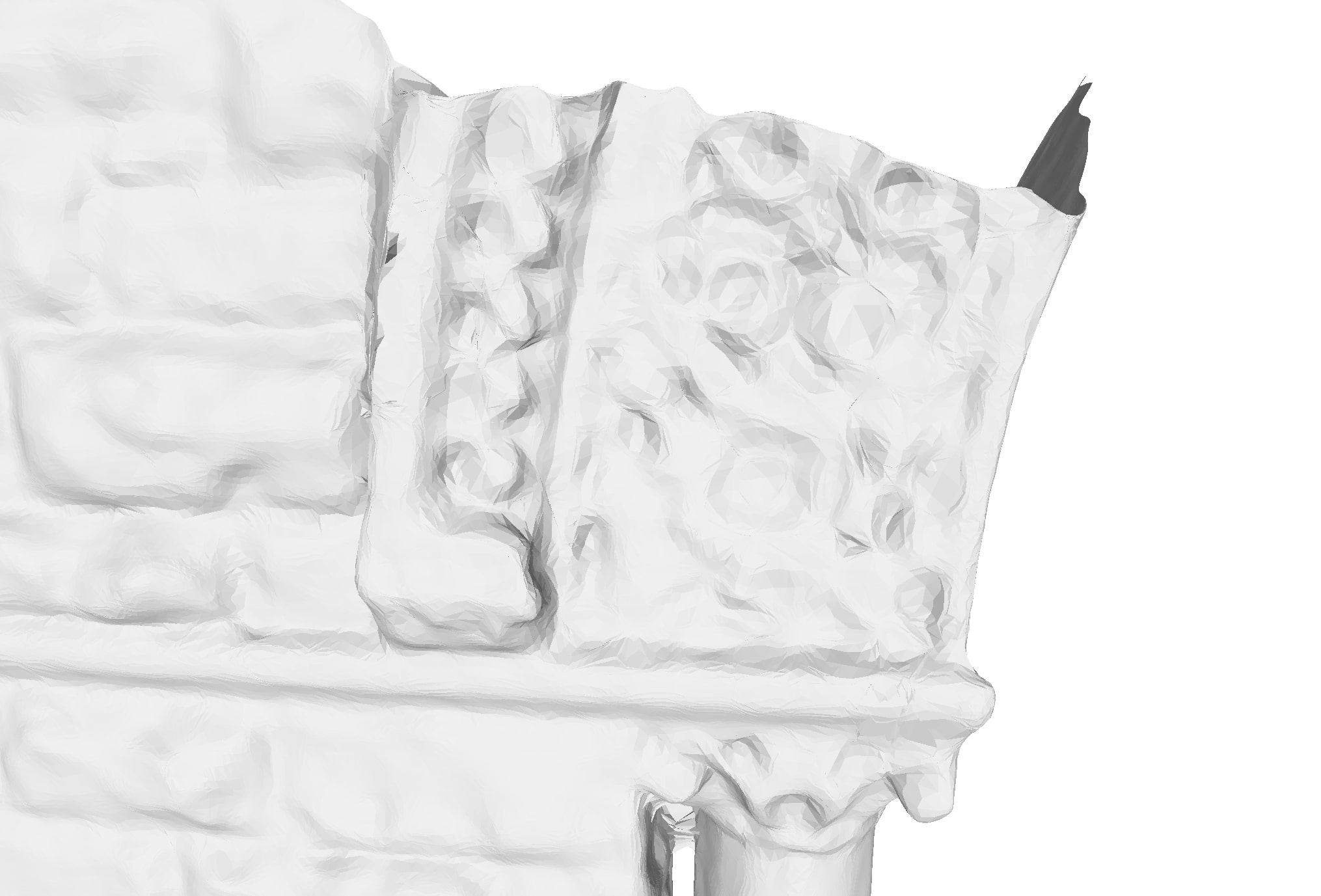}
        \end{minipage}
    }
    \subfigure[]{
        \begin{minipage}[b]{0.065\textwidth}
            \includegraphics[width=1\textwidth]{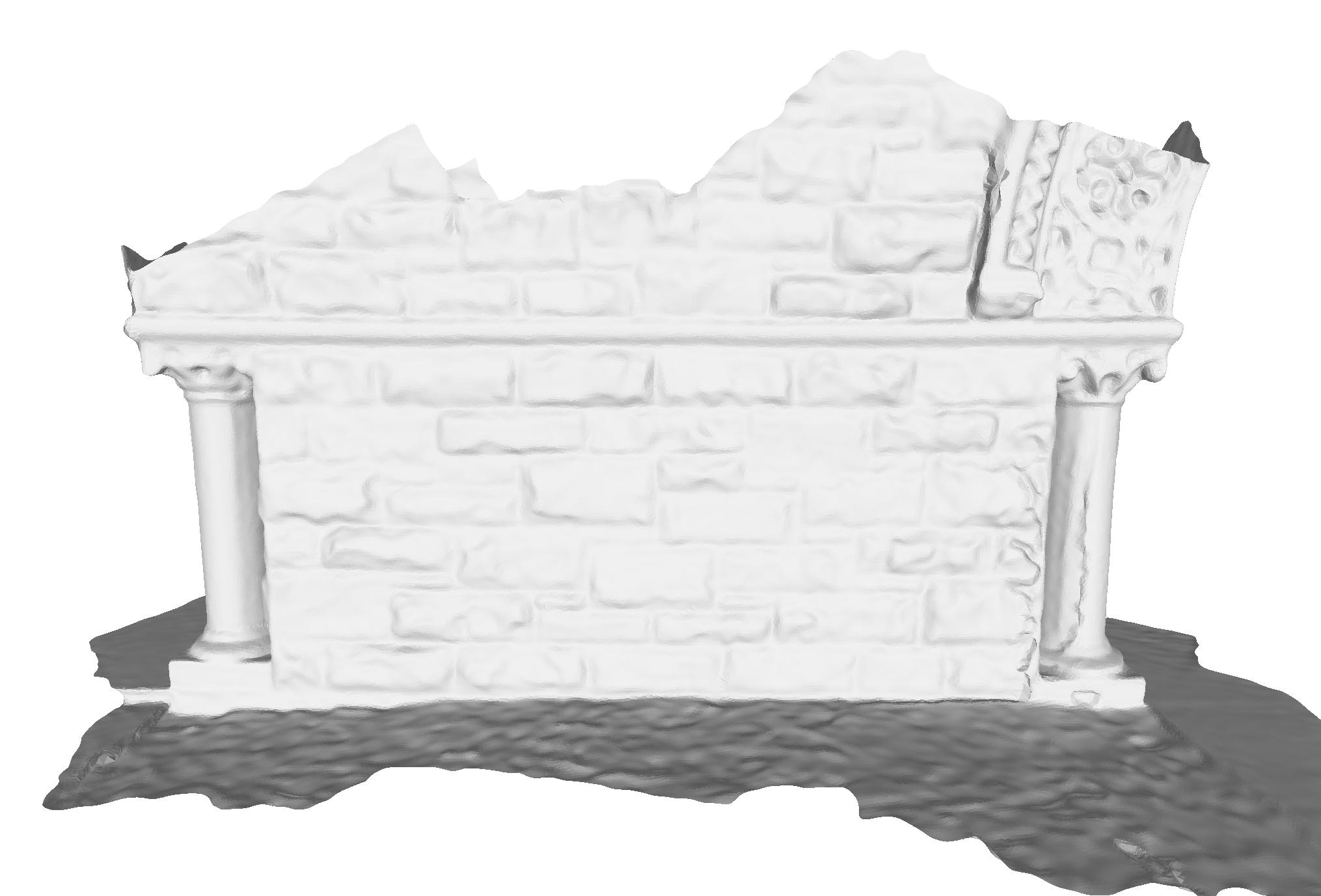}\\
            \includegraphics[width=1\textwidth]{images/detail/stonewall/newours_part_sto01.jpg}
        \end{minipage}
    }
    \caption{Visual results of the ablation studies: (a) Applying the absolute value to the output of the SIREN network. (b) Applying the softplus function to the output of the SIREN network. (c) Using uniform Eikonal constraints. (d) Removing normal alignment. (e) Replacing estimated normals with random vector. (f) Applying all components.
    }
    \label{fig:ablation}
\end{figure}

\paragraph{Unconditioned MLPs}
We assess the impact of using an unconditioned SIREN network on the performance of our method by comparing it to other versions of the SIREN network that utilize absolute value and softplus function in the output layer, respectively. As shown in Figure~\ref{fig:ablation} and Table~\ref{tab:ablation}, the absolute output layer results in large fitting error. On the other hand, due to the vanishing gradient effect of softplus, the reconstructed mesh using the SIREN network with a softplus output is typically over-smoothed.

\paragraph{Normal alignment}
To evaluate the contribution of normal alignment to reconstruct geometric details, we compare our method with and without the normal alignment loss. In addition, we test the robustness of our method when the estimated normals are unreliable. Figure~\ref{fig:ablation} and Table~\ref{tab:ablation} show the effects of the normal alignment loss. Interestingly, we observe that even fully random normals could also improve the results significantly. While the normals are unreliable, the $\mathcal{L}_{normal}$ also locates the local minimal distance surface between the two sample points $\mathbf{q}^1_i$ and $\mathbf{q}^2_i$. The is also beneficial to minimize the error.

\begin{table}[!htbp]
    \centering
    \setlength{\tabcolsep}{1pt}
    \footnotesize	
    \begin{tabular}{ccccc}
        \toprule
          & CD-mean &  CD-median & $F1^{0.005}$ & $F1^{0.0025}$\\
        \midrule
         SIREN+Abs. & 5.27 & 3.07  & 87.65 &  77.87 \\
         SIREN+softplus & 2.91 & 2.71  & 99.79 &  91.08\\
         w/o weighted Eikonal & 2.87 & 2.54  &99.26 &  90.98\\
         w/o normal alignment & 3.20 & 2.97  & 99.36 &  88.68\\
         random normal alignment & 2.72 & 2.53  & 99.87 & 92.12\\

        \midrule
         DEUDF & 2.69 &  2.51 & 99.95 & 93.52 \\
        \bottomrule
    \end{tabular}
    \caption{ Ablation studies on the model ``Stonewall'' (Figure~\ref{fig:ablation}) of the Stanford 3D Scene Dataset.}
    \label{tab:ablation}
\end{table}

\paragraph{Weighted Eikonal}
We explore the effects of different configurations of the Eikonal loss. Specifically, we compare our method using a standard Eikonal loss that is applied uniformly to all sample points with our adaptively weighted Eikonal loss. We observe that the standard Eikonal loss results in learned UDFs with lower accuracy near the zero level sets, leading to numerous small holes in the extracted meshes. In contrast, our adaptively weighted Eikonal loss more effectively addresses the vanishing gradient problem, and stabilizes the learning process, thereby yielding meshes with higher quality. In Figure~\ref{fig:ablation} and Table~\ref{tab:ablation}, due to the uniform Eikonal constraint that impedes the UDF learning process at the zero level set, slightly higher distance values are observed around the target surface, which can result in holes in the extracted mesh, even though a larger DCUDF~\cite{Hou2023} threshold has be applied in this example.

\section{Conclusions and Limitations}
\label{sec:conclusions}

This paper presents an improved UDF learning method for high fidelity 3D surface reconstruction. The method integrates novel UDF representation, normal alignment, adaptively weighted Eikonal constraint and SIREN network to learn more accurate UDFs. Our DEUDF can not only lean geometry details but also keep boundaries, thereby maintaining better topology. Extensive experiments illustrate that our method produces low Chamfer distances and better topology outperforming state-of-the-art methods.

Our method is primarily designed for dense, evenly sampled point clouds, as sparse point clouds often fail to capture fine geometric details. If the input point cloud is highly uneven, in some sparse locations, the learned unsigned distance values could be high resulting in small holes in the reconstructed mesh. In the future, we aim to extend DEUDF to handle point clouds with non-uniform densities, where points are sparse in smooth areas and dense in regions with fine details.

\section{Acknowledgments}
This work was supported in part by the National Key R\&D Program of China under Grant 2023YFB3002901, the Basic Research Project of ISCAS under Grant ISCAS-JCMS-202303, the Major Research Project of ISCAS under Grant ISCAS-ZD-202401, the Ministry of Education, Singapore, under its Academic Research Fund Grants (MOE-T2EP20220-0005 \& RT19/22), and an OPPO gift fund.

\bibliography{arxiv_aaai}

\begin{thebibliography}{38}
\providecommand{\natexlab}[1]{#1}

\bibitem[{Amenta and Bern(1998)}]{Amenta1998}
Amenta, N.; and Bern, M. 1998.
\newblock Surface Reconstruction by Voronoi Filtering.
\newblock In \emph{Proceedings of SoCG}, 39--48.

\bibitem[{Chabra et~al.(2020)Chabra, Lenssen, Ilg, Schmidt, Straub, Lovegrove, and Newcombe}]{Chabra2020}
Chabra, R.; Lenssen, J.~E.; Ilg, E.; Schmidt, T.; Straub, J.; Lovegrove, S.; and Newcombe, R. 2020.
\newblock Deep Local Shapes: Learning Local SDF Priors for Detailed 3D Reconstruction.
\newblock In \emph{Proc. of ECCV}, 608--625.

\bibitem[{Chang et~al.(2015)Chang, Funkhouser, Guibas, Hanrahan, Huang, Li, Savarese, Savva, Song, Su, Xiao, Yi, and Yu}]{Chang2015}
Chang, A.~X.; Funkhouser, T.; Guibas, L.; Hanrahan, P.; Huang, Q.; Li, Z.; Savarese, S.; Savva, M.; Song, S.; Su, H.; Xiao, J.; Yi, L.; and Yu, F. 2015.
\newblock ShapeNet: An Information-Rich 3D Model Repository.
\newblock arXiv:1512.03012.

\bibitem[{Chibane, Alldieck, and Pons-Moll(2020)}]{Chibane2020IFNET}
Chibane, J.; Alldieck, T.; and Pons-Moll, G. 2020.
\newblock Implicit Functions in Feature Space for 3D Shape Reconstruction and Completion.
\newblock In \emph{Proc. of CVPR}, 6968--6979.

\bibitem[{Chibane, Mir, and Pons-Moll(2020)}]{Chibane2020NDF}
Chibane, J.; Mir, A.; and Pons-Moll, G. 2020.
\newblock Neural Unsigned Distance Fields for Implicit Function Learning.
\newblock In \emph{Proc. of NeurIPS}, 21638--21652.

\bibitem[{Deng et~al.(2024)Deng, Hou, Chen, Wang, and He}]{2SUDF}
Deng, J.; Hou, F.; Chen, X.; Wang, W.; and He, Y. 2024.
\newblock 2S-UDF: A Novel Two-stage UDF Learning Method for Robust Non-watertight Model Reconstruction from Multi-view Images.
\newblock In \emph{Proc. of CVPR}, 5084--5093.

\bibitem[{Dey and Goswami(2003)}]{Dey2003}
Dey, T.~K.; and Goswami, S. 2003.
\newblock Tight Cocone: A Water-Tight Surface Reconstructor.
\newblock In \emph{Proc. of ACM SMA}, 127--134.

\bibitem[{Fainstein, Siless, and Iarussi(2024)}]{fainstein2024dudf}
Fainstein, M.; Siless, V.; and Iarussi, E. 2024.
\newblock {DUDF:} Differentiable Unsigned Distance Fields with Hyperbolic Scaling.
\newblock In \emph{Proc. of CVPR}, 4484--4493.

\bibitem[{Guillard, Stella, and Fua(2022)}]{Guillard2022MeshUDF}
Guillard, B.; Stella, F.; and Fua, P. 2022.
\newblock MeshUDF: Fast and Differentiable Meshing of Unsigned Distance Field Networks.
\newblock In \emph{Proc. of ECCV}, 576--592.

\bibitem[{Hoppe et~al.(1992)Hoppe, DeRose, Duchamp, McDonald, and Stuetzle}]{Hoppe1992}
Hoppe, H.; DeRose, T.; Duchamp, T.; McDonald, J.; and Stuetzle, W. 1992.
\newblock Surface reconstruction from unorganized points.
\newblock \emph{SIGGRAPH Comput. Graph.}, 26(2): 71–78.

\bibitem[{Hou et~al.(2023)Hou, Chen, Wang, Qin, and He}]{Hou2023}
Hou, F.; Chen, X.; Wang, W.; Qin, H.; and He, Y. 2023.
\newblock Robust Zero Level-Set Extraction from Unsigned Distance Fields Based on Double Covering.
\newblock \emph{ACM Trans. Graph.}, 42(6).

\bibitem[{Hou et~al.(2022)Hou, Wang, Wang, Qin, Qian, and He}]{Hou2022}
Hou, F.; Wang, C.; Wang, W.; Qin, H.; Qian, C.; and He, Y. 2022.
\newblock Iterative Poisson Surface Reconstruction (iPSR) for Unoriented Points.
\newblock \emph{ACM Trans. Graph.}, 41(4).

\bibitem[{Kazhdan, Bolitho, and Hoppe(2006)}]{Kazhdan2006}
Kazhdan, M.; Bolitho, M.; and Hoppe, H. 2006.
\newblock Poisson Surface Reconstruction.
\newblock In \emph{Proc. of SGP}, 61--70.

\bibitem[{Kazhdan and Hoppe(2013)}]{Kazhdan2013}
Kazhdan, M.; and Hoppe, H. 2013.
\newblock Screened Poisson Surface Reconstruction.
\newblock \emph{ACM Trans. Graph.}, 32(3).

\bibitem[{Kingma and Ba(2015)}]{2014Adam}
Kingma, D.~P.; and Ba, J. 2015.
\newblock Adam: {A} Method for Stochastic Optimization.
\newblock In \emph{Proc. of ICLR}.

\bibitem[{Liu et~al.(2024)Liu, Li, Chen, Hou, Xin, Wang, Wu, Qian, and He}]{DWG}
Liu, W.; Li, J.; Chen, X.; Hou, F.; Xin, S.; Wang, X.; Wu, Z.; Qian, C.; and He, Y. 2024.
\newblock Diffusing Winding Gradients {(DWG):} {A} Parallel and Scalable Method for 3D Reconstruction from Unoriented Point Clouds.
\newblock \emph{arXiv:2405.13839}.

\bibitem[{Liu et~al.(2023)Liu, Wang, Yang, Chen, Meng, Yang, and Gao}]{Liu2023}
Liu, Y.-T.; Wang, L.; Yang, J.; Chen, W.; Meng, X.; Yang, B.; and Gao, L. 2023.
\newblock {NeUDF}: Learning Neural Unsigned Distance Fields with Volume Rendering.
\newblock In \emph{Proc. of CVPR}, 237--247.

\bibitem[{Long et~al.(2023)Long, Lin, Liu, Liu, Wang, Theobalt, Komura, and Wang}]{Long2023}
Long, X.; Lin, C.; Liu, L.; Liu, Y.; Wang, P.; Theobalt, C.; Komura, T.; and Wang, W. 2023.
\newblock {NeuralUDF}: Learning Unsigned Distance Fields for Multi-view Reconstruction of Surfaces with Arbitrary Topologies.
\newblock In \emph{Proc. of CVPR}, 20834--20843.

\bibitem[{Loshchilov and Hutter(2017)}]{2017SGDR}
Loshchilov, I.; and Hutter, F. 2017.
\newblock {SGDR:} Stochastic Gradient Descent with Warm Restarts.
\newblock In \emph{Proc. of ICLR}.

\bibitem[{Mescheder et~al.(2019)Mescheder, Oechsle, Niemeyer, Nowozin, and Geiger}]{Mescheder2019}
Mescheder, L.; Oechsle, M.; Niemeyer, M.; Nowozin, S.; and Geiger, A. 2019.
\newblock Occupancy Networks: Learning 3D Reconstruction in Function Space.
\newblock In \emph{Proc. of CVPR}, 4455--4465.

\bibitem[{Ohtake et~al.(2003)Ohtake, Belyaev, Alexa, Turk, and Seidel}]{Ohtake2003}
Ohtake, Y.; Belyaev, A.; Alexa, M.; Turk, G.; and Seidel, H.-P. 2003.
\newblock Multi-Level Partition of Unity Implicits.
\newblock \emph{ACM Trans. Graph.}, 22(3): 463--470.

\bibitem[{Park et~al.(2019)Park, Florence, Straub, Newcombe, and Lovegrove}]{Park2019}
Park, J.~J.; Florence, P.; Straub, J.; Newcombe, R.; and Lovegrove, S. 2019.
\newblock DeepSDF: Learning Continuous Signed Distance Functions for Shape Representation.
\newblock In \emph{Proc. of CVPR}, 165--174.

\bibitem[{Peng et~al.(2020)Peng, Niemeyer, Mescheder, Pollefeys, and Geiger}]{Peng2020}
Peng, S.; Niemeyer, M.; Mescheder, L.; Pollefeys, M.; and Geiger, A. 2020.
\newblock Convolutional Occupancy Networks.
\newblock In \emph{Proc. of ECCV}, 523--540.

\bibitem[{Ren et~al.(2023)Ren, Hou, Chen, He, and Wang}]{Ren2023}
Ren, S.; Hou, J.; Chen, X.; He, Y.; and Wang, W. 2023.
\newblock GeoUDF: Surface Reconstruction from 3D Point Clouds via Geometry-guided Distance Representation.
\newblock In \emph{Proc. of ICCV}, 14214--14224.

\bibitem[{Sitzmann et~al.(2020)Sitzmann, Martel, Bergman, Lindell, and Wetzstein}]{Sitzmann2020SIREN}
Sitzmann, V.; Martel, J. N.~P.; Bergman, A.~W.; Lindell, D.~B.; and Wetzstein, G. 2020.
\newblock Implicit neural representations with periodic activation functions.
\newblock In \emph{Proc. of NeurIPS}, 7462--7473.

\bibitem[{Wang et~al.(2021)Wang, Liu, Liu, Theobalt, Komura, and Wang}]{wang2021neus}
Wang, P.; Liu, L.; Liu, Y.; Theobalt, C.; Komura, T.; and Wang, W. 2021.
\newblock NeuS: Learning Neural Implicit Surfaces by Volume Rendering for Multi-view Reconstruction.
\newblock In \emph{Proc. of NeurIPS}, 27171--27183.

\bibitem[{Wang et~al.(2023{\natexlab{a}})Wang, Wang, Zhang, Chen, Xin, Tu, and Wang}]{AlignGraidentandHessian}
Wang, R.; Wang, Z.; Zhang, Y.; Chen, S.; Xin, S.; Tu, C.; and Wang, W. 2023{\natexlab{a}}.
\newblock Aligning Gradient and Hessian for Neural Signed Distance Function.
\newblock In \emph{Proc. of NeurIPS}, 63515--63528.

\bibitem[{Wang, Rahmann, and Sorkine{-}Hornung(2022)}]{Wang2022IDF}
Wang, Y.; Rahmann, L.; and Sorkine{-}Hornung, O. 2022.
\newblock Geometry-Consistent Neural Shape Representation with Implicit Displacement Fields.
\newblock In \emph{Proc. of ICLR}.

\bibitem[{Wang et~al.(2023{\natexlab{b}})Wang, Zhang, Xu, Zhang, Wang, Chen, Xin, Wang, and Tu}]{NSH}
Wang, Z.; Zhang, Y.; Xu, R.; Zhang, F.; Wang, P.; Chen, S.; Xin, S.; Wang, W.; and Tu, C. 2023{\natexlab{b}}.
\newblock Neural-Singular-Hessian: Implicit Neural Representation of Unoriented Point Clouds by Enforcing Singular Hessian.
\newblock \emph{{ACM} Trans. Graph.}, 42(6): 274:1--274:14.

\bibitem[{Xu et~al.(2023)Xu, Dou, Wang, Xin, Chen, Jiang, Guo, Wang, and Tu}]{Rui2023GCNO}
Xu, R.; Dou, Z.; Wang, N.; Xin, S.; Chen, S.; Jiang, M.; Guo, X.; Wang, W.; and Tu, C. 2023.
\newblock Globally Consistent Normal Orientation for Point Clouds by Regularizing the Winding-Number Field.
\newblock \emph{ACM Trans. Graph.}, 42(4).

\bibitem[{Yang et~al.(2023)Yang, Lin, Chen, and Zhou}]{Yang_2023_CVPR}
Yang, X.; Lin, G.; Chen, Z.; and Zhou, L. 2023.
\newblock Neural Vector Fields: Implicit Representation by Explicit Learning.
\newblock In \emph{Proc. of CVPR}, 16727--16738.

\bibitem[{Yariv et~al.(2021)Yariv, Gu, Kasten, and Lipman}]{VolSDF}
Yariv, L.; Gu, J.; Kasten, Y.; and Lipman, Y. 2021.
\newblock Volume Rendering of Neural Implicit Surfaces.
\newblock In \emph{Proc. of NeurIPS}, 4805--4815.

\bibitem[{Ye et~al.(2022)Ye, Chen, Wang, and Wang}]{Ye2022}
Ye, J.; Chen, Y.; Wang, N.; and Wang, X. 2022.
\newblock {GIFS}: Neural Implicit Function for General Shape Representation.
\newblock In \emph{Proc. of CVPR}, 12819--12829.

\bibitem[{Yu et~al.(2025)Yu, Dou, Long, Lin, Li, Liu, M{\"u}ller, Komura, Habermann, Theobalt et~al.}]{yu2025surfd}
Yu, Z.; Dou, Z.; Long, X.; Lin, C.; Li, Z.; Liu, Y.; M{\"u}ller, N.; Komura, T.; Habermann, M.; Theobalt, C.; et~al. 2025.
\newblock Surf-D: Generating High-Quality Surfaces of Arbitrary Topologies Using Diffusion Models.
\newblock In \emph{European Conference on Computer Vision (ECCV2024)}, 419--438. Springer.

\bibitem[{Zhou et~al.(2023)Zhou, Ma, Li, Liu, and Han}]{Zhou2023levelset}
Zhou, J.; Ma, B.; Li, S.; Liu, Y.-S.; and Han, Z. 2023.
\newblock Learning a More Continuous Zero Level Set in Unsigned Distance Fields through Level Set Projection.
\newblock In \emph{Proc. of ICCV}, 3158--3169.

\bibitem[{Zhou et~al.(2022)Zhou, Ma, Liu, Fang, and Han}]{Zhou2022CAPUDF}
Zhou, J.; Ma, B.; Liu, Y.-S.; Fang, Y.; and Han, Z. 2022.
\newblock Learning Consistency-Aware Unsigned Distance Functions Progressively from Raw Point Clouds.
\newblock In \emph{Proc. of NeurIPS}, 16481--16494.

\bibitem[{Zhou et~al.(2024)Zhou, Zhang, Ma, Shi, Liu, and Han}]{udiff}
Zhou, J.; Zhang, W.; Ma, B.; Shi, K.; Liu, Y.-S.; and Han, Z. 2024.
\newblock UDiFF: Generating Conditional Unsigned Distance Fields with Optimal Wavelet Diffusion.
\newblock In \emph{Proc. of CVPR}, 21496--21506.

\bibitem[{Zhou and Koltun(2013)}]{3DSceneDataset}
Zhou, Q.-Y.; and Koltun, V. 2013.
\newblock Dense Scene Reconstruction with Points of Interest.
\newblock \emph{ACM Trans. Graph.}, 32.

\end{thebibliography}

\clearpage

\twocolumn[
\begin{@twocolumnfalse}
\section*{\centering{Supplementary Material for \\ \emph{Details Enhancement in Unsigned Distance Field Learning for High-fidelity 3D Surface Reconstruction\\[25pt]}}}
\end{@twocolumnfalse}
]

\section{Appendix}
\begin{figure*}[htbp]
    \centering
    \subfigure[CAP-UDF]
    {
        \begin{minipage}[b]{0.23\textwidth}
		  \includegraphics[width=1\textwidth]{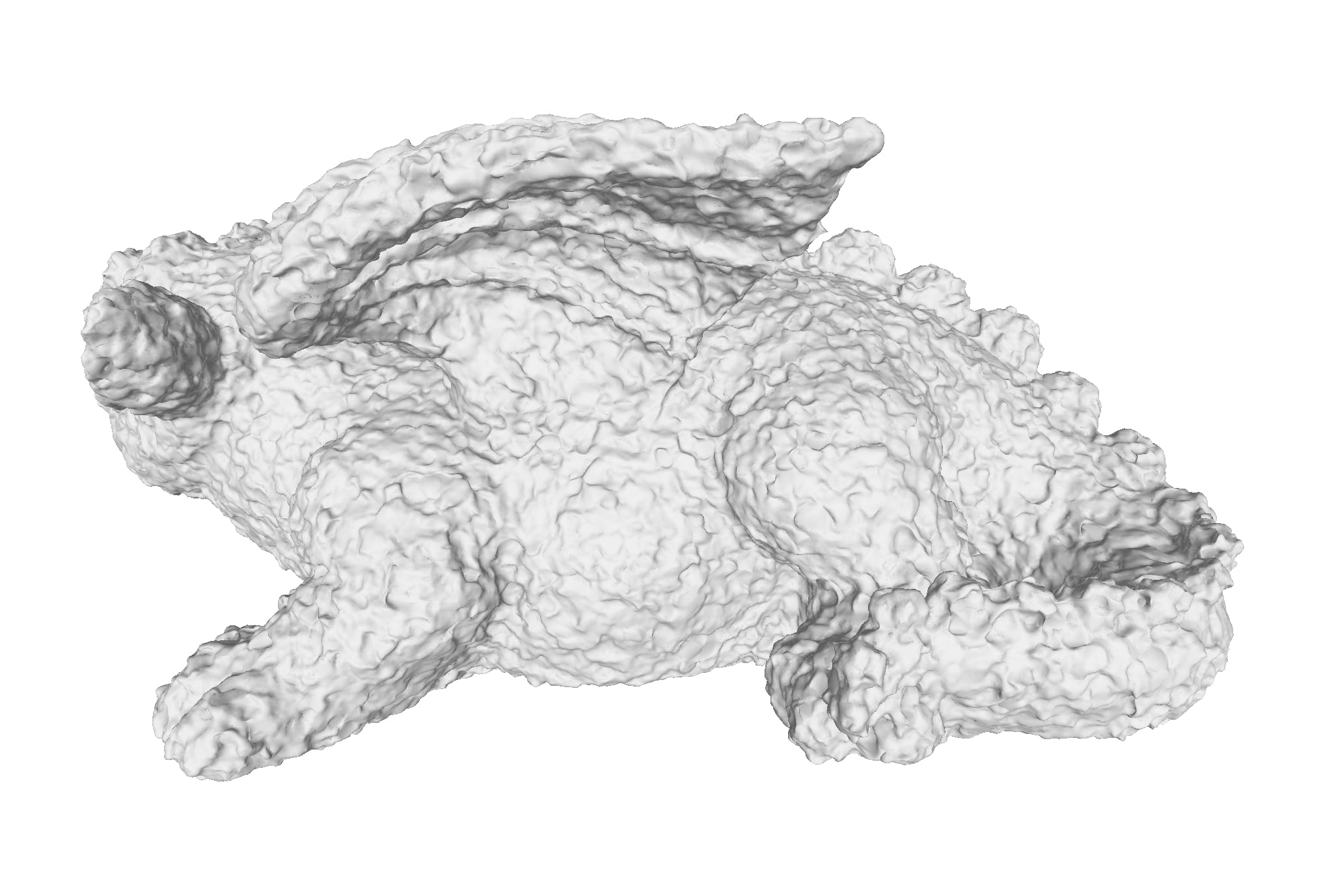} \\
		  \includegraphics[width=1\textwidth]{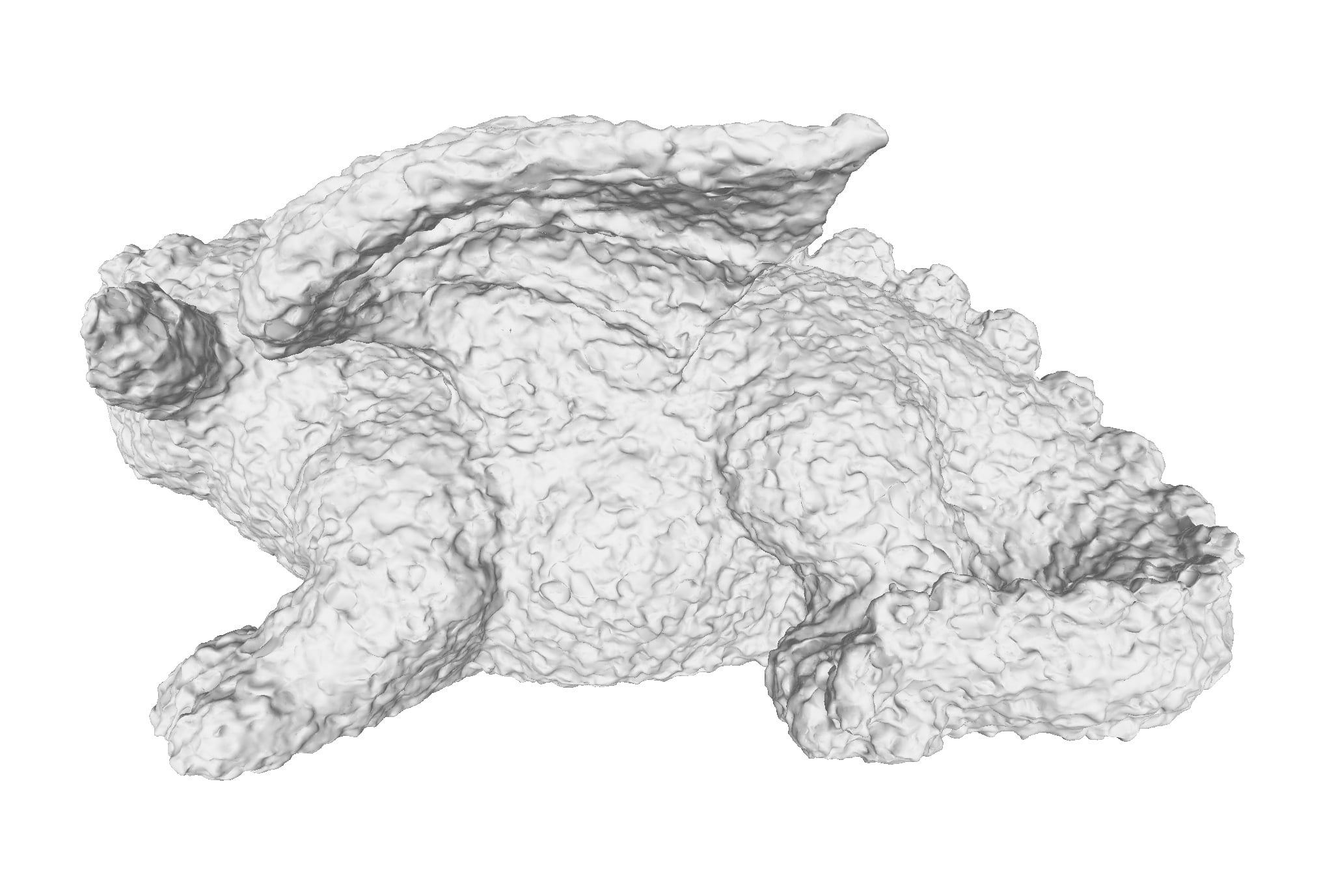} \\
            \includegraphics[width=1\textwidth]{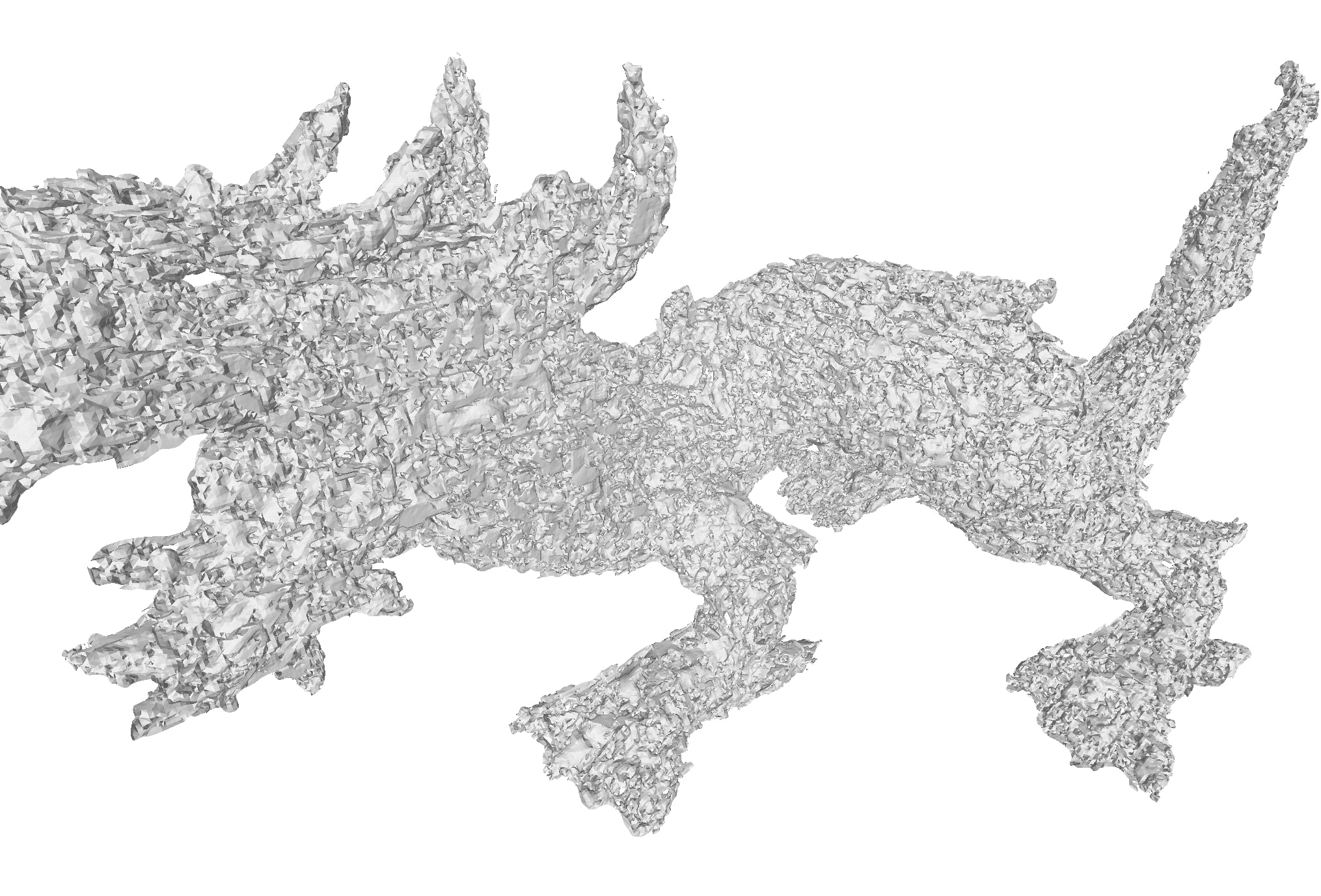} \\
		  \includegraphics[width=1\textwidth]{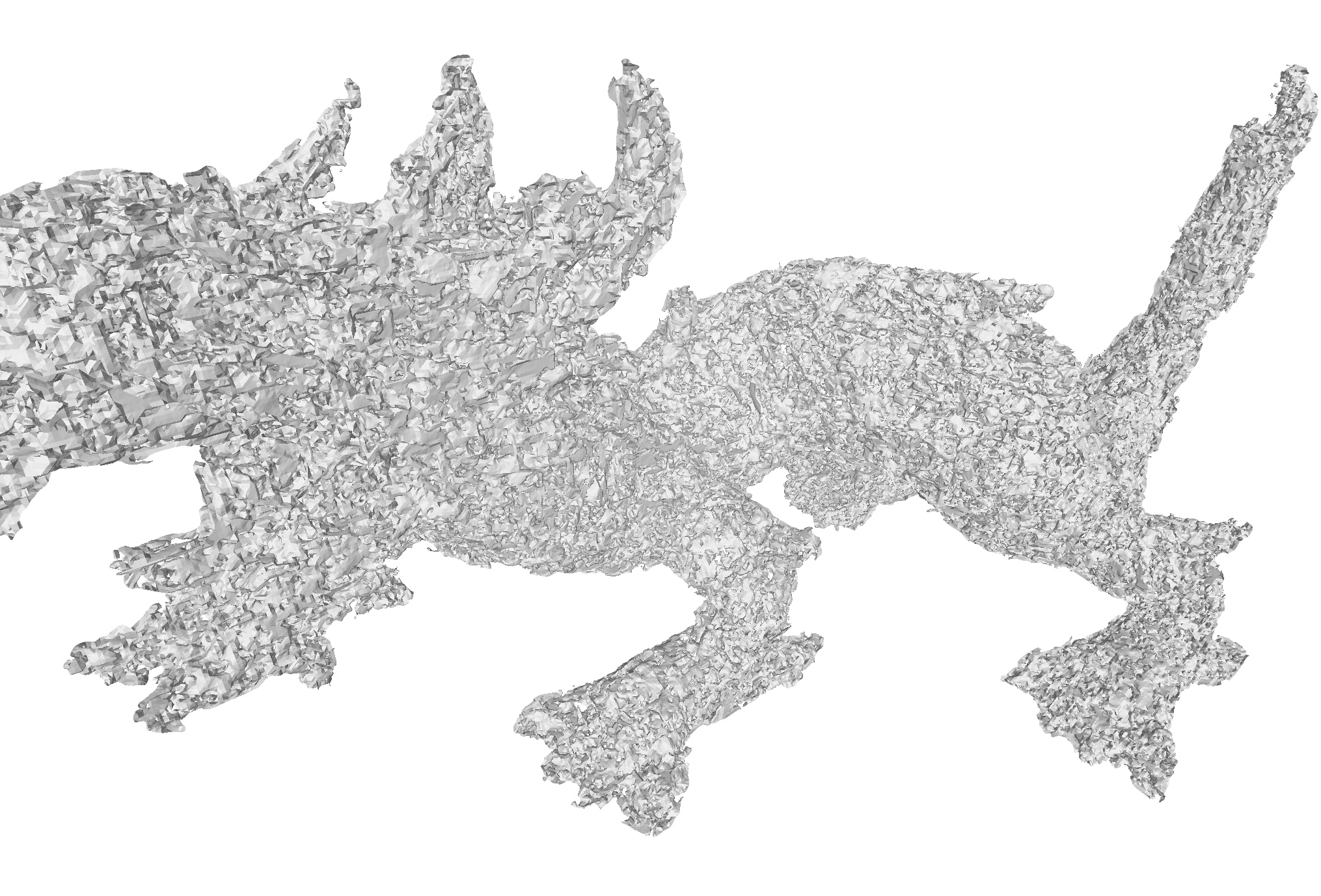} \\
    	\includegraphics[width=1\textwidth]{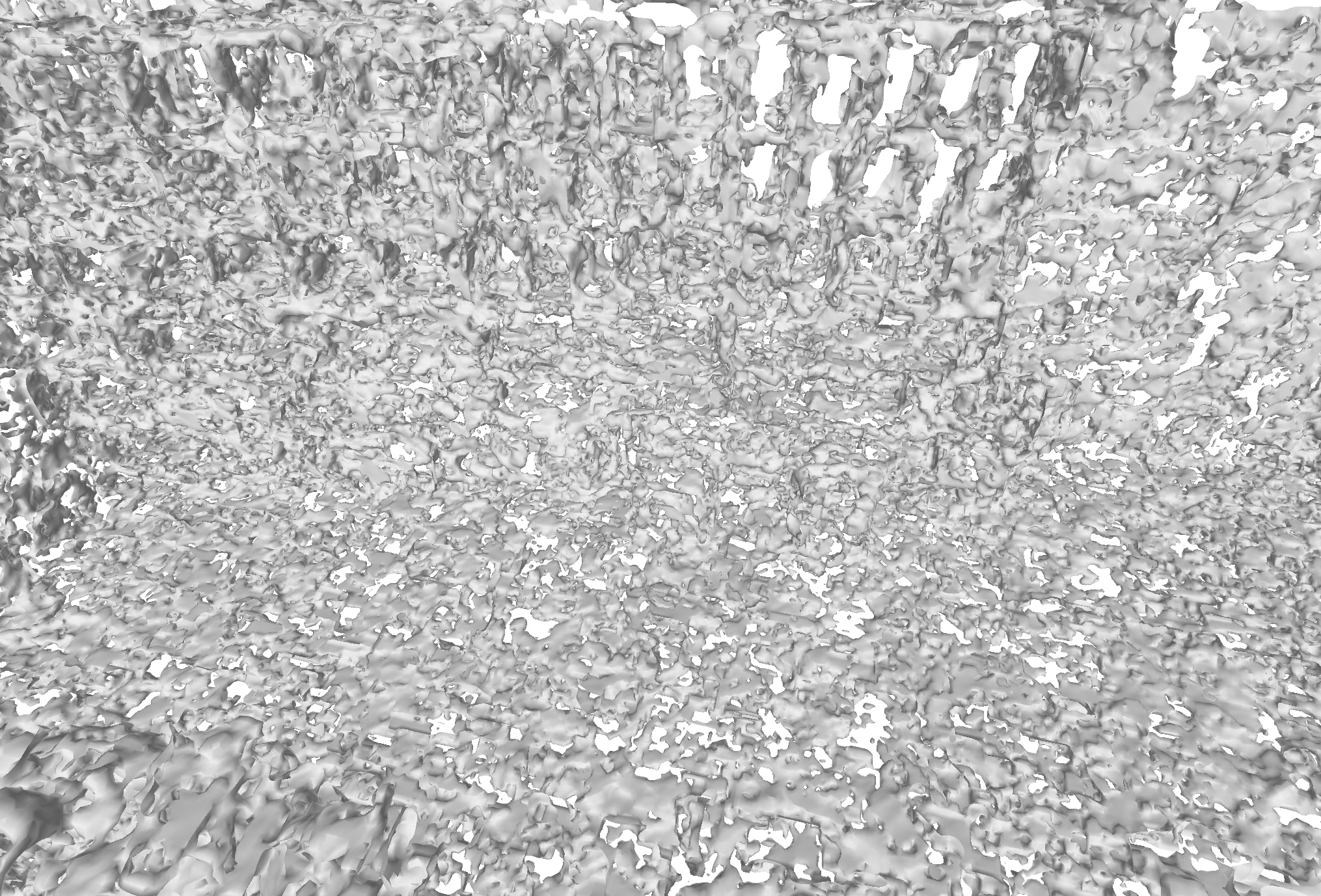} \\
		  \includegraphics[width=1\textwidth]{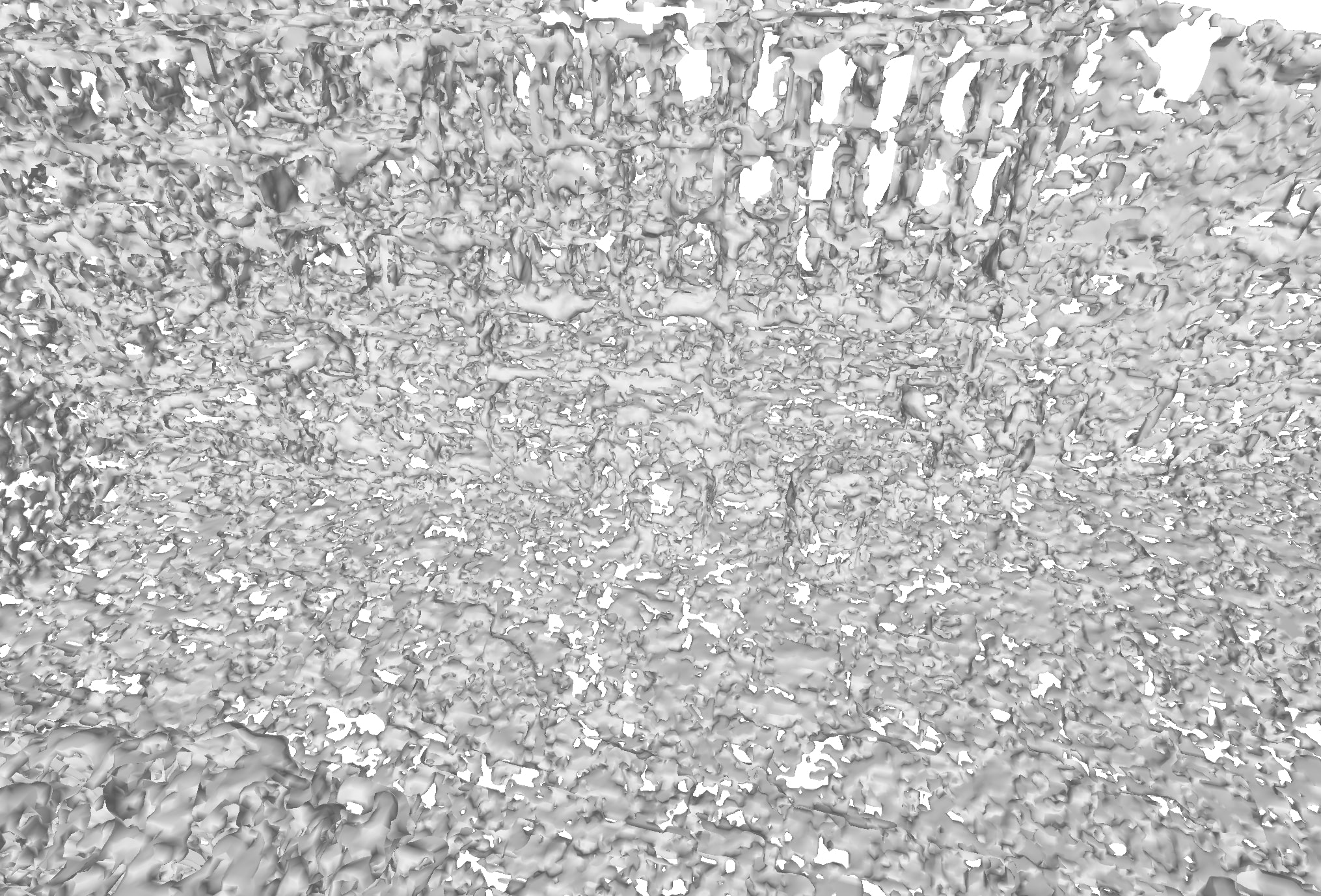} \\
    	\includegraphics[width=1\textwidth]{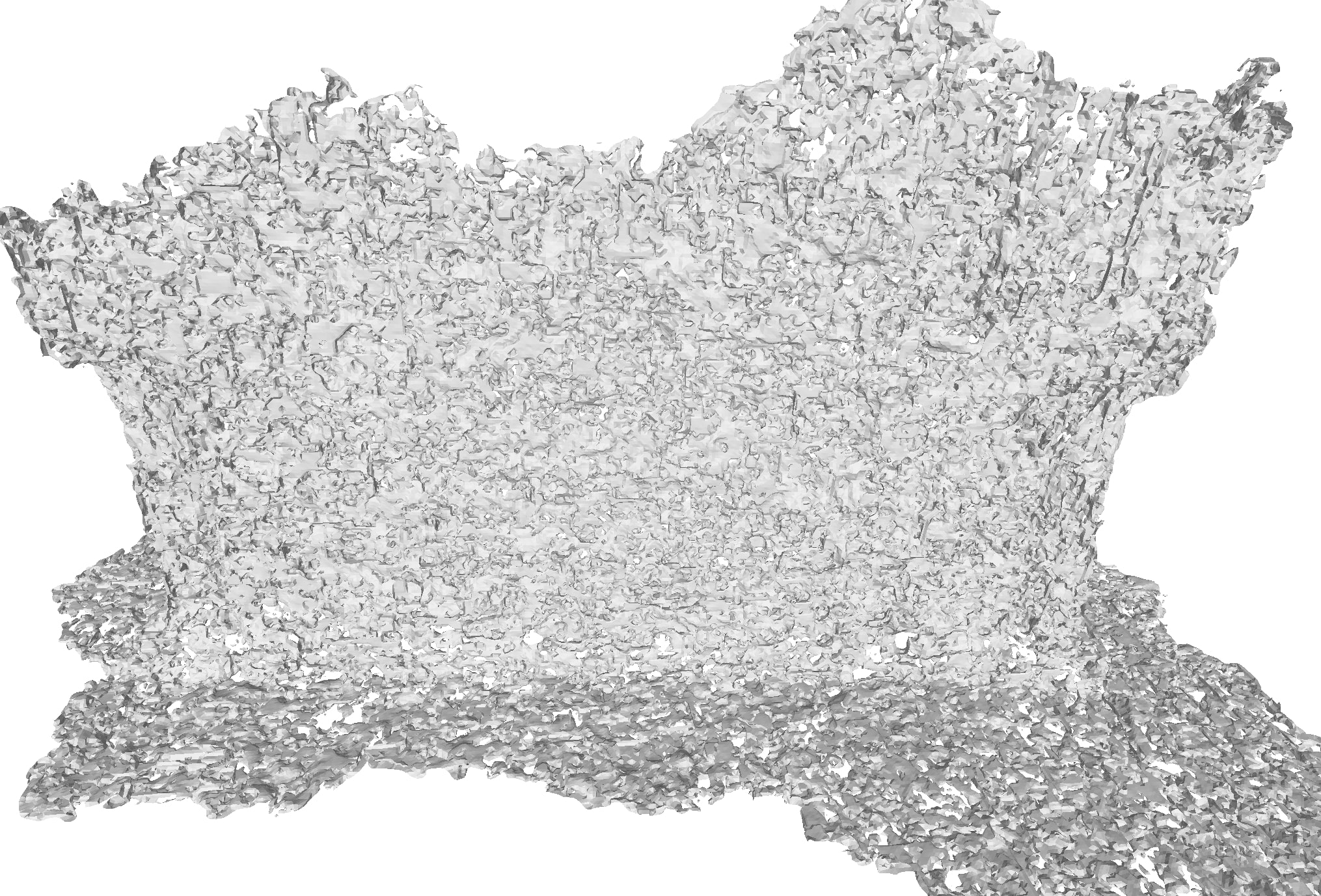} \\
		  \includegraphics[width=1\textwidth]{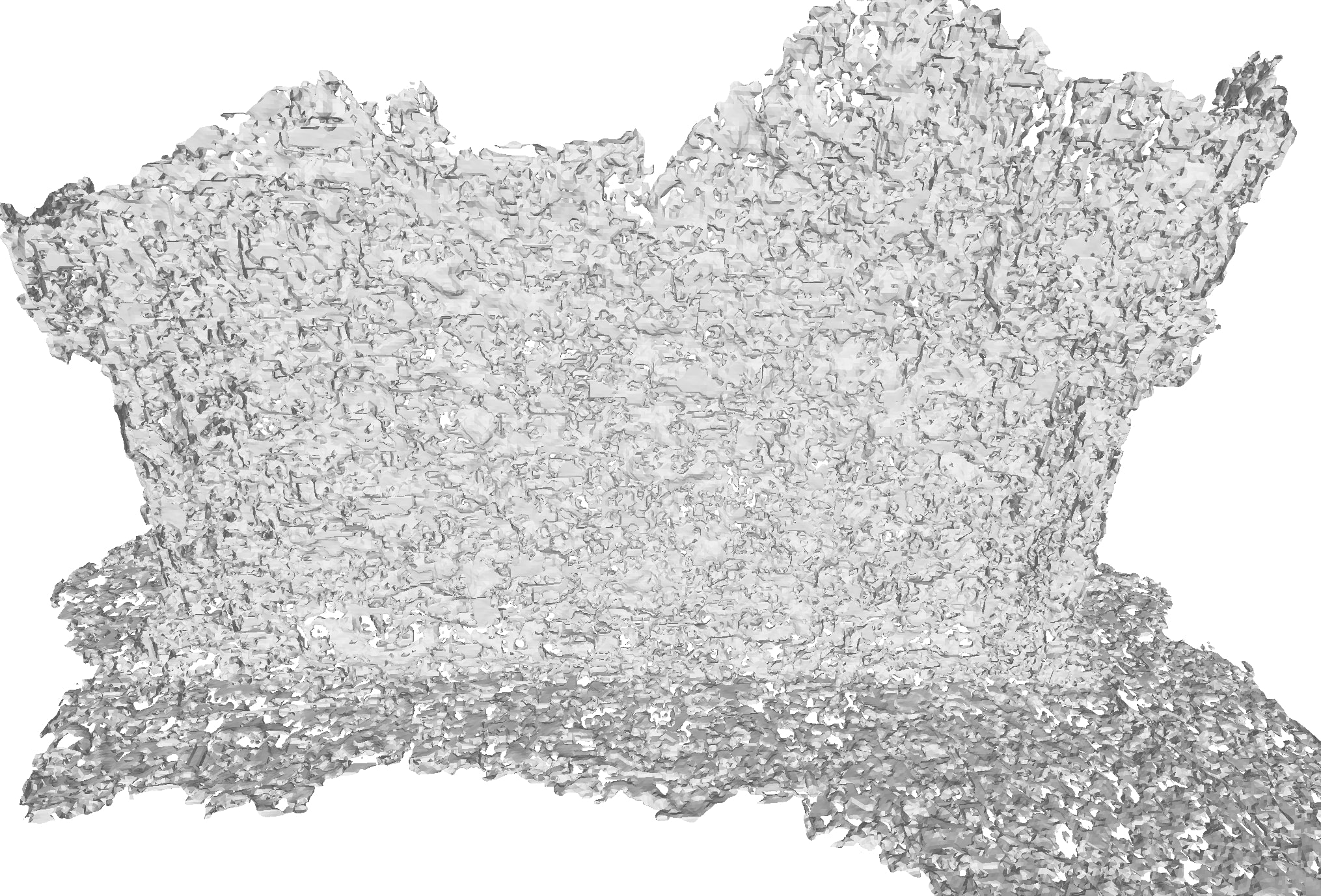} 
        \end{minipage}
    }
    \subfigure[LevelSetUDF]
    {
        \begin{minipage}[b]{0.23\textwidth}
		  \includegraphics[width=1\textwidth]{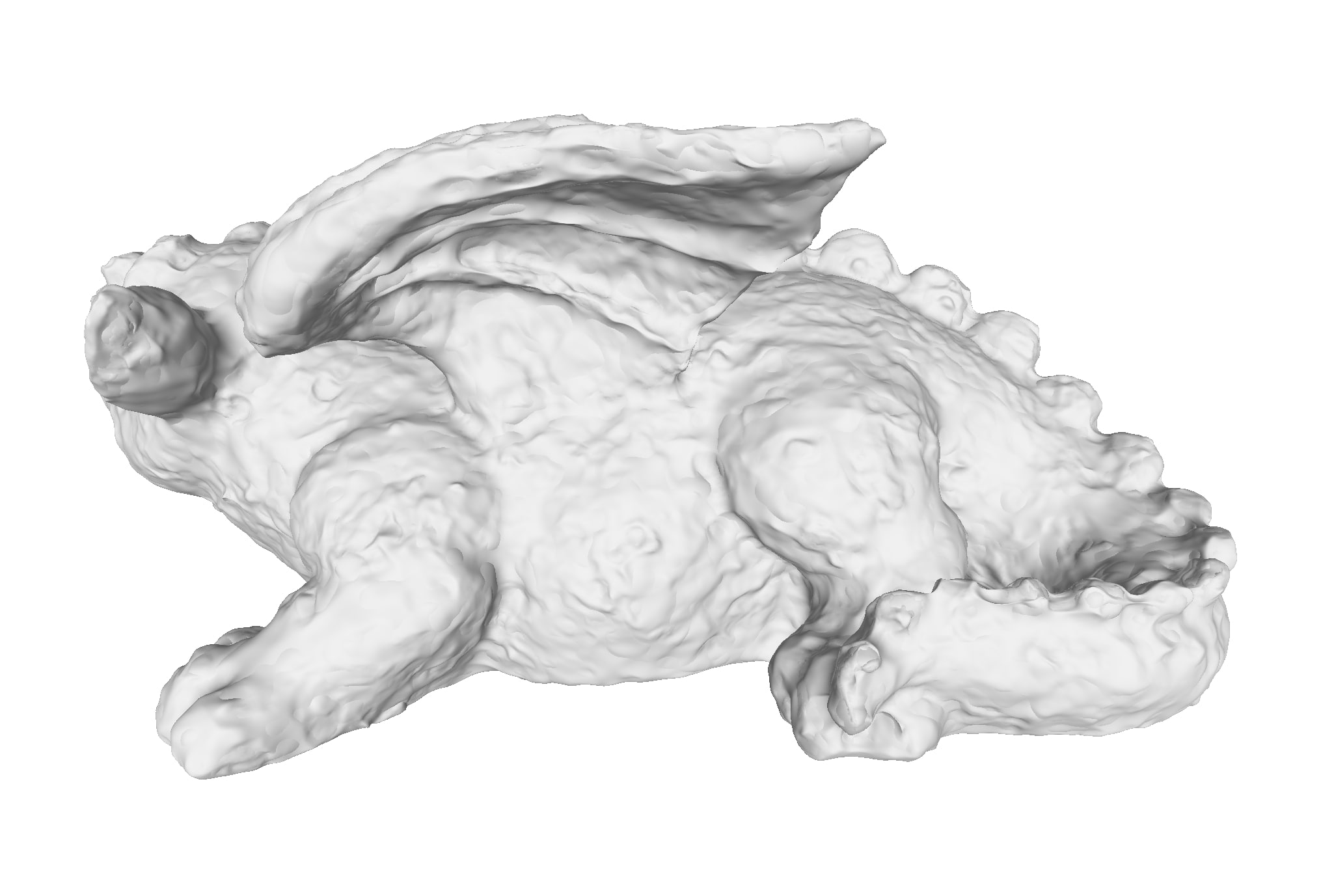} \\
		  \includegraphics[width=1\textwidth]{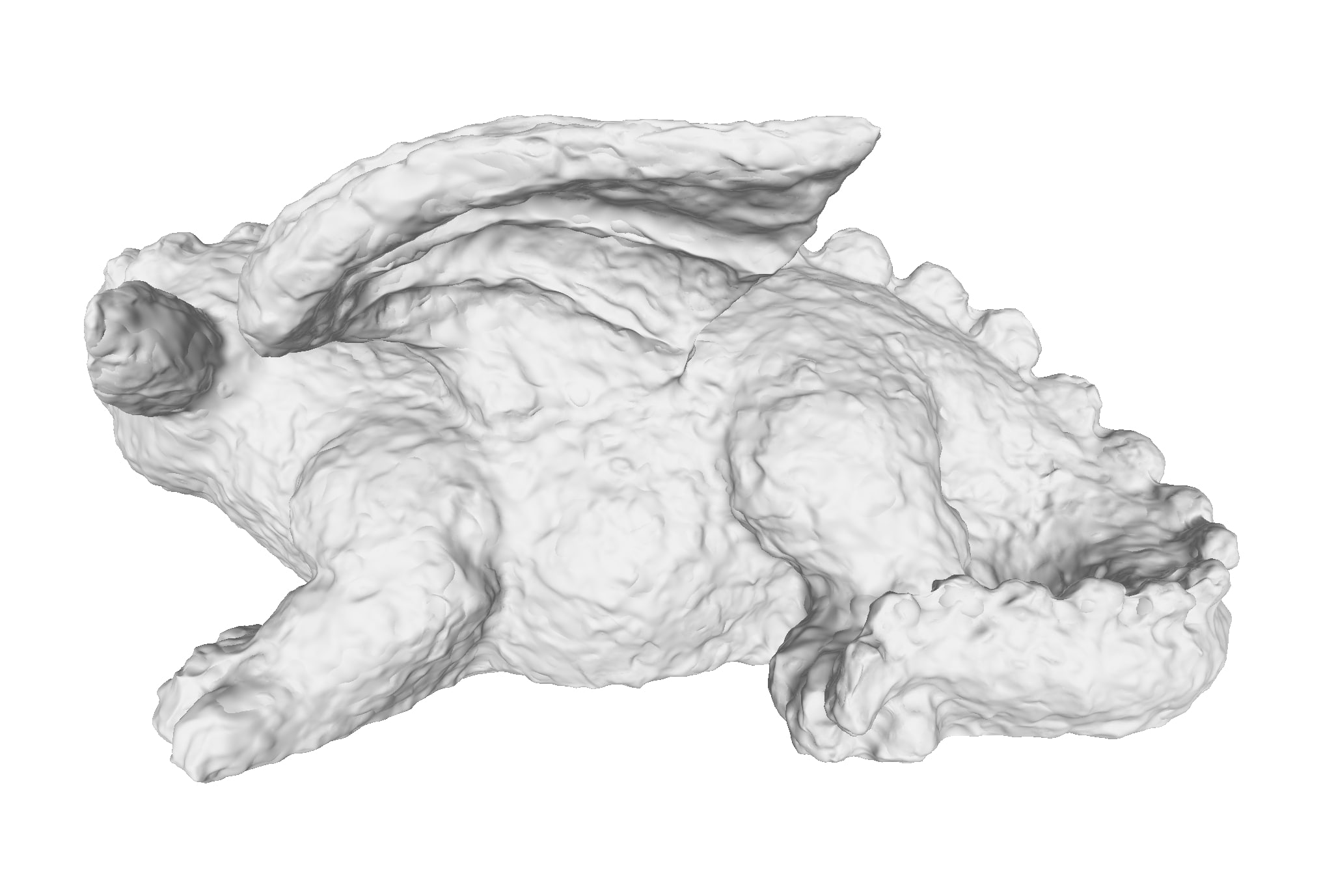} \\
            \includegraphics[width=1\textwidth]{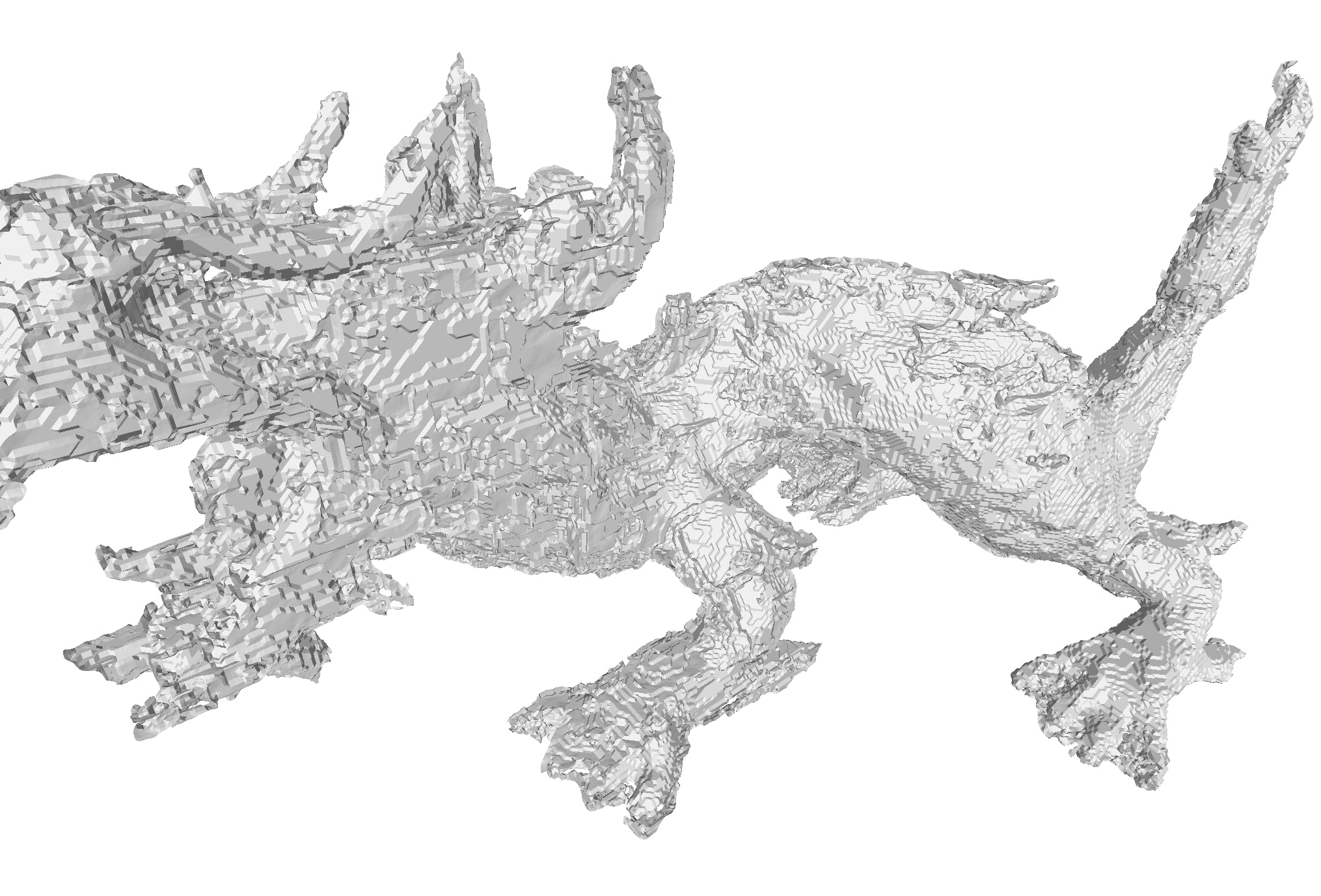} \\
		  \includegraphics[width=1\textwidth]{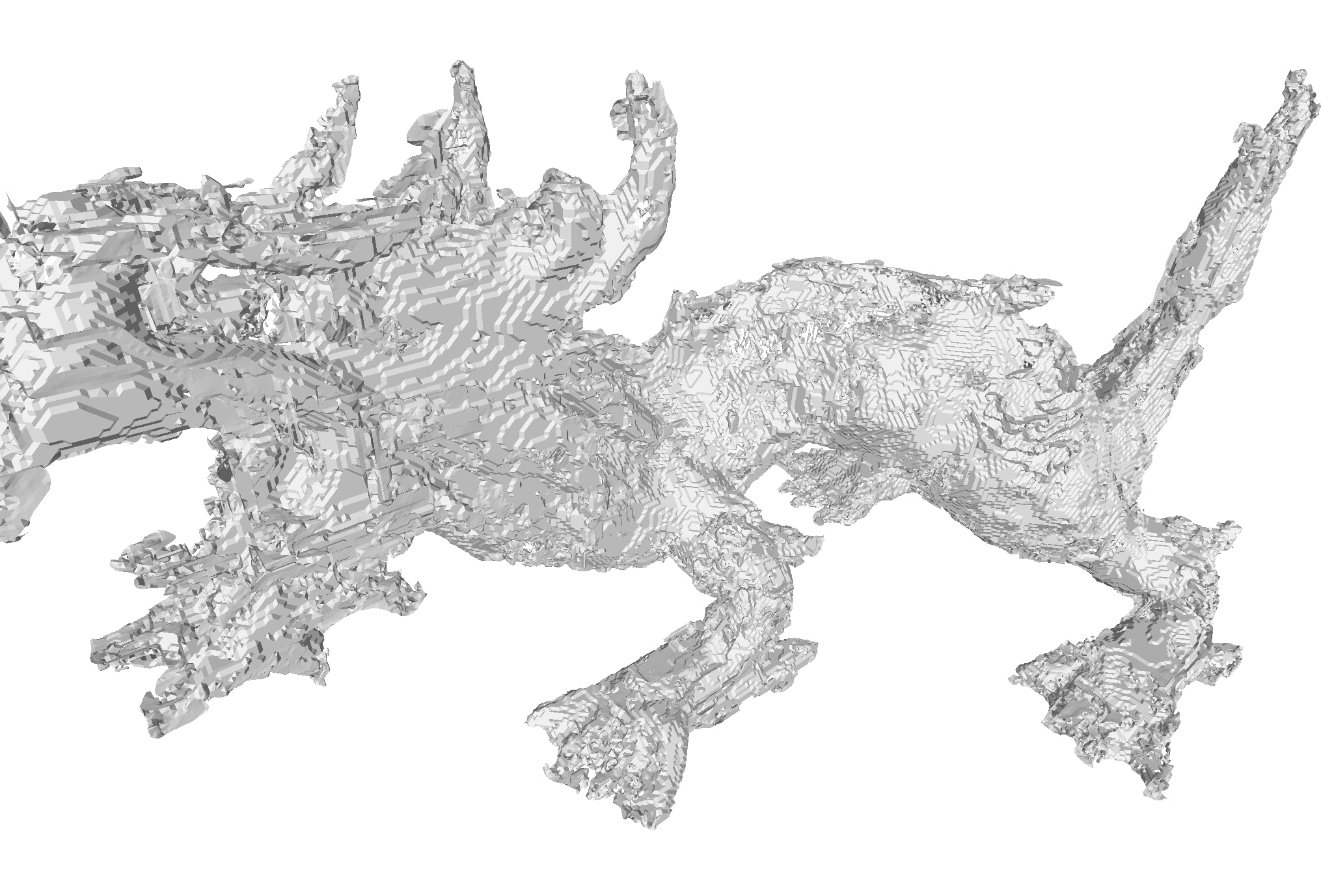} \\
            \includegraphics[width=1\textwidth]{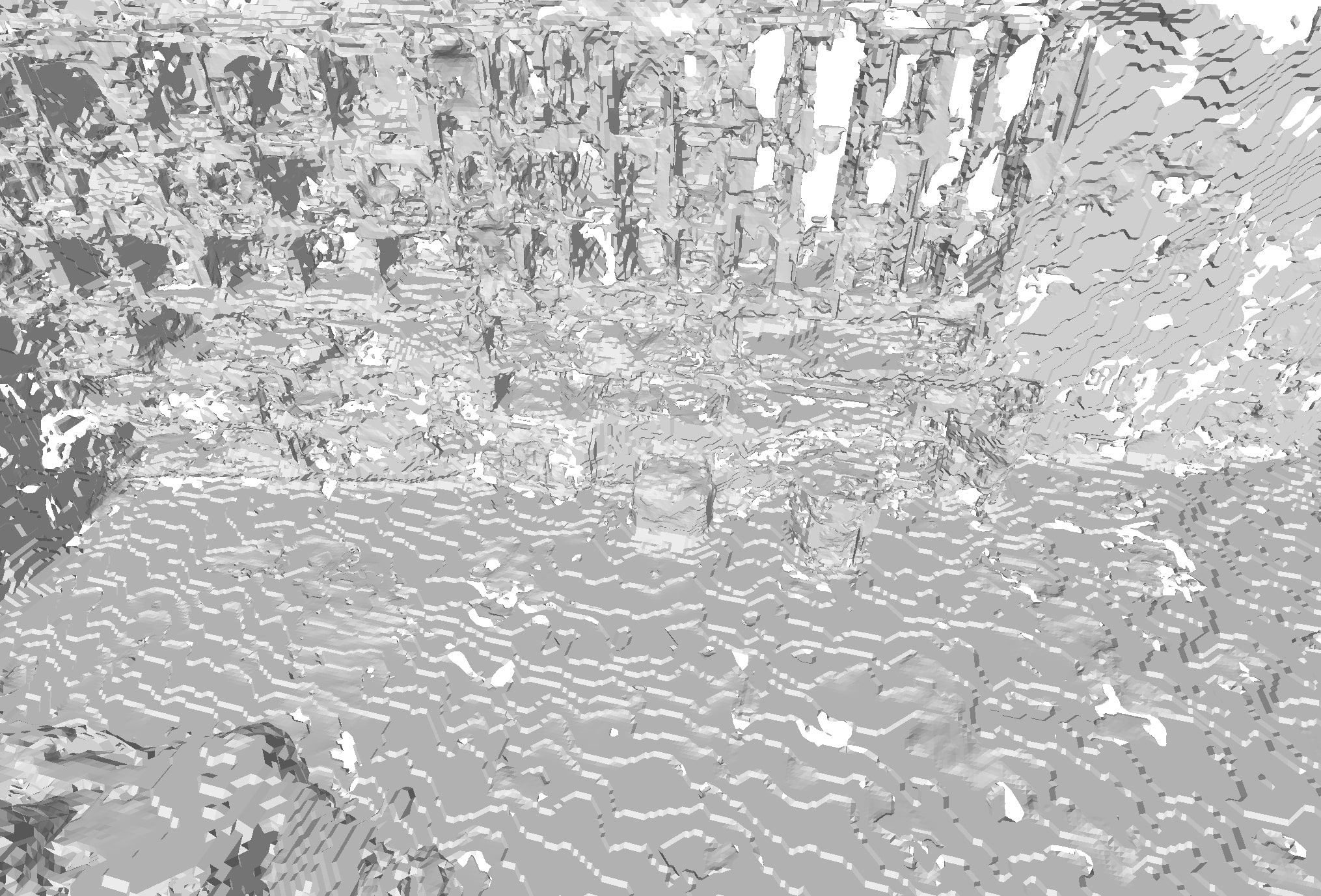} \\
		  \includegraphics[width=1\textwidth]{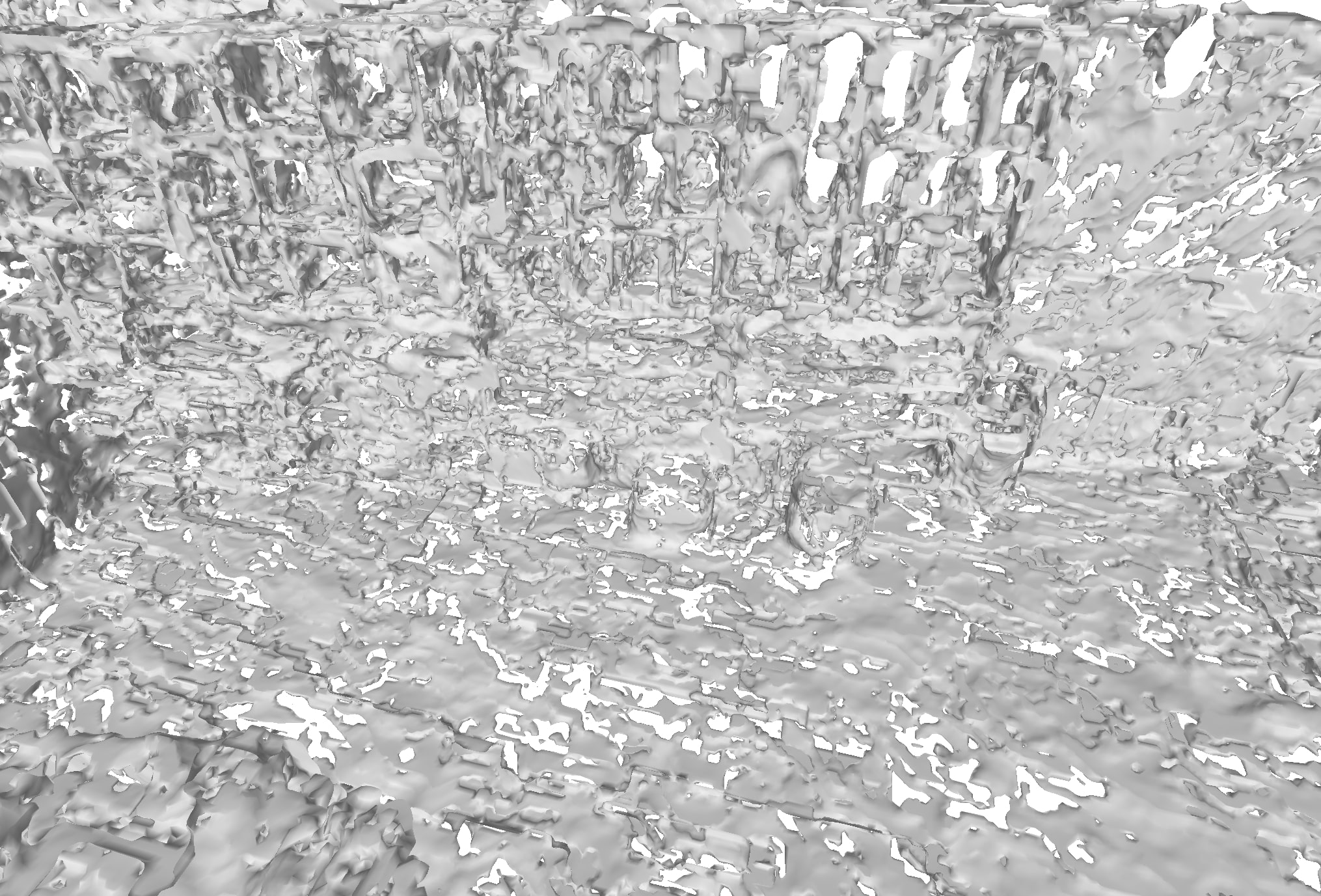} \\
    	\includegraphics[width=1\textwidth]{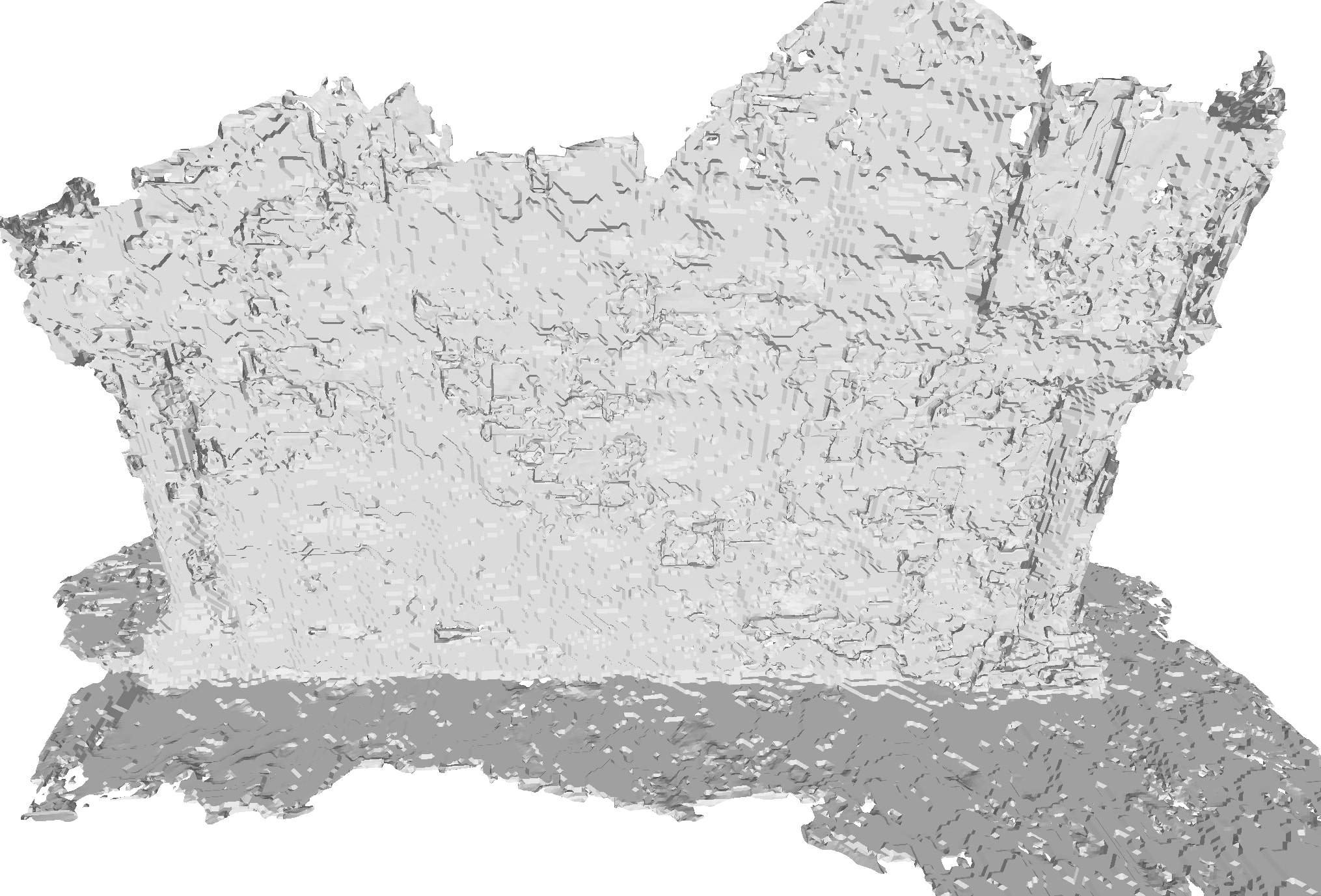} \\
		  \includegraphics[width=1\textwidth]{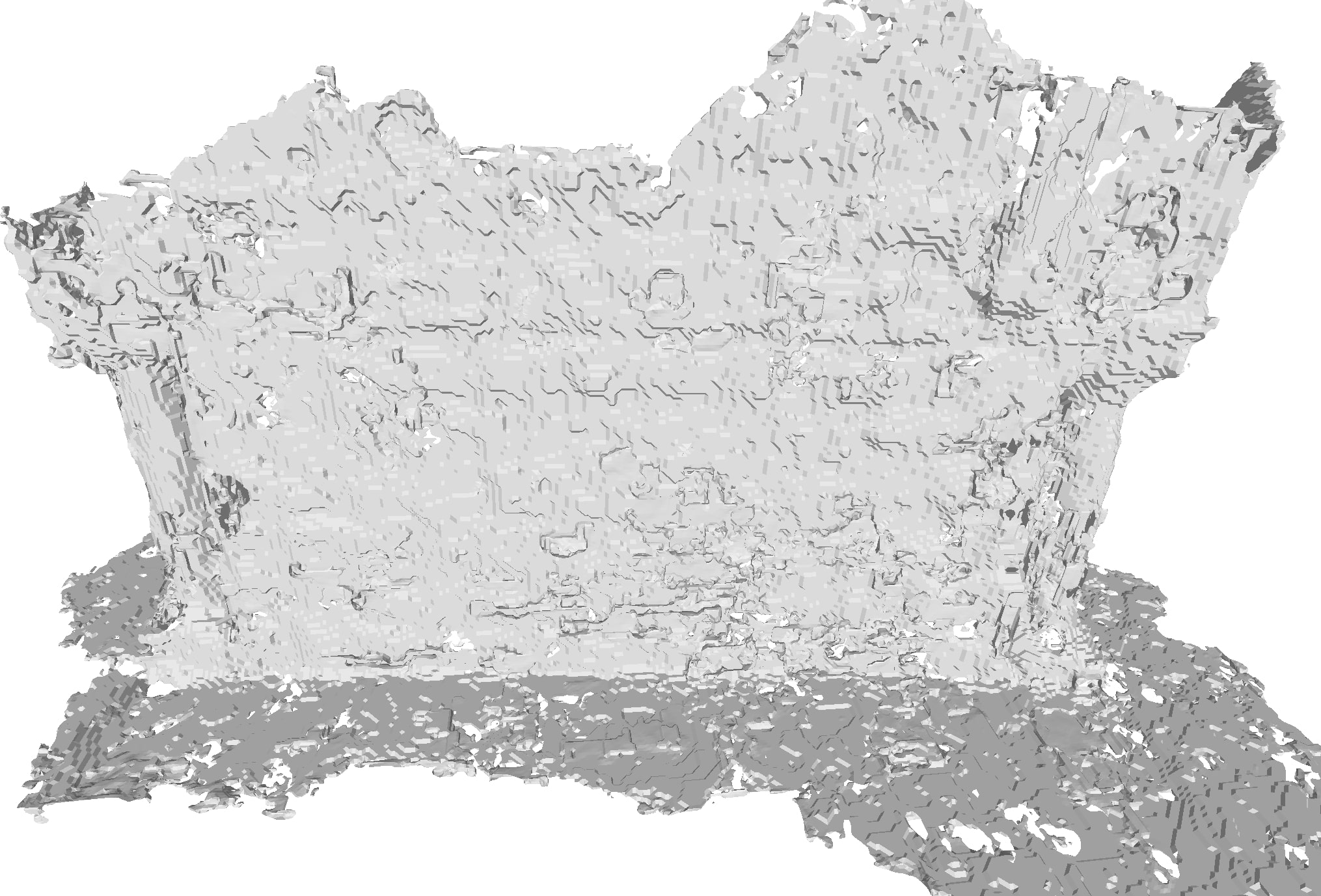} 
        \end{minipage}
    }
    \subfigure[DUDF]{
        \begin{minipage}[b]{0.23\textwidth}
		  \includegraphics[width=1\textwidth]{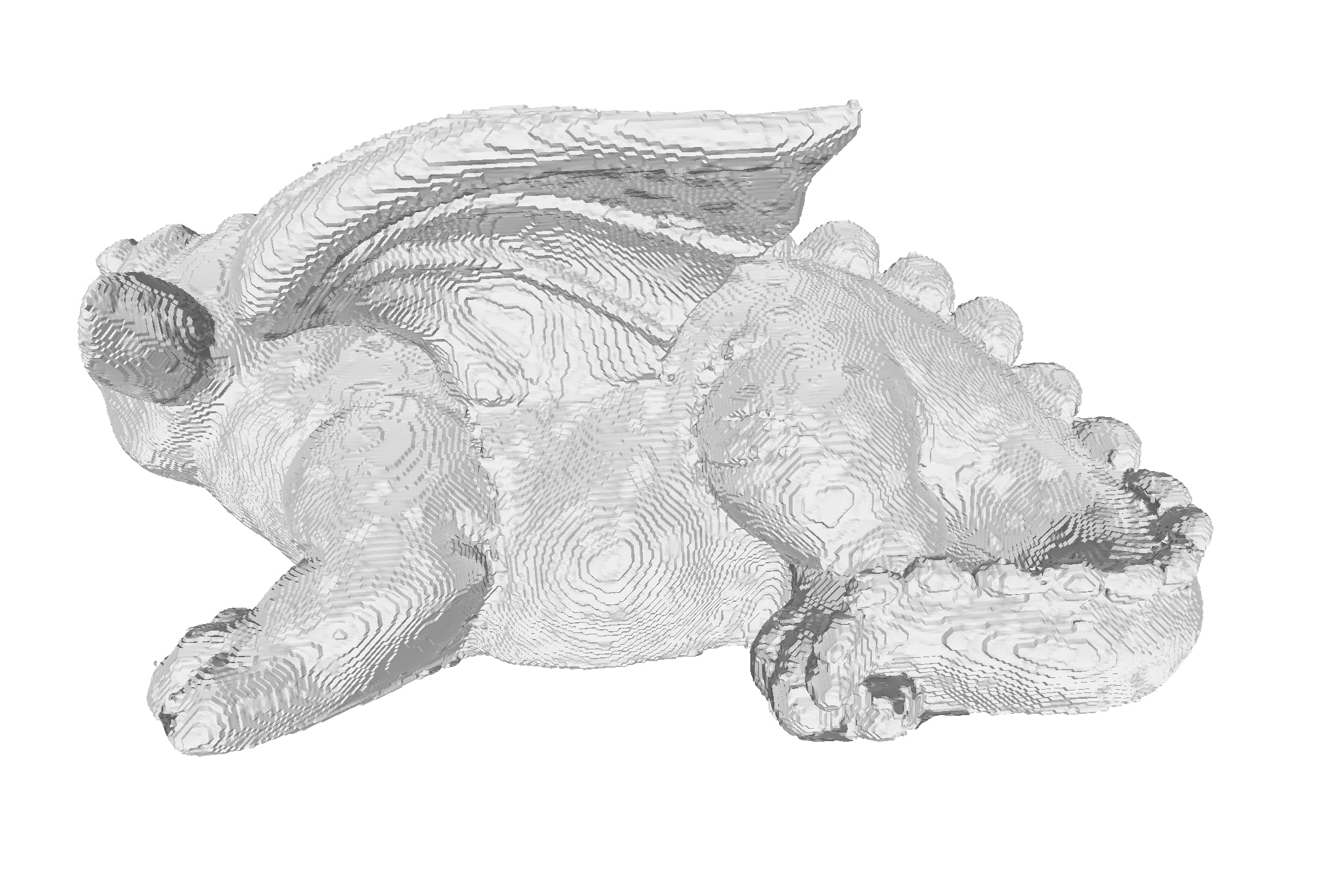} \\
		  \includegraphics[width=1\textwidth]{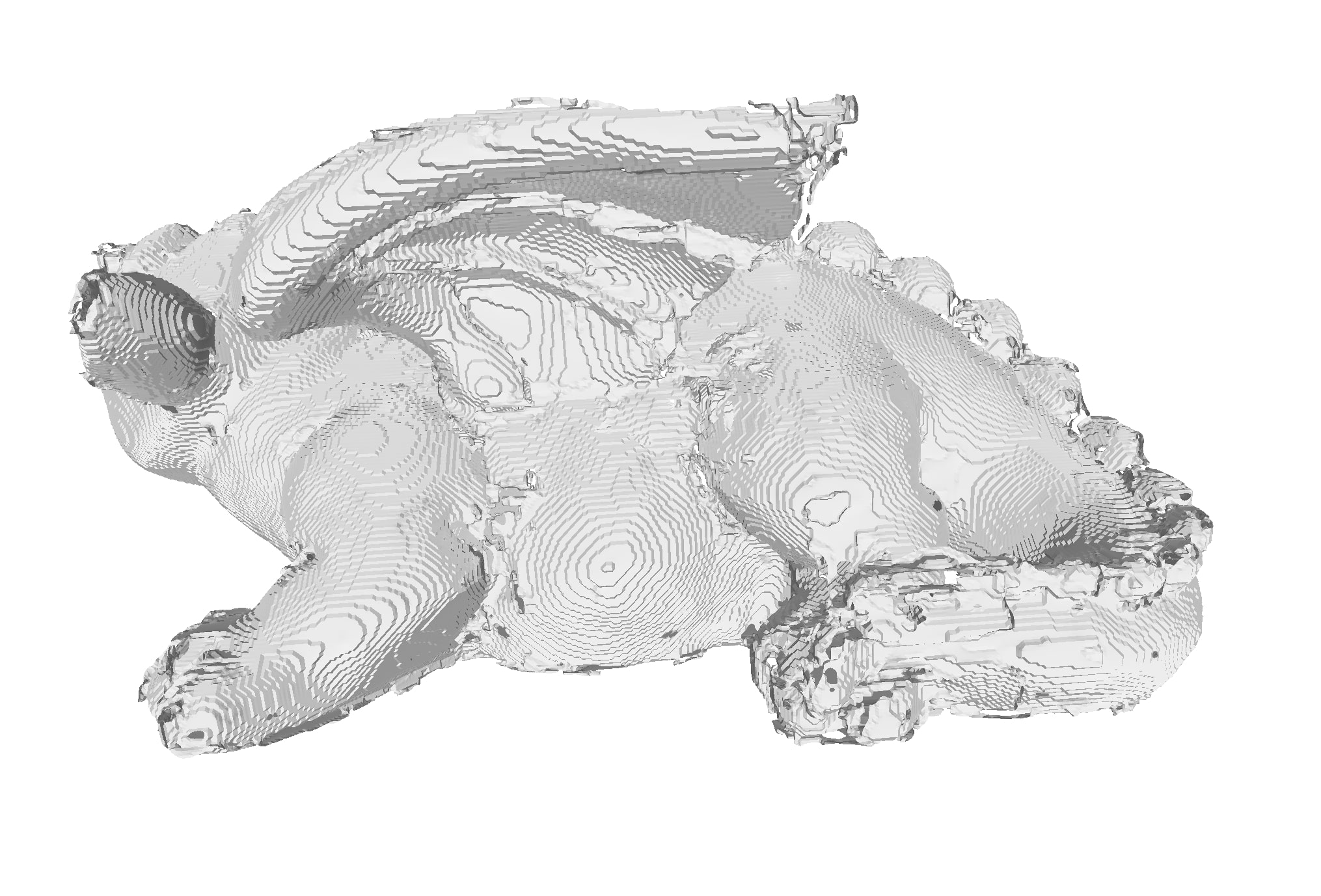} \\
            \includegraphics[width=1\textwidth]{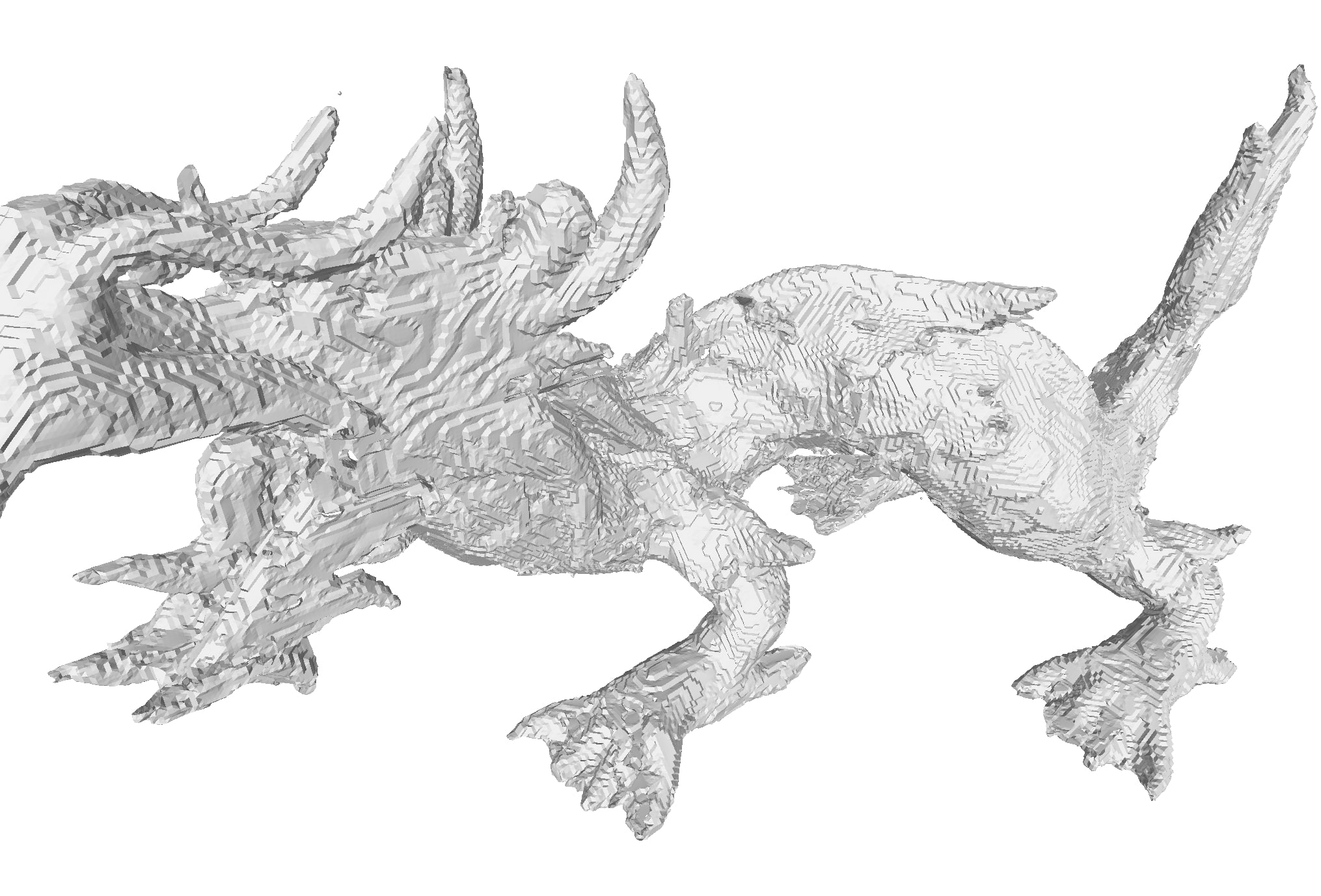} \\
		  \includegraphics[width=1\textwidth]{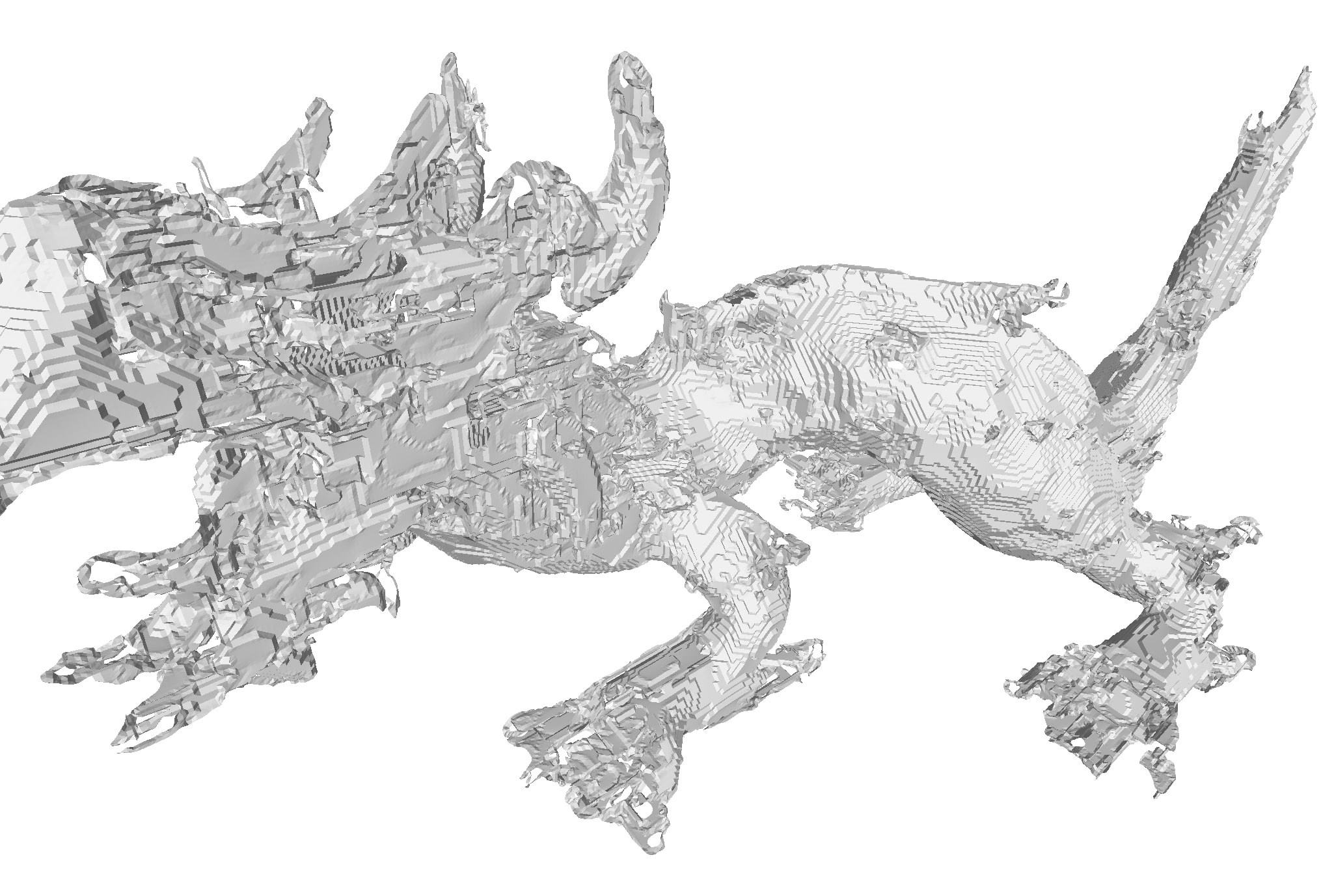} \\
            \includegraphics[width=1\textwidth]{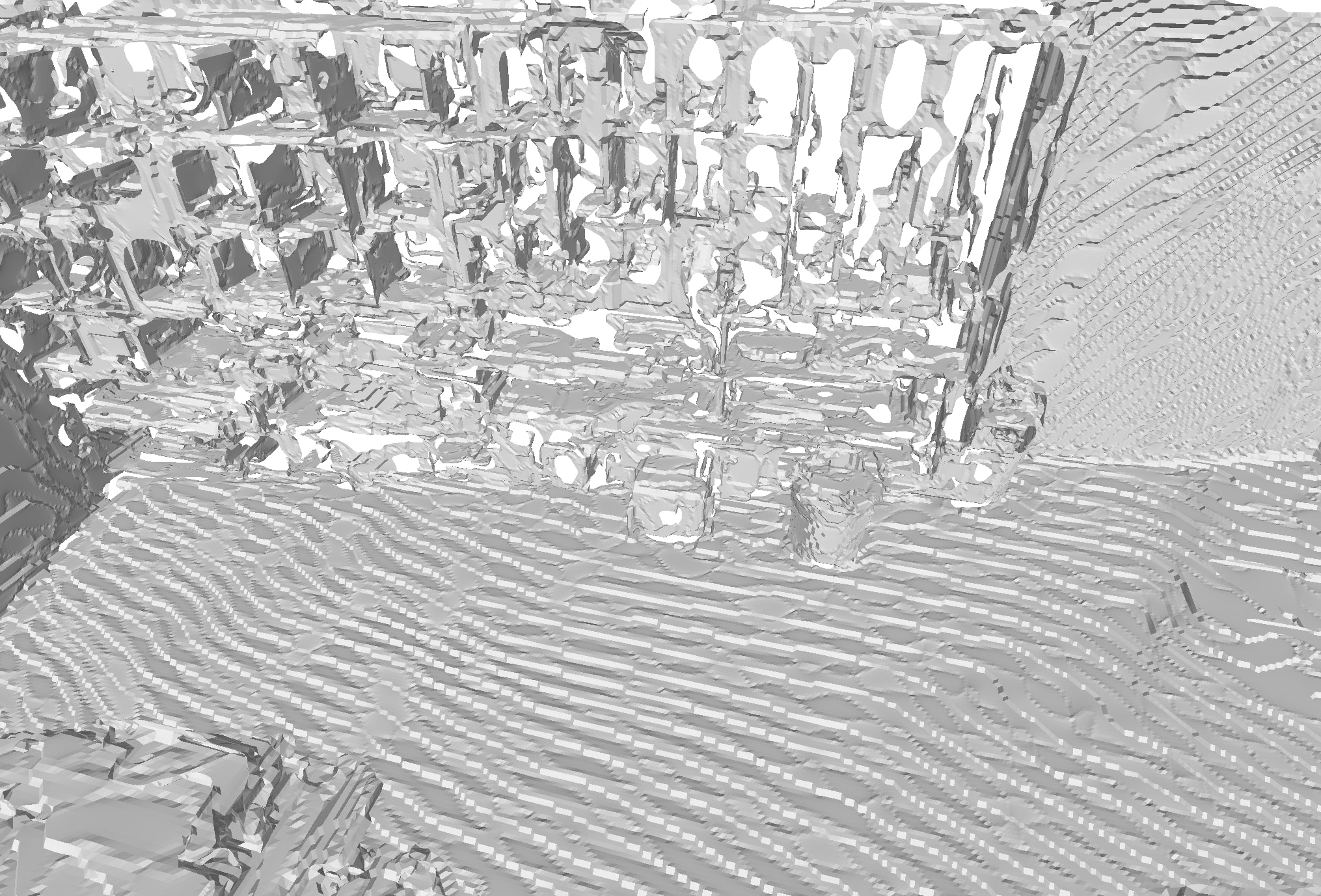} \\
		  \includegraphics[width=1\textwidth]{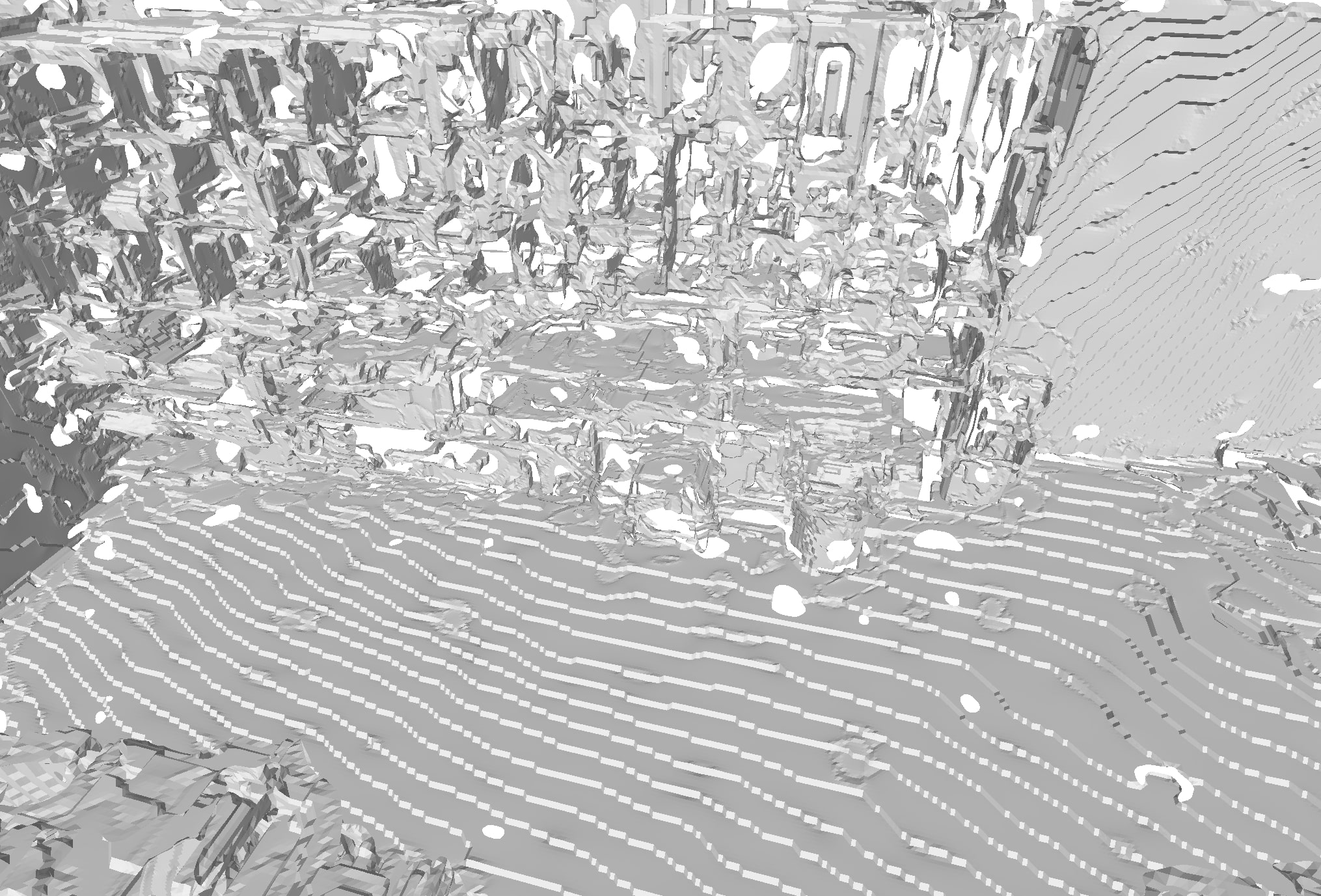} \\
    	\includegraphics[width=1\textwidth]{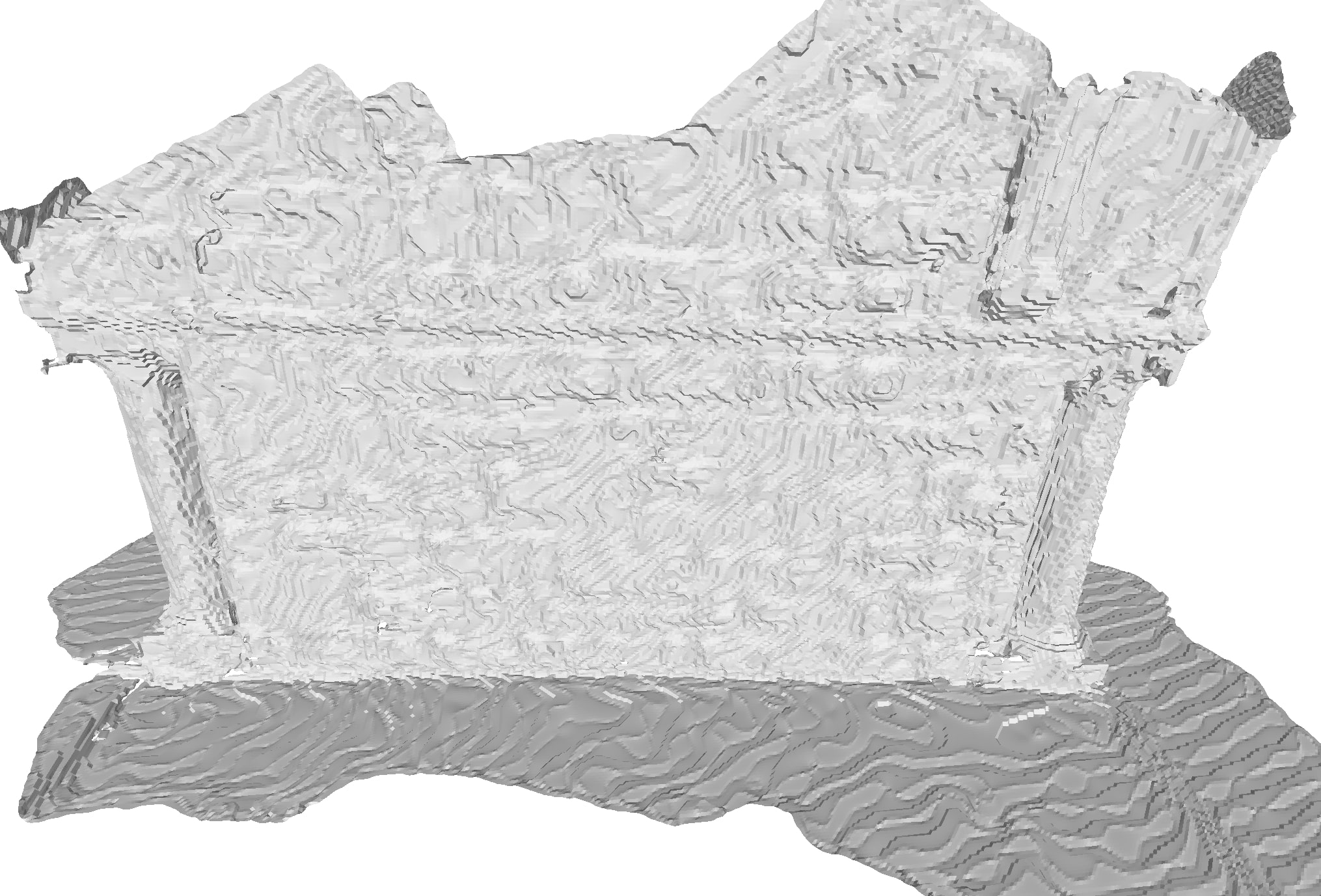} \\
		  \includegraphics[width=1\textwidth]{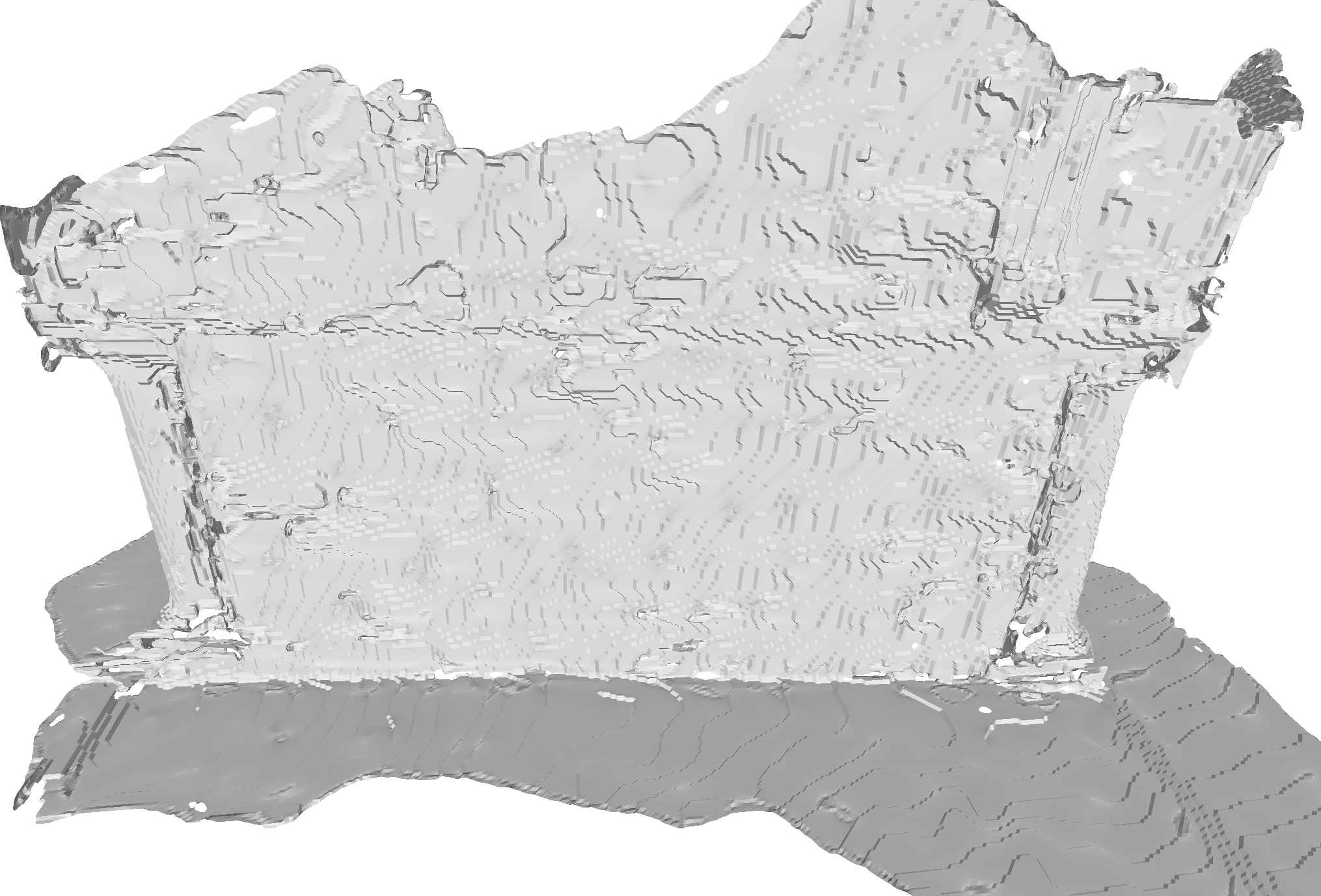} 
        \end{minipage}
    }
    \subfigure[ours-30Freq]{
        \begin{minipage}[b]{0.23\textwidth}
		  \includegraphics[width=1\textwidth]{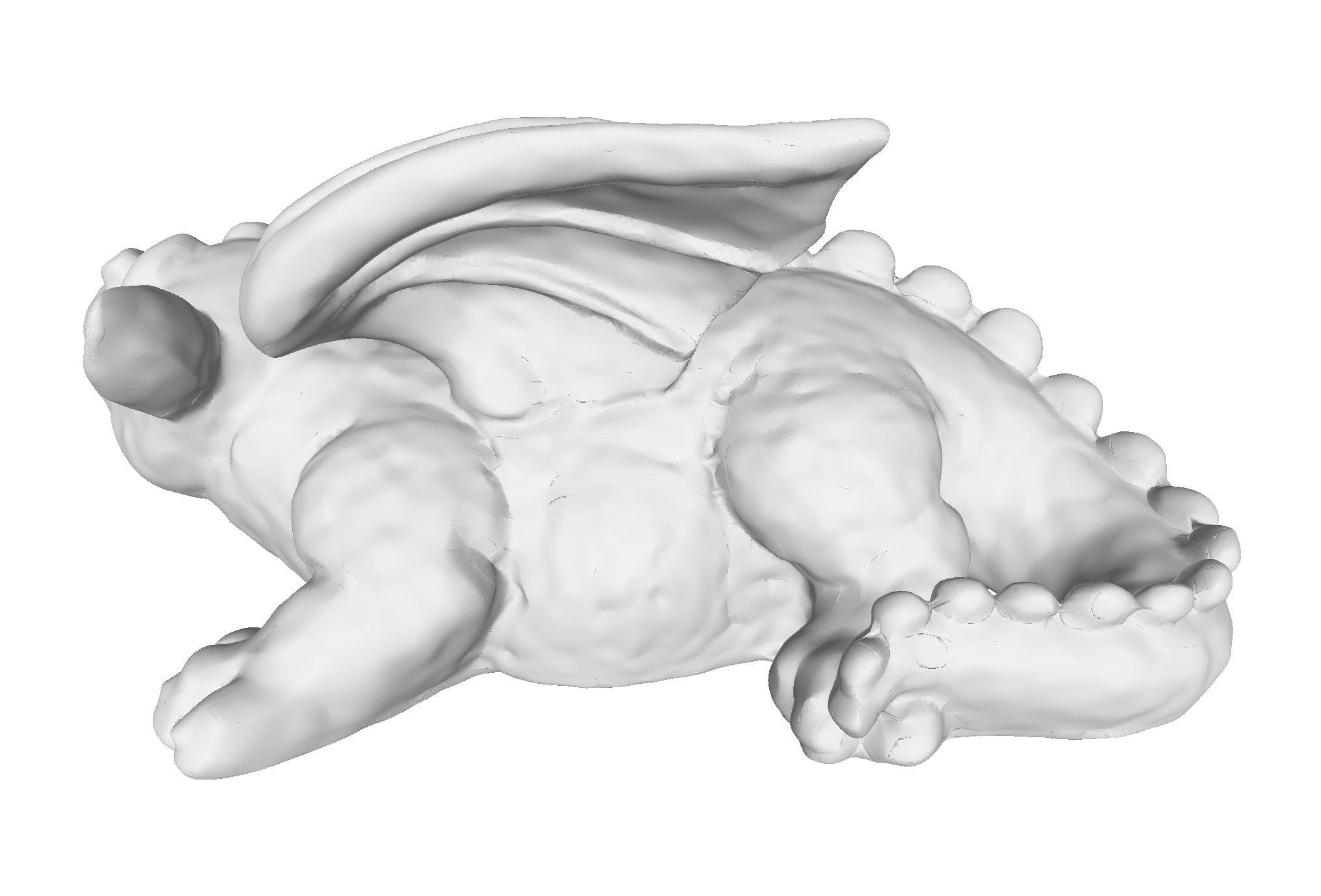} \\
		  \includegraphics[width=1\textwidth]{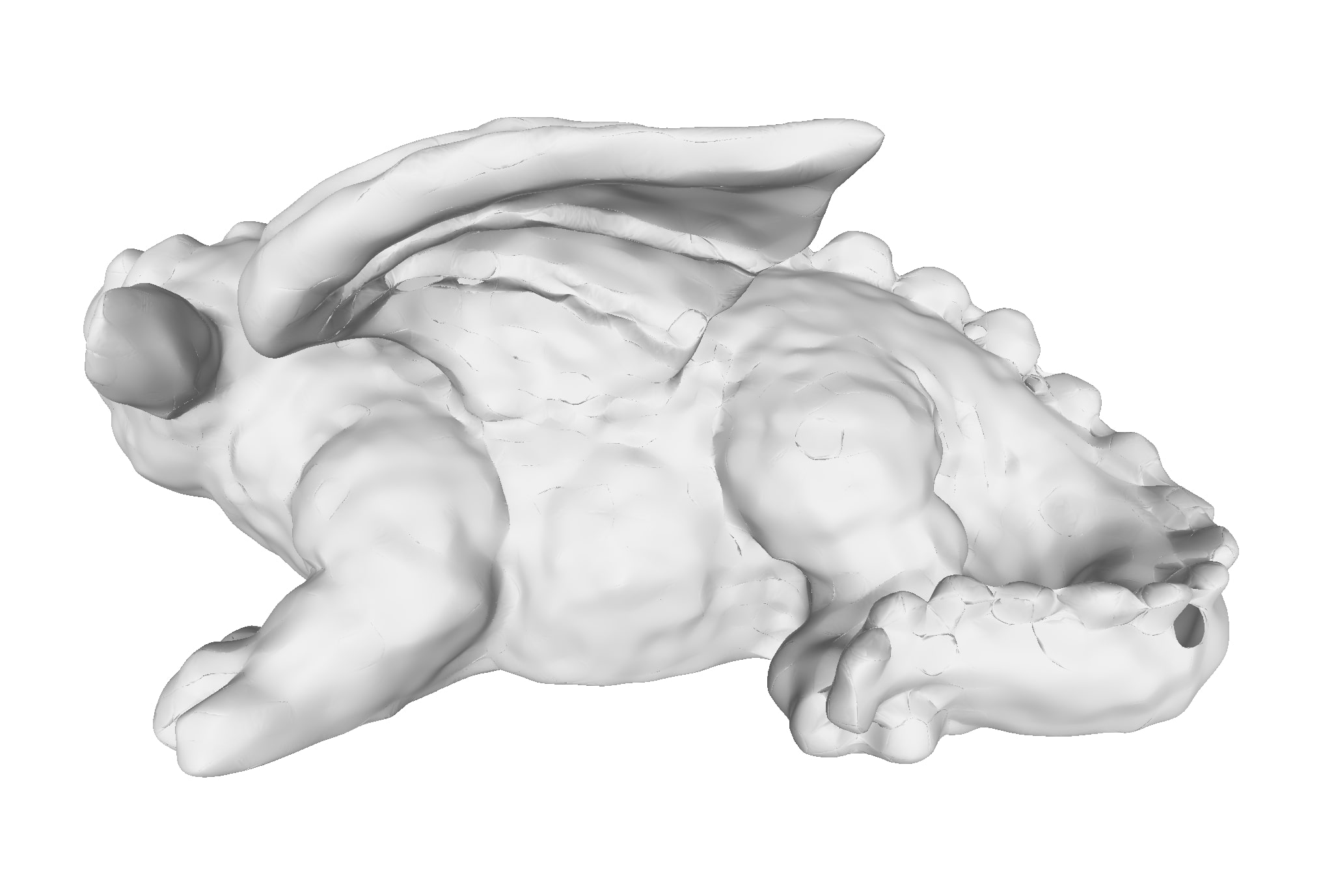} \\
            \includegraphics[width=1\textwidth]{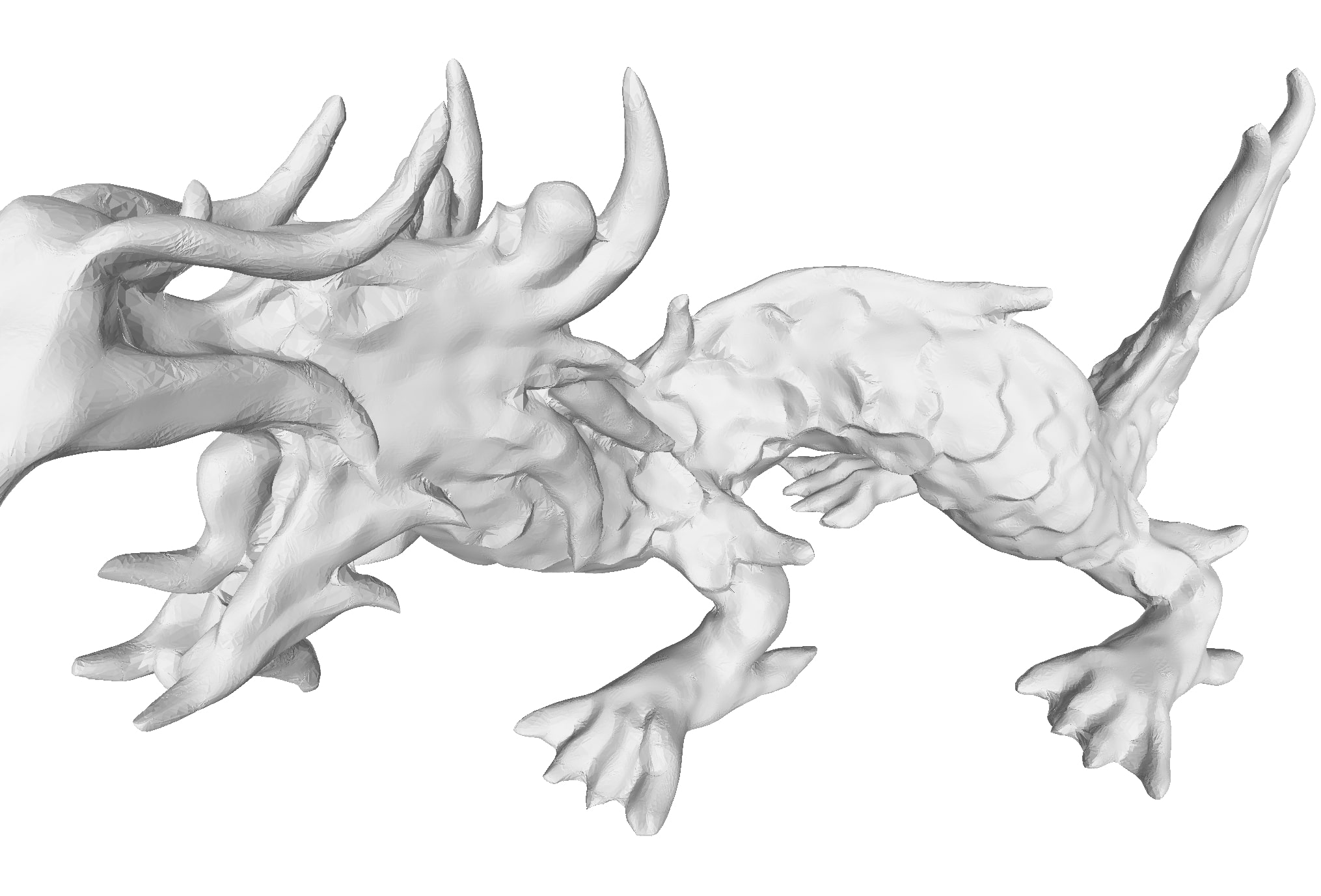} \\
		  \includegraphics[width=1\textwidth]{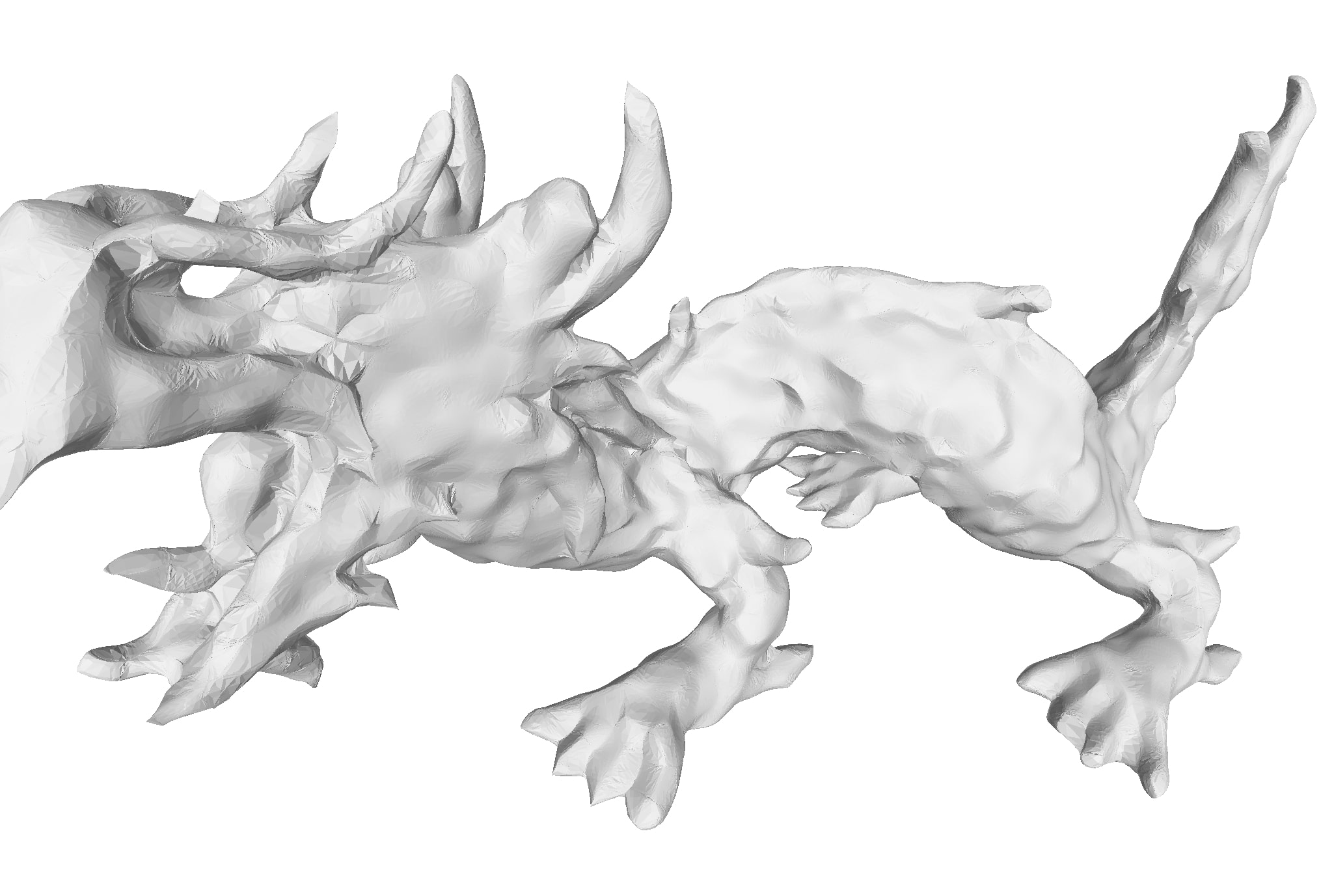} \\
            \includegraphics[width=1\textwidth]{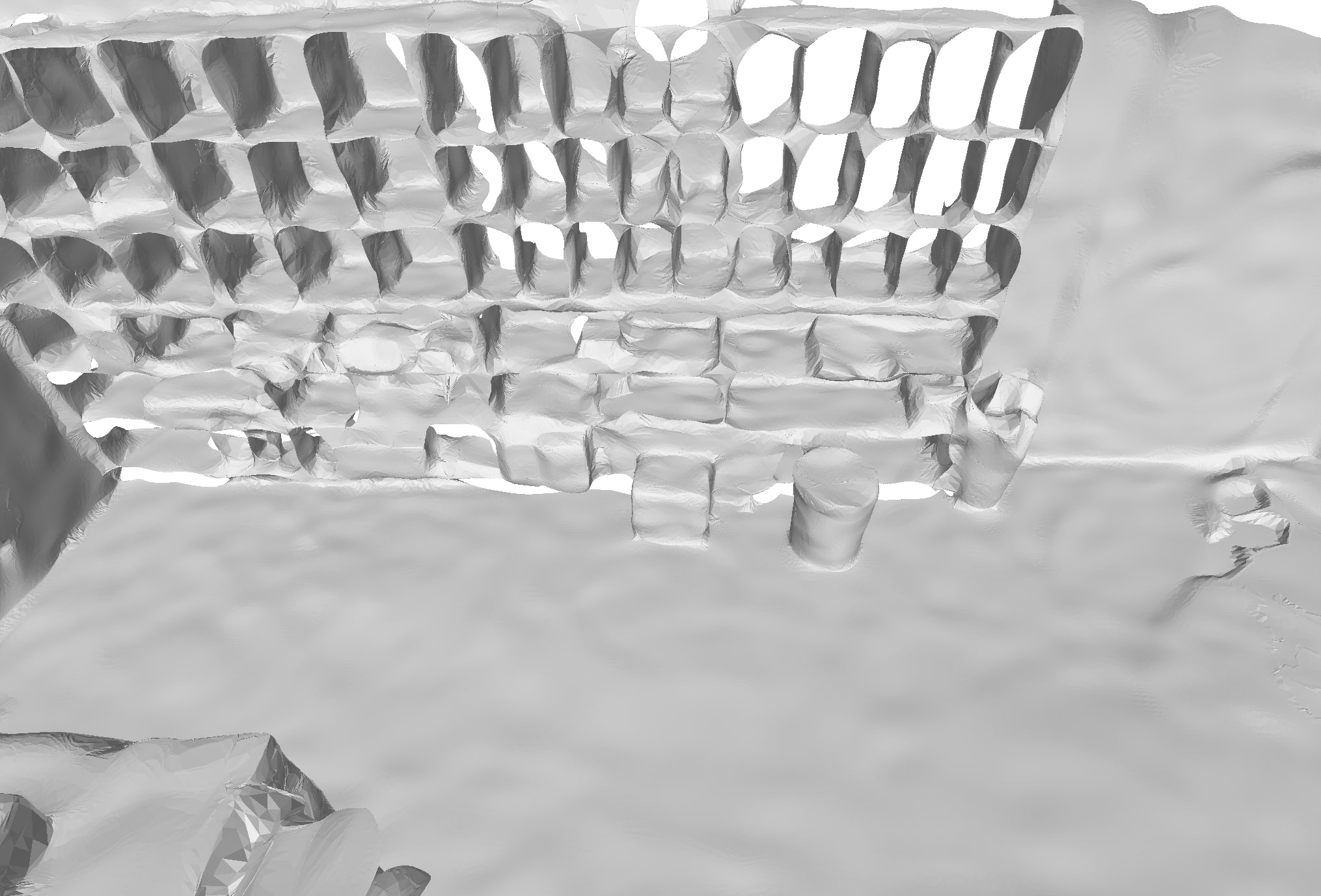} \\
		  \includegraphics[width=1\textwidth]{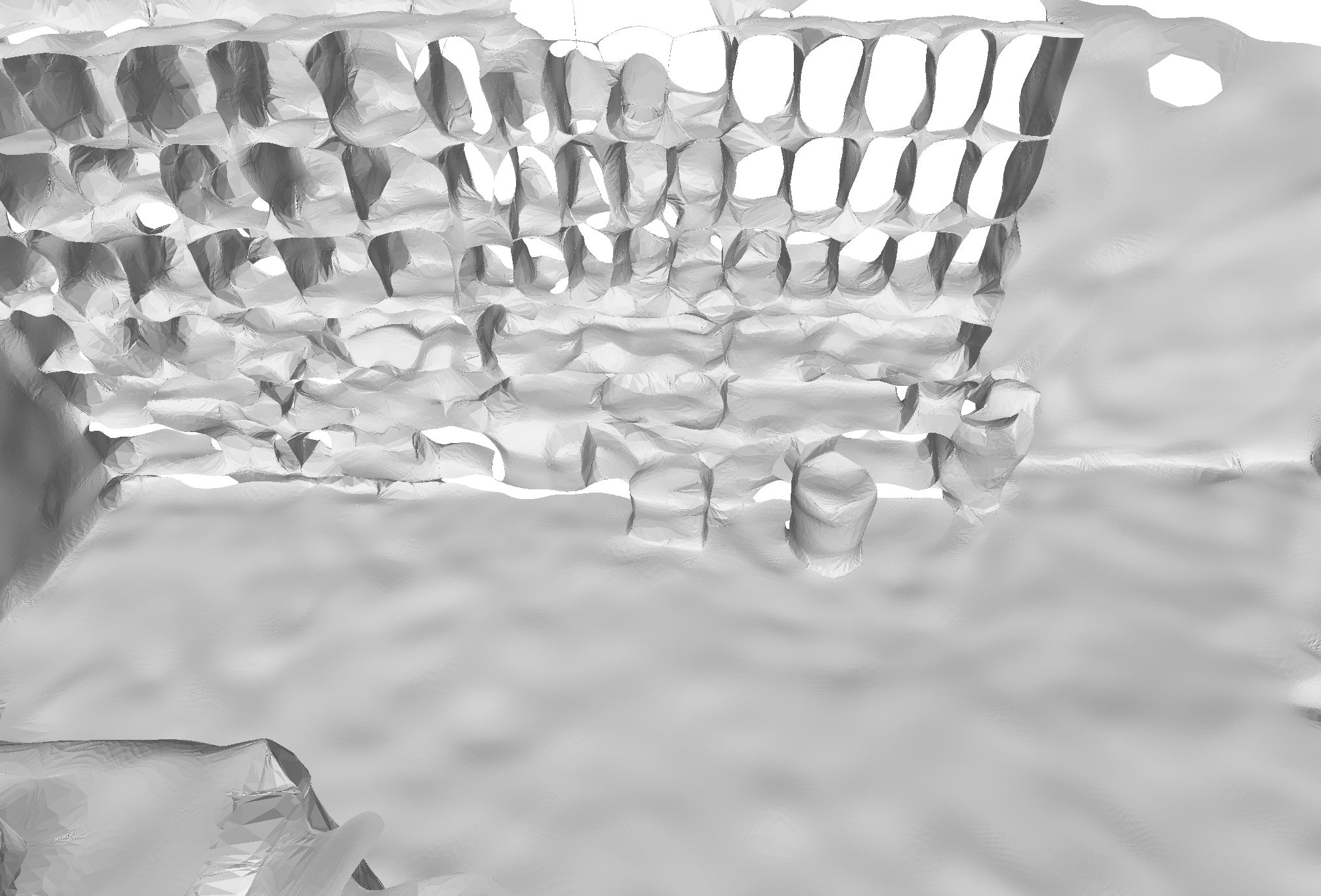} \\
    	\includegraphics[width=1\textwidth]{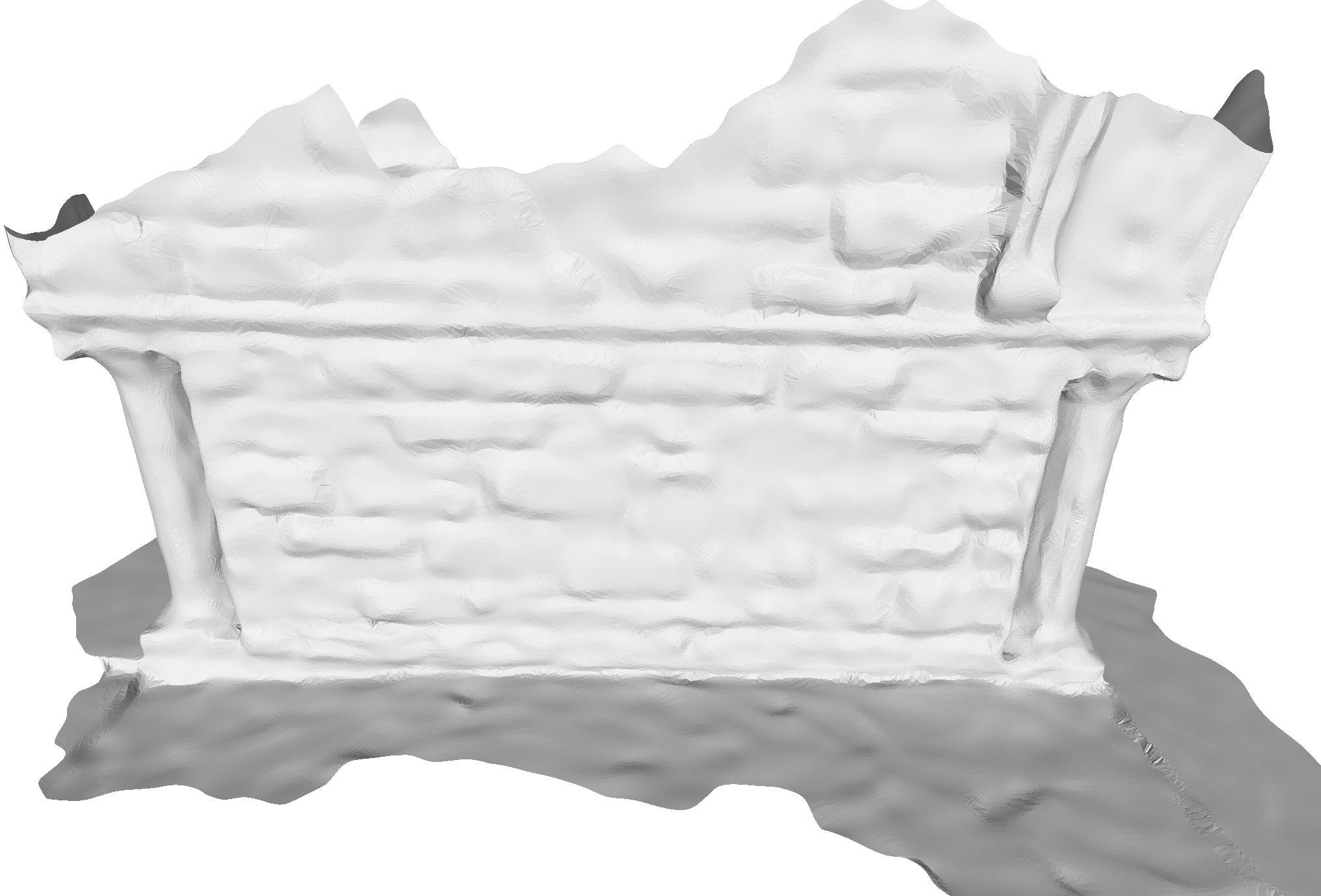} \\
		  \includegraphics[width=1\textwidth]{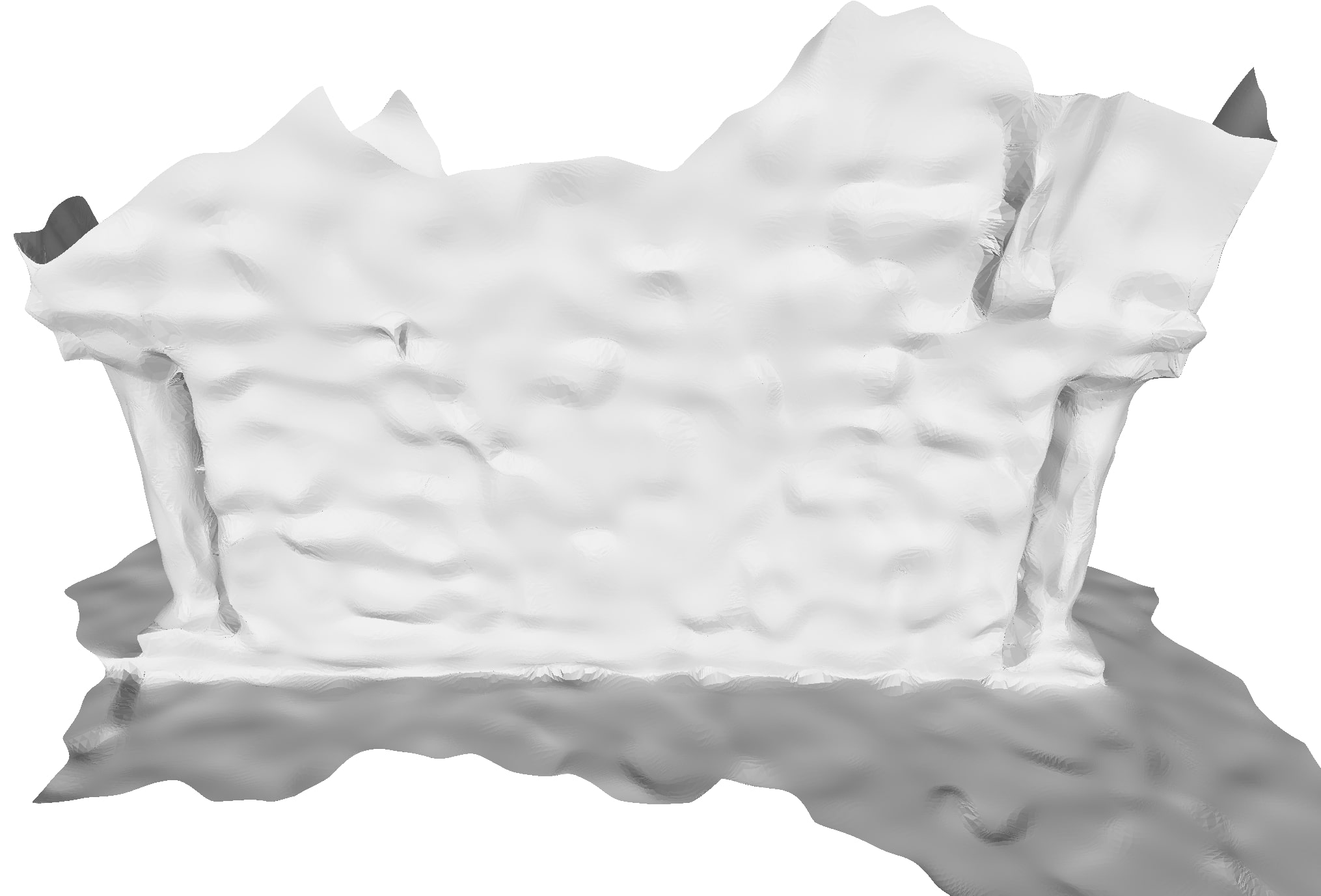} 
        \end{minipage}
    }
    \caption{Visual results for reconstruction on noisy point clouds. We separately add $N(0, 0.0025)$, $N(0, 0.005)$ Gaussian noise to input point cloud.
    Specially, we changed our method to SIREN with frequency 30 for better performance.
    }
    \label{fig:noise-comp}
\end{figure*}

\begin{figure*}[!htbp]
    \centering
    \subfigure[CAP-UDF]{
        \begin{minipage}[b]{0.18\textwidth}
		  \includegraphics[width=1\textwidth]{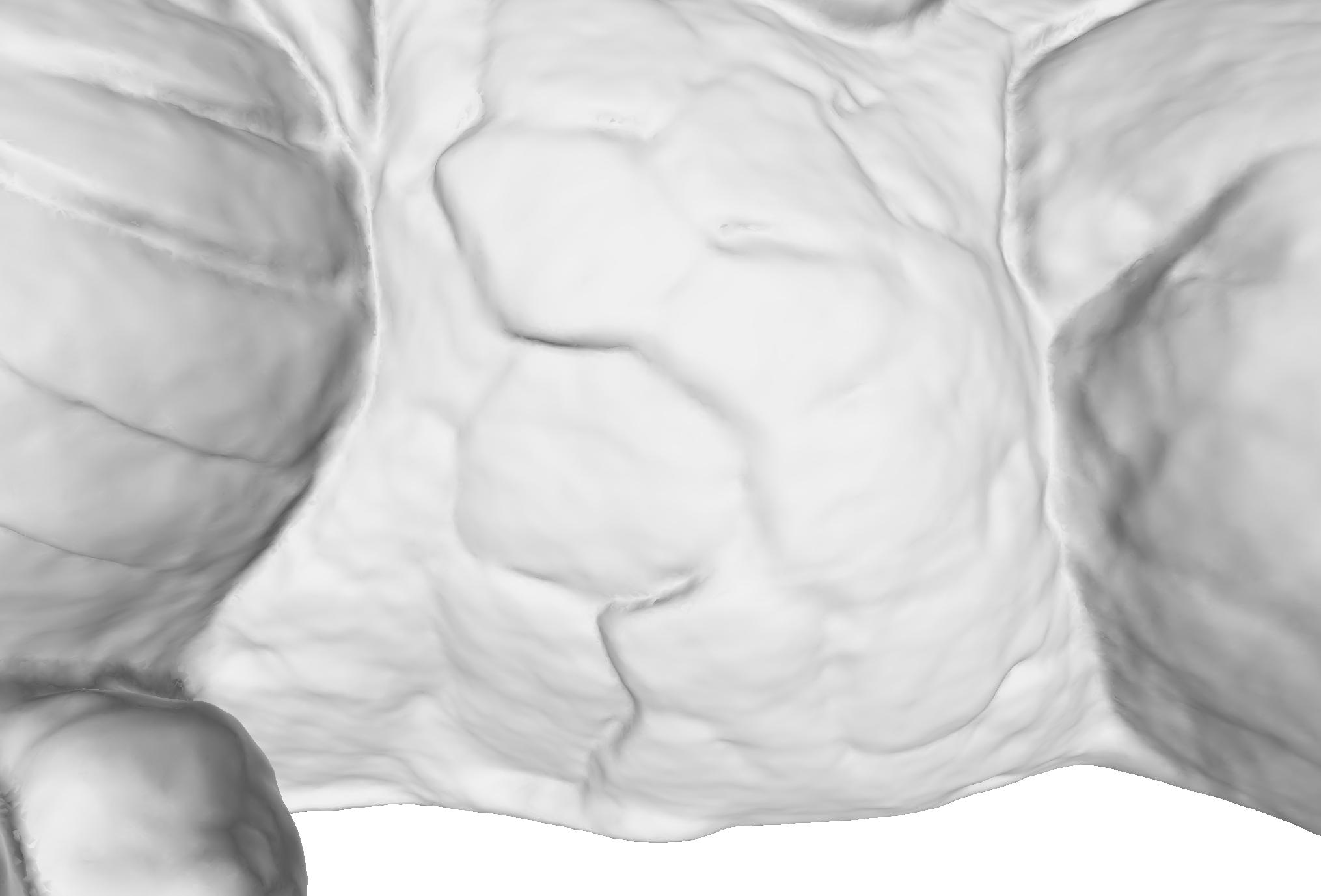} \\
		  \includegraphics[width=1\textwidth]{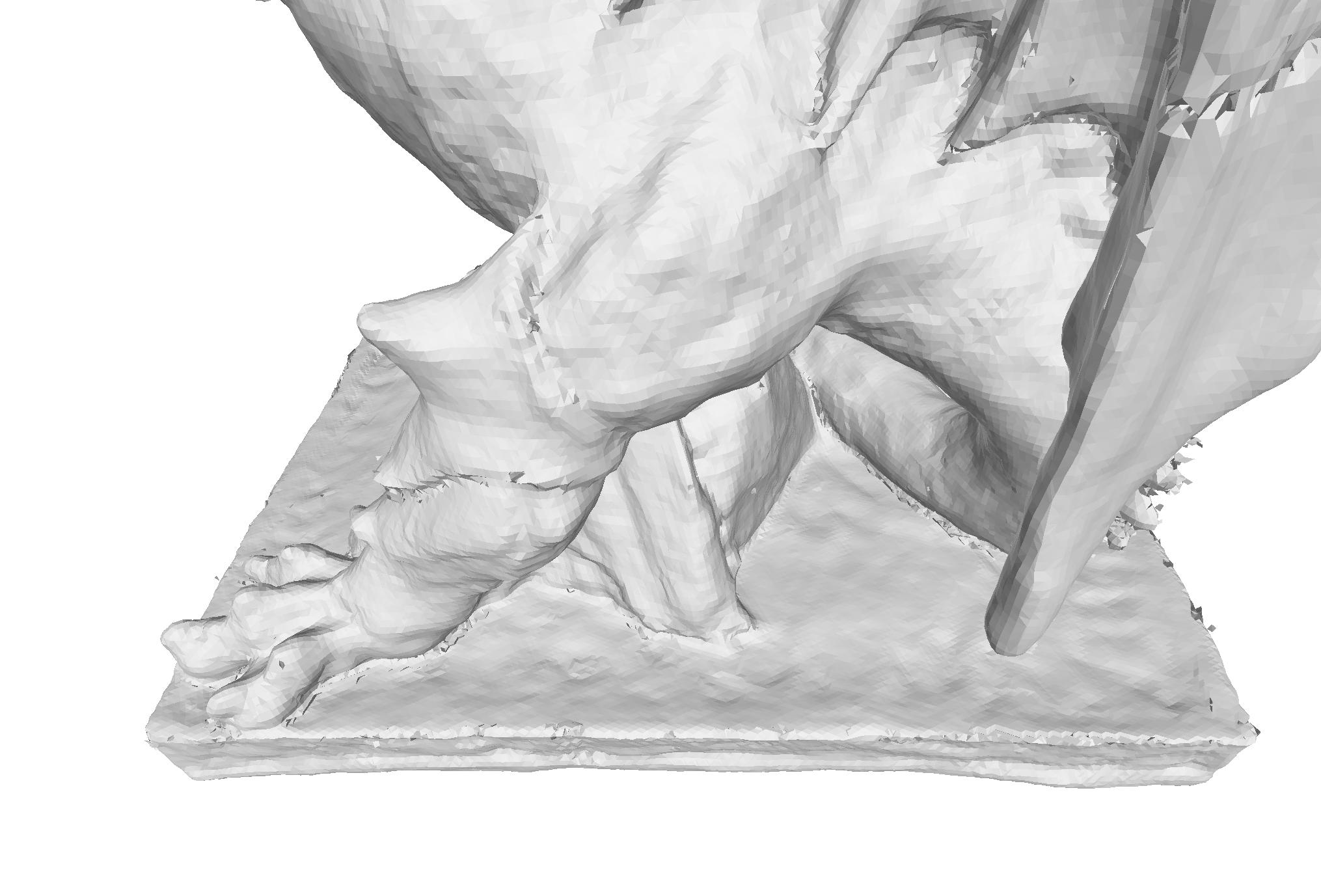} \\
		  \includegraphics[width=1\textwidth]{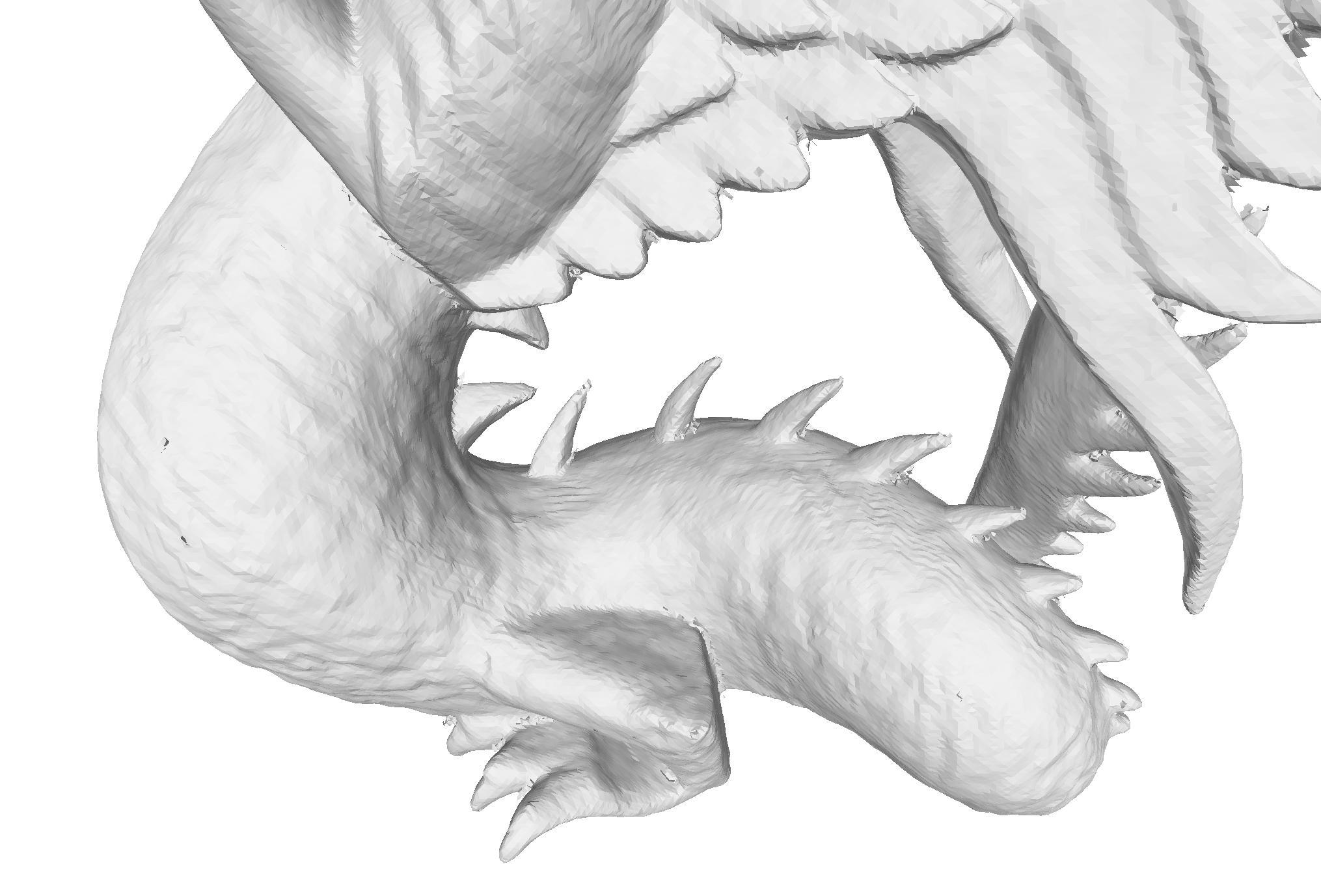} \\
		  \includegraphics[width=1\textwidth]{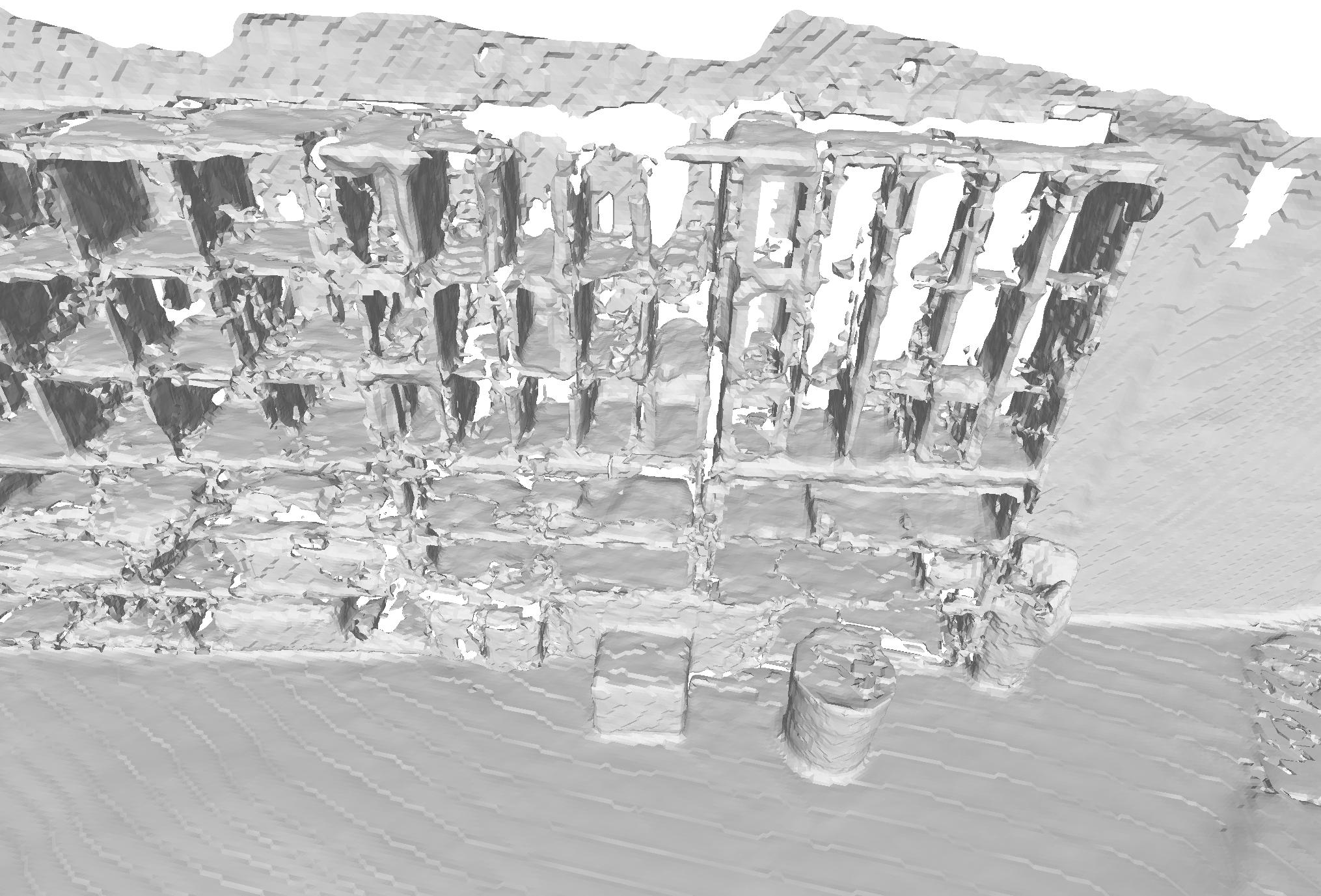} \\
 		  \includegraphics[width=1\textwidth]{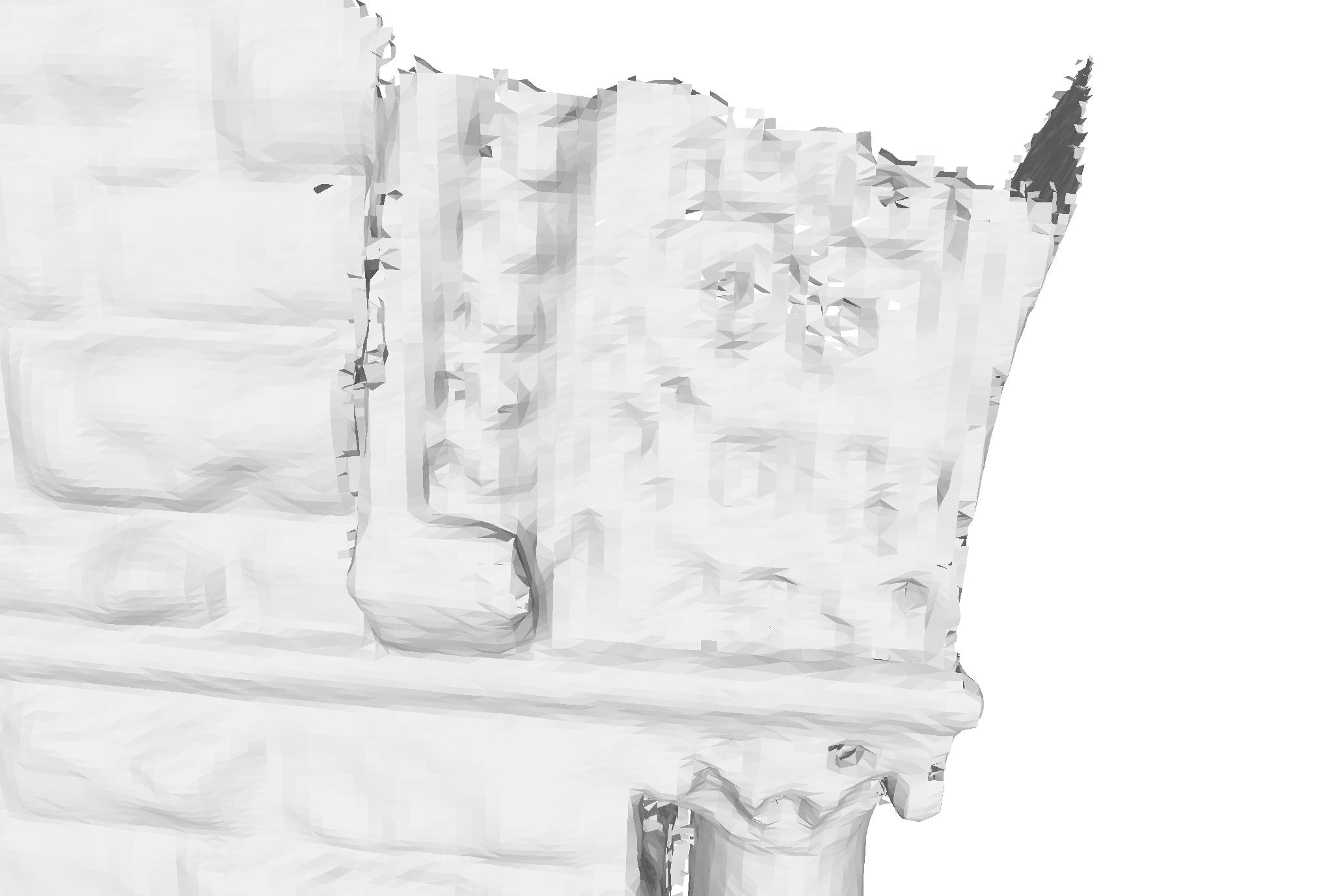} \\
 		  \includegraphics[width=1\textwidth]{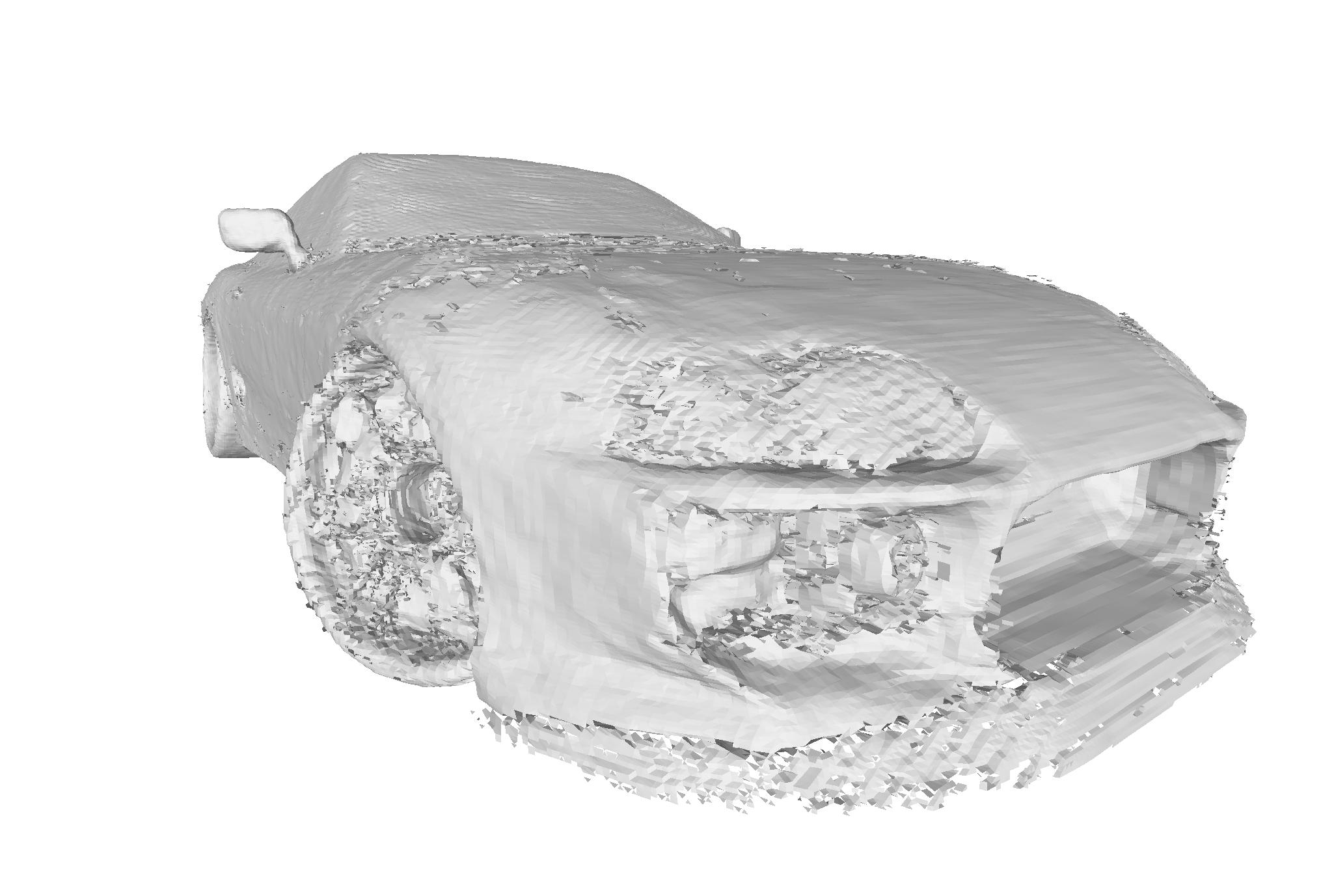} \\
 		  \includegraphics[width=1\textwidth]{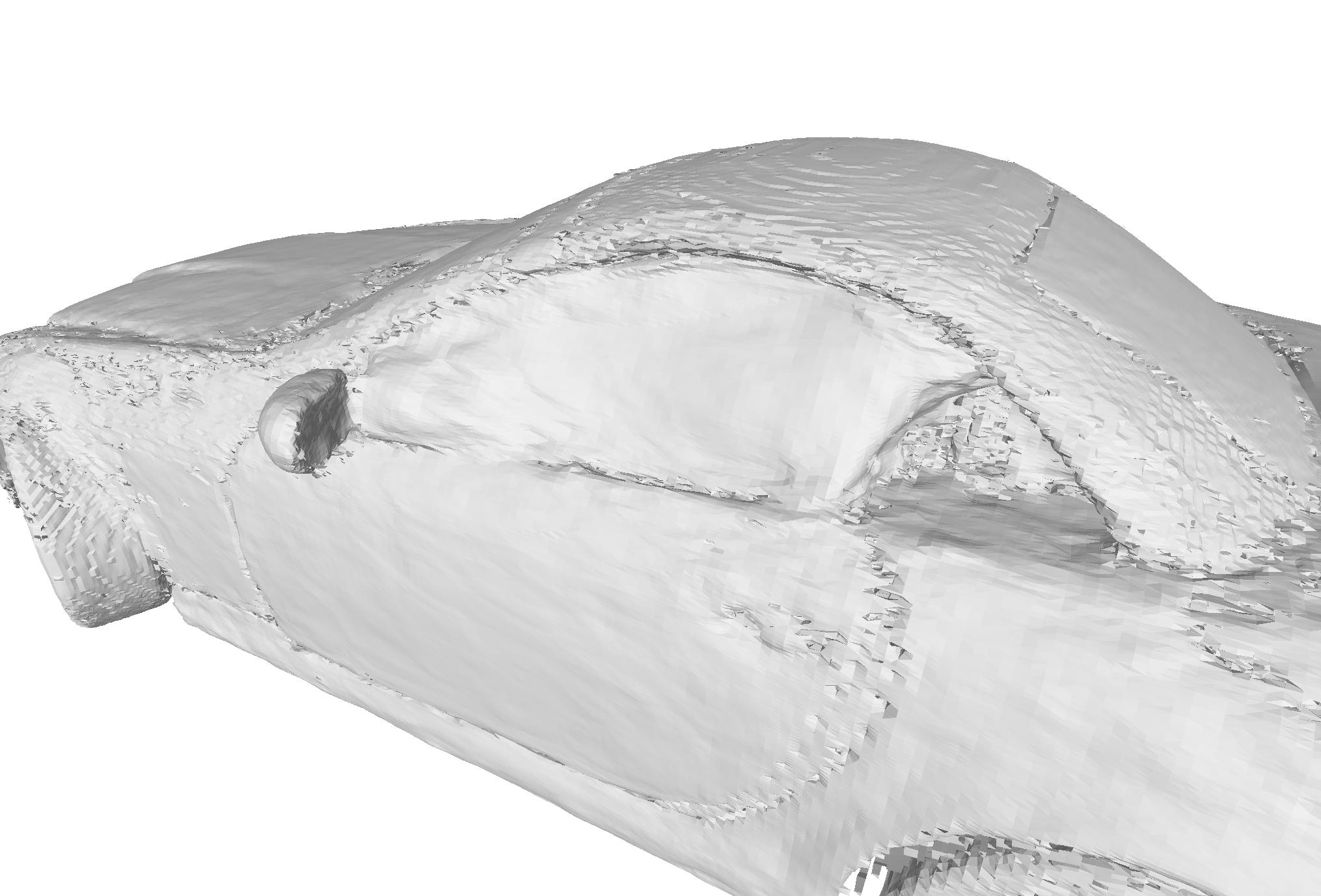} \\
 		  \includegraphics[width=1\textwidth]{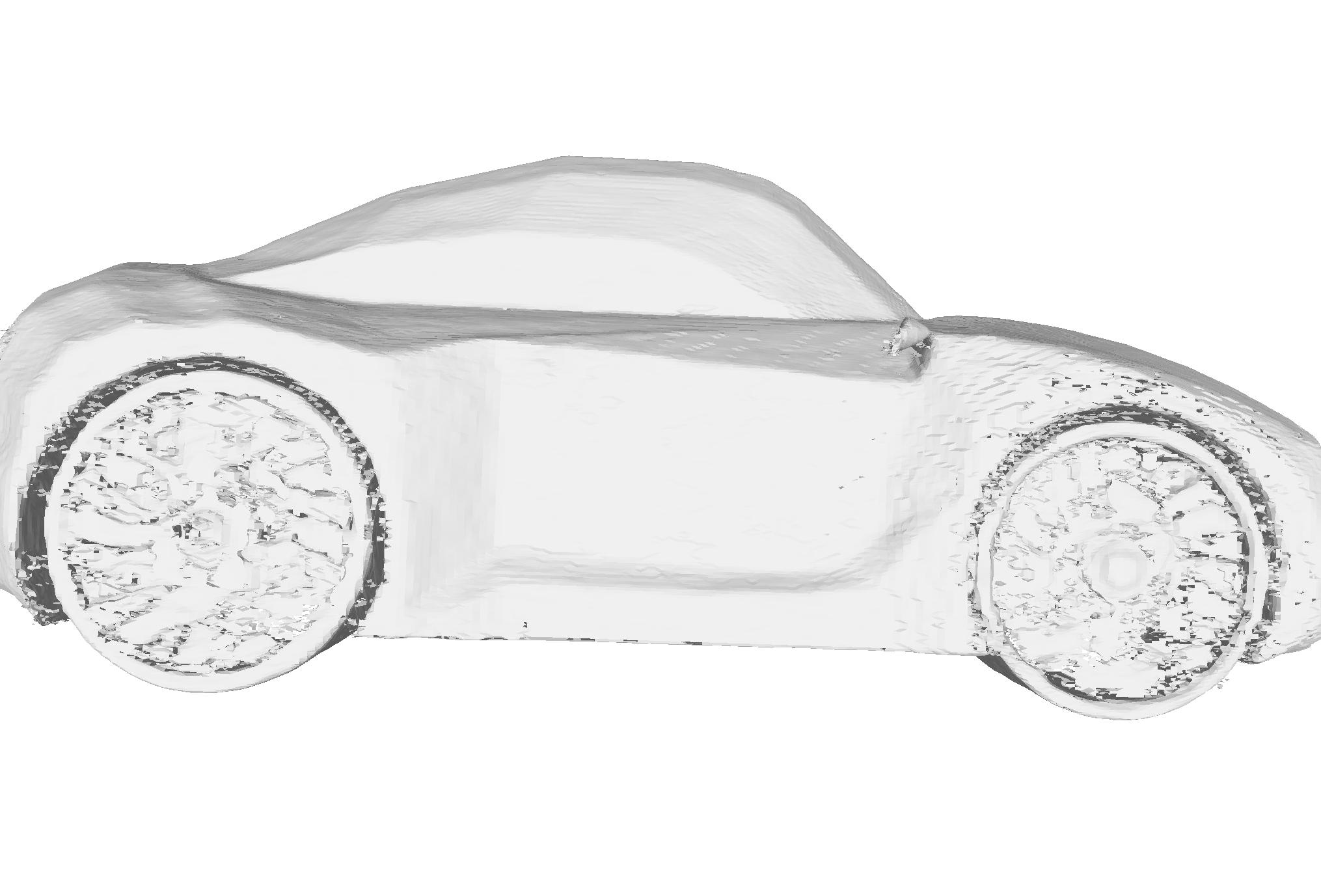} \\
 		  \includegraphics[width=1\textwidth]{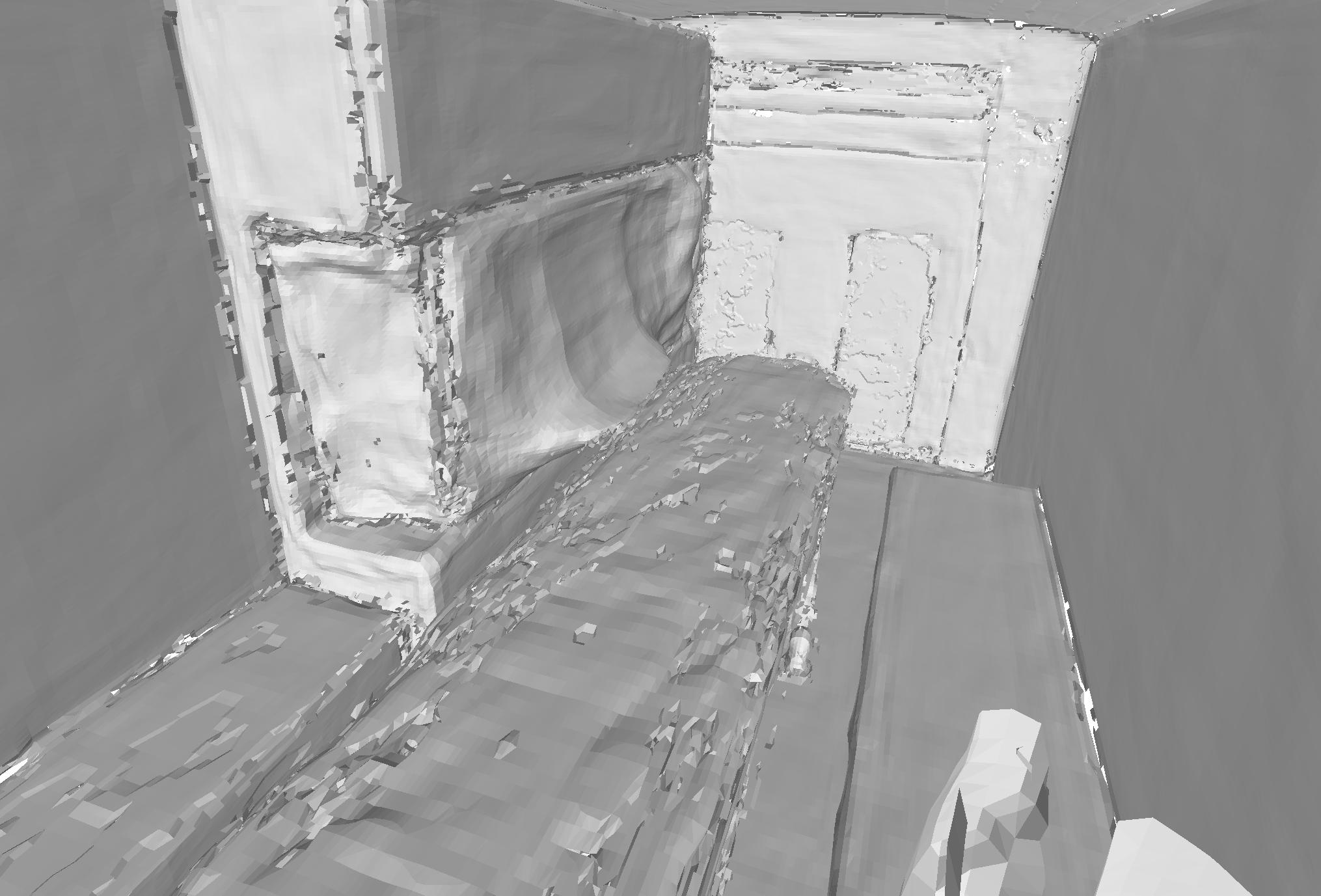} \\
        \end{minipage}
    }
    \subfigure[DUDF]{
        \begin{minipage}[b]{0.18\textwidth}
		  \includegraphics[width=1\textwidth]{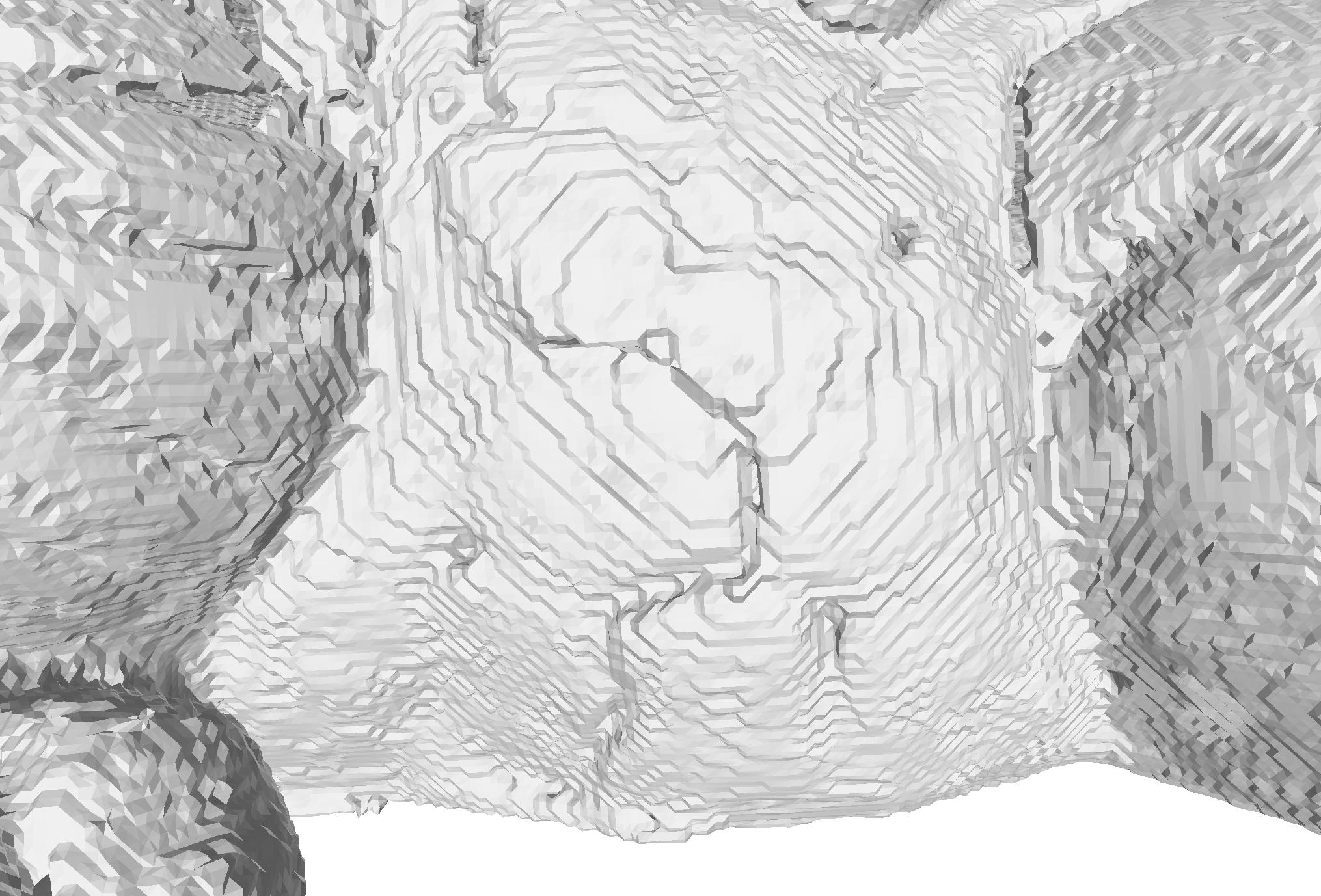} \\
		  \includegraphics[width=1\textwidth]{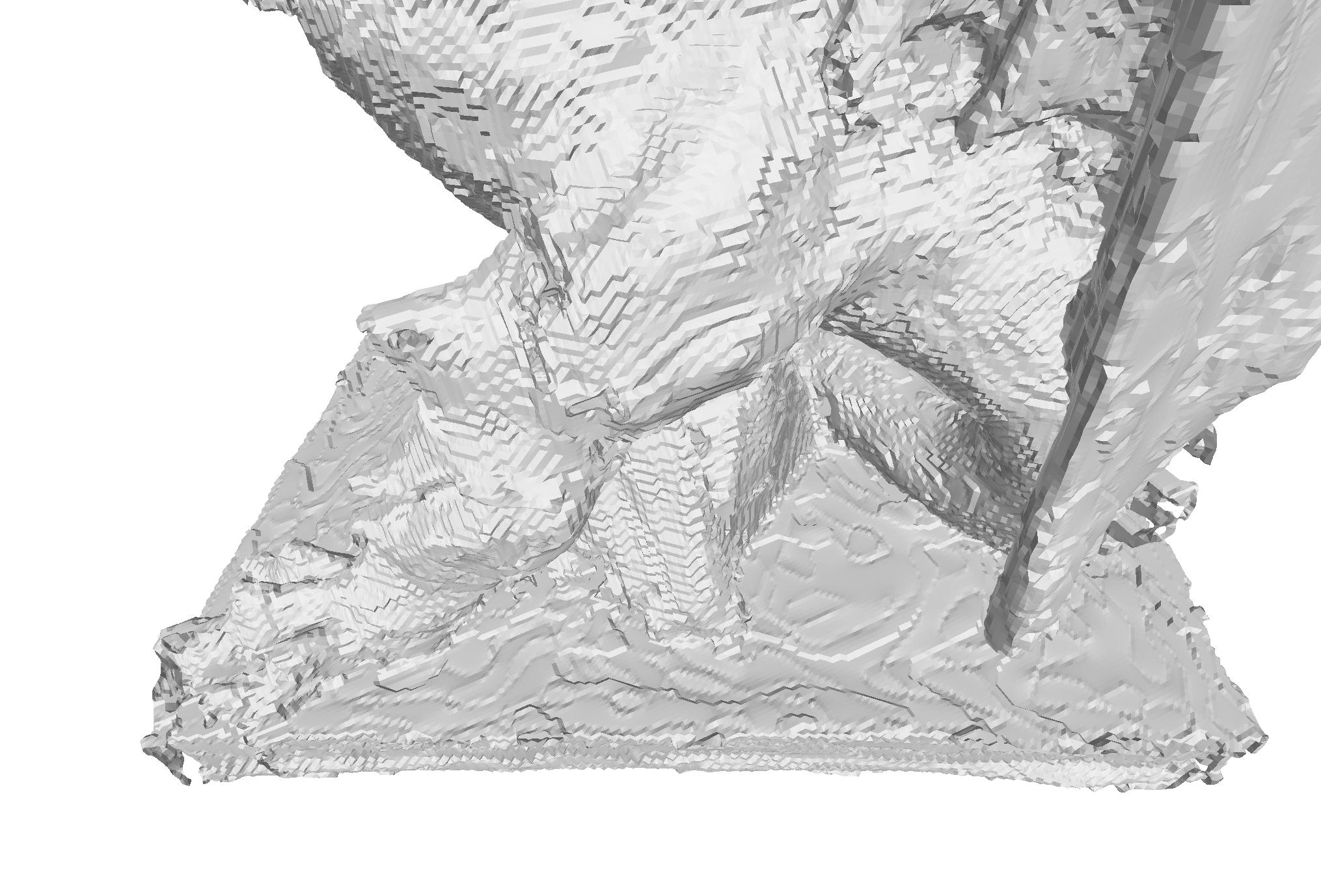} \\
		  \includegraphics[width=1\textwidth]{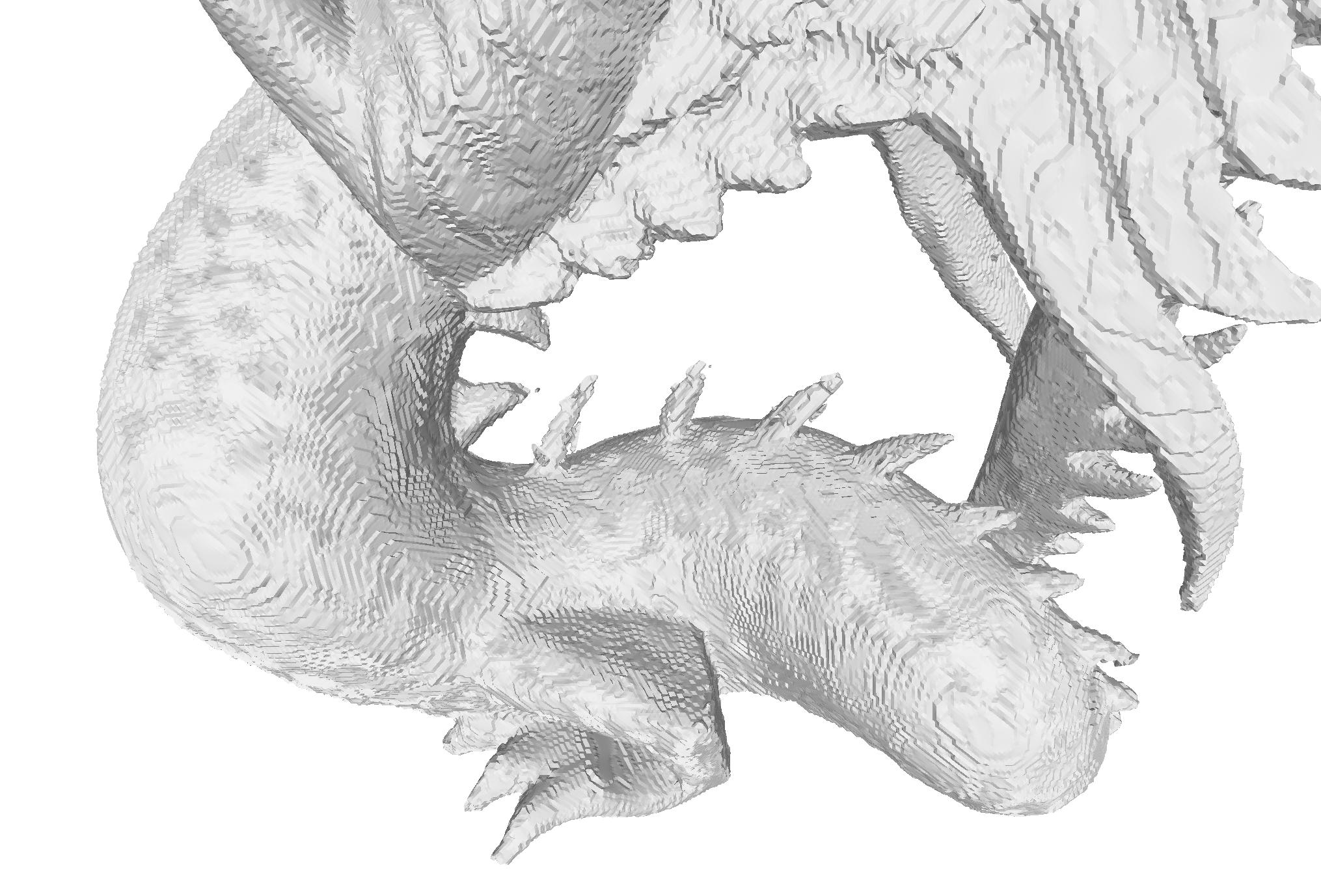} \\
		  \includegraphics[width=1\textwidth]{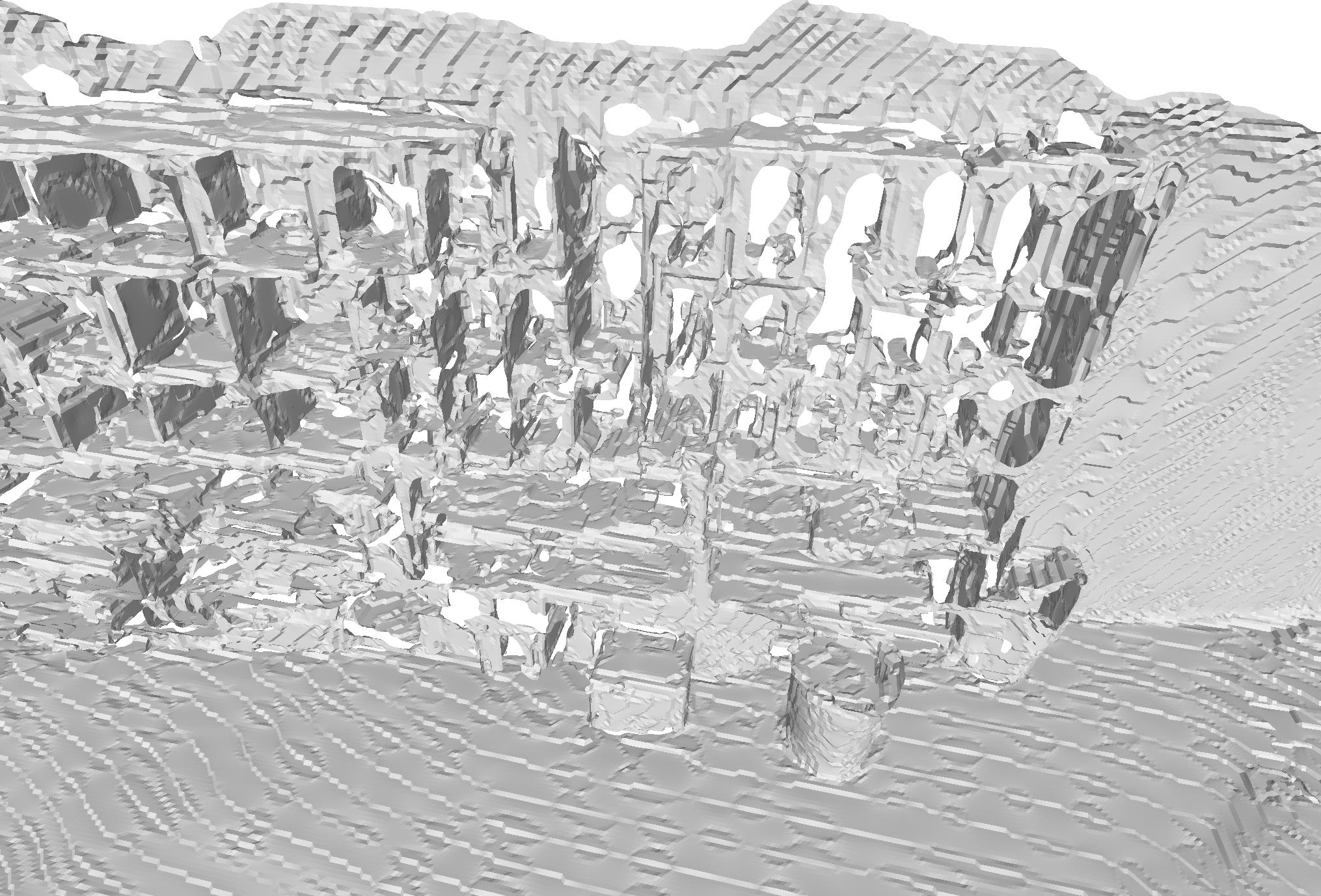} \\
 		  \includegraphics[width=1\textwidth]{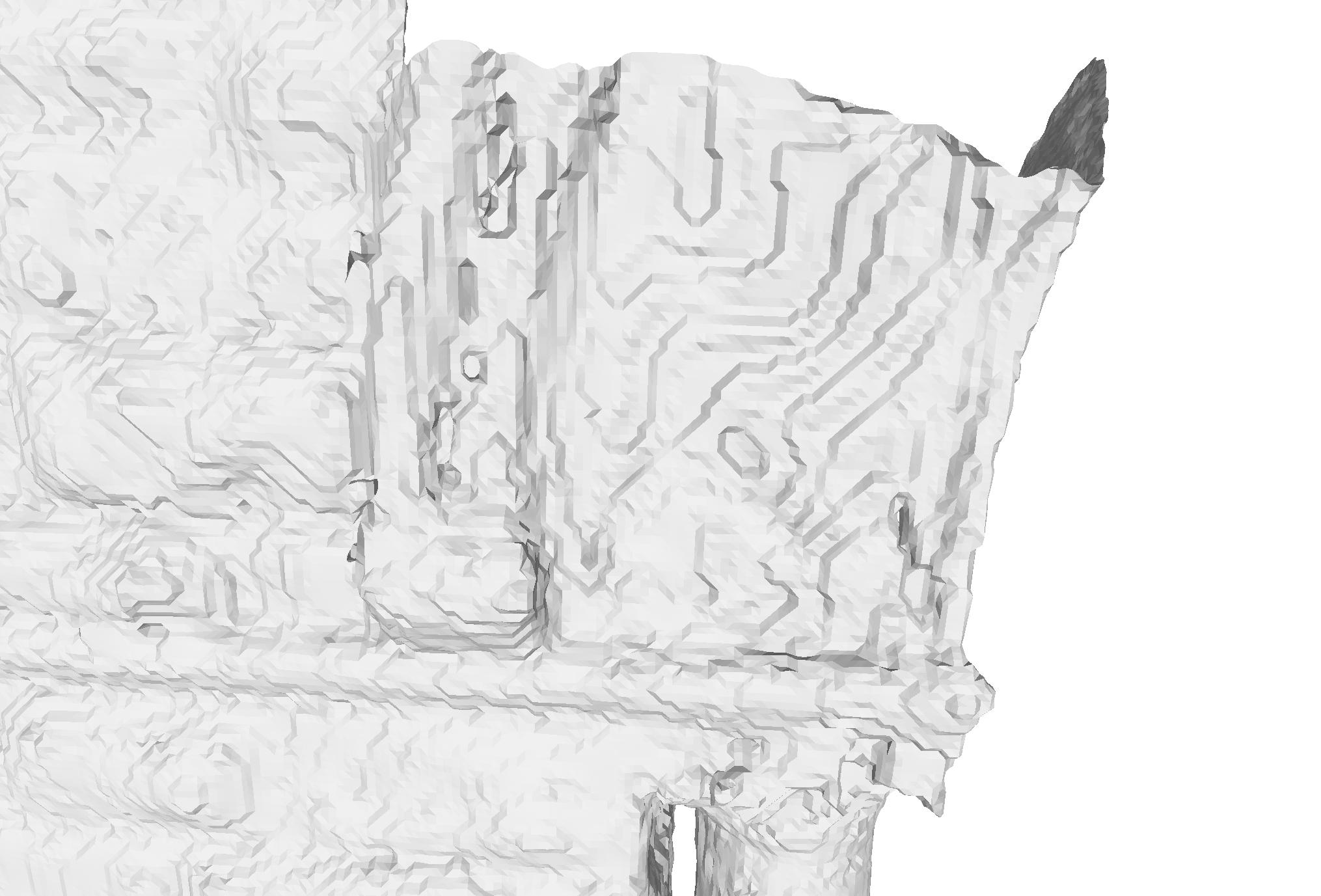} \\
 		  \includegraphics[width=1\textwidth]{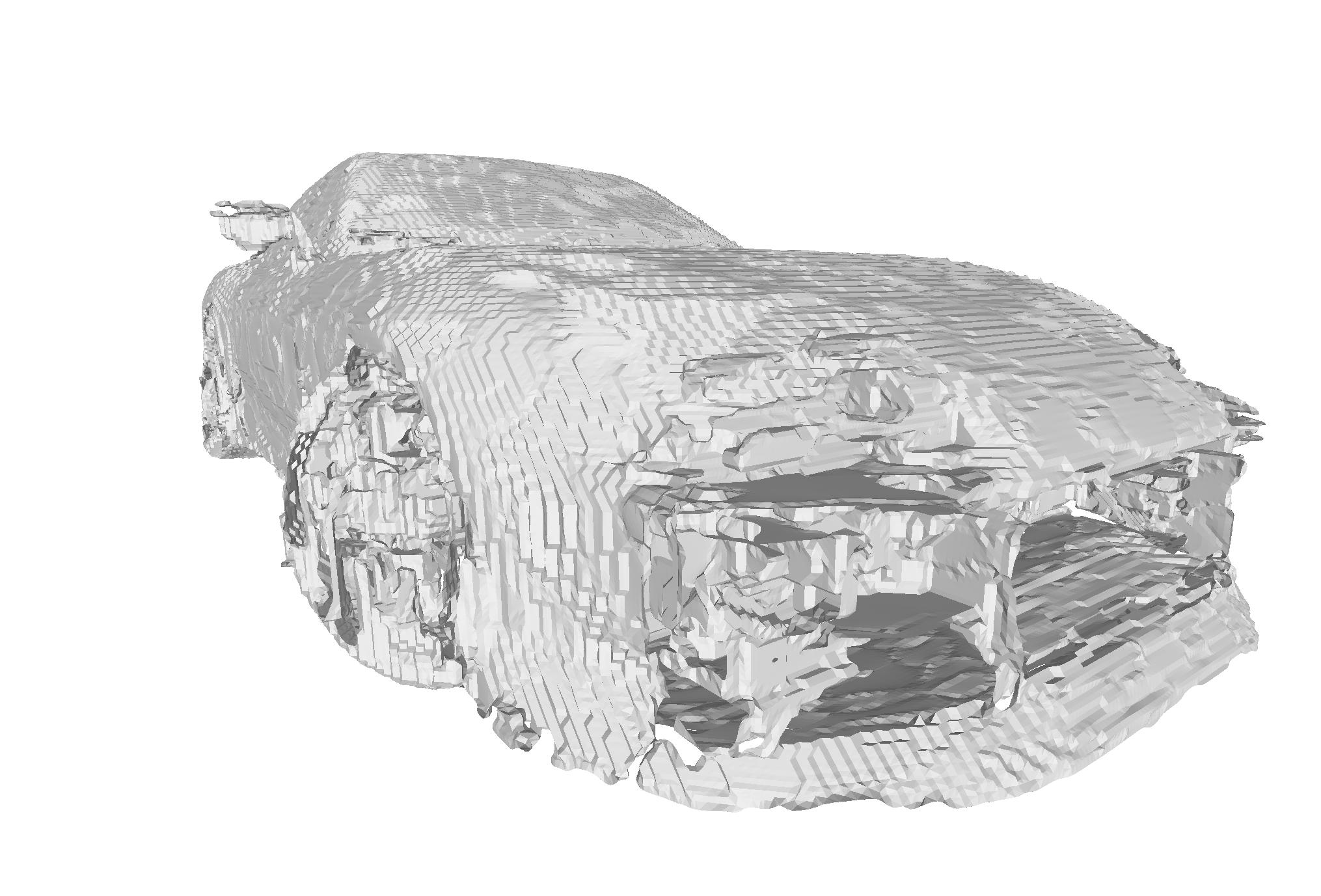} \\
 		  \includegraphics[width=1\textwidth]{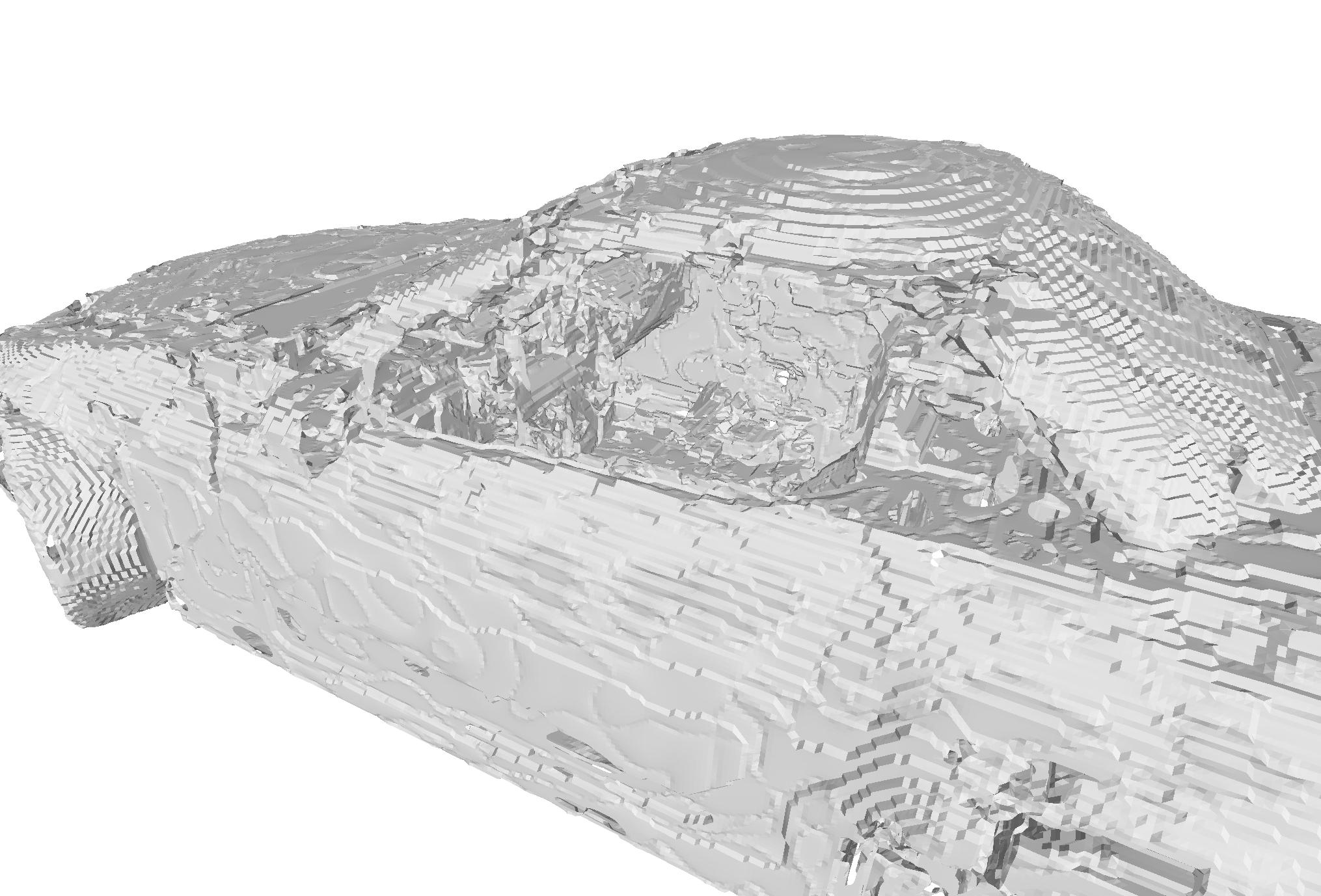} \\
 		  \includegraphics[width=1\textwidth]{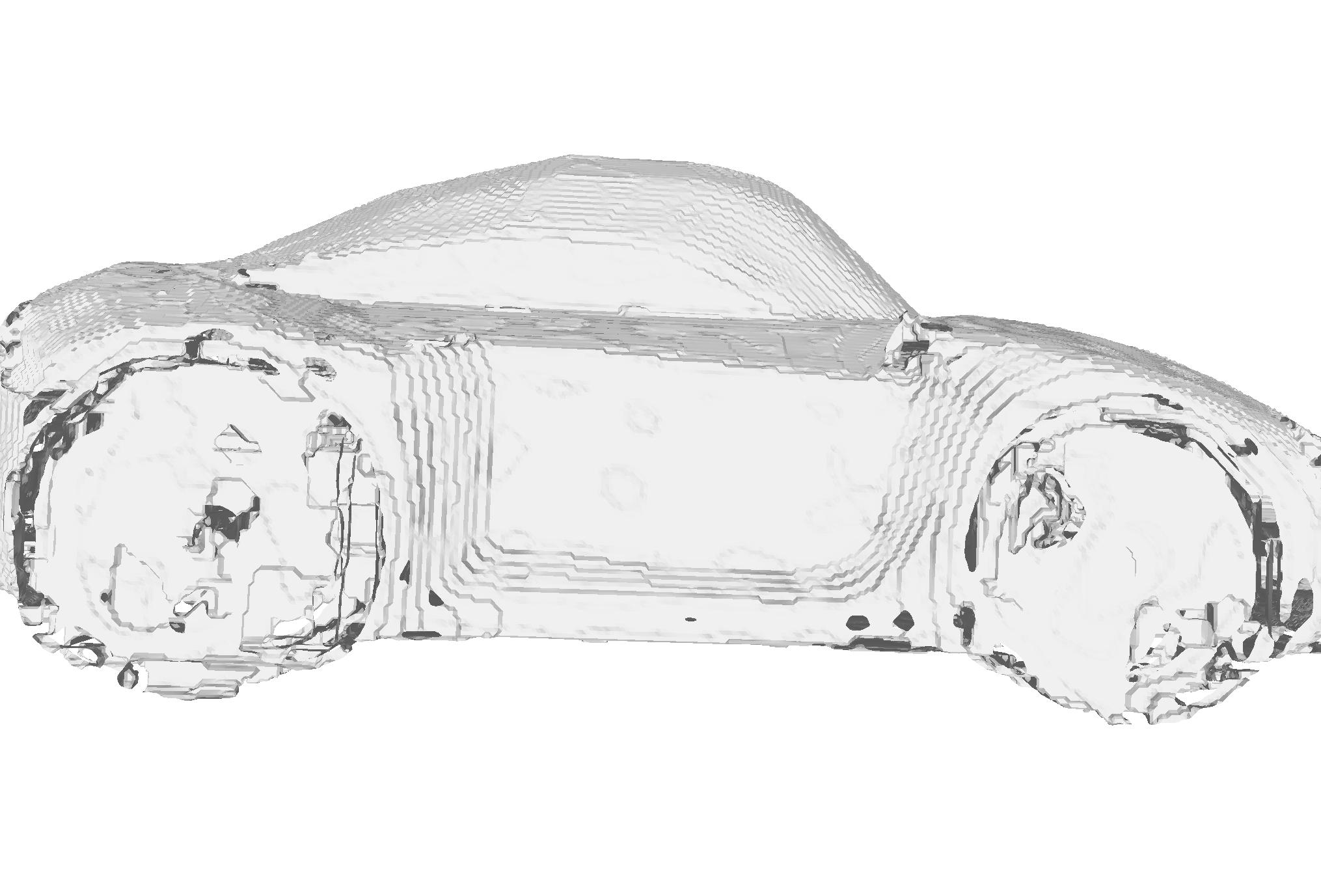} \\
 		  \includegraphics[width=1\textwidth]{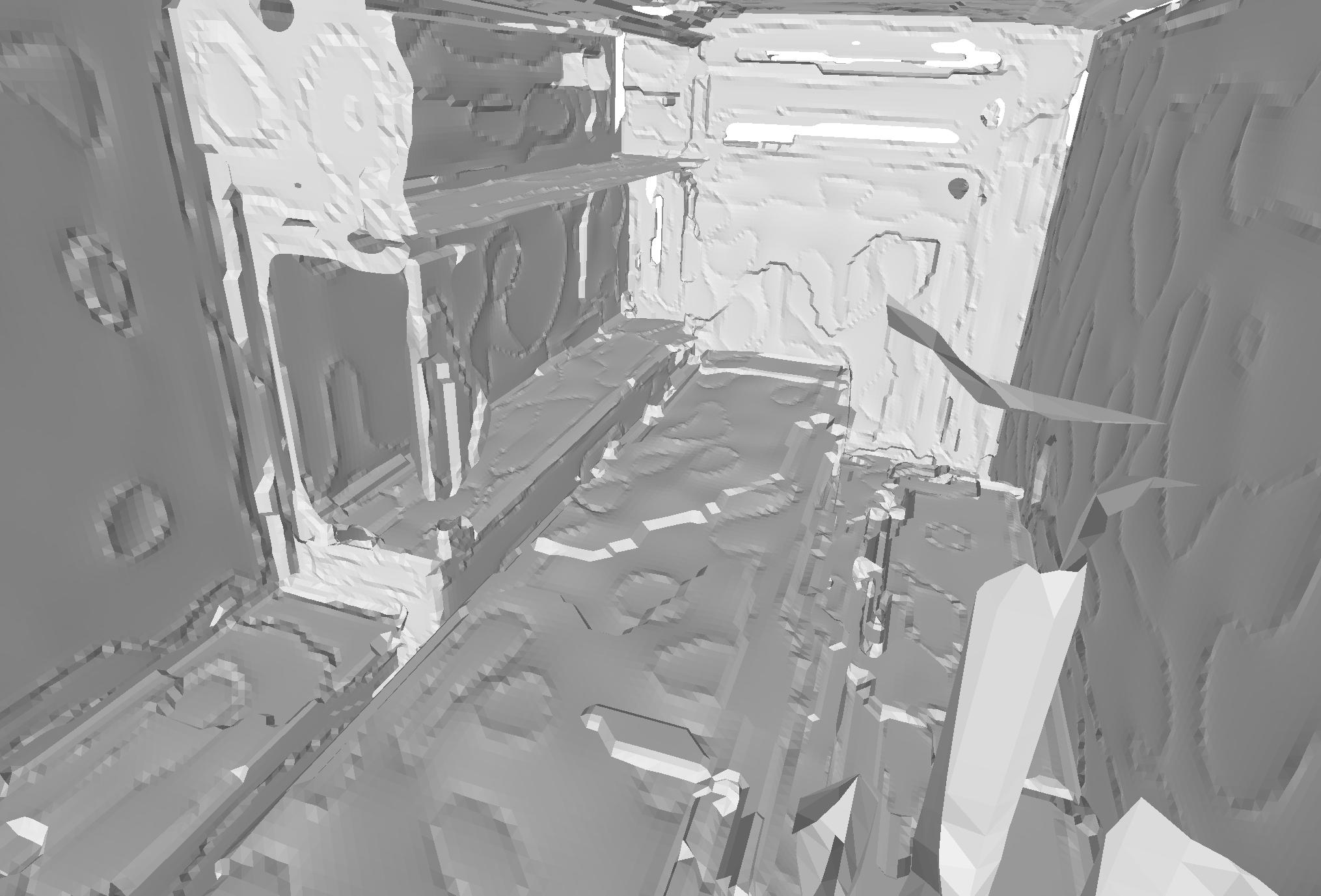} \\
        \end{minipage}
    }
    \subfigure[LevelSetUDF]{
        \begin{minipage}[b]{0.18\textwidth}
		  \includegraphics[width=1\textwidth]{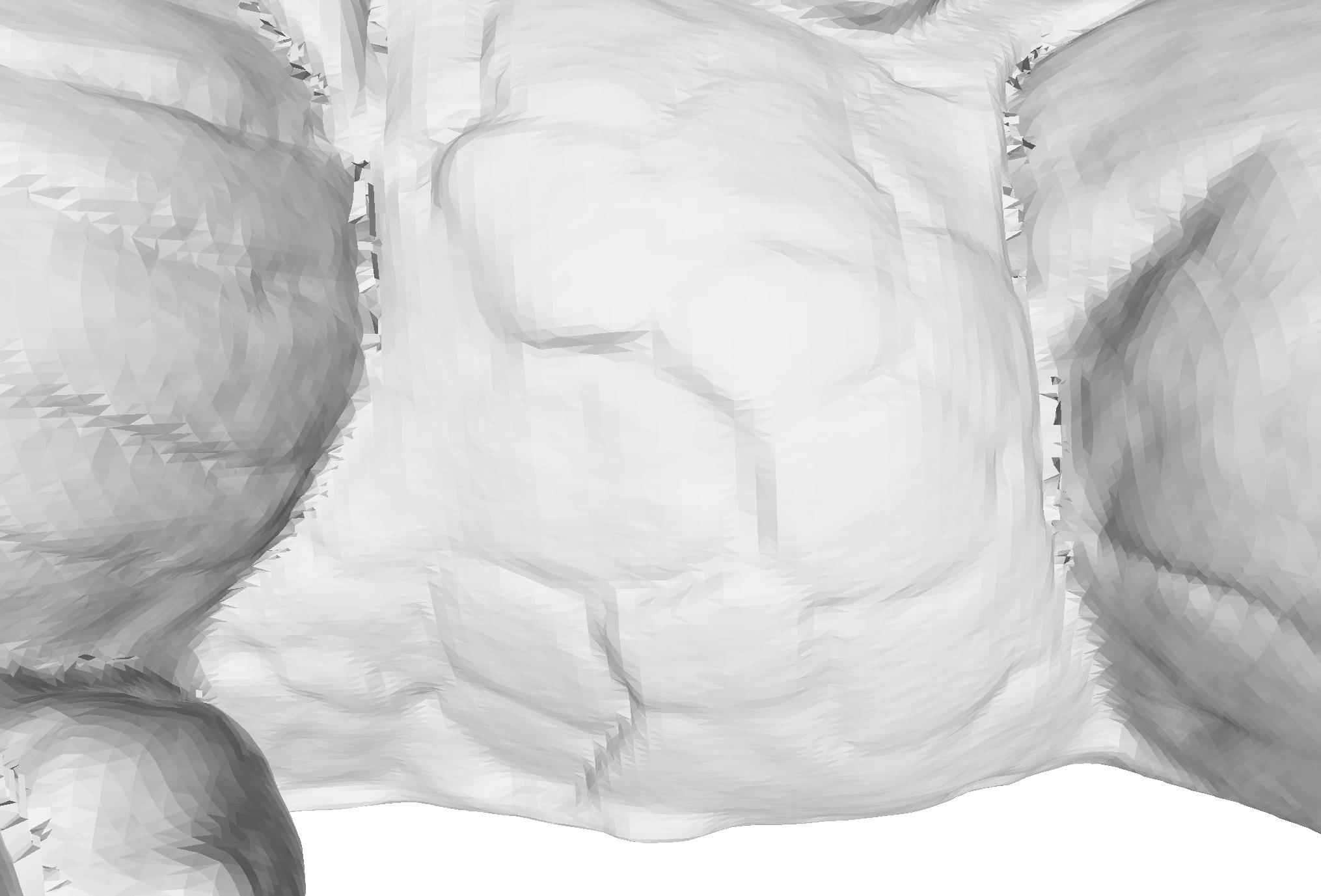} \\
		  \includegraphics[width=1\textwidth]{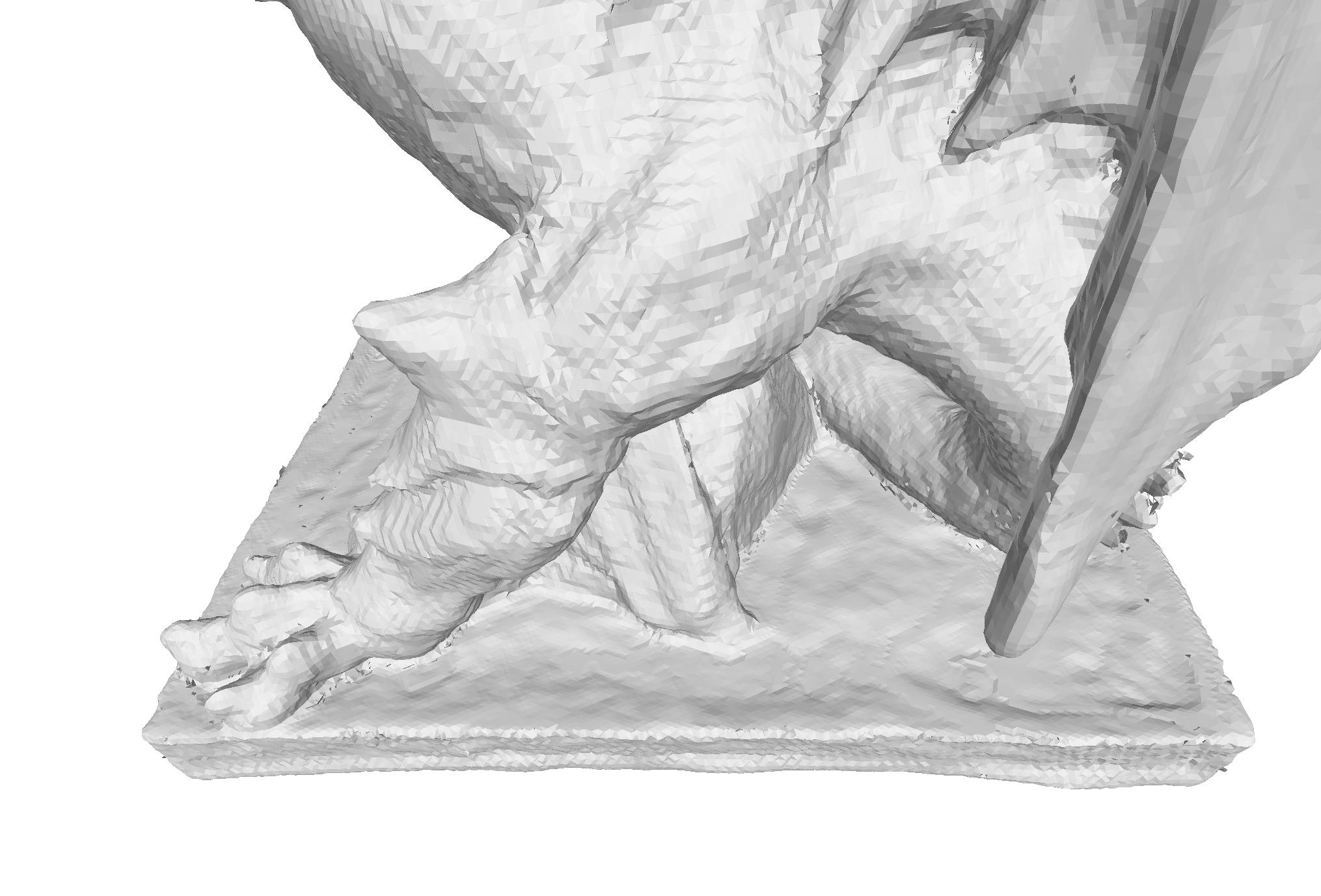} \\
		  \includegraphics[width=1\textwidth]{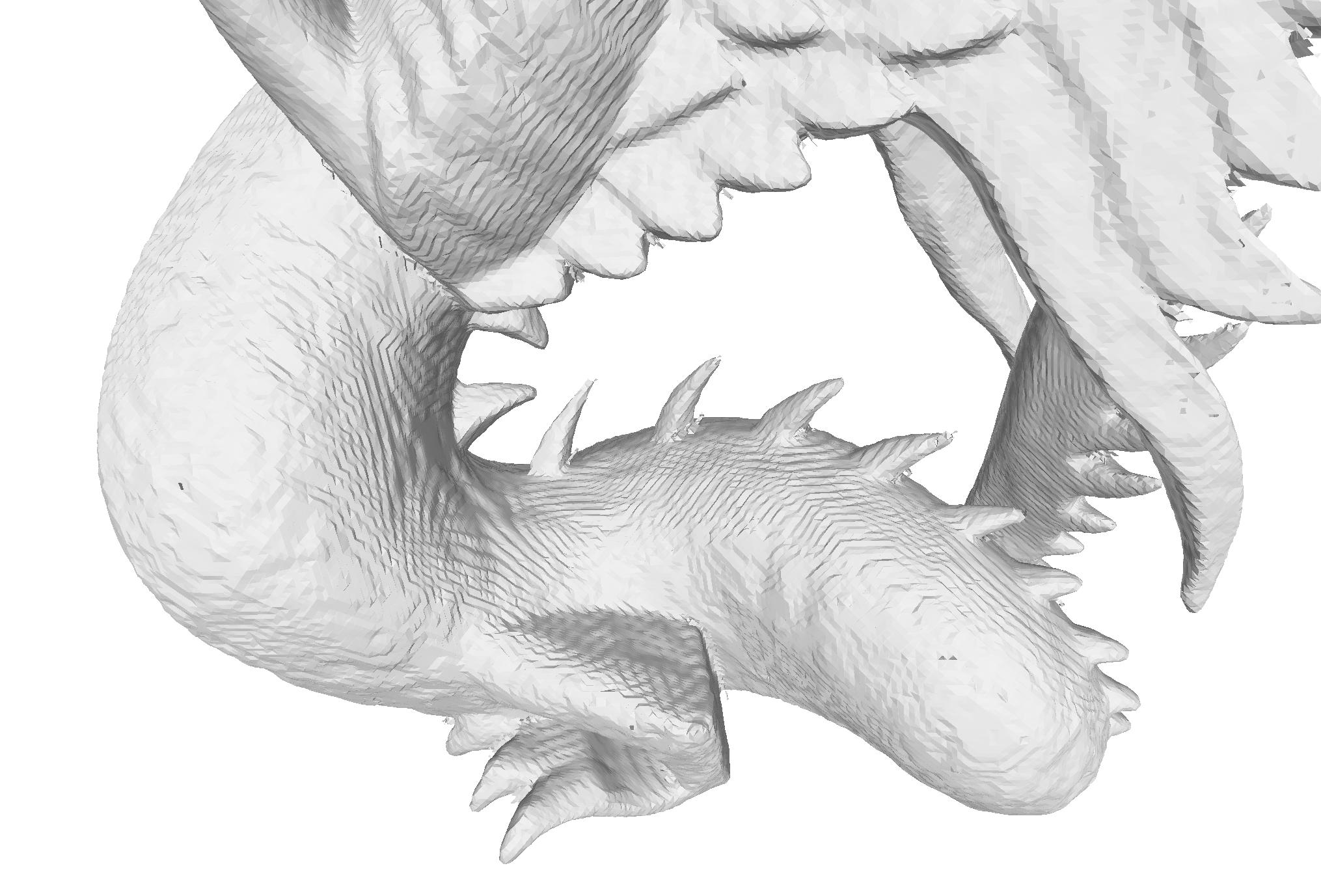} \\
		  \includegraphics[width=1\textwidth]{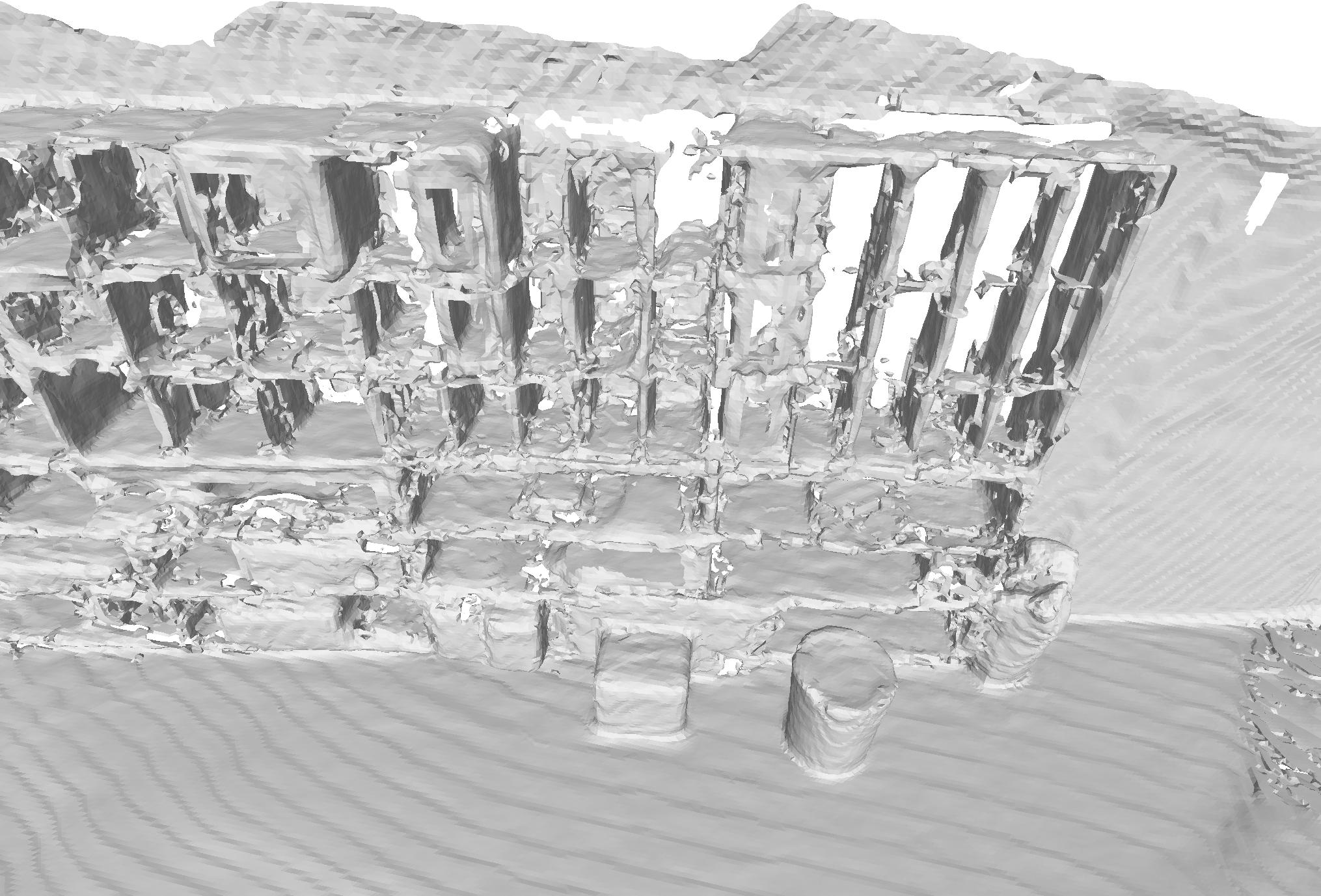} \\
 		  \includegraphics[width=1\textwidth]{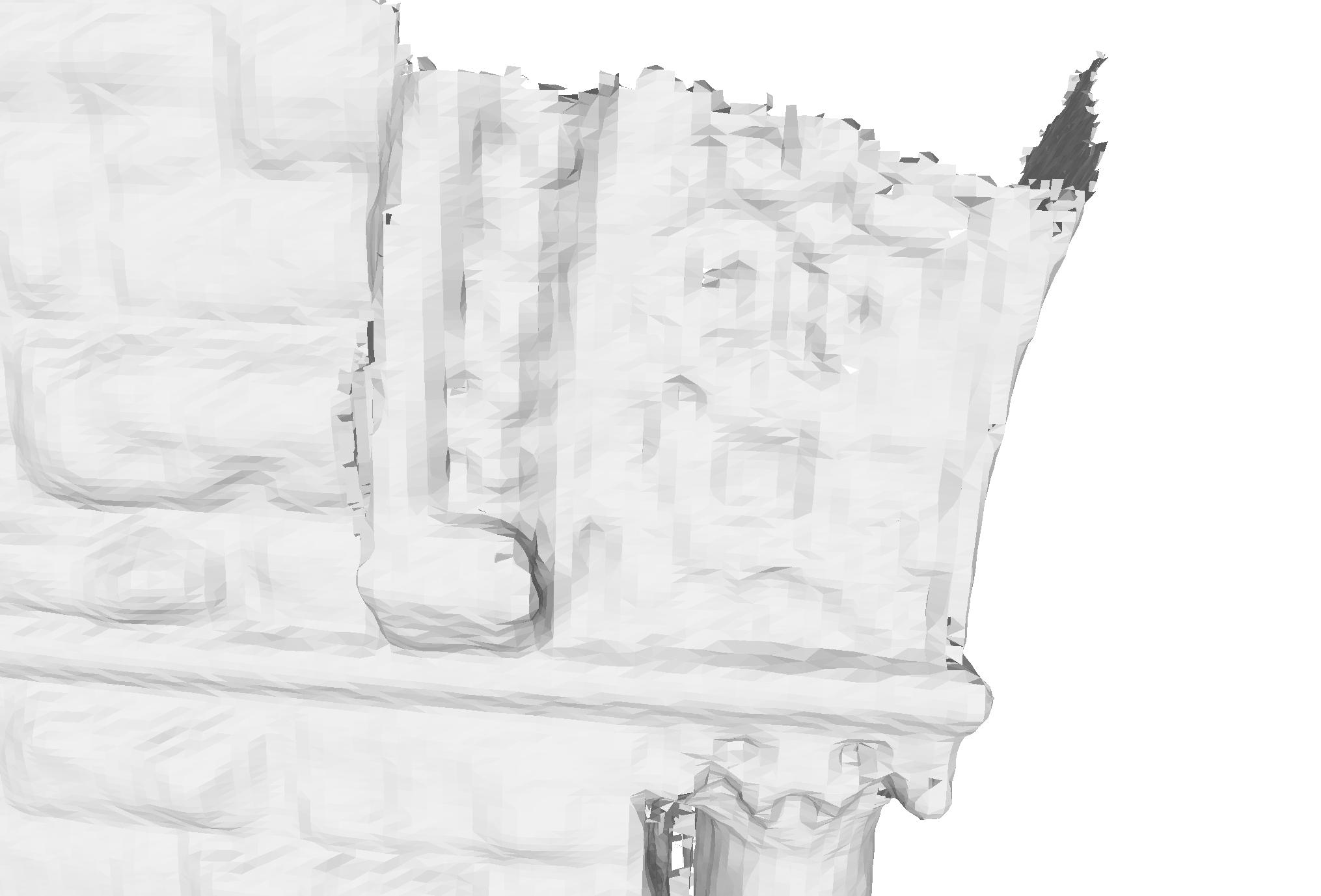} \\
 		  \includegraphics[width=1\textwidth]{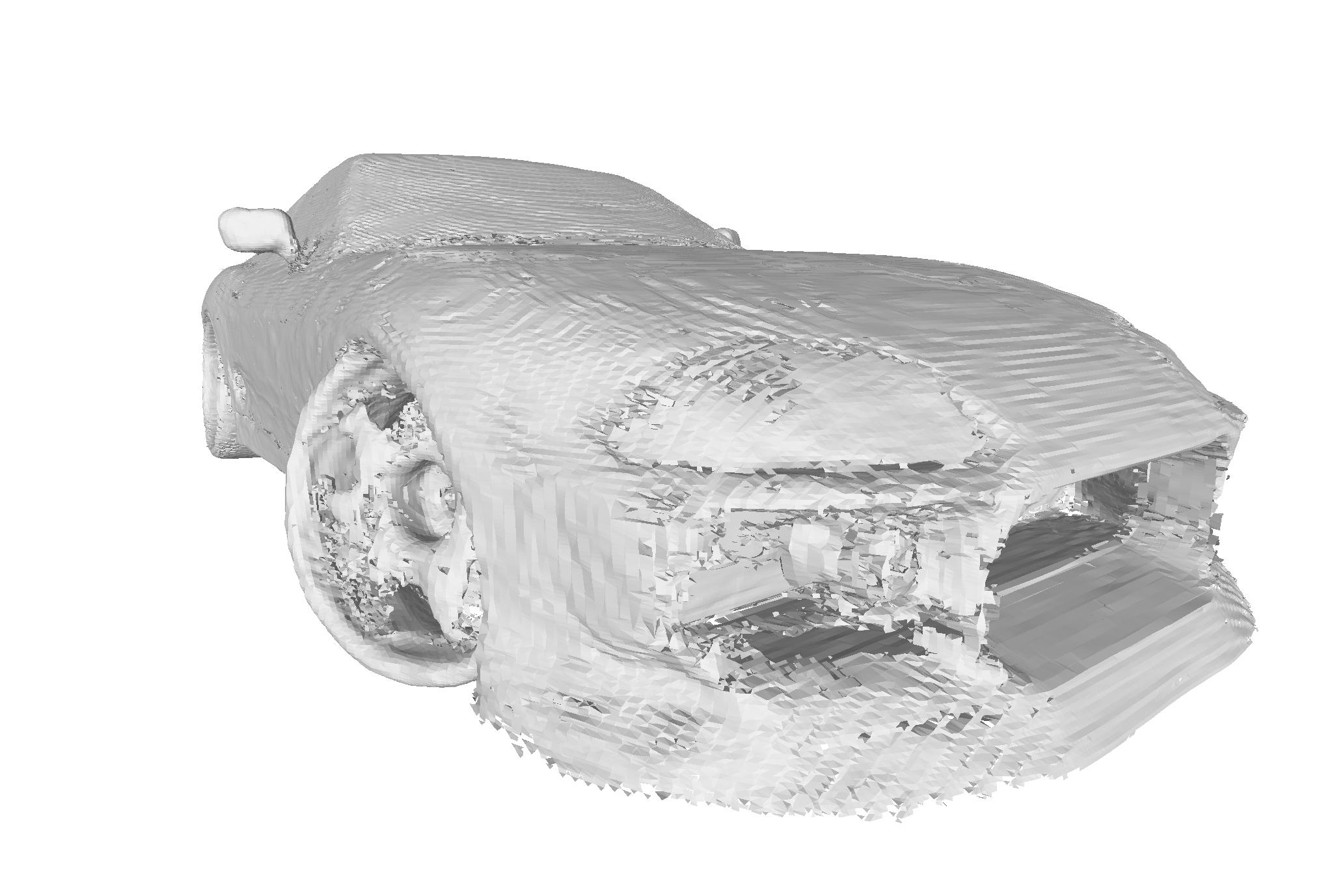} \\
 		  \includegraphics[width=1\textwidth]{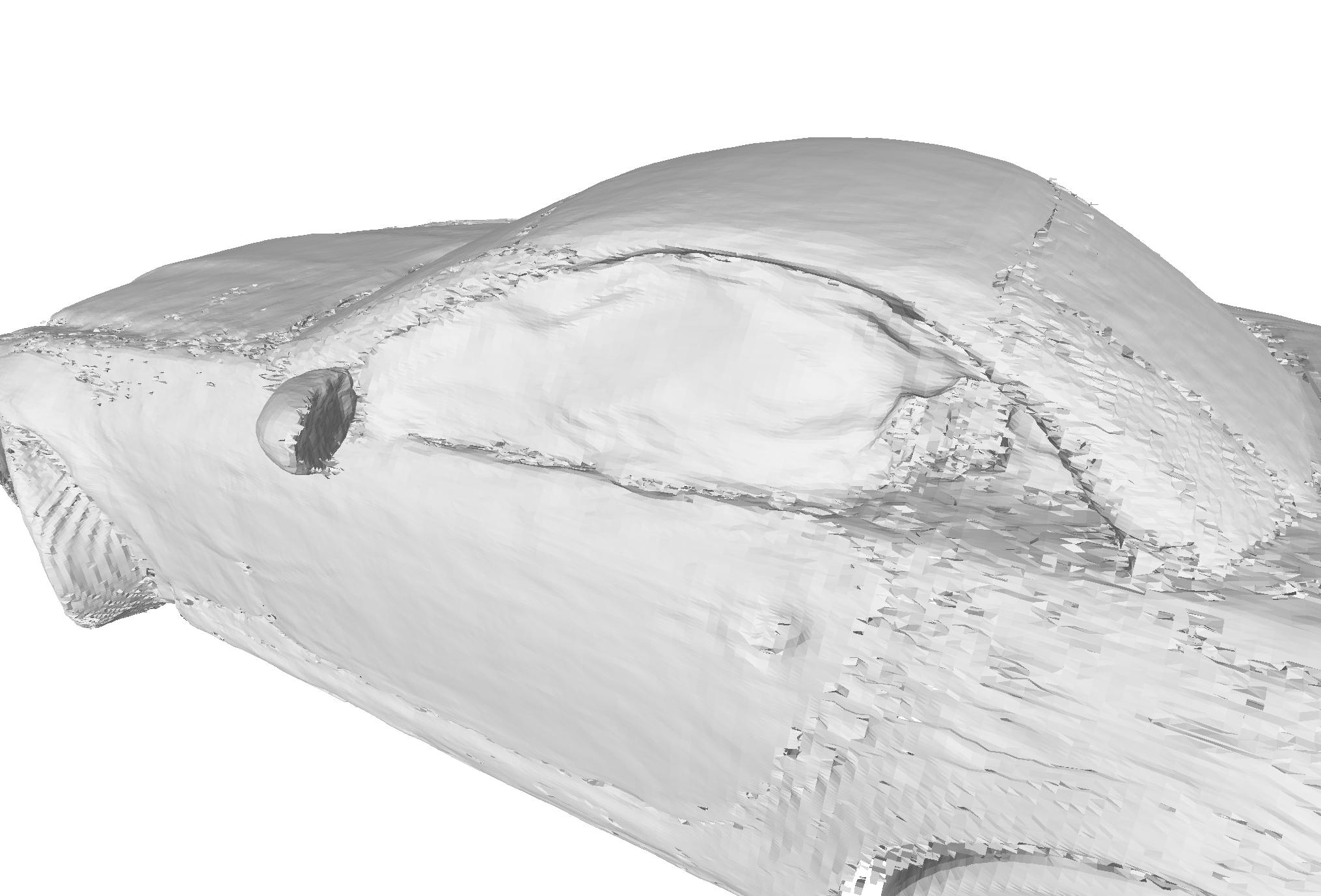} \\
 		  \includegraphics[width=1\textwidth]{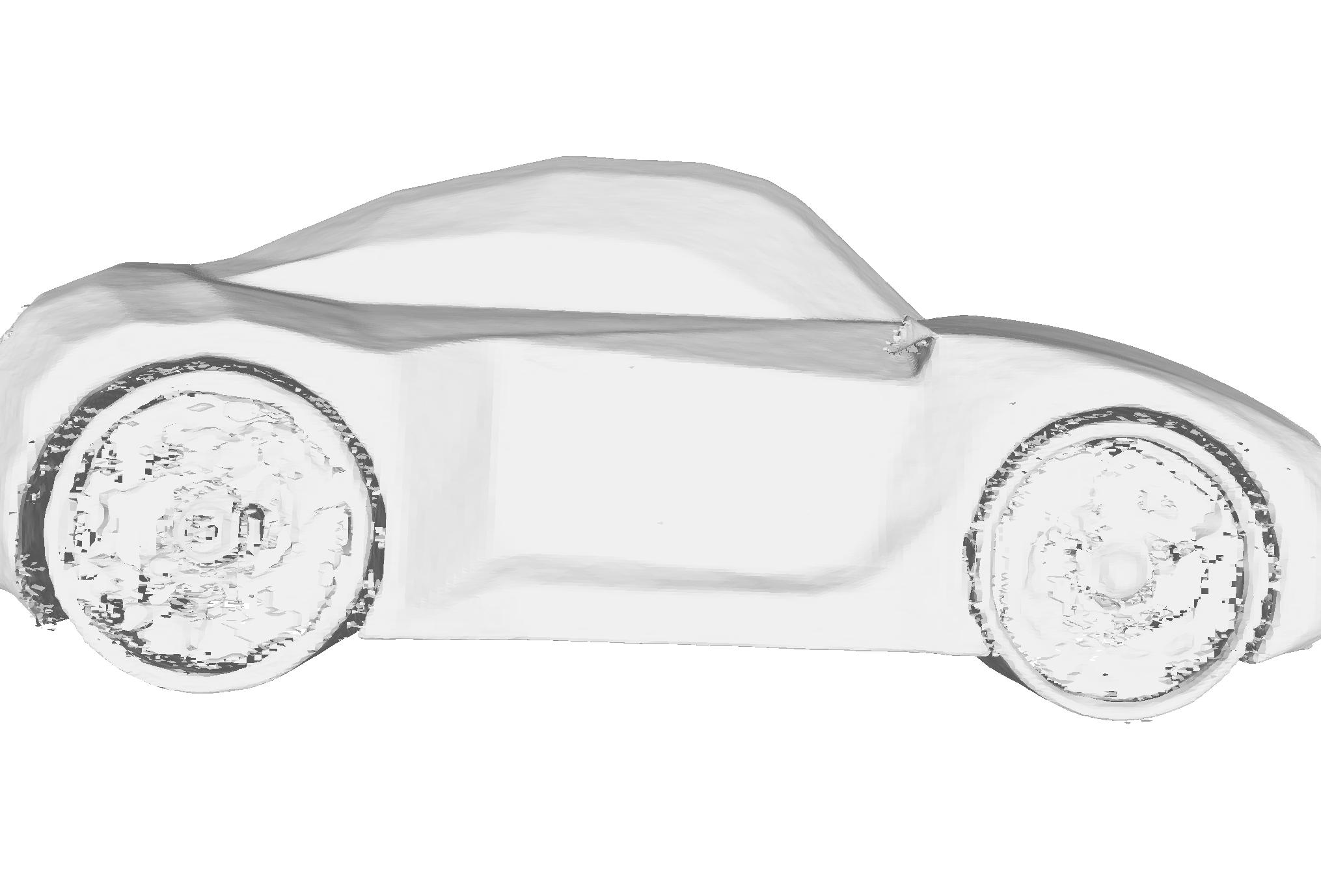} \\
 		  \includegraphics[width=1\textwidth]{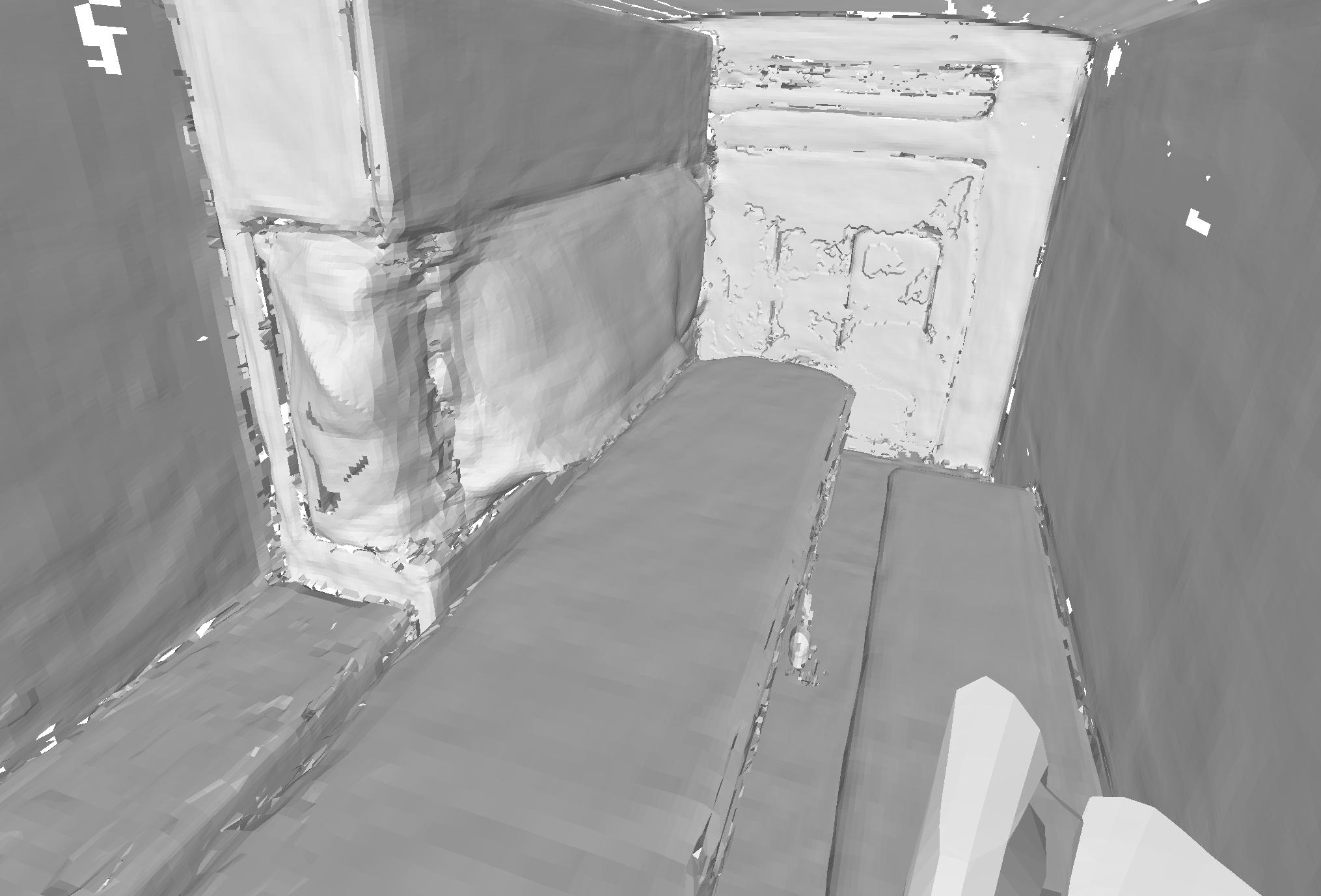} \\
        \end{minipage}
    }
    \subfigure[Ours]{
        \begin{minipage}[b]{0.18\textwidth}
		  \includegraphics[width=1\textwidth]{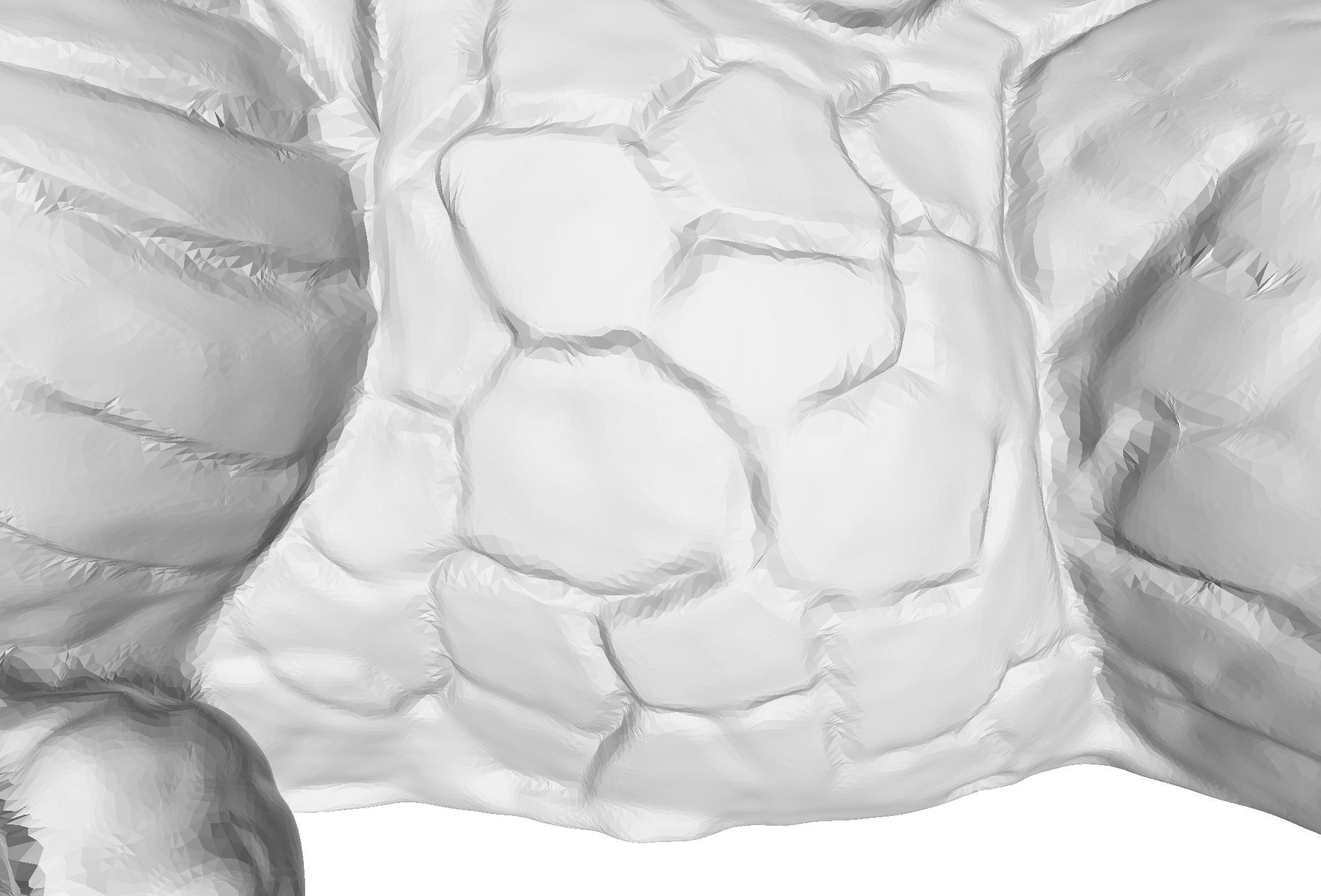} \\
		  \includegraphics[width=1\textwidth]{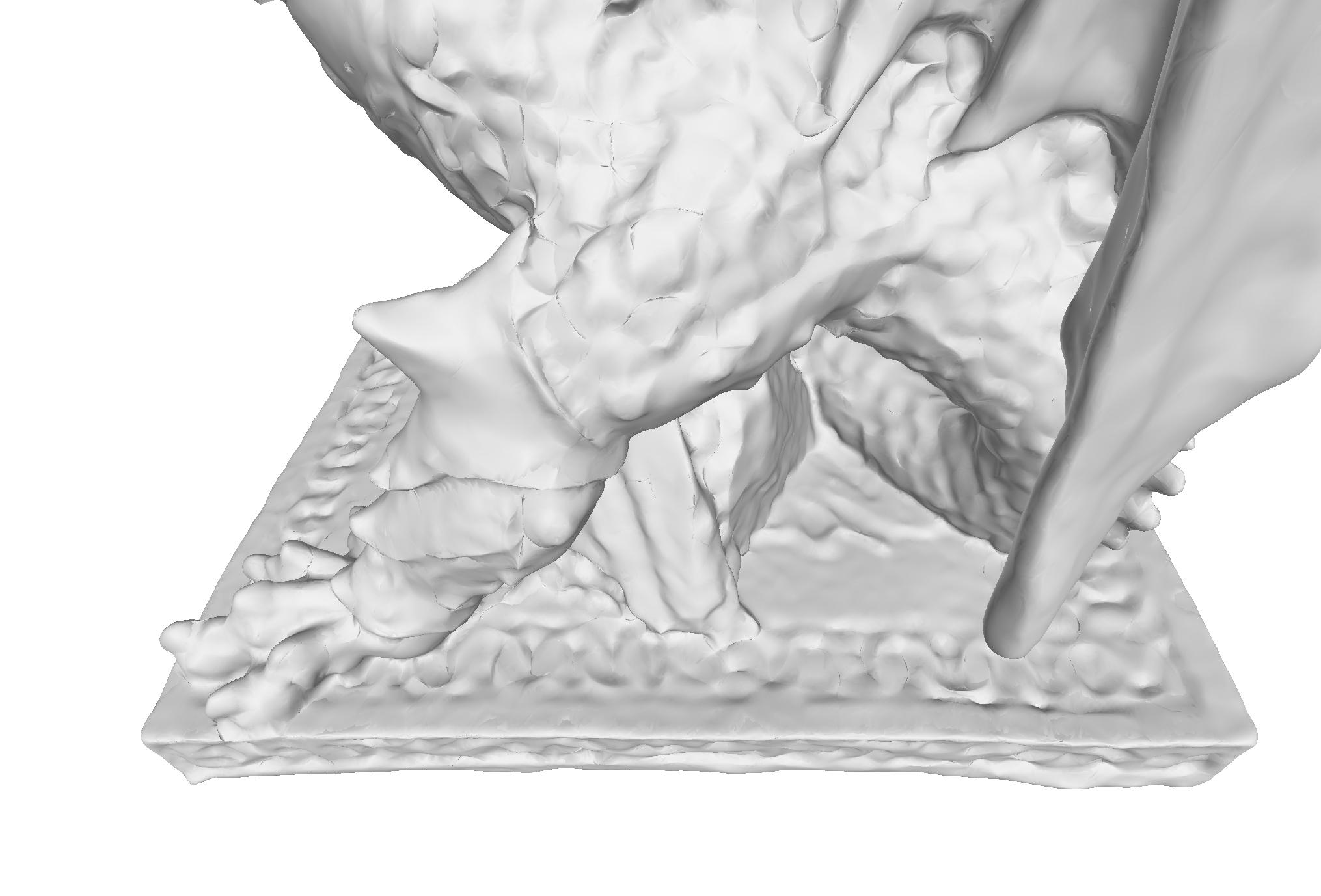} \\
		  \includegraphics[width=1\textwidth]{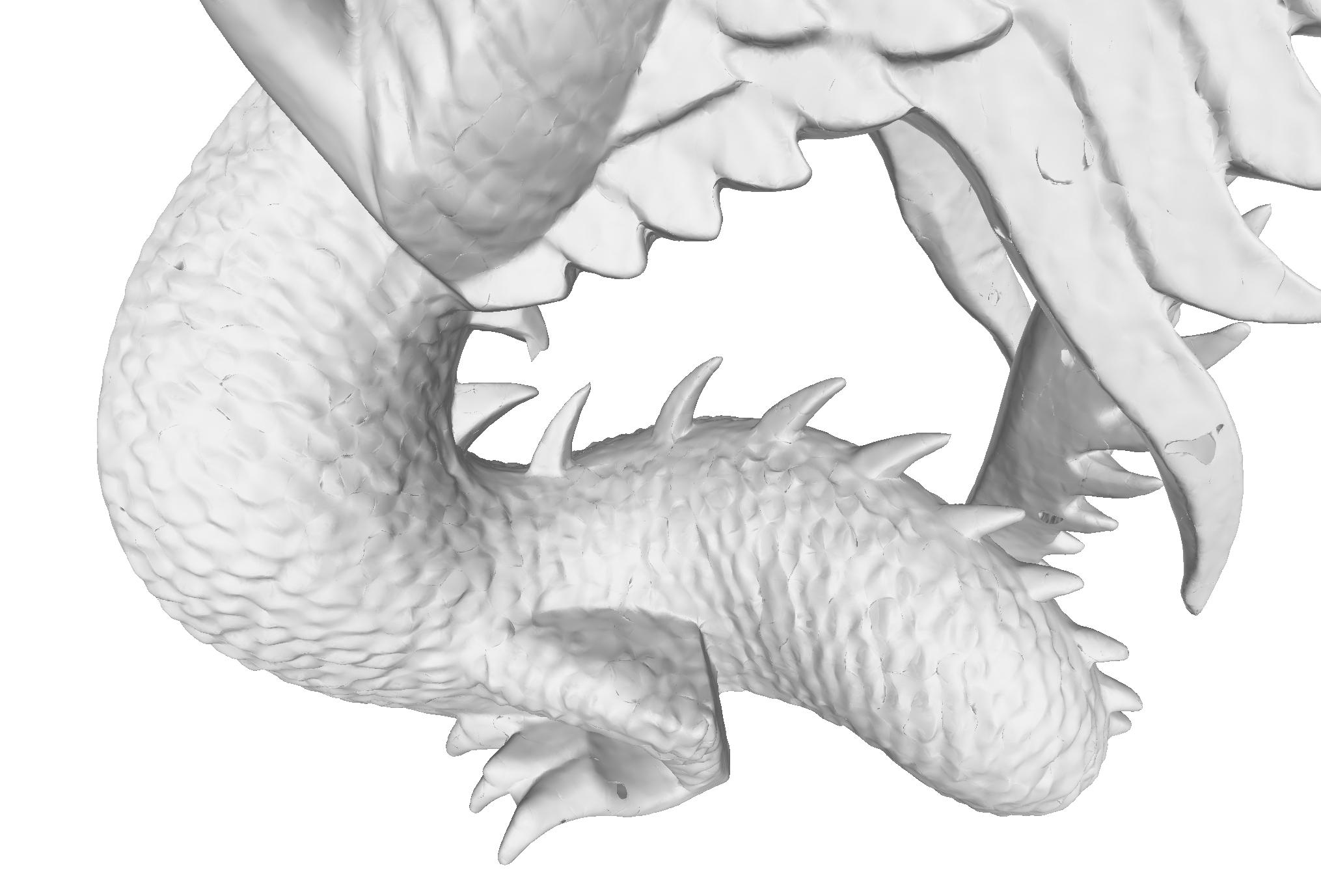} \\
		  \includegraphics[width=1\textwidth]{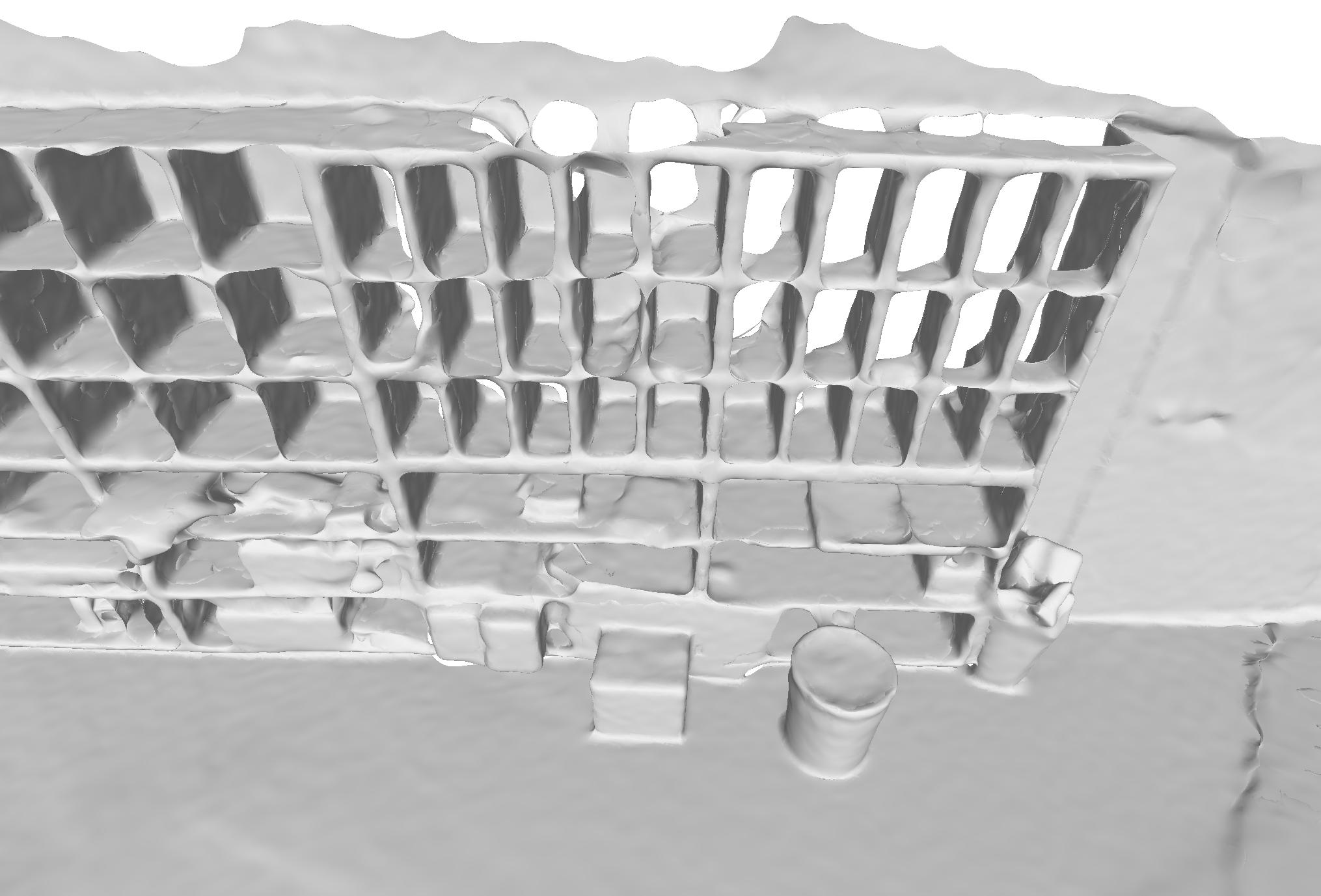} \\
 		  \includegraphics[width=1\textwidth]{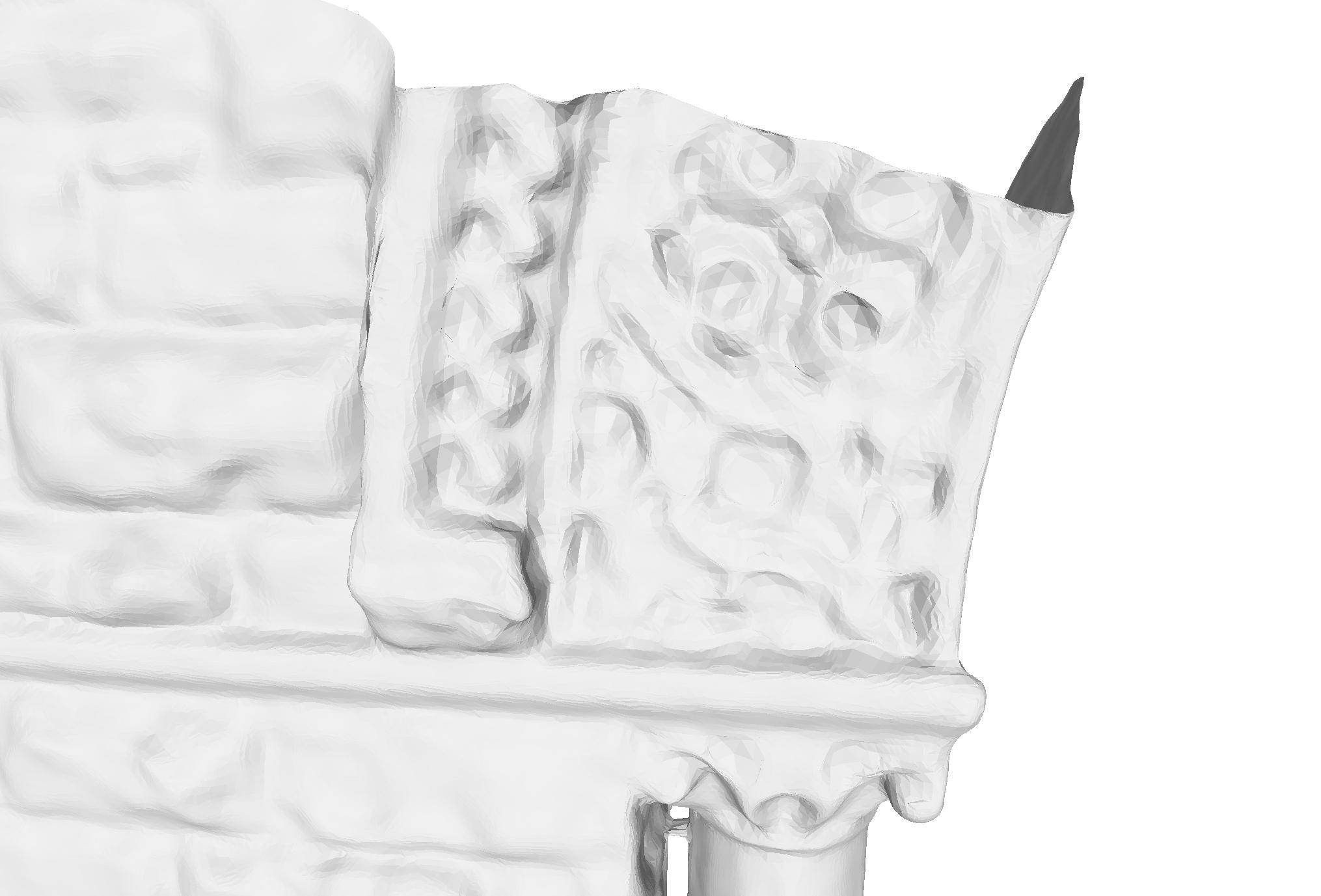} \\
 		  \includegraphics[width=1\textwidth]{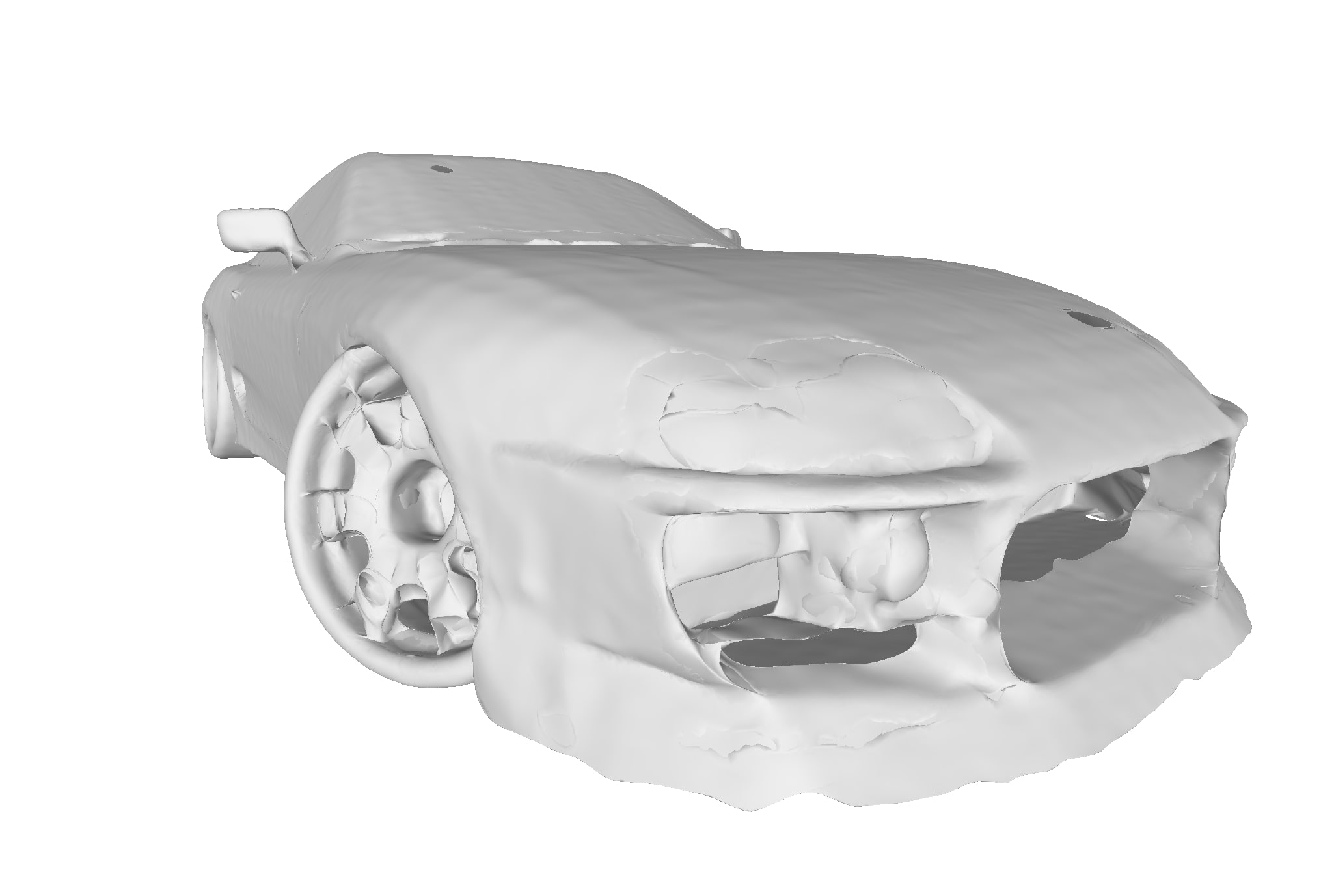} \\
 		  \includegraphics[width=1\textwidth]{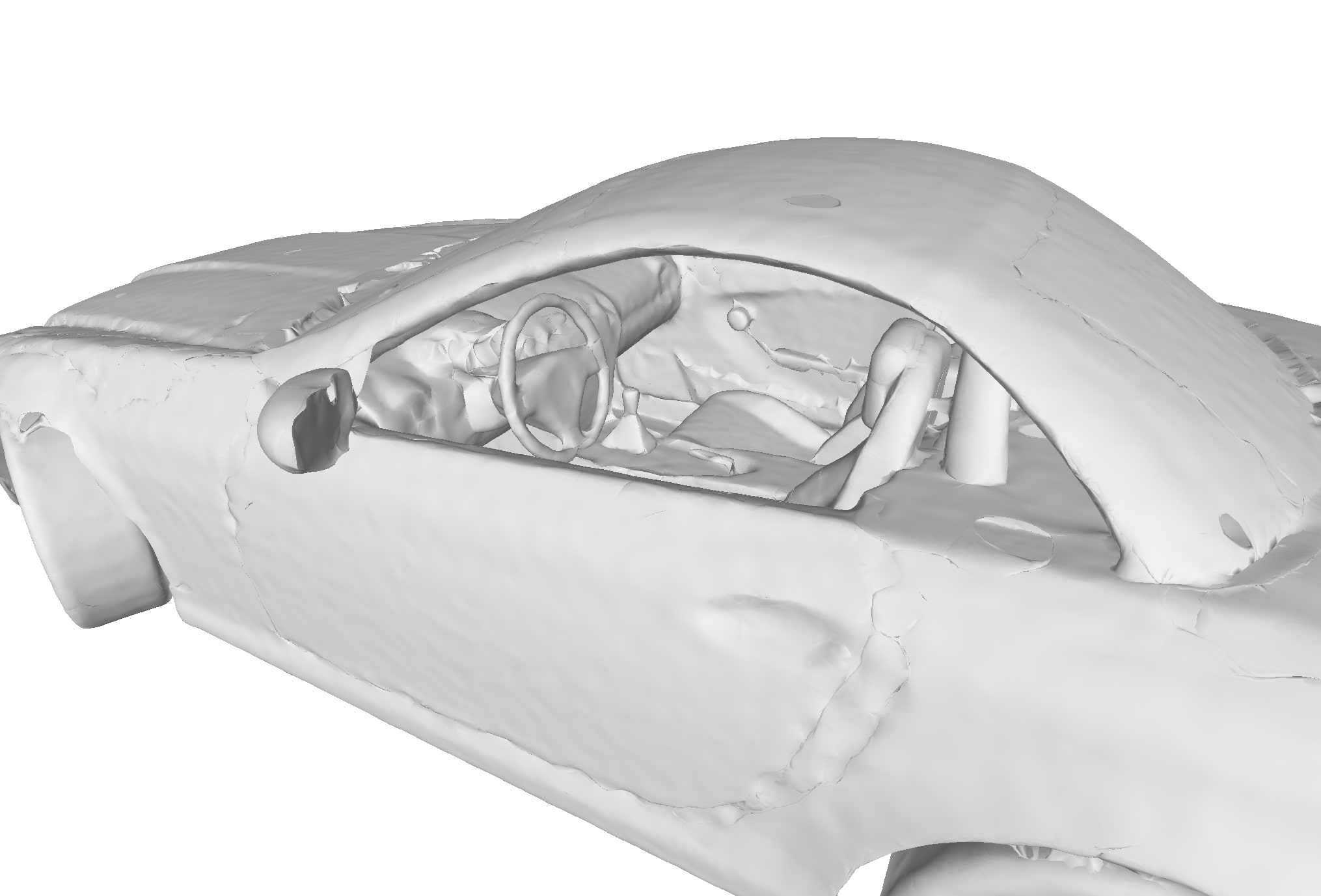} \\
 		  \includegraphics[width=1\textwidth]{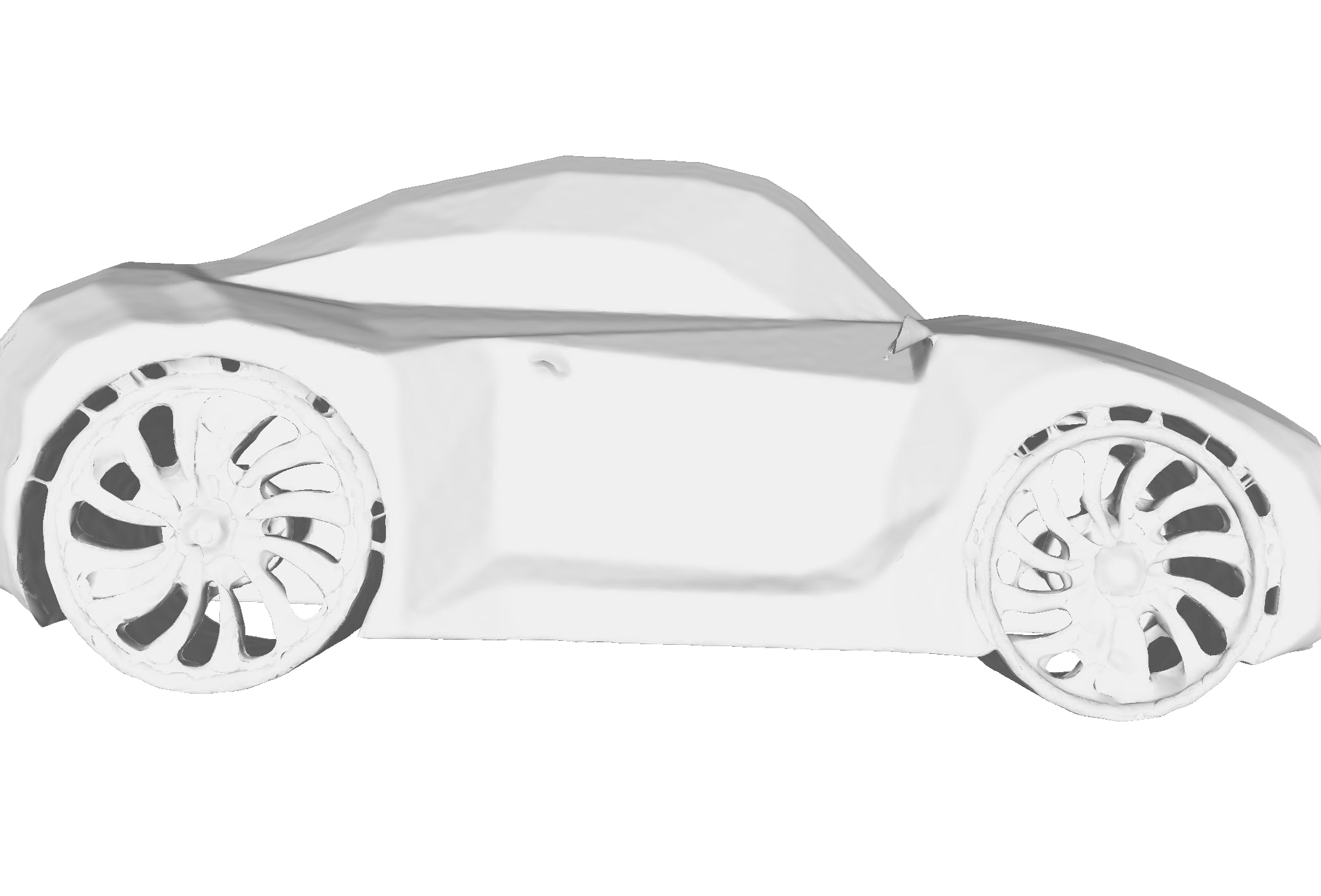} \\
 		  \includegraphics[width=1\textwidth]{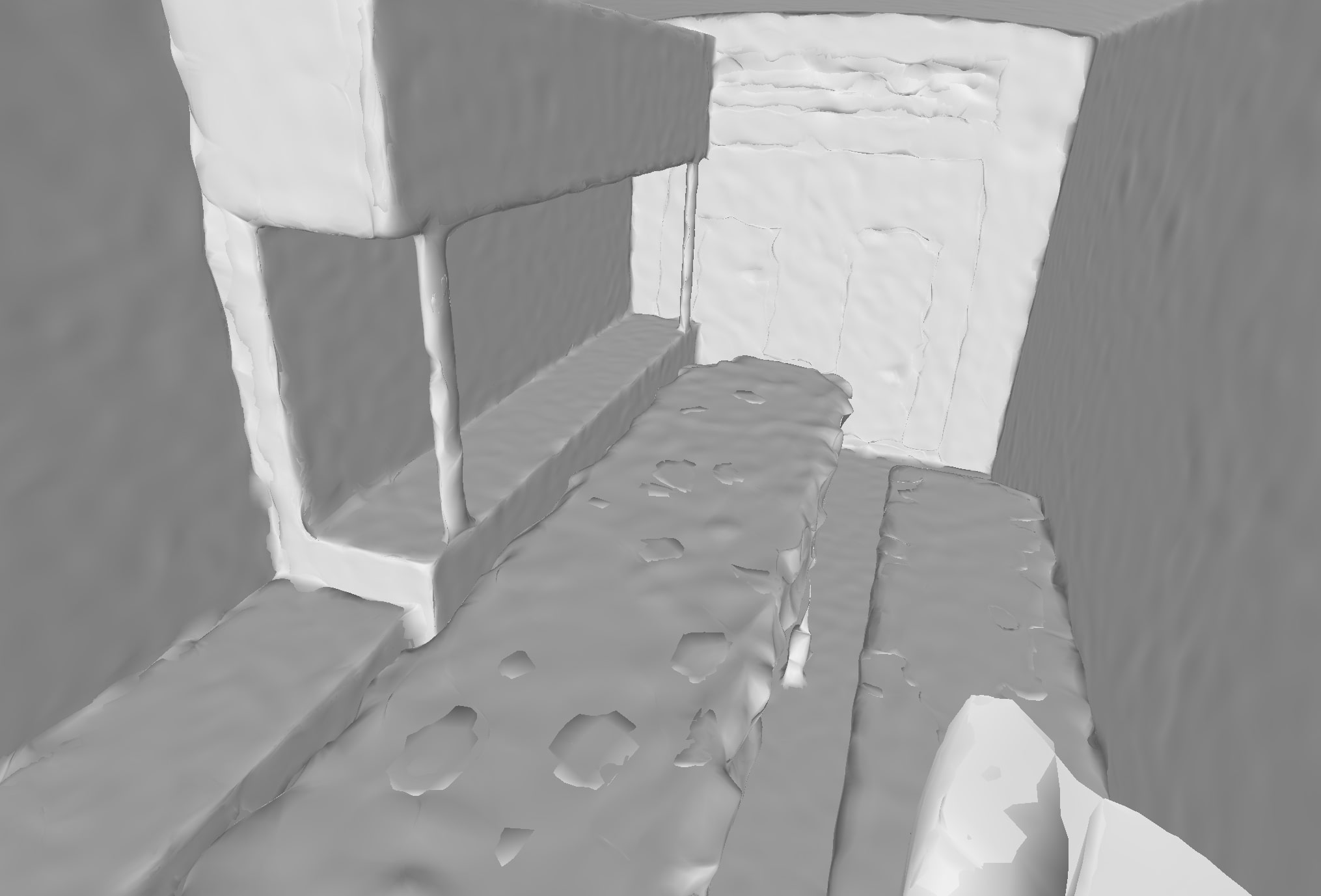} \\
        \end{minipage}
    }
    \subfigure[GT]{
        \begin{minipage}[b]{0.18\textwidth}
		  \includegraphics[width=1\textwidth]{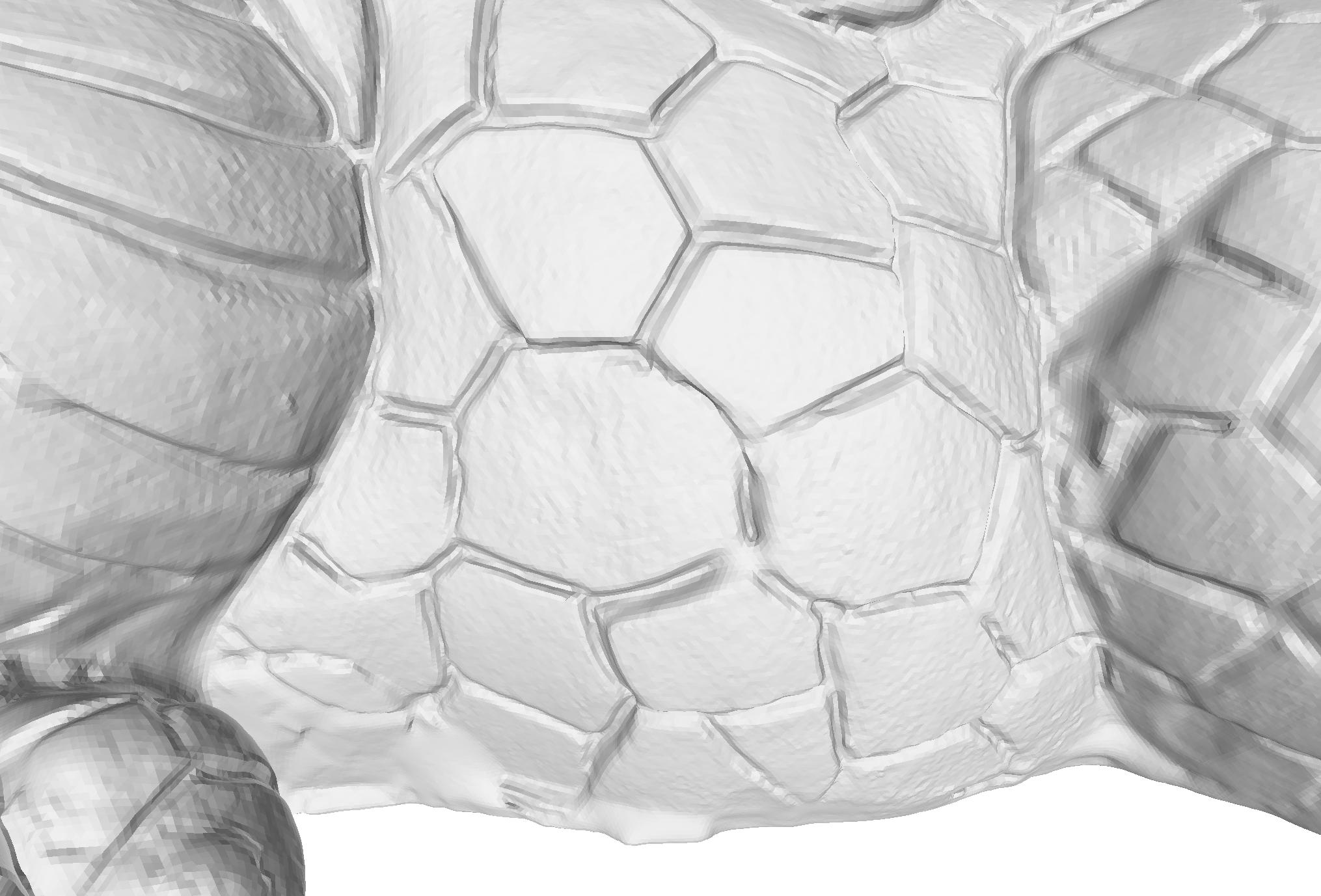} \\
		  \includegraphics[width=1\textwidth]{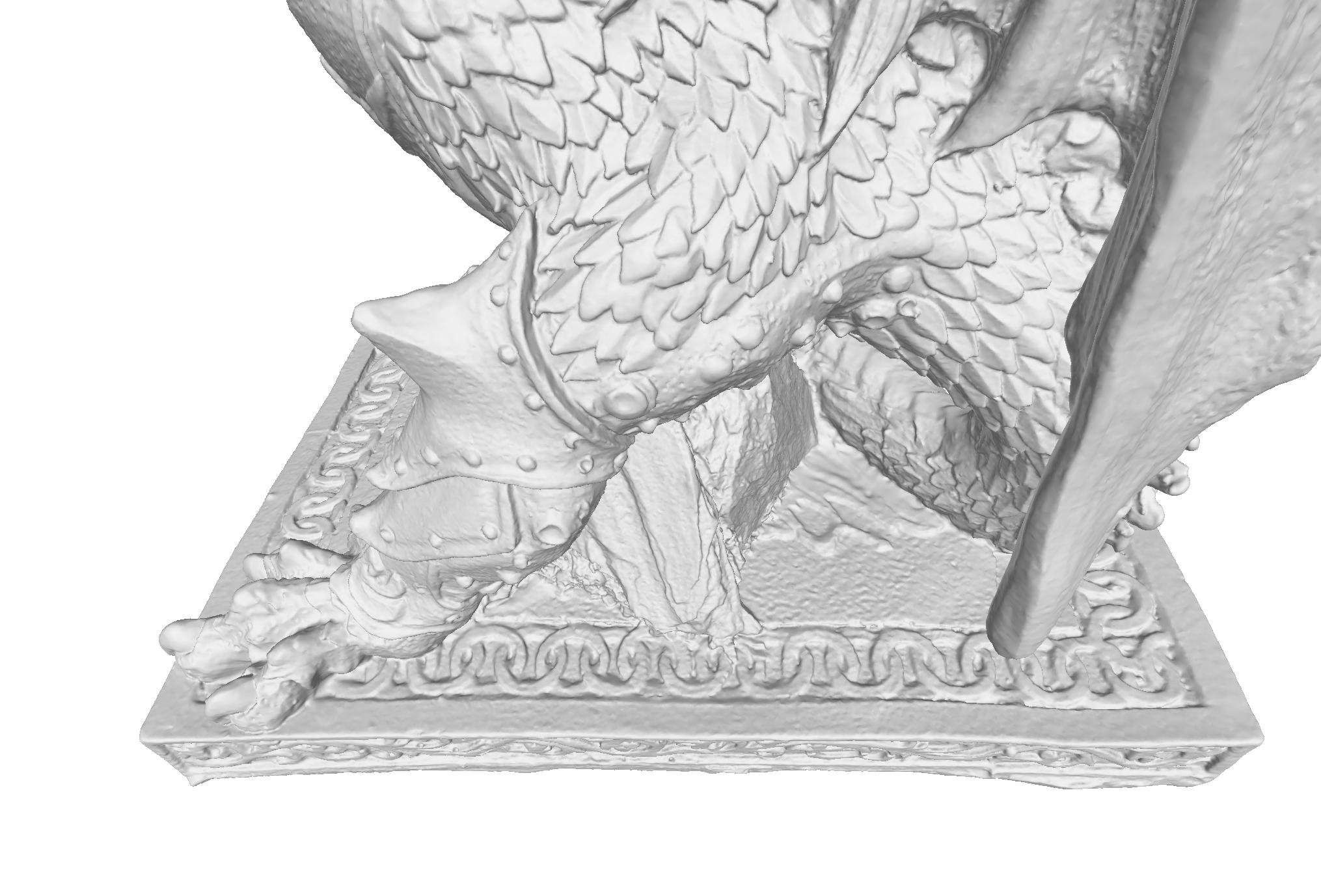} \\
		  \includegraphics[width=1\textwidth]{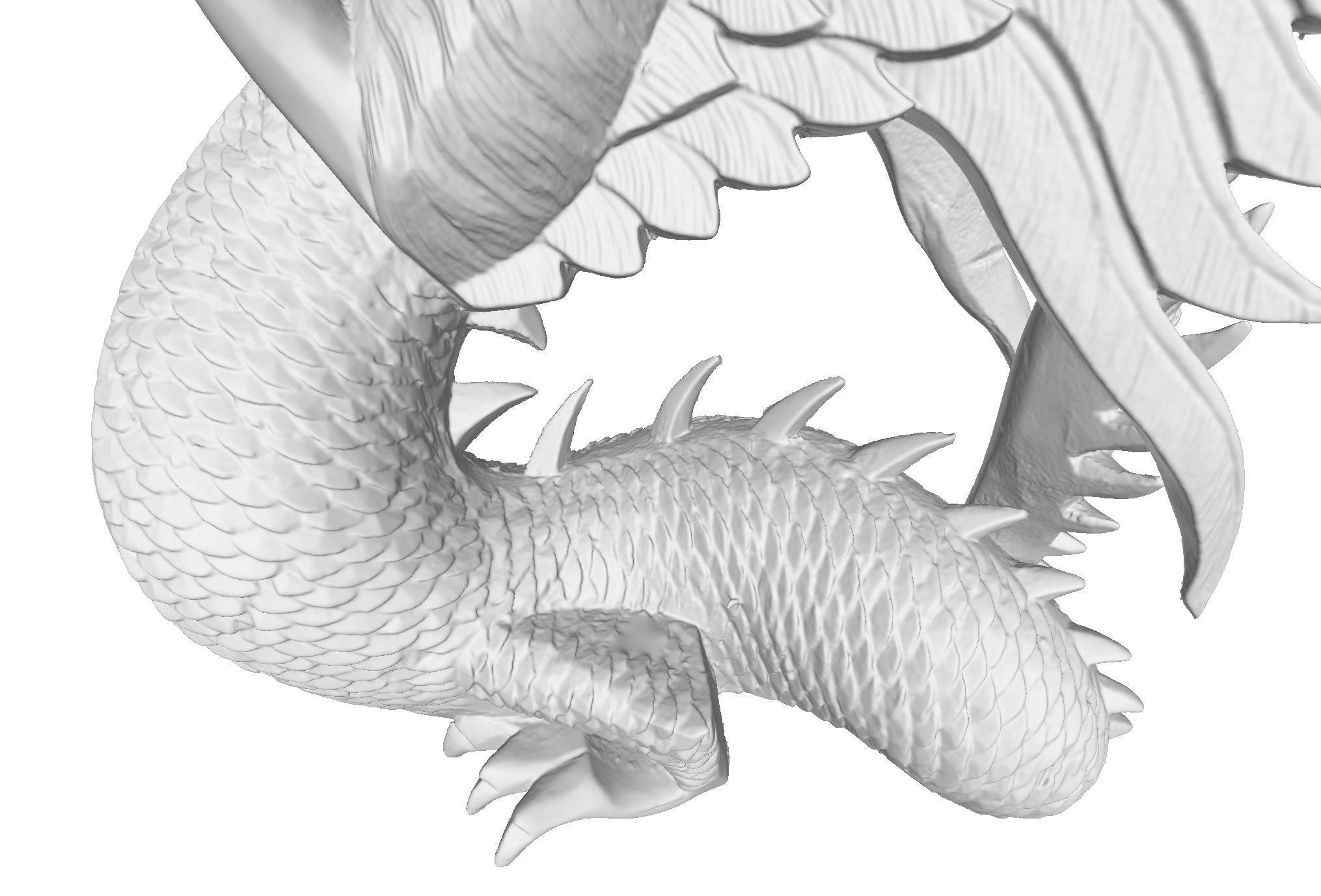} \\
		  \includegraphics[width=1\textwidth]{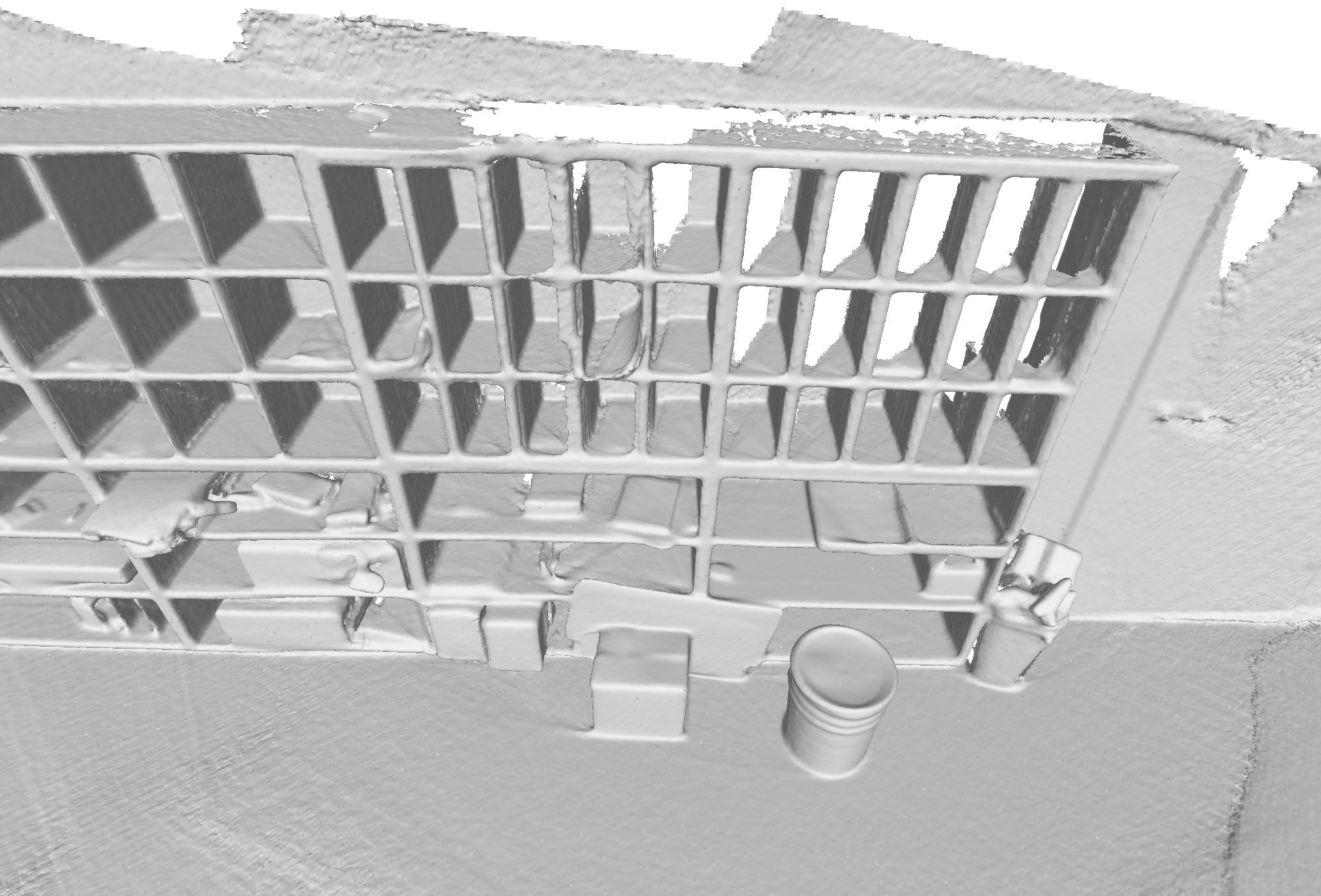} \\
 		  \includegraphics[width=1\textwidth]{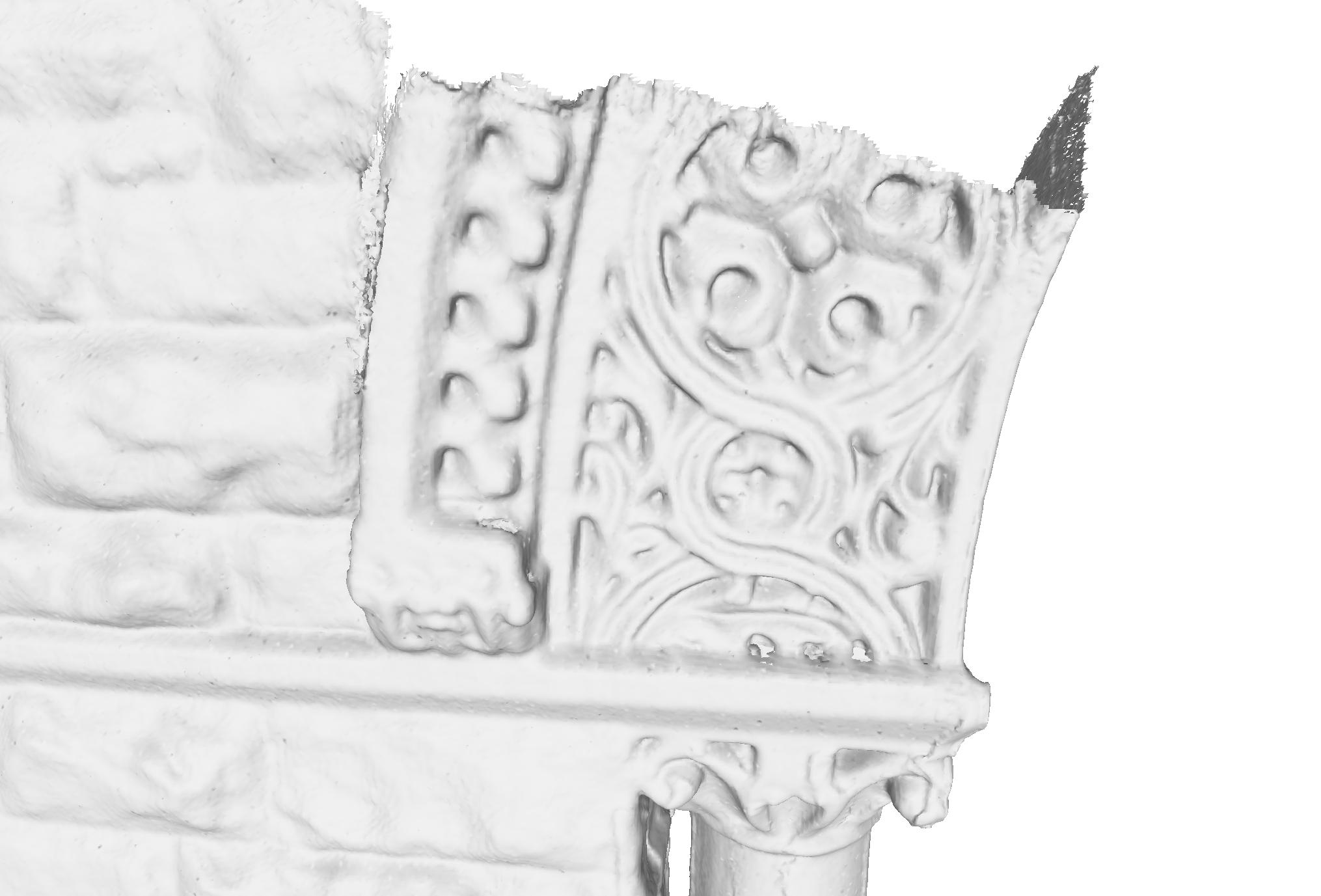} \\
 		  \includegraphics[width=1\textwidth]{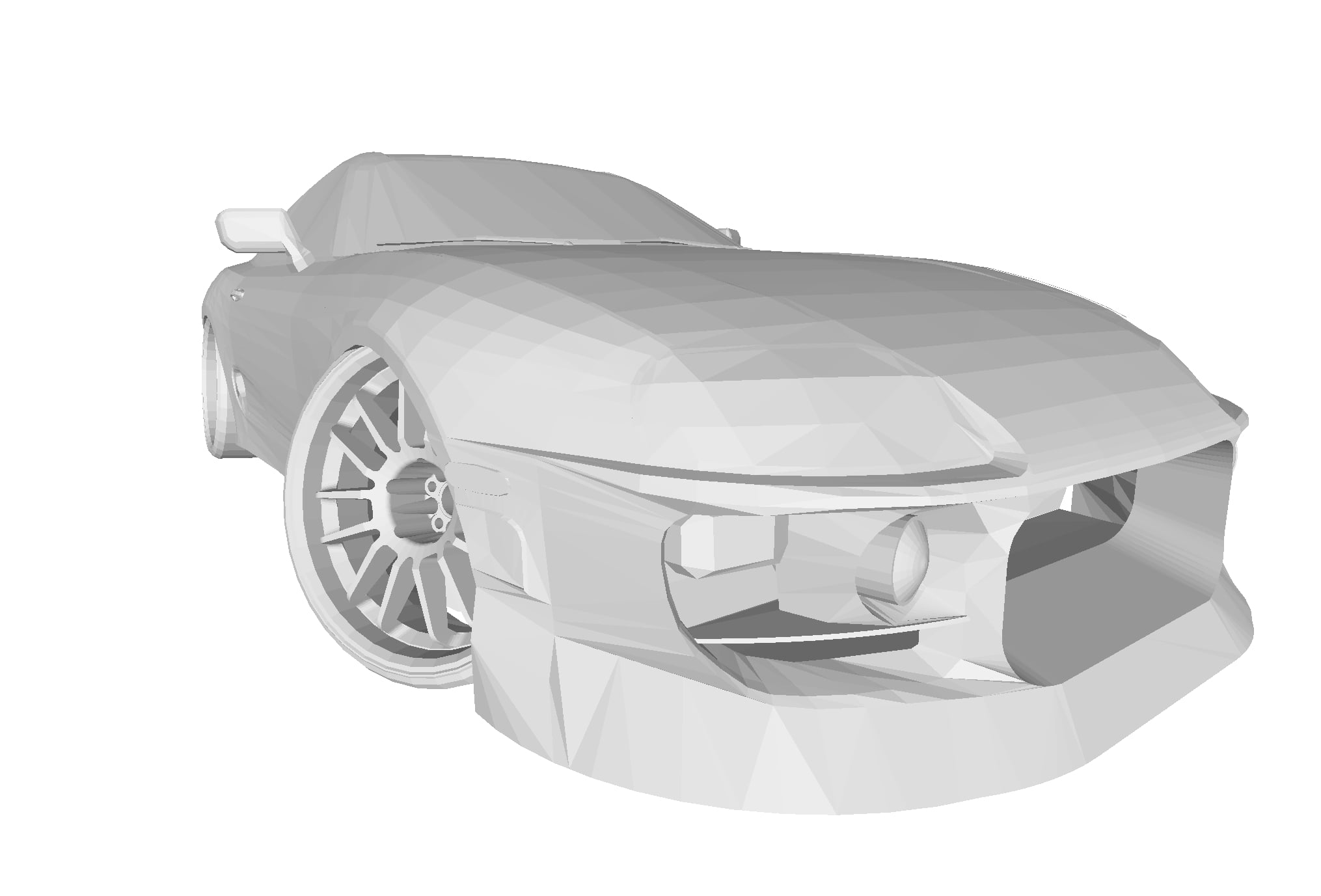} \\
 		  \includegraphics[width=1\textwidth]{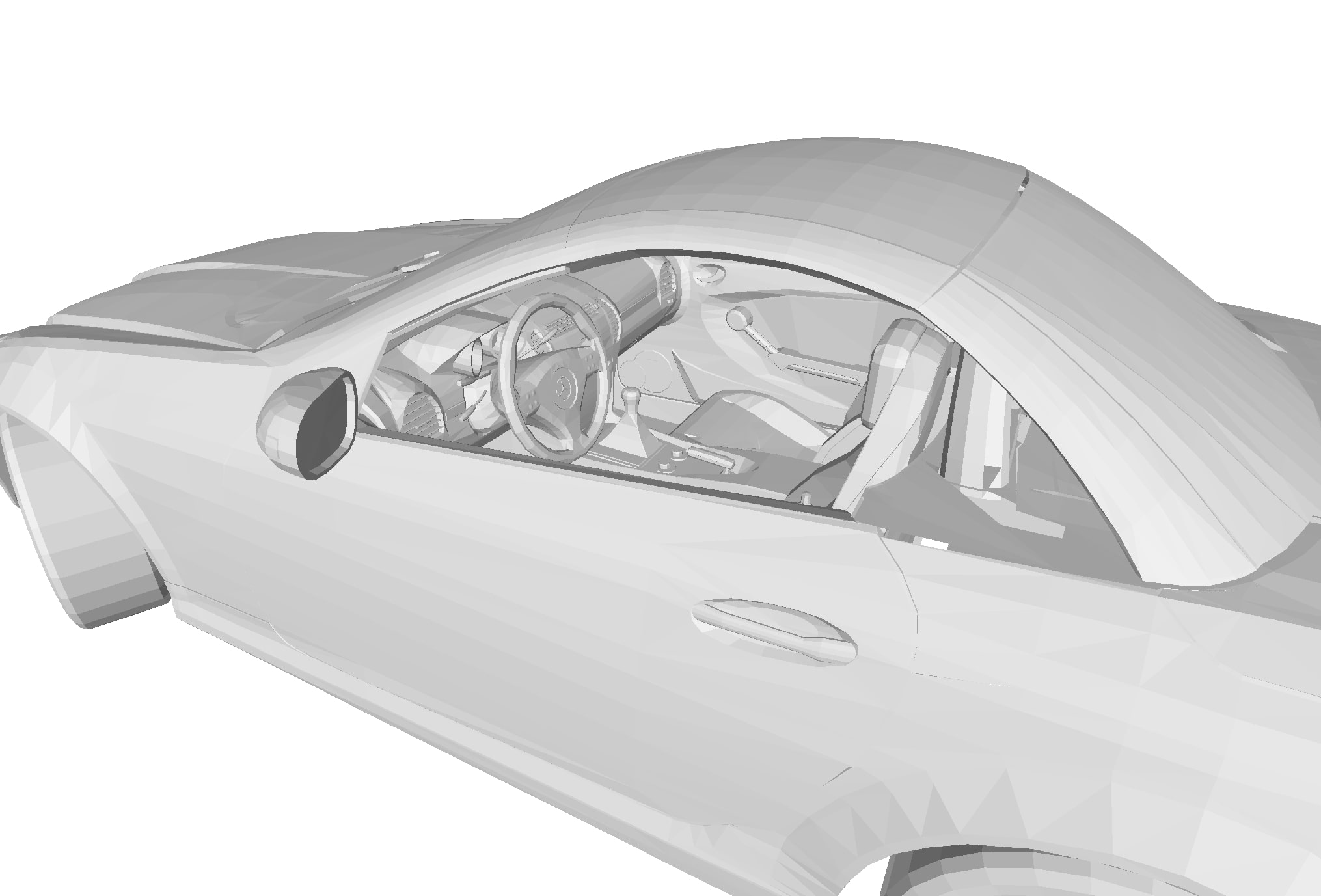} \\
 		  \includegraphics[width=1\textwidth]{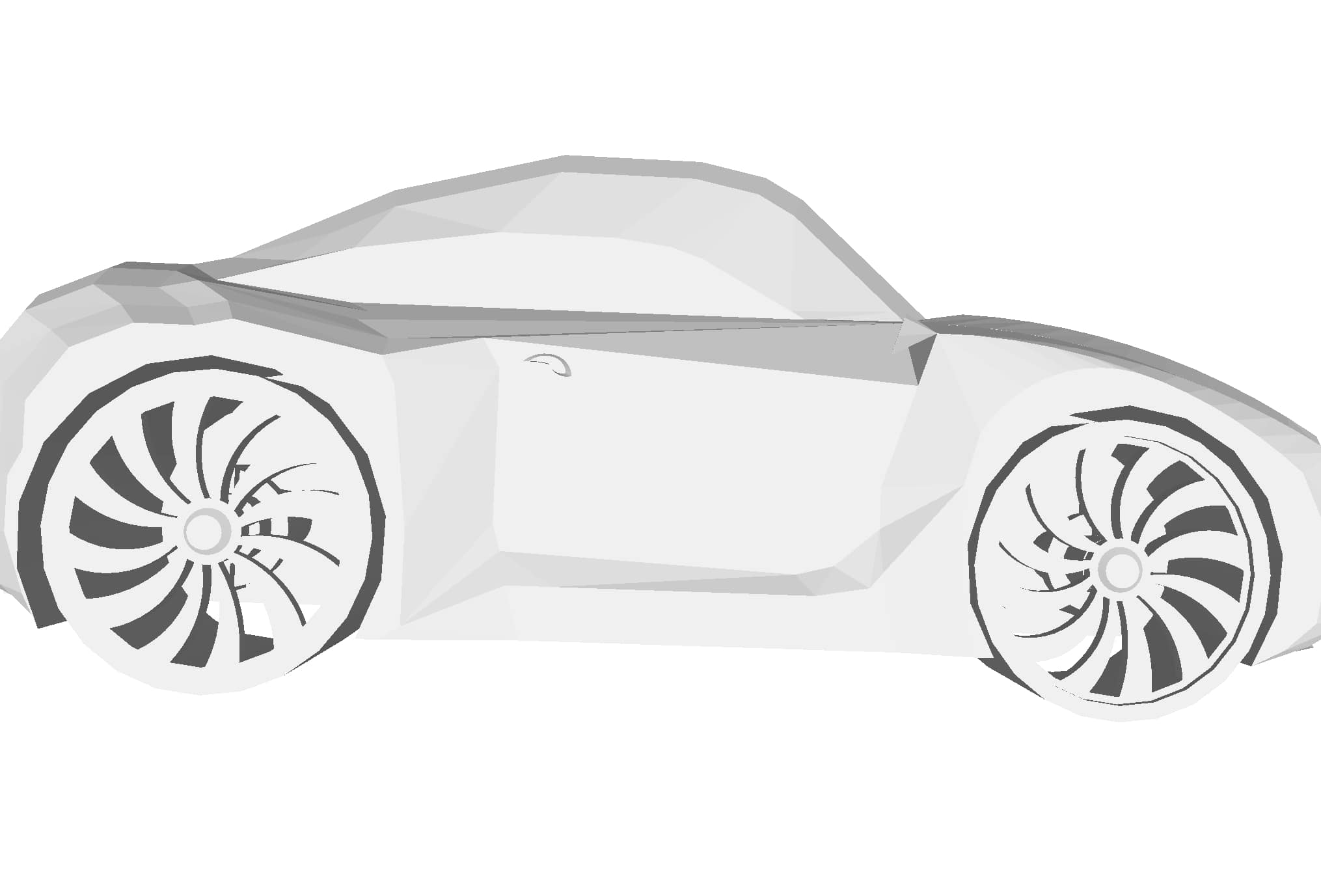} \\
 		  \includegraphics[width=1\textwidth]{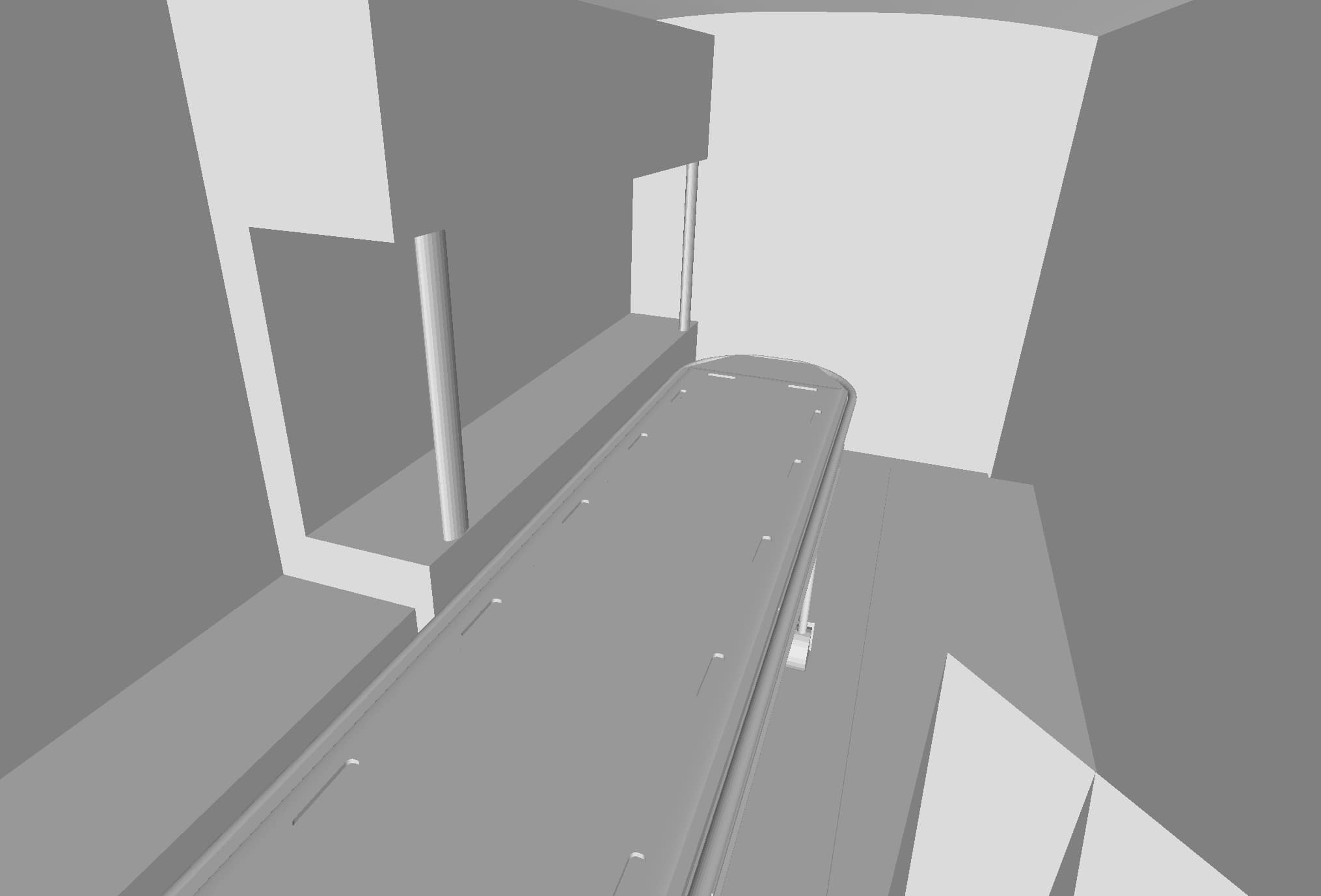} \\
        \end{minipage}
    }
    \caption{Visual comparison with CAP-UDF, DUDF, LevelSetUDF, and our DEUDF across various test models. To eliminate the impact of adopting DCUDF for extracting zero level sets from the learned UDF, we utilize the same zero level set extraction technique as originally proposed/used for each method. Still, our method consistently delivers results with higher quality, characterized by more detailed geometric features and smoother shape boundaries.}
    \label{fig:officail-comparison}

\end{figure*}

\begin{figure*}[!htbp]
    \centering
        \subfigure[DUDF]{
        \begin{minipage}[b]{0.18\textwidth}
		  \includegraphics[width=1\textwidth]{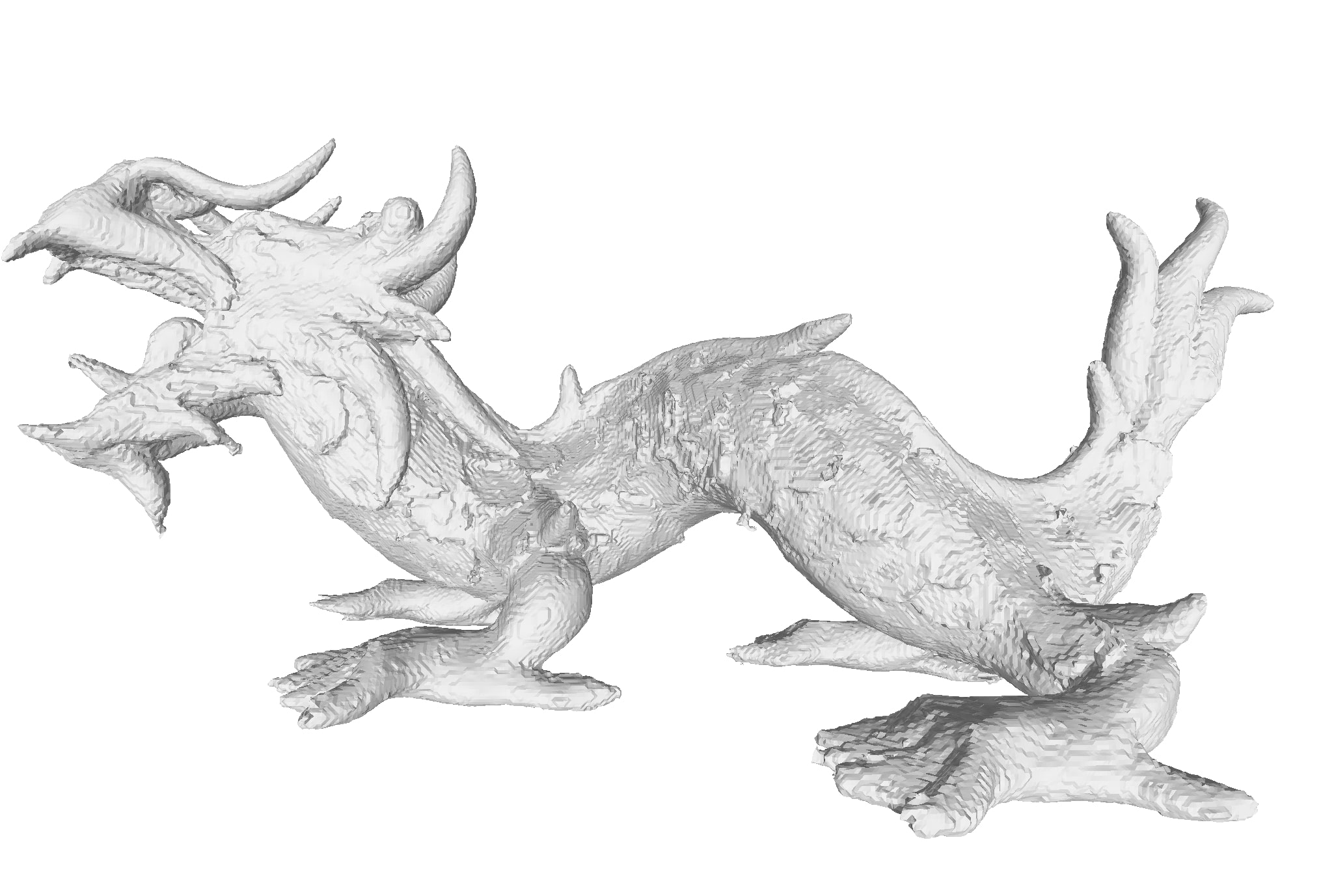} \\
		  \includegraphics[width=1\textwidth]{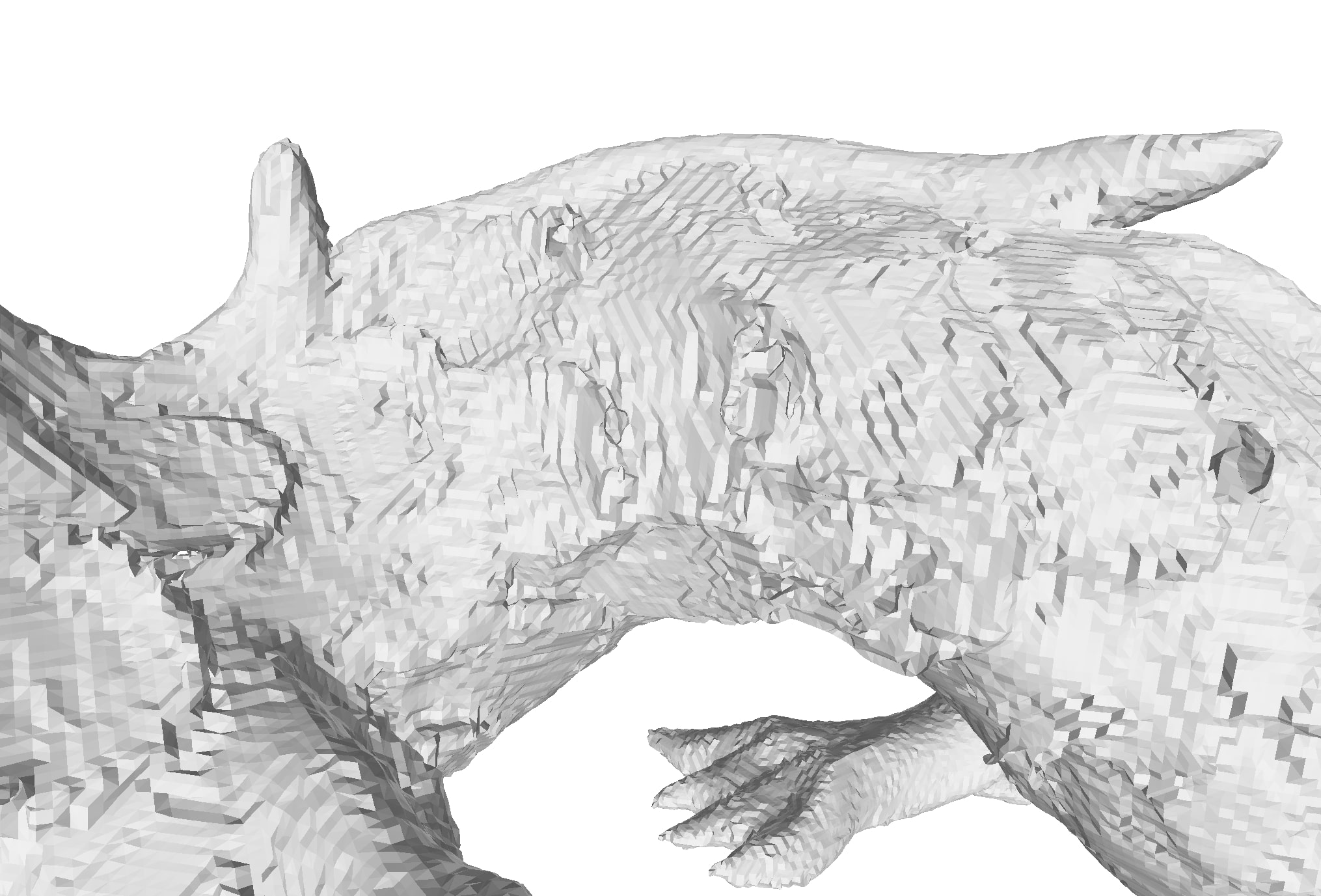} \\
    	\includegraphics[width=1\textwidth]{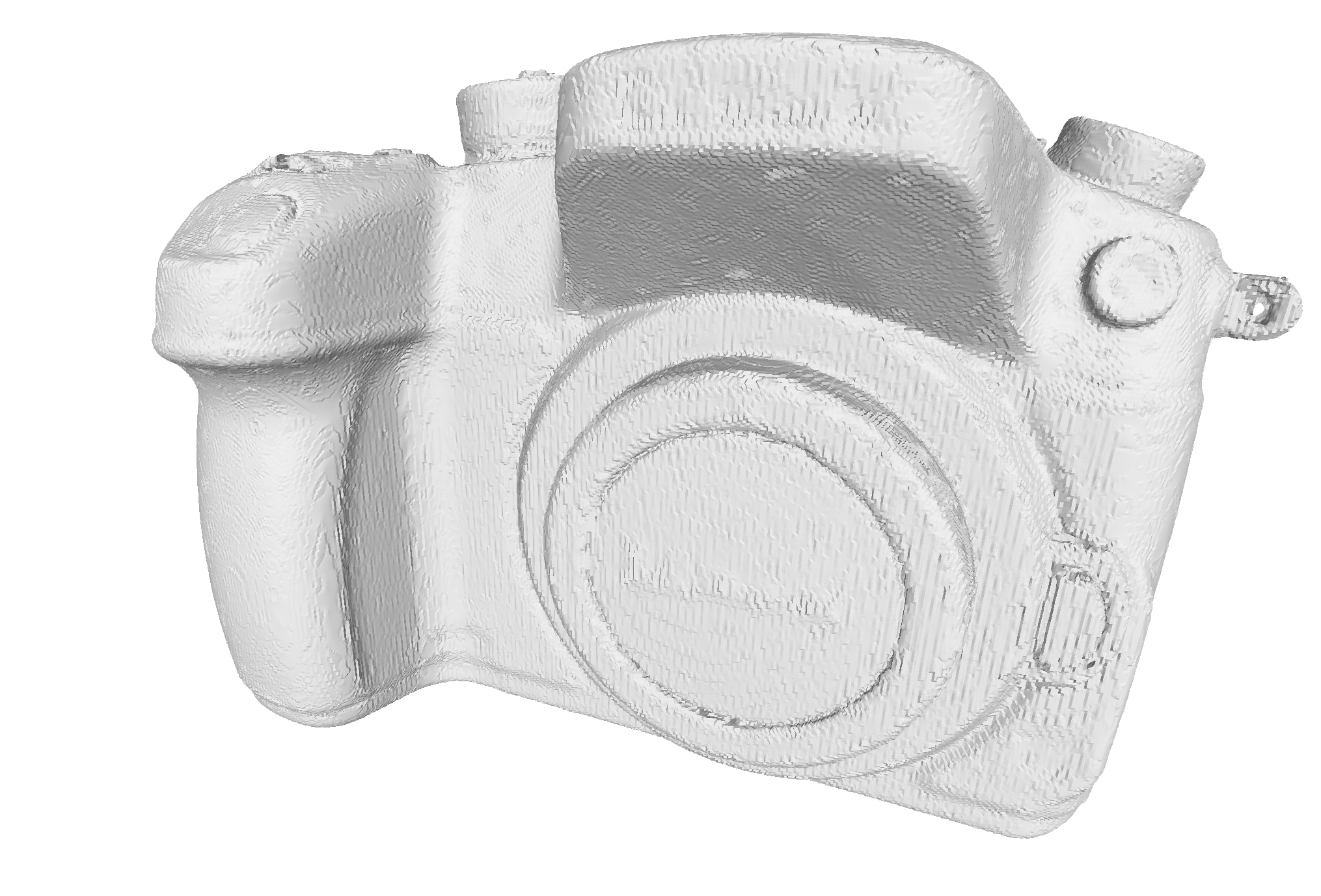} \\
            \includegraphics[width=1\textwidth]{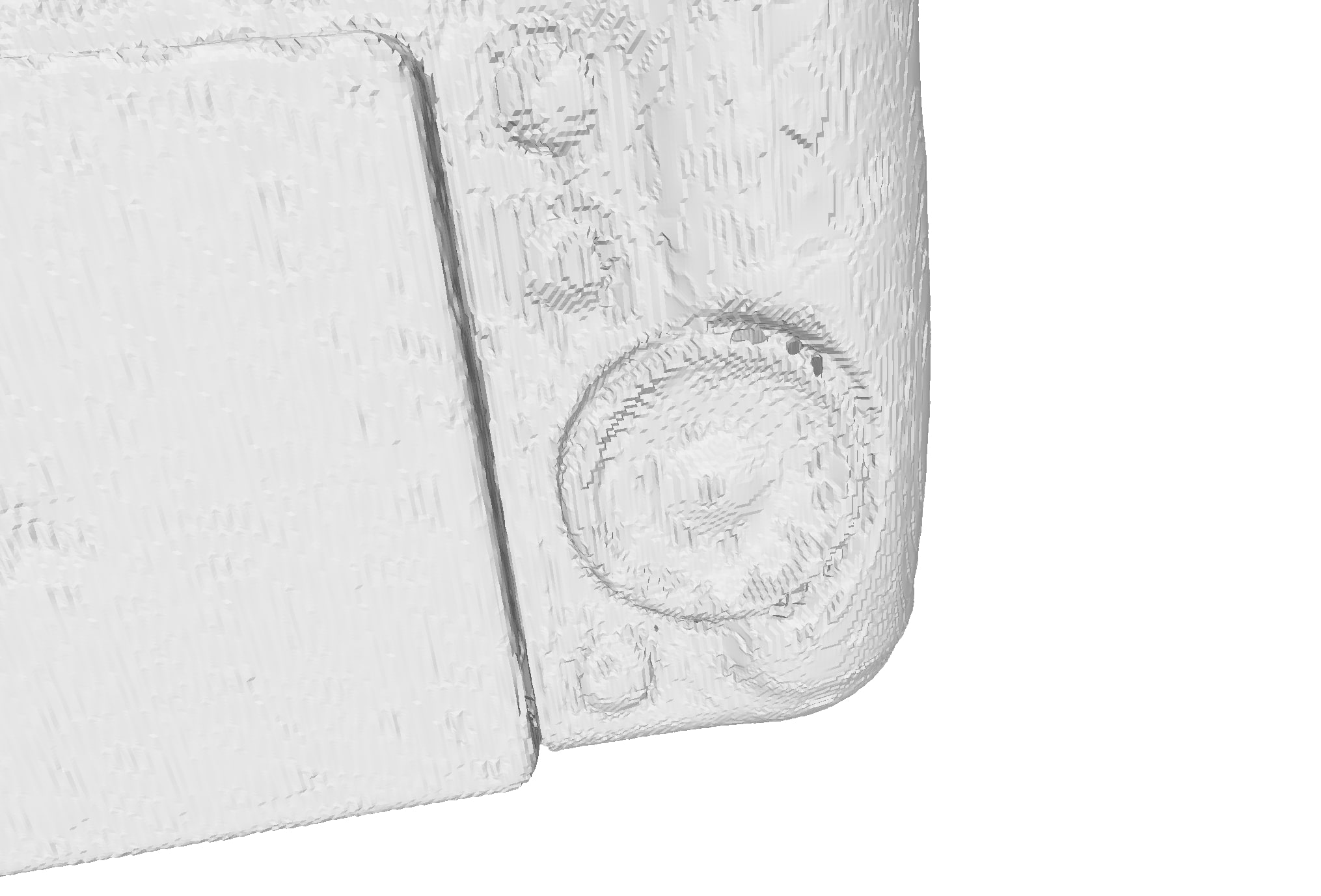} \\
    	\includegraphics[width=1\textwidth]{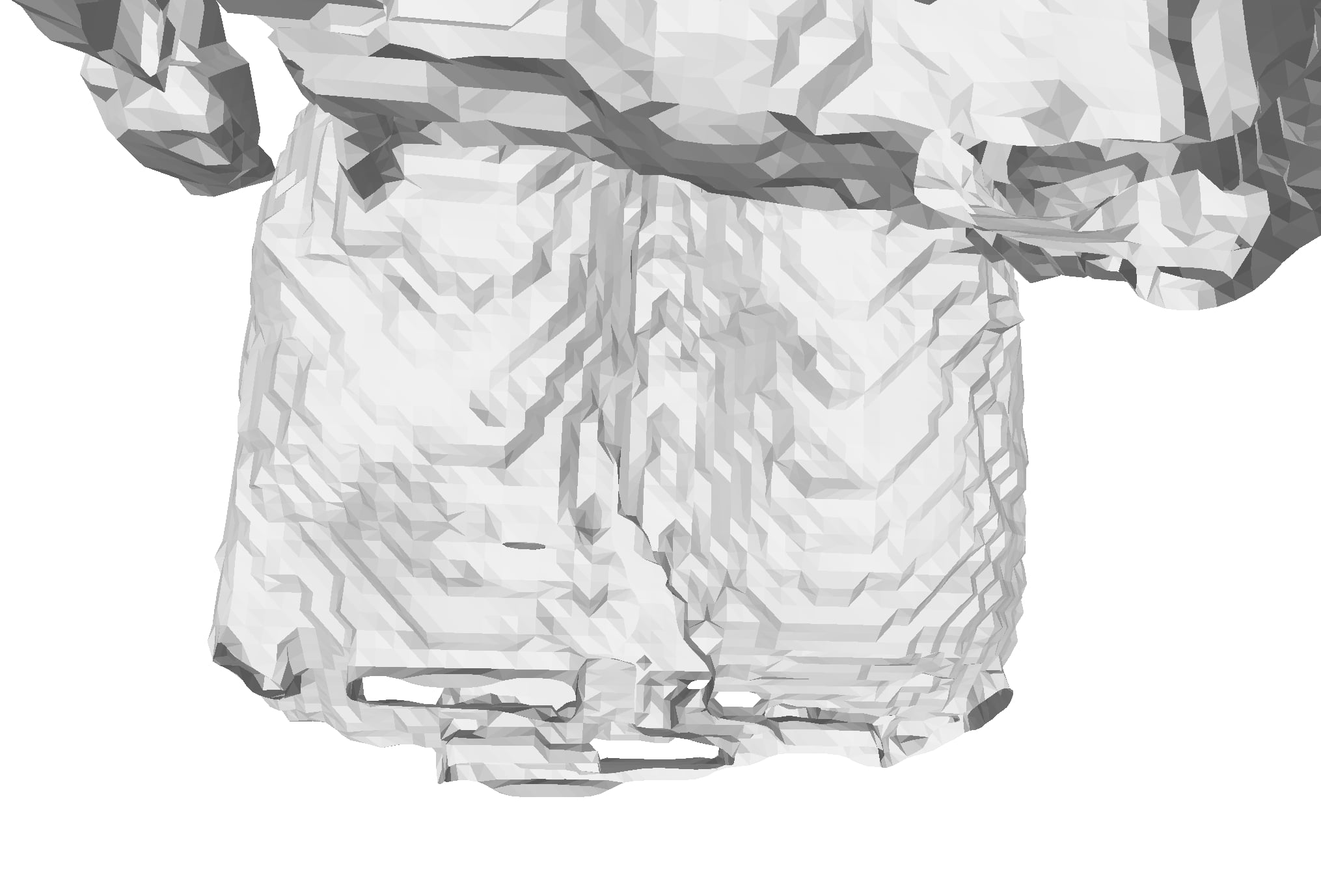} \\
            \includegraphics[width=1\textwidth]{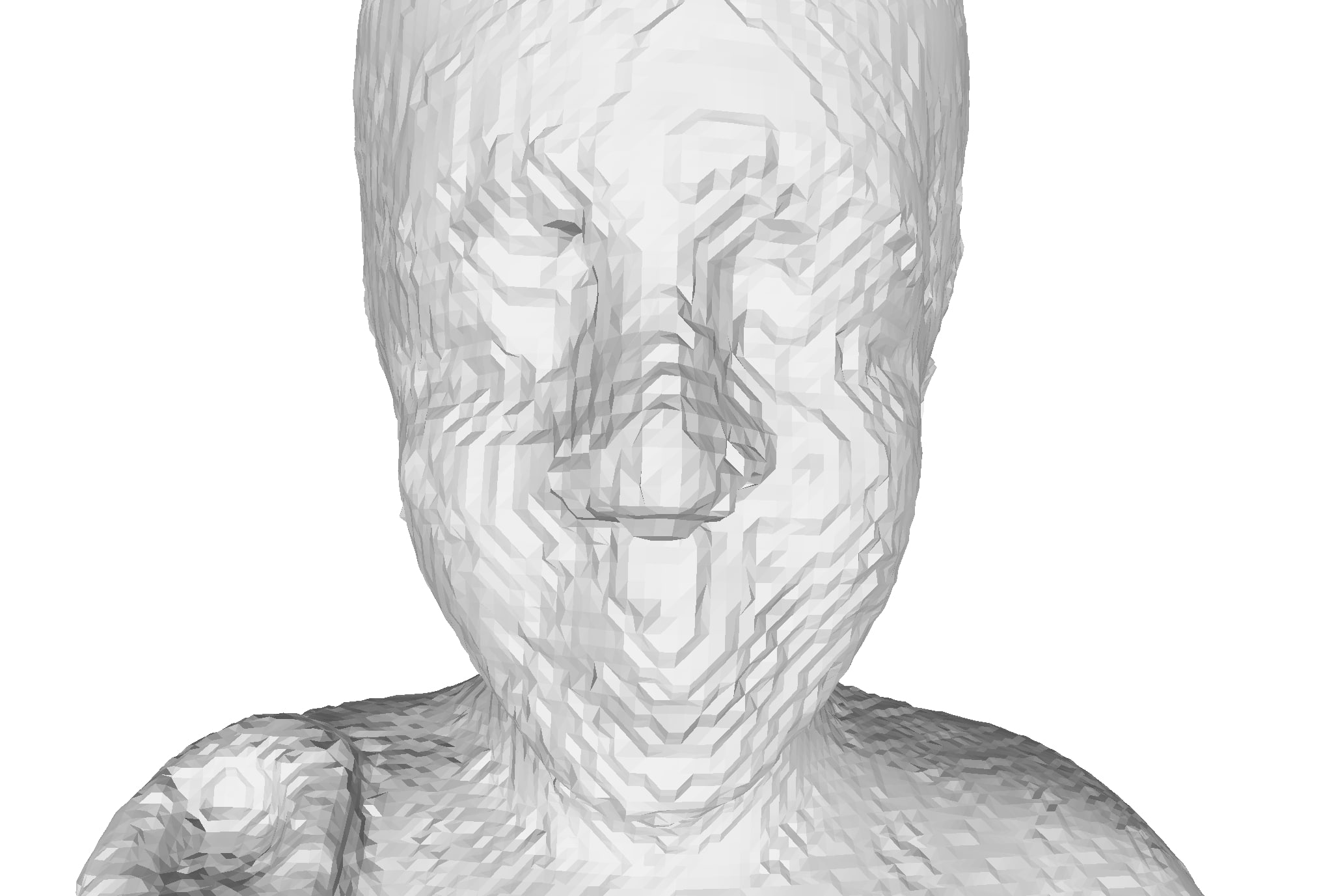} \\
    	\includegraphics[width=1\textwidth]{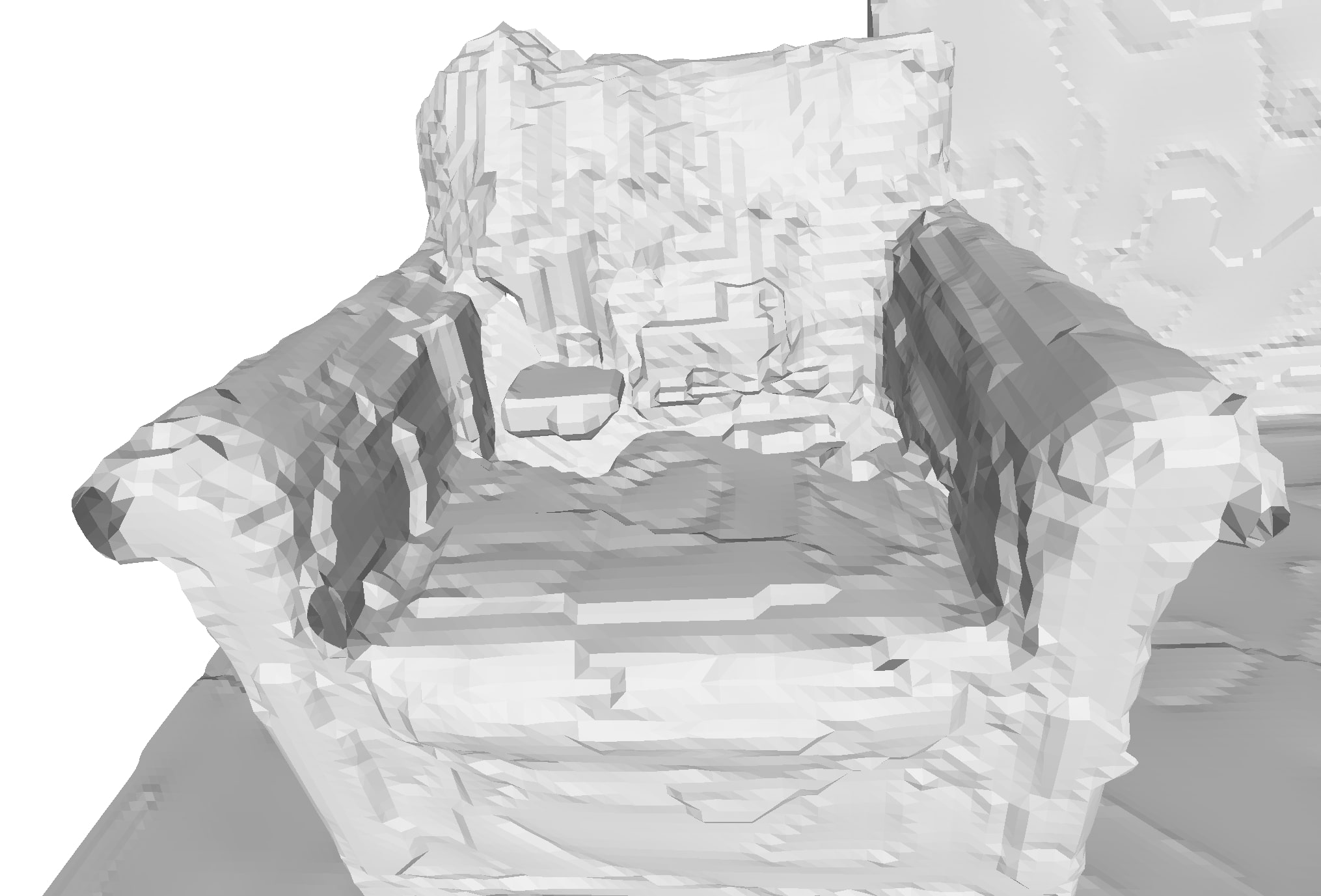} 
            \includegraphics[width=1\textwidth]{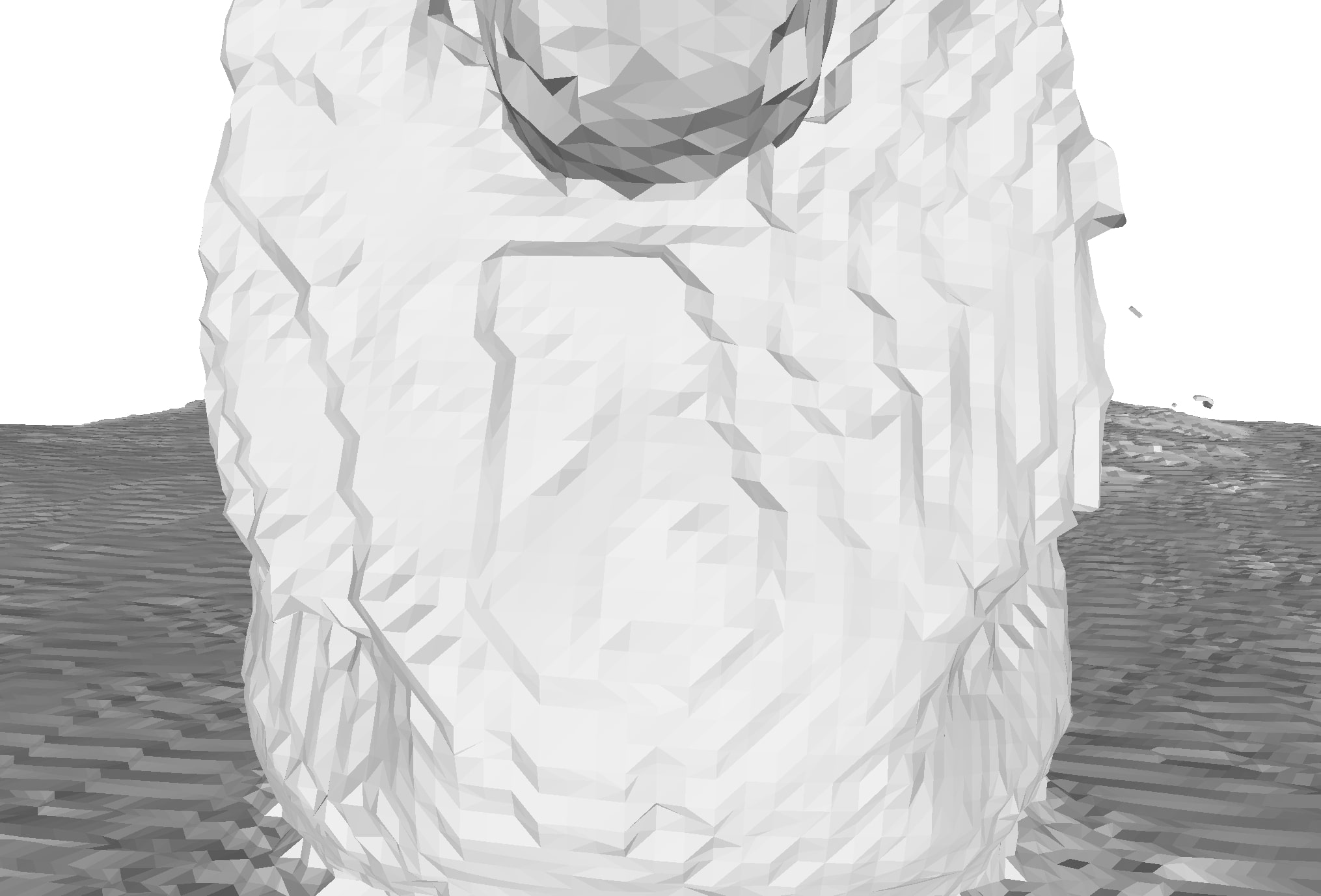} 

        \end{minipage}
    }
    \subfigure[CAP-UDF]{
        \begin{minipage}[b]{0.18\textwidth}
		  \includegraphics[width=1\textwidth]{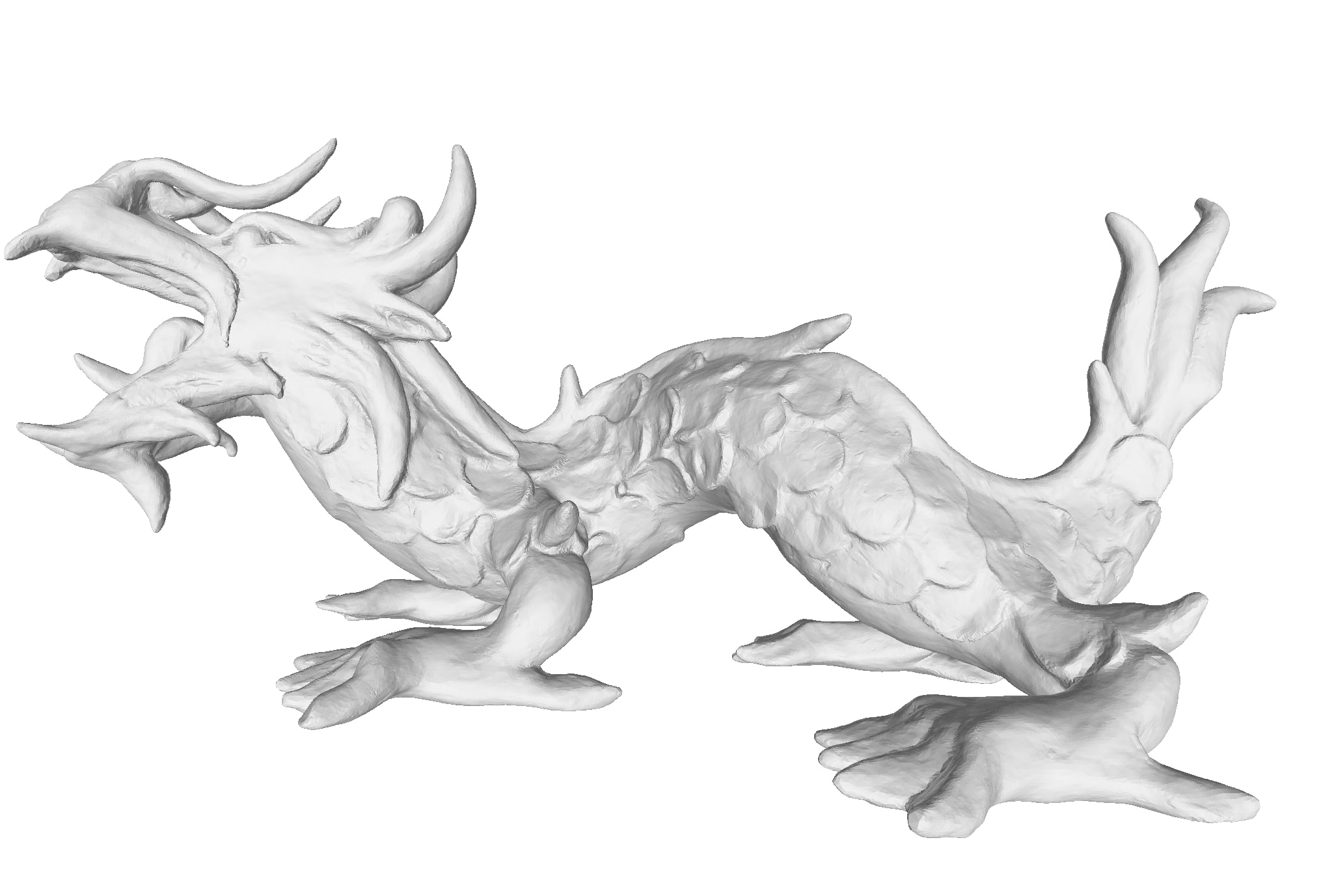} \\
		  \includegraphics[width=1\textwidth]{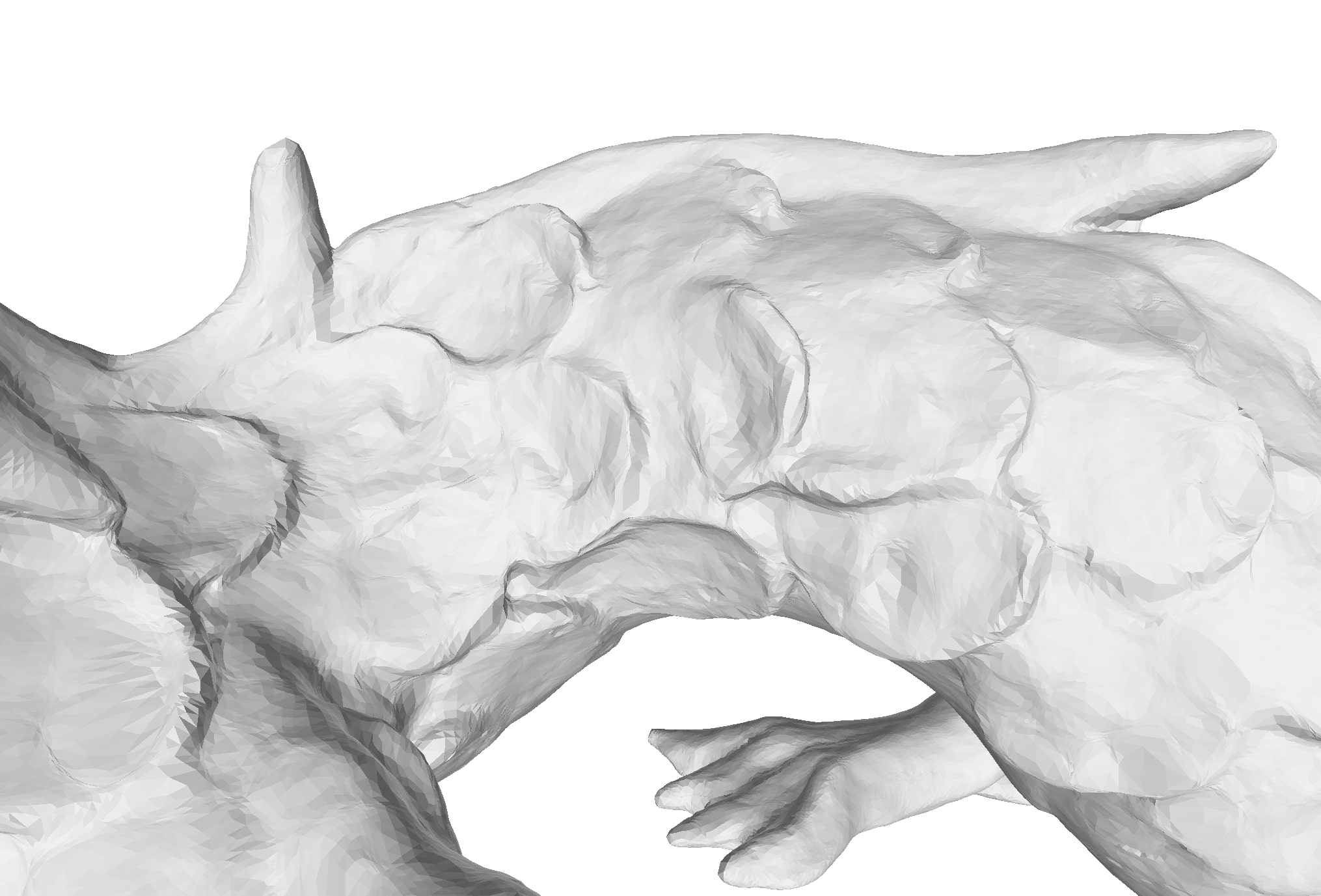} \\
    	\includegraphics[width=1\textwidth]{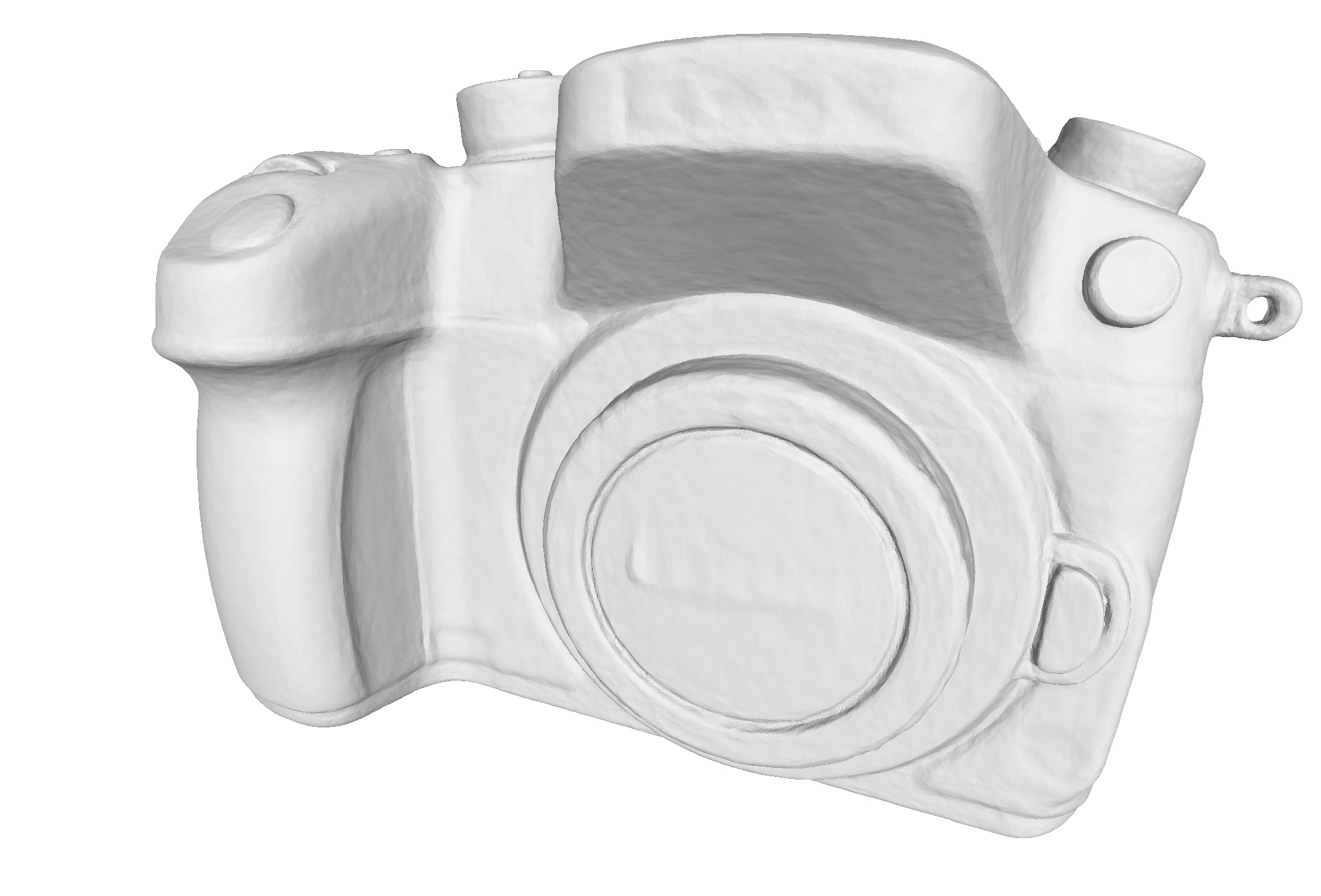} \\
            \includegraphics[width=1\textwidth]{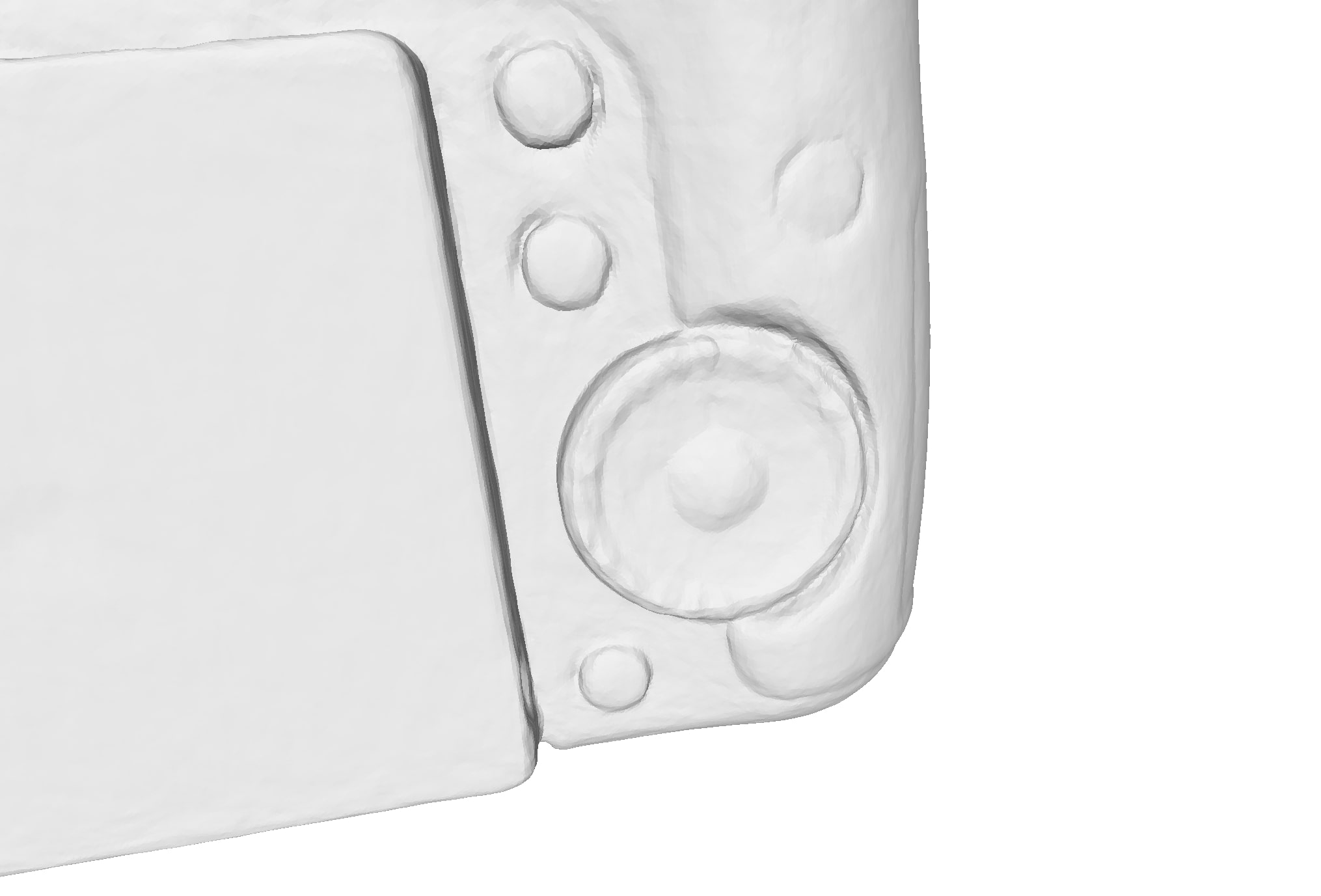} \\
    	\includegraphics[width=1\textwidth]{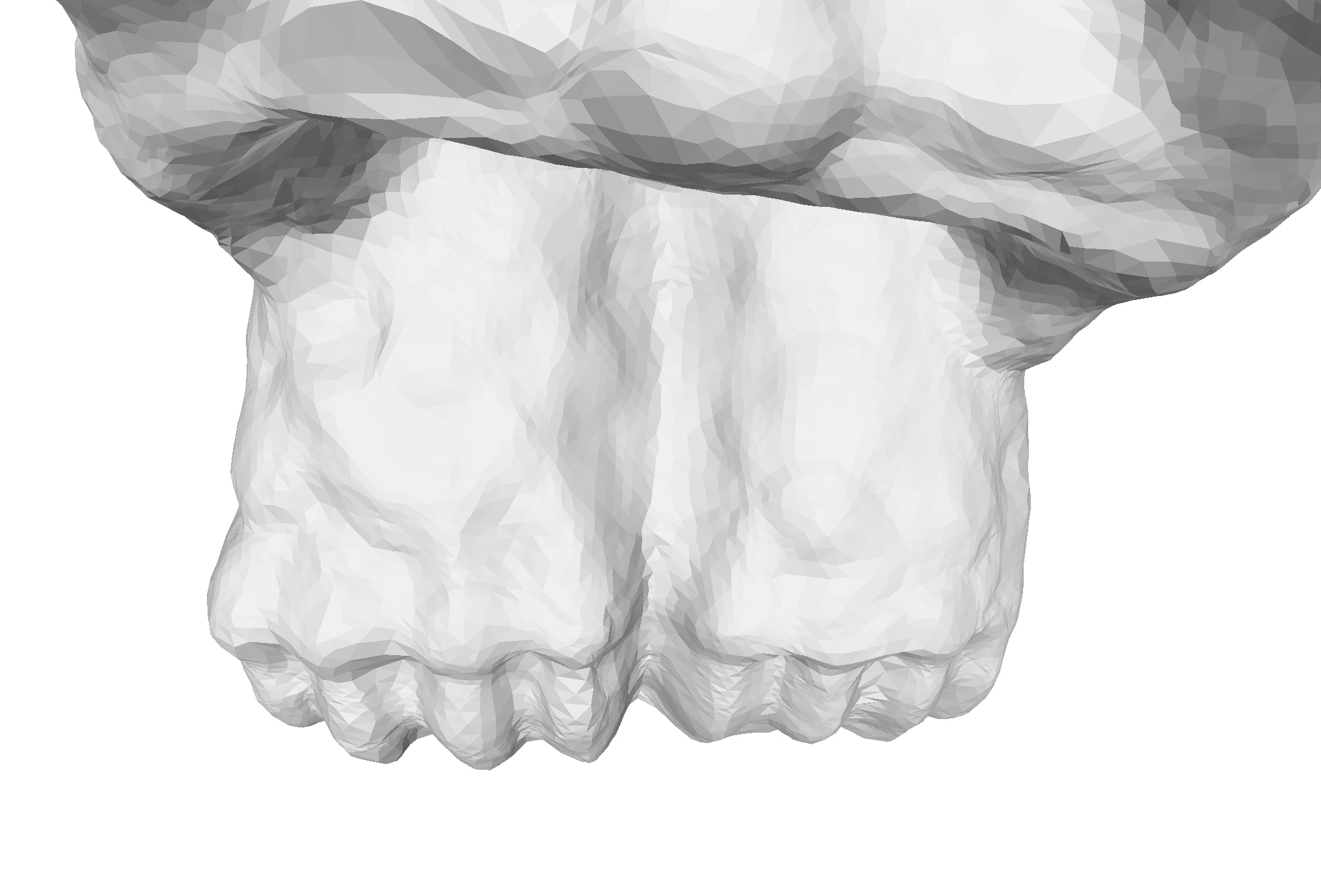} \\
            \includegraphics[width=1\textwidth]{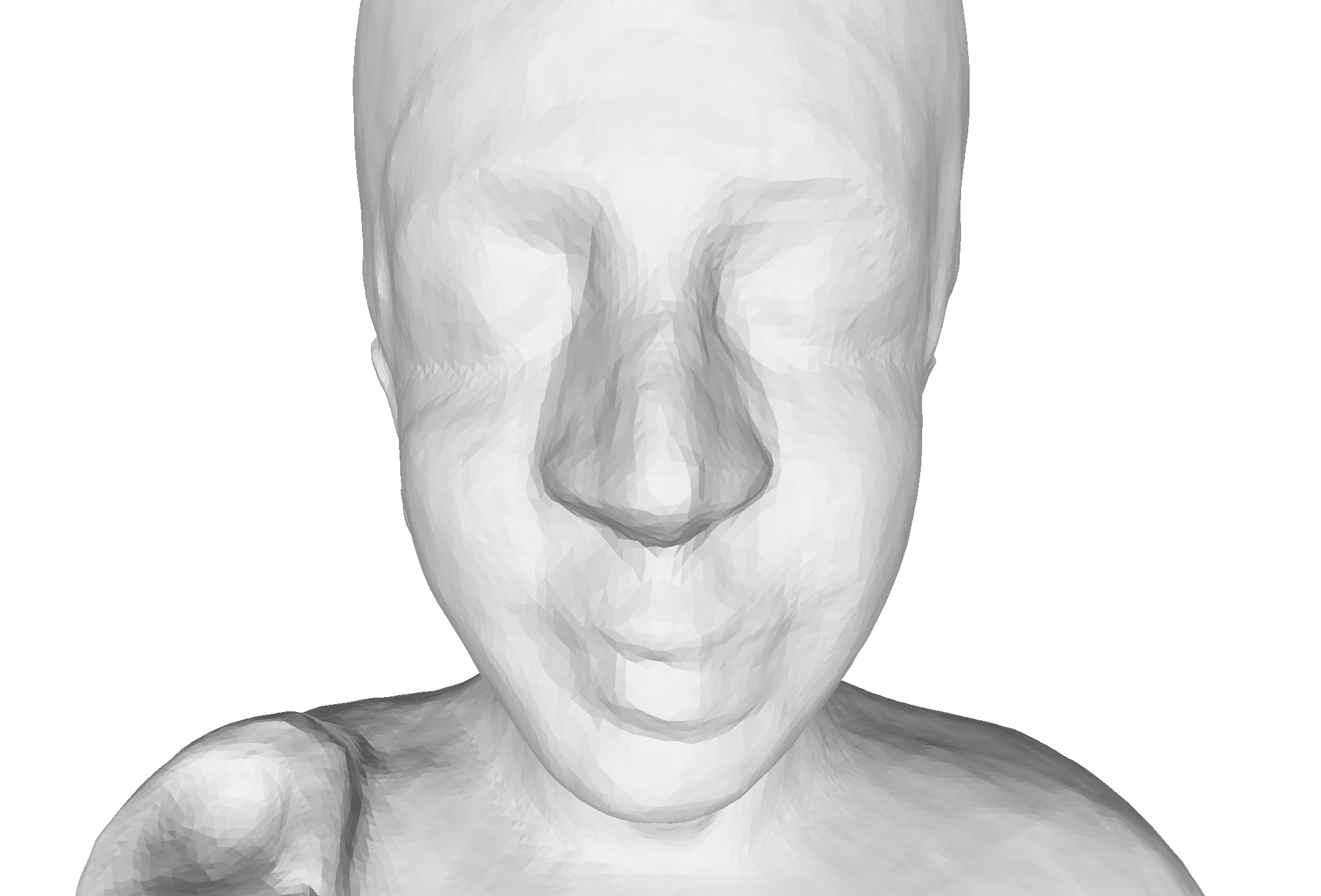} \\
    	\includegraphics[width=1\textwidth]{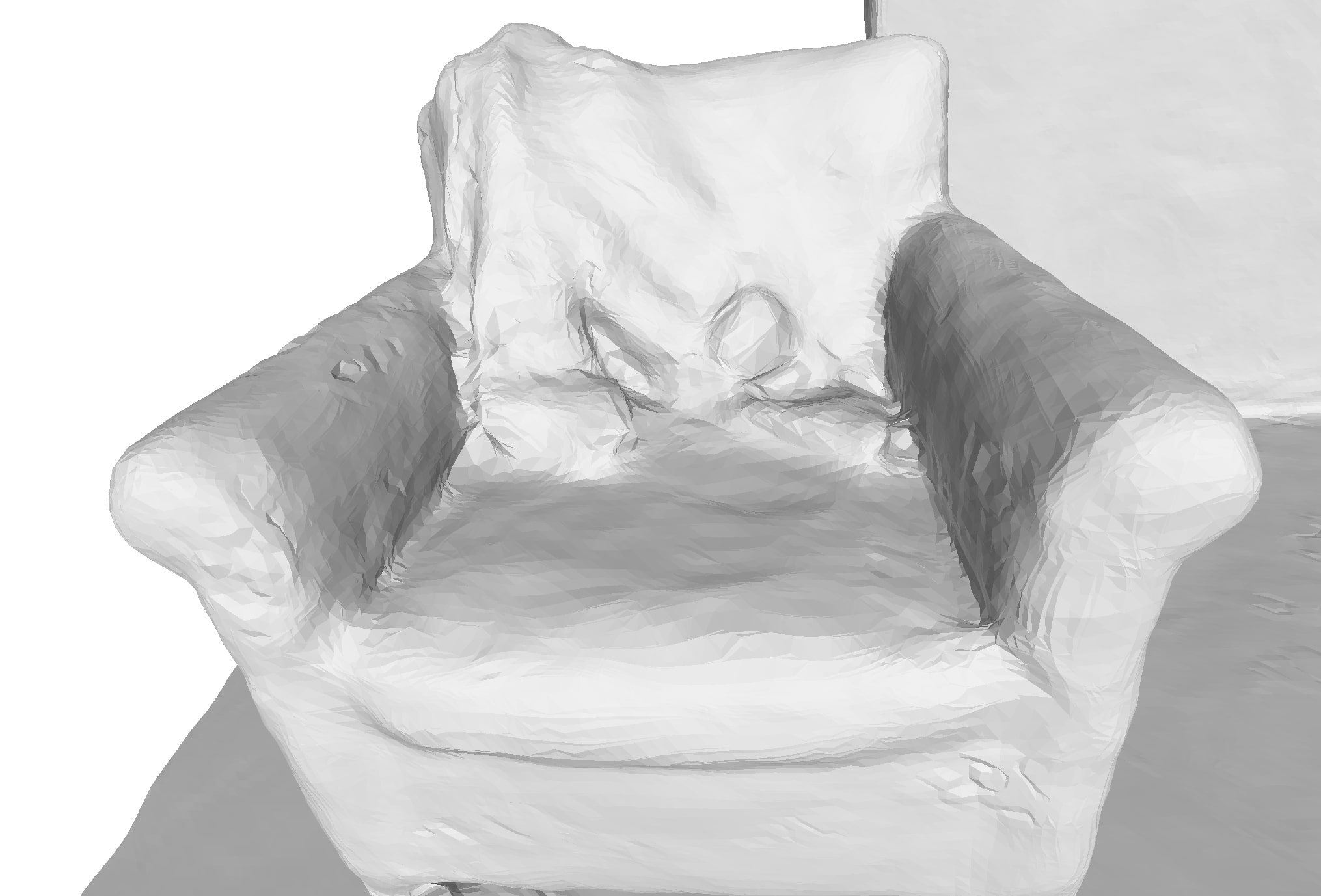} 
            \includegraphics[width=1\textwidth]{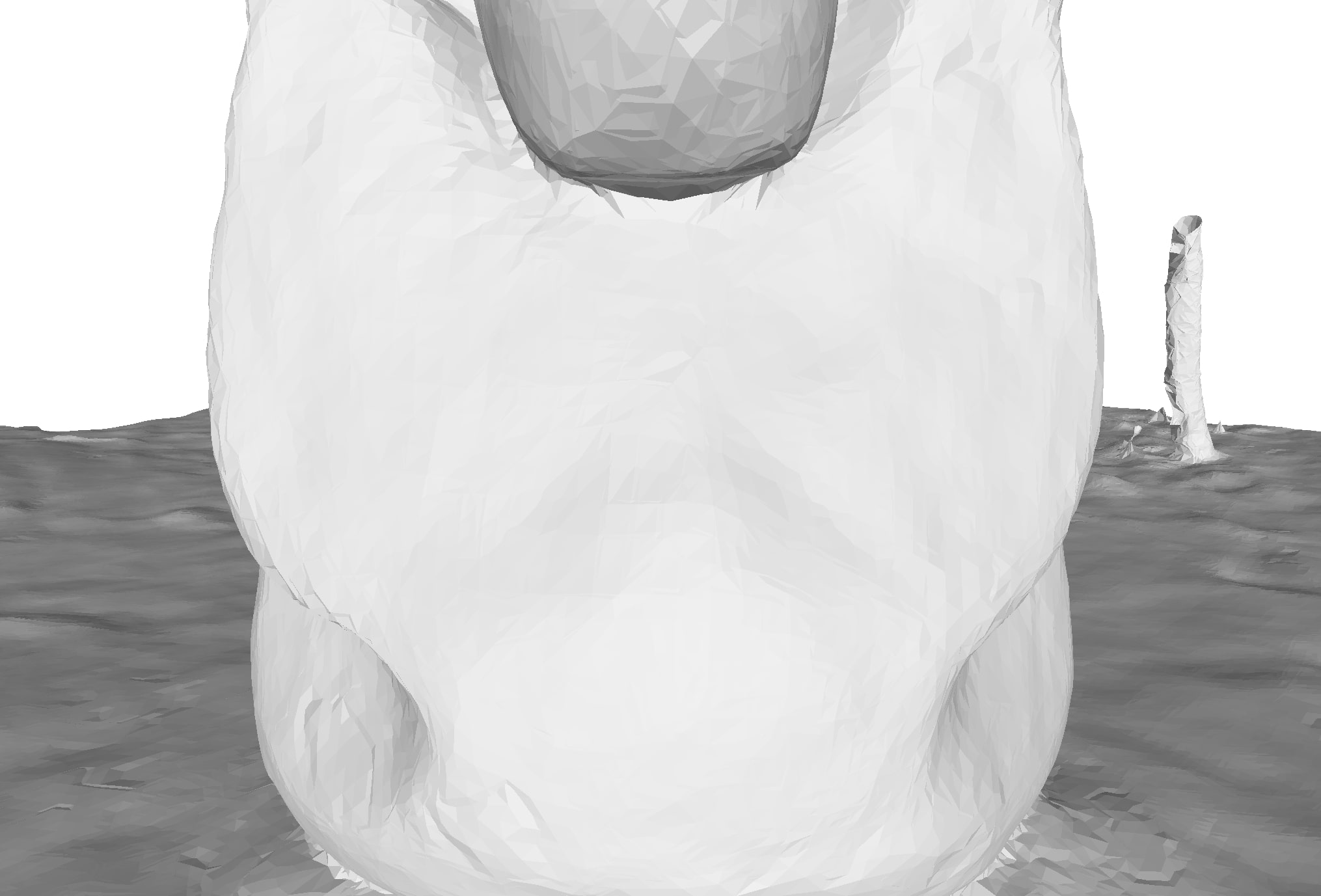} 
        \end{minipage}
    }
    \subfigure[LevelSetUDF]{
        \begin{minipage}[b]{0.18\textwidth}
		  \includegraphics[width=1\textwidth]{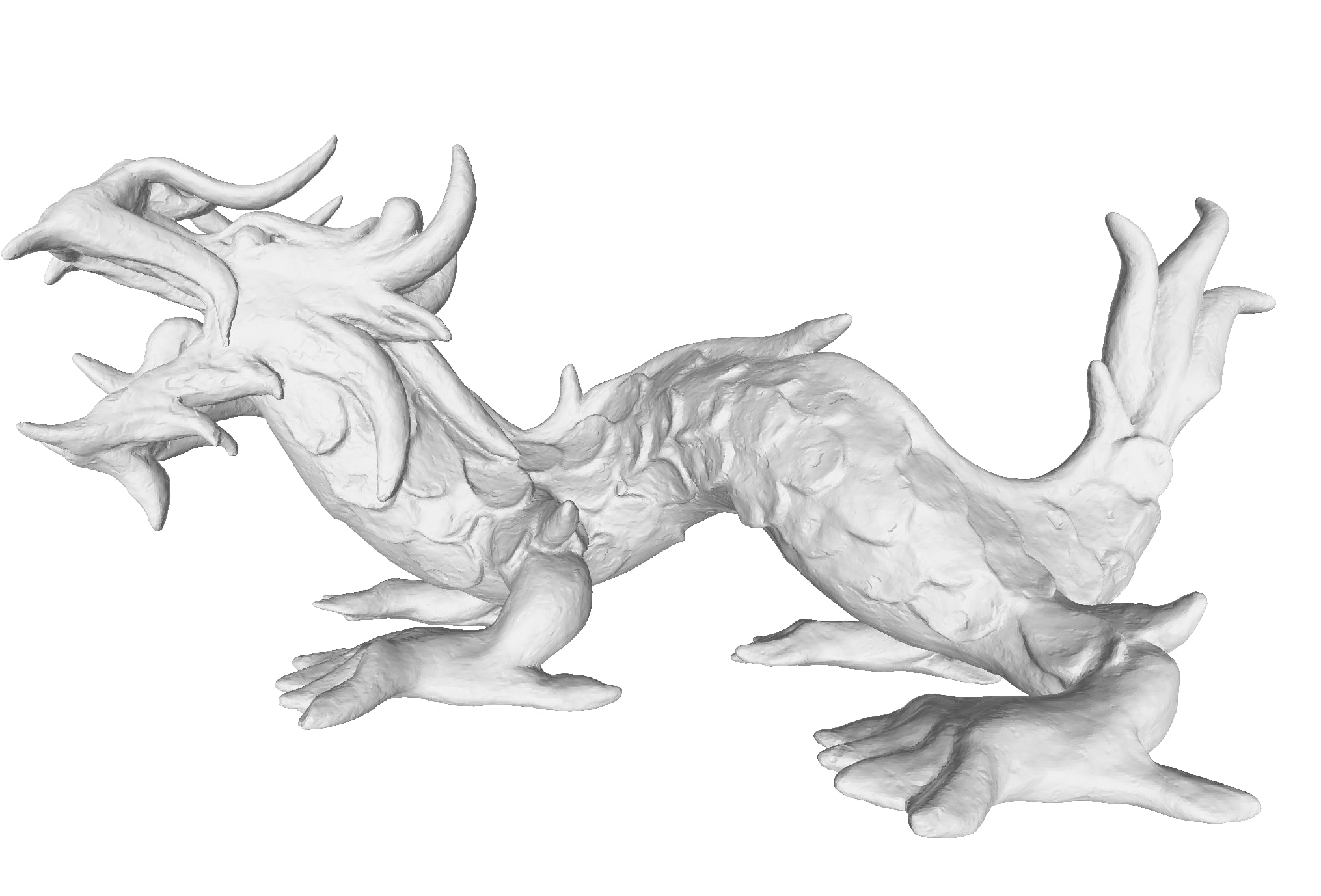} \\
		  \includegraphics[width=1\textwidth]{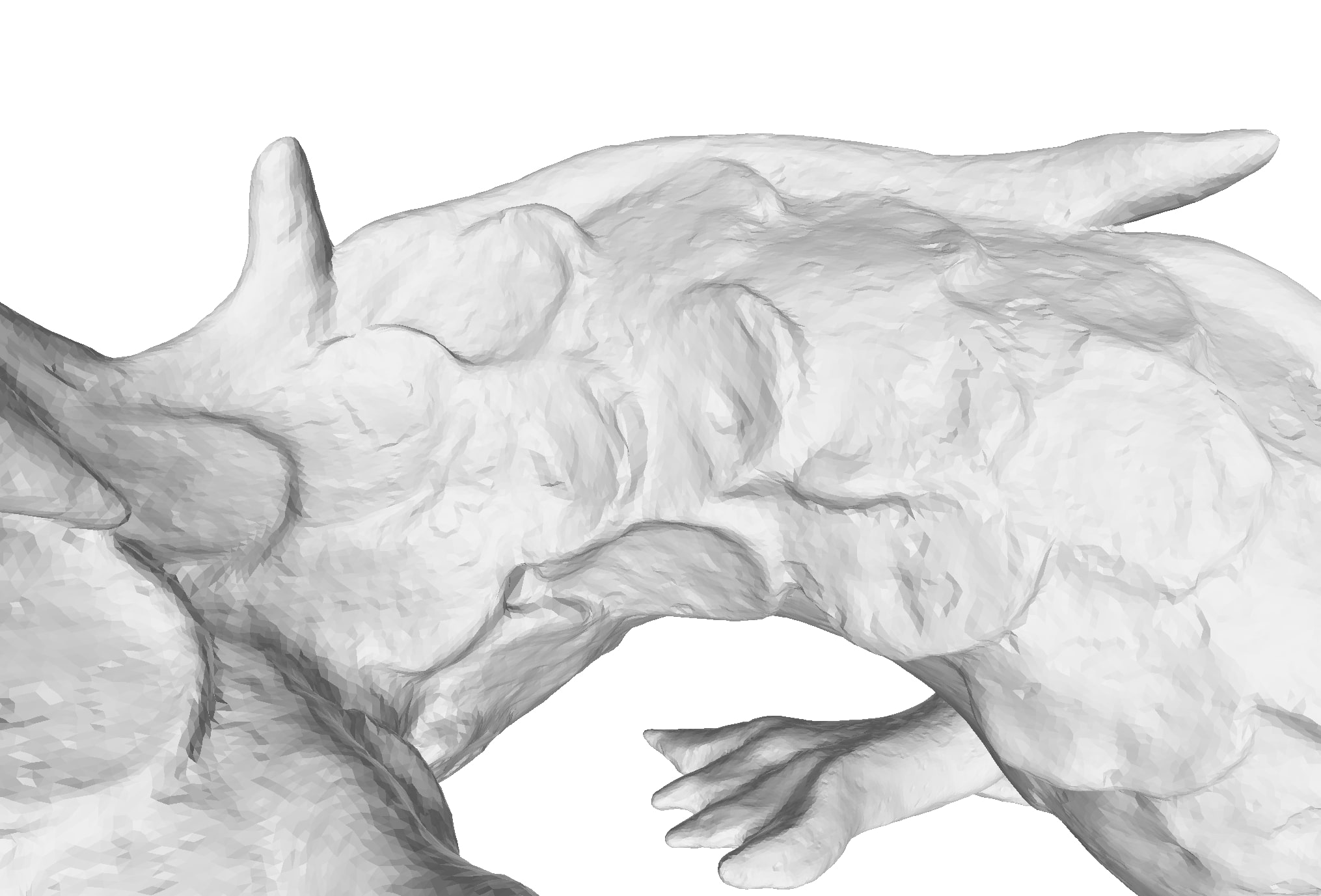} \\
    	\includegraphics[width=1\textwidth]{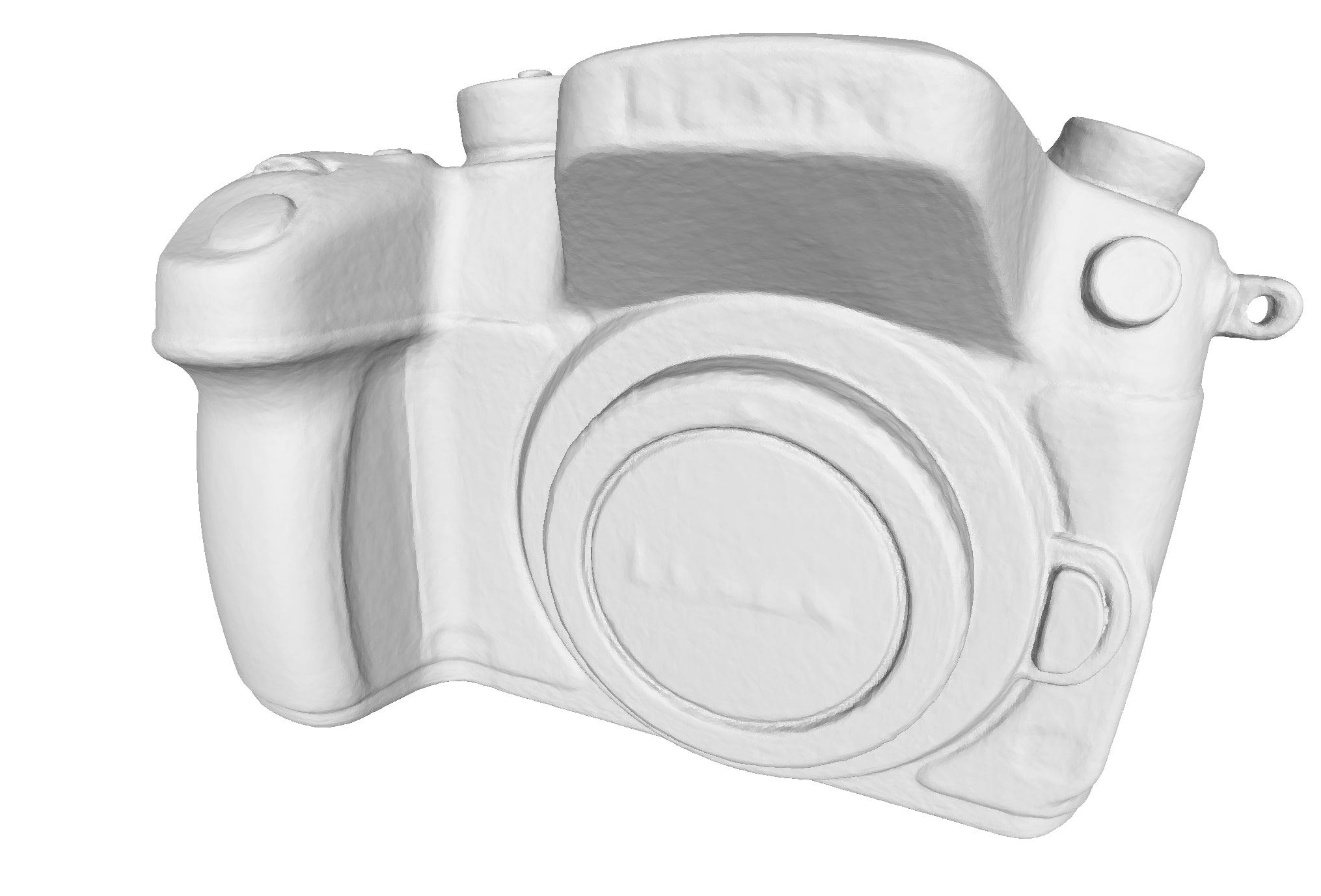} \\
            \includegraphics[width=1\textwidth]{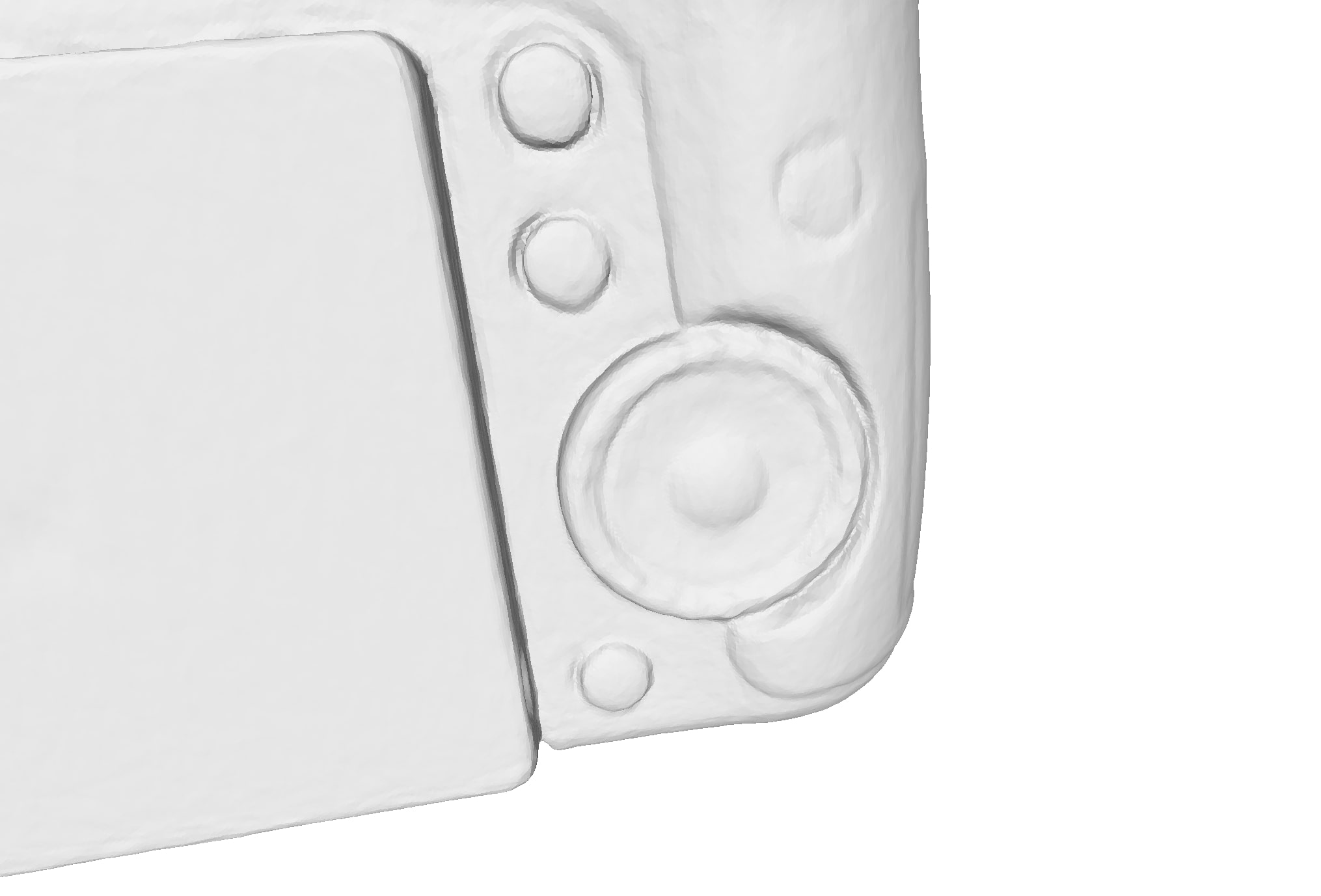} \\
    	\includegraphics[width=1\textwidth]{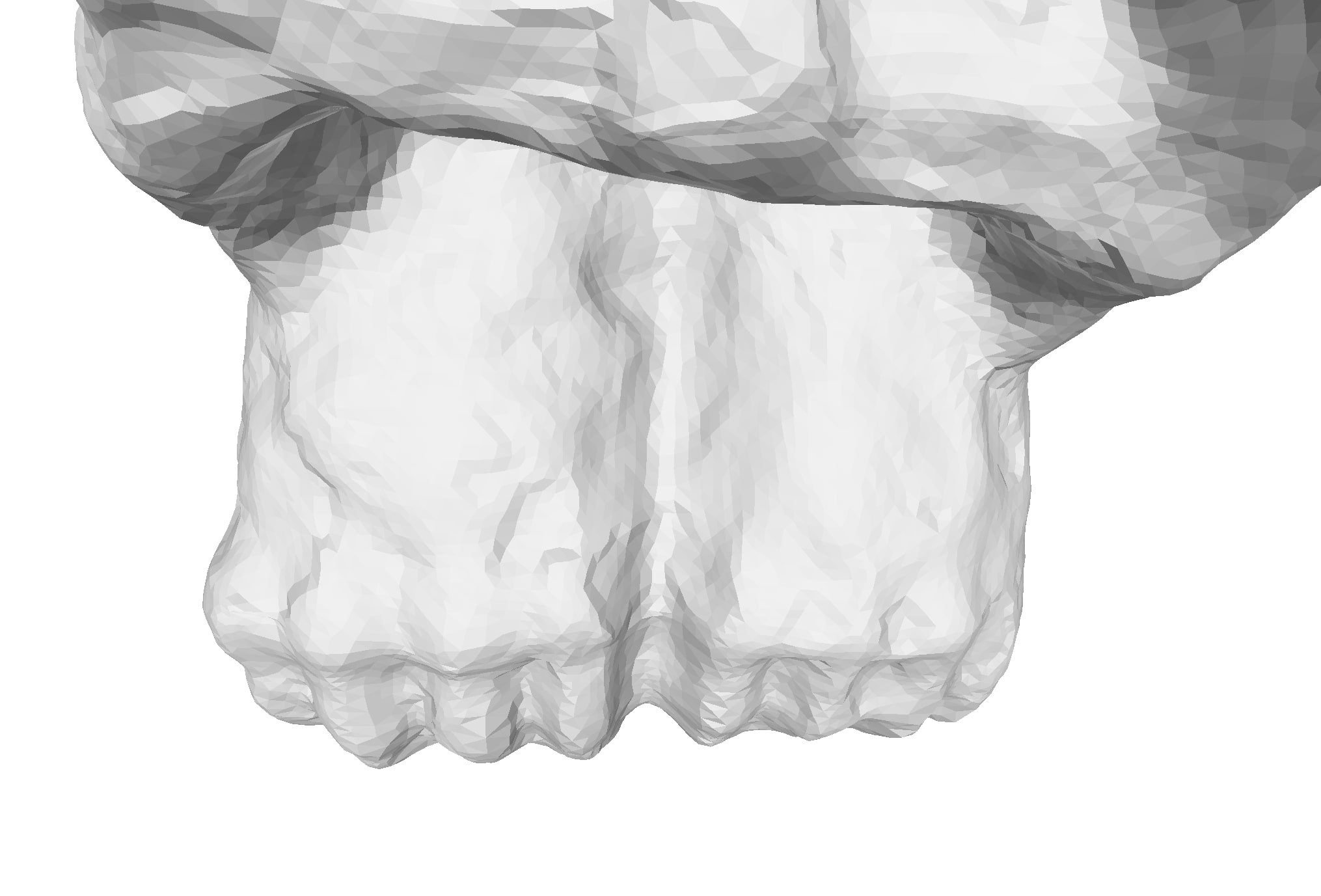} \\
            \includegraphics[width=1\textwidth]{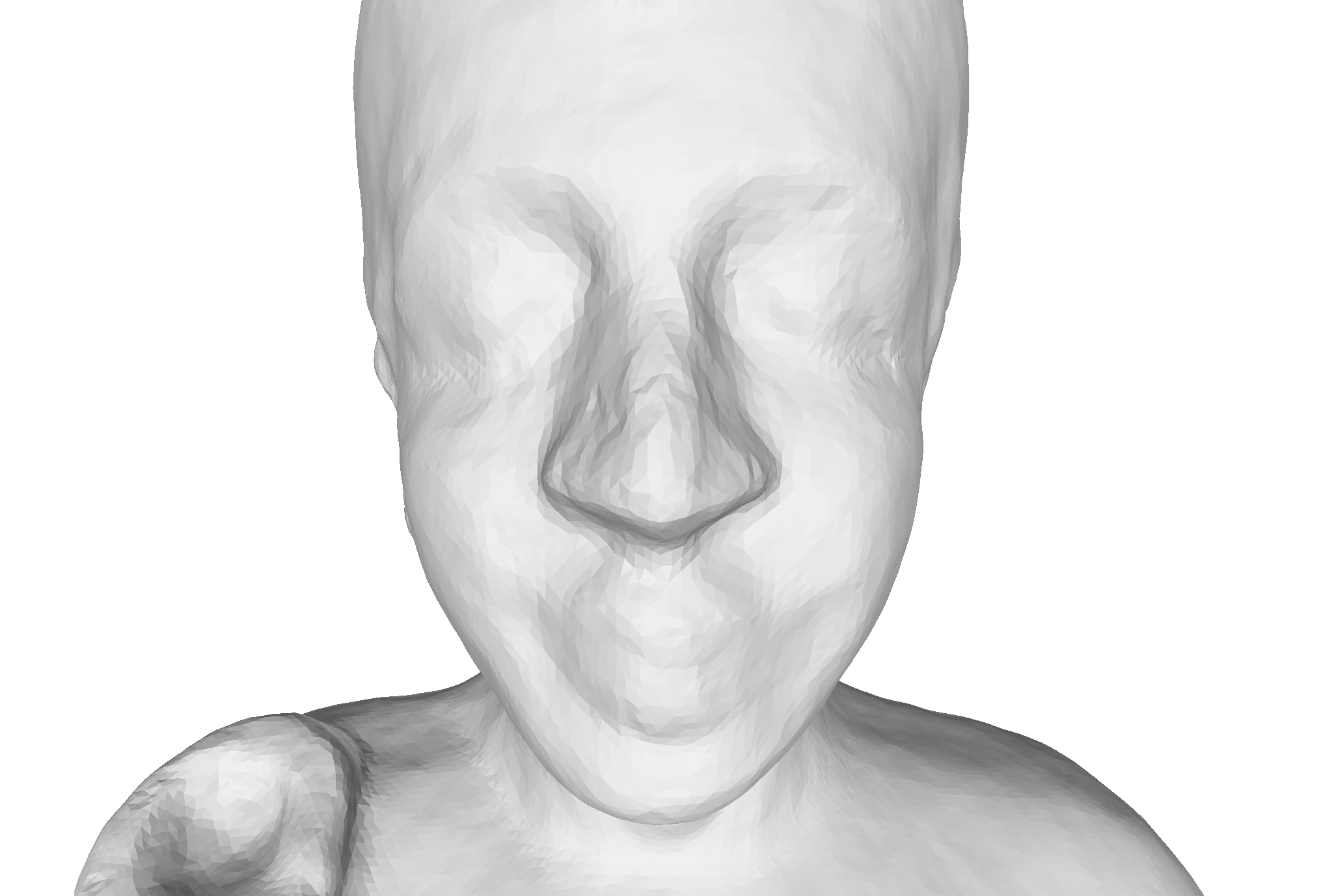} \\
    	\includegraphics[width=1\textwidth]{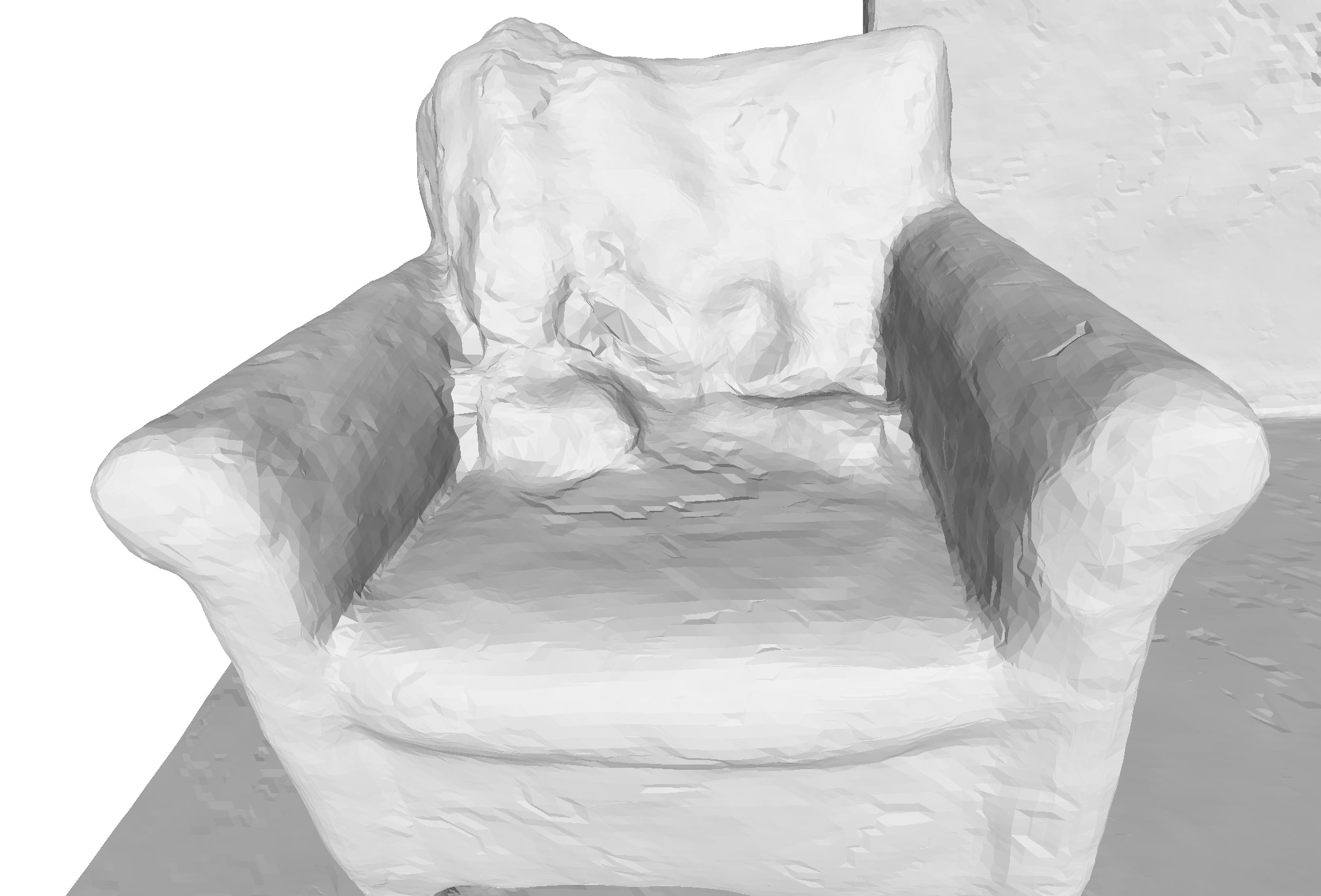} 
            \includegraphics[width=1\textwidth]{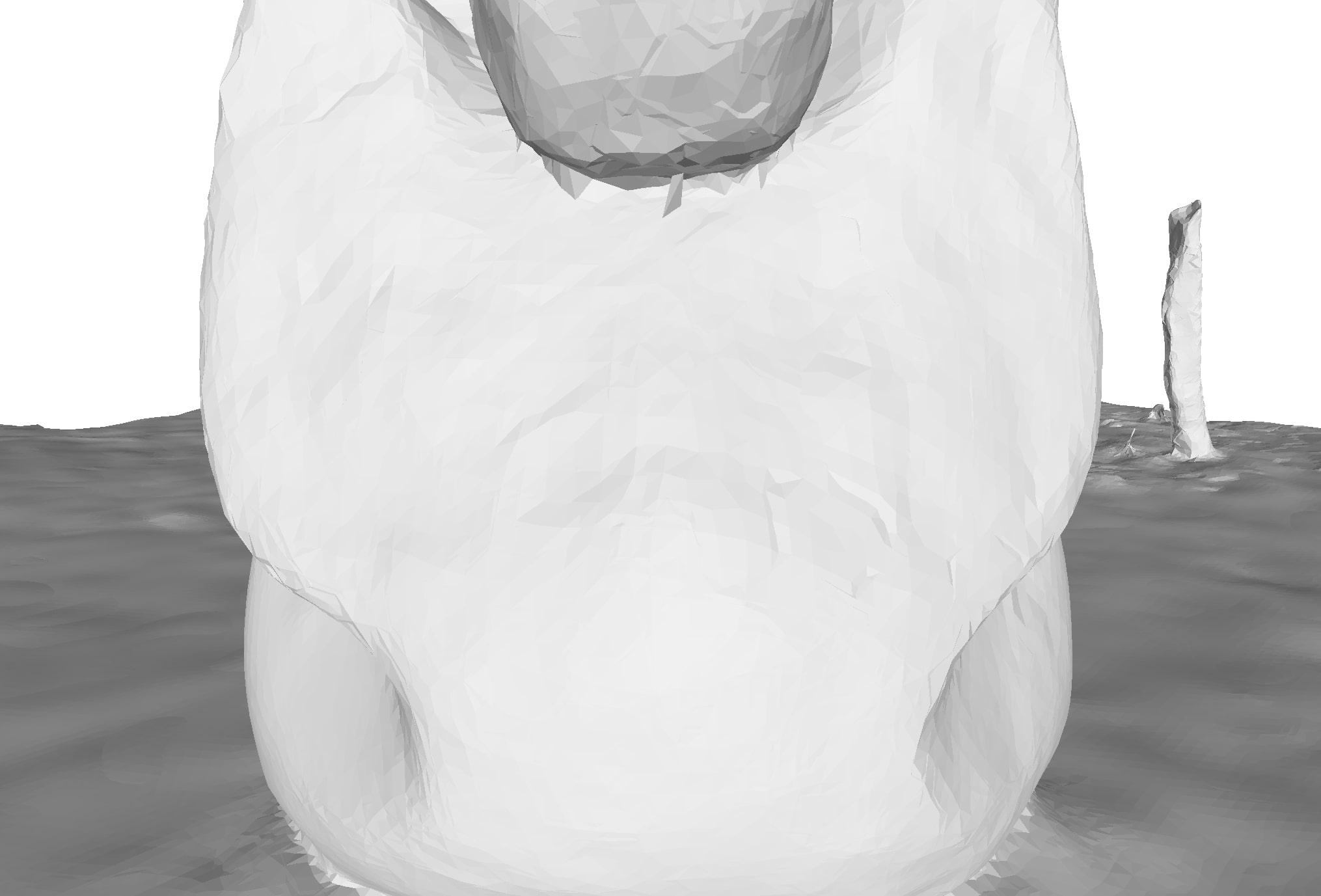} 
        \end{minipage}
    }
    \subfigure[Ours]{
        \begin{minipage}[b]{0.18\textwidth}
		  \includegraphics[width=1\textwidth]{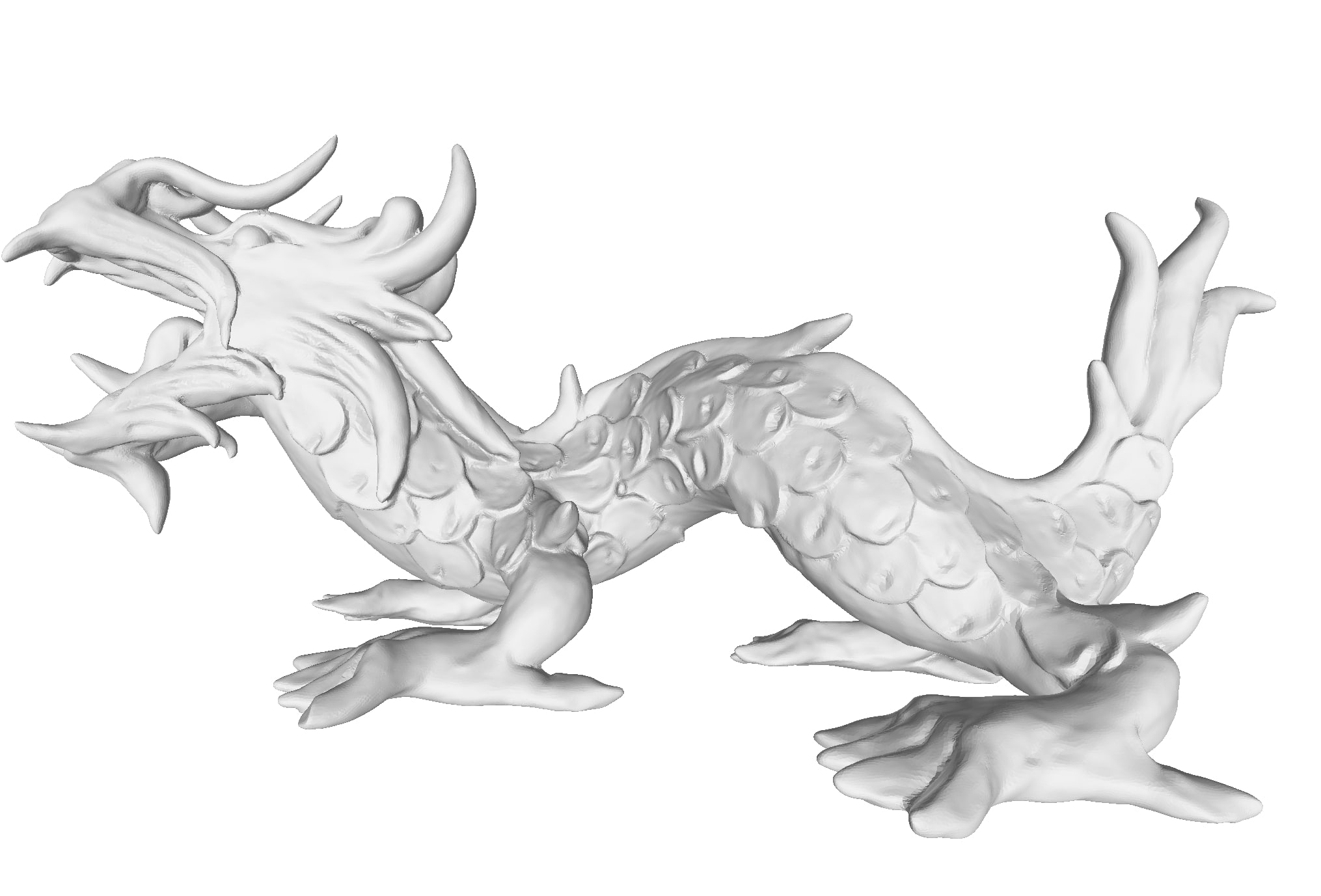} \\
		  \includegraphics[width=1\textwidth]{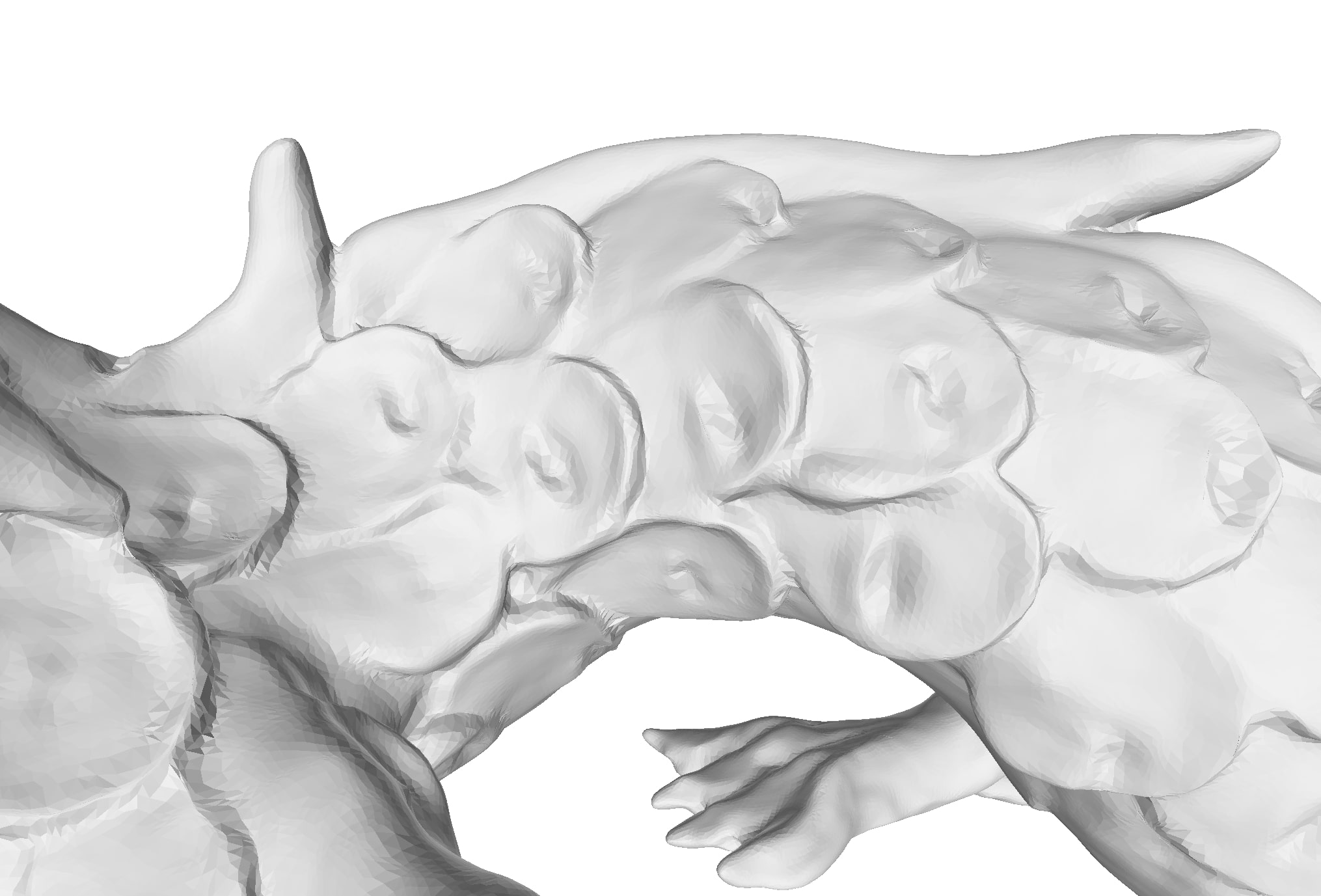} \\
    	\includegraphics[width=1\textwidth]{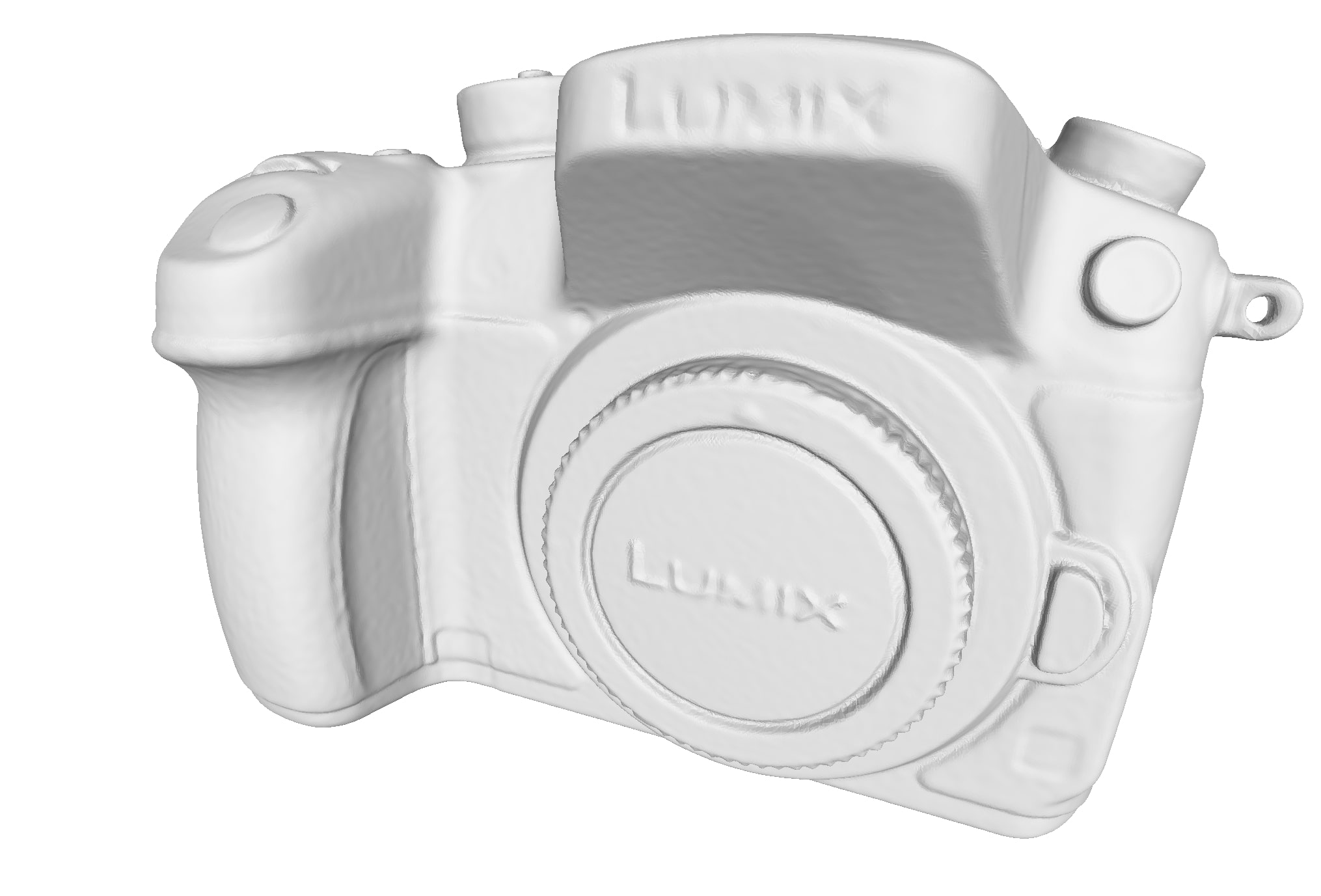} \\
            \includegraphics[width=1\textwidth]{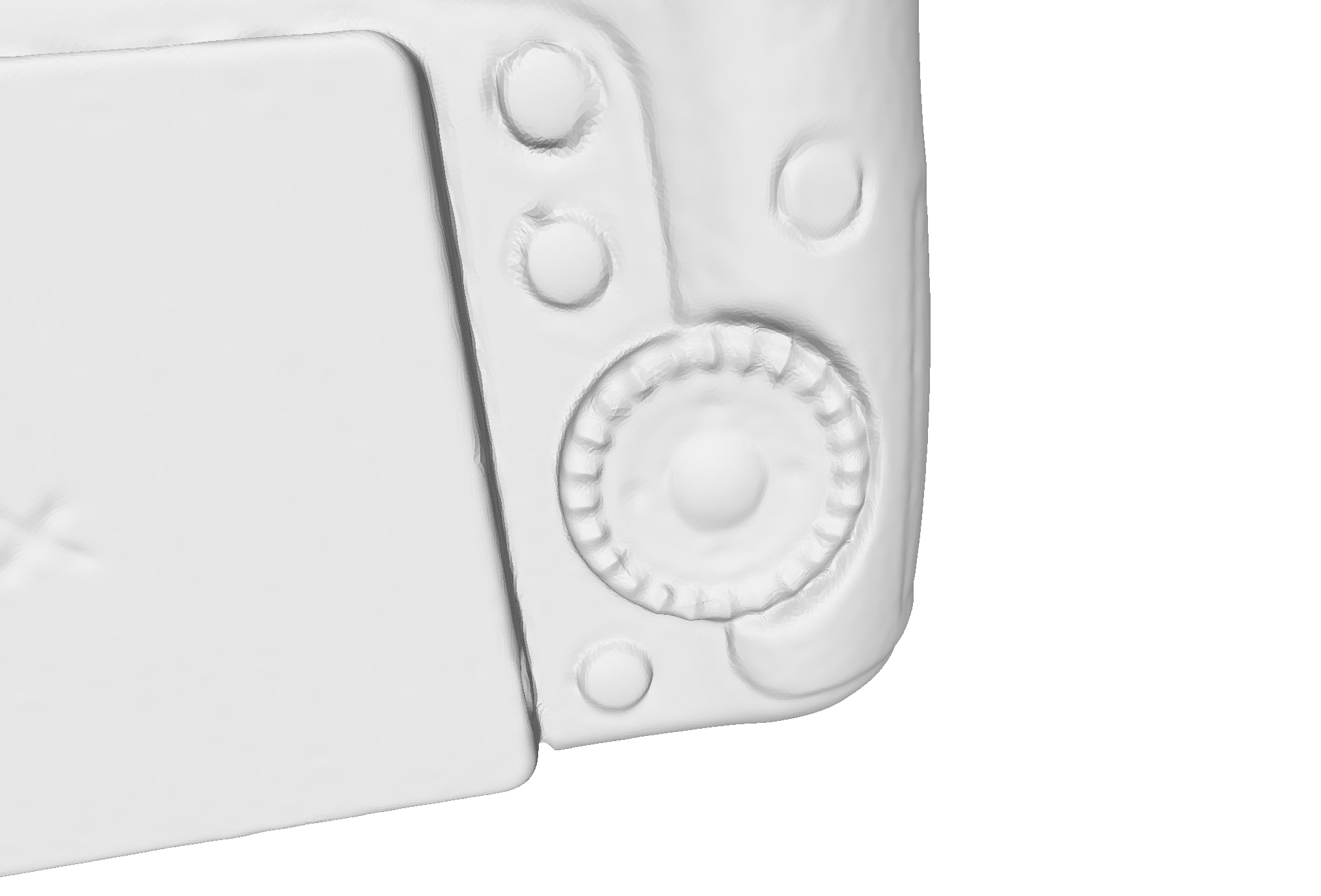} \\
    	\includegraphics[width=1\textwidth]{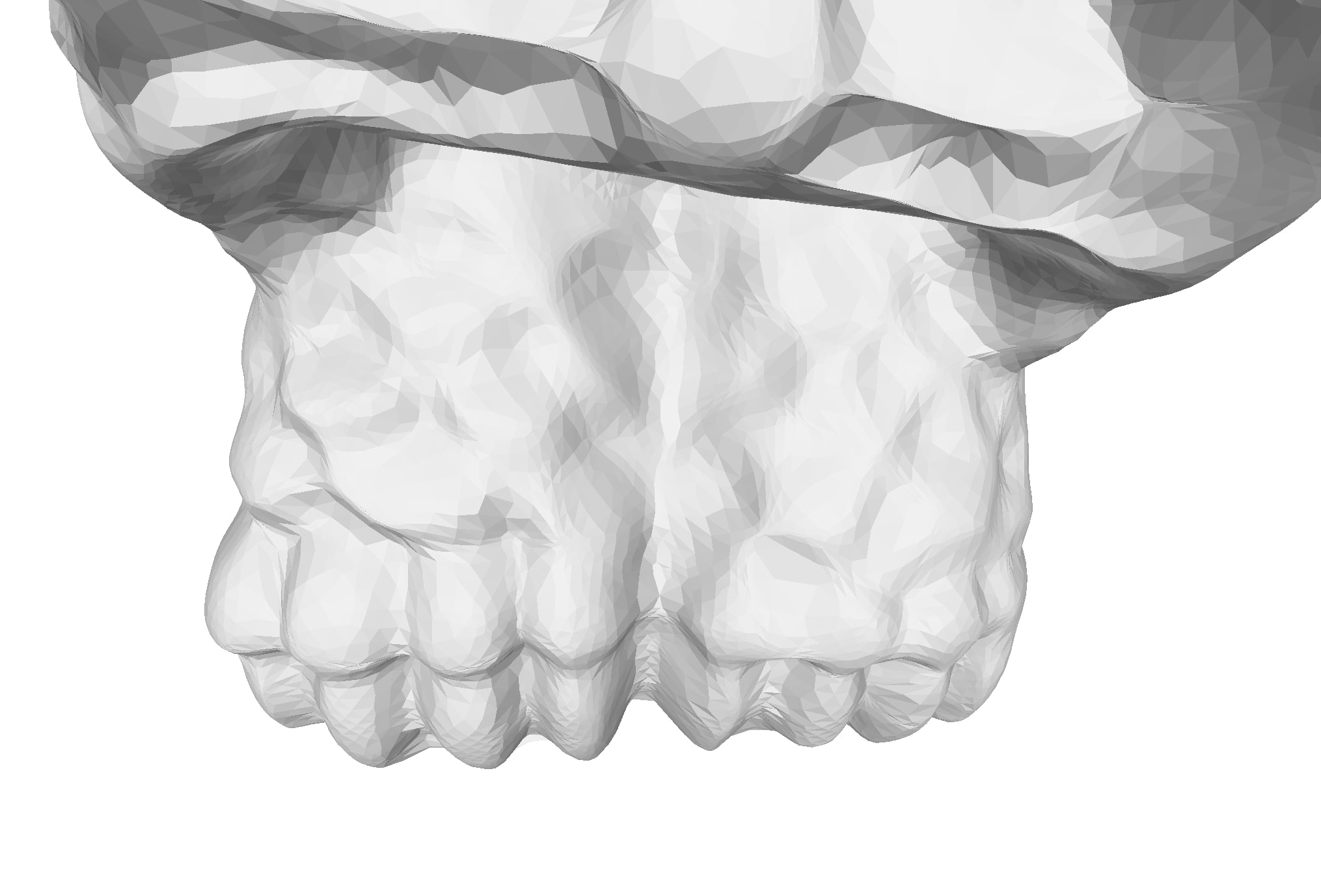} \\
            \includegraphics[width=1\textwidth]{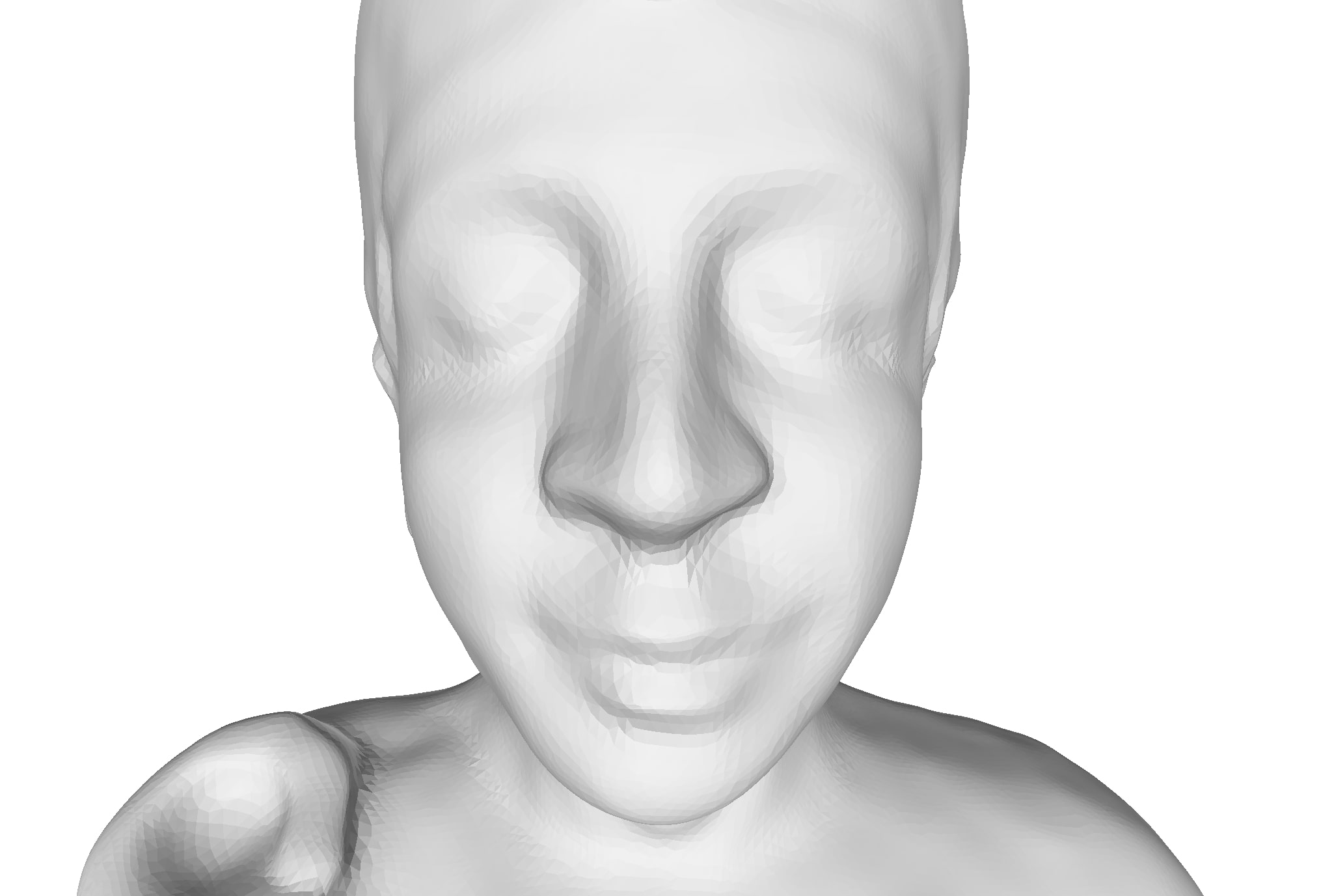} \\
    	\includegraphics[width=1\textwidth]{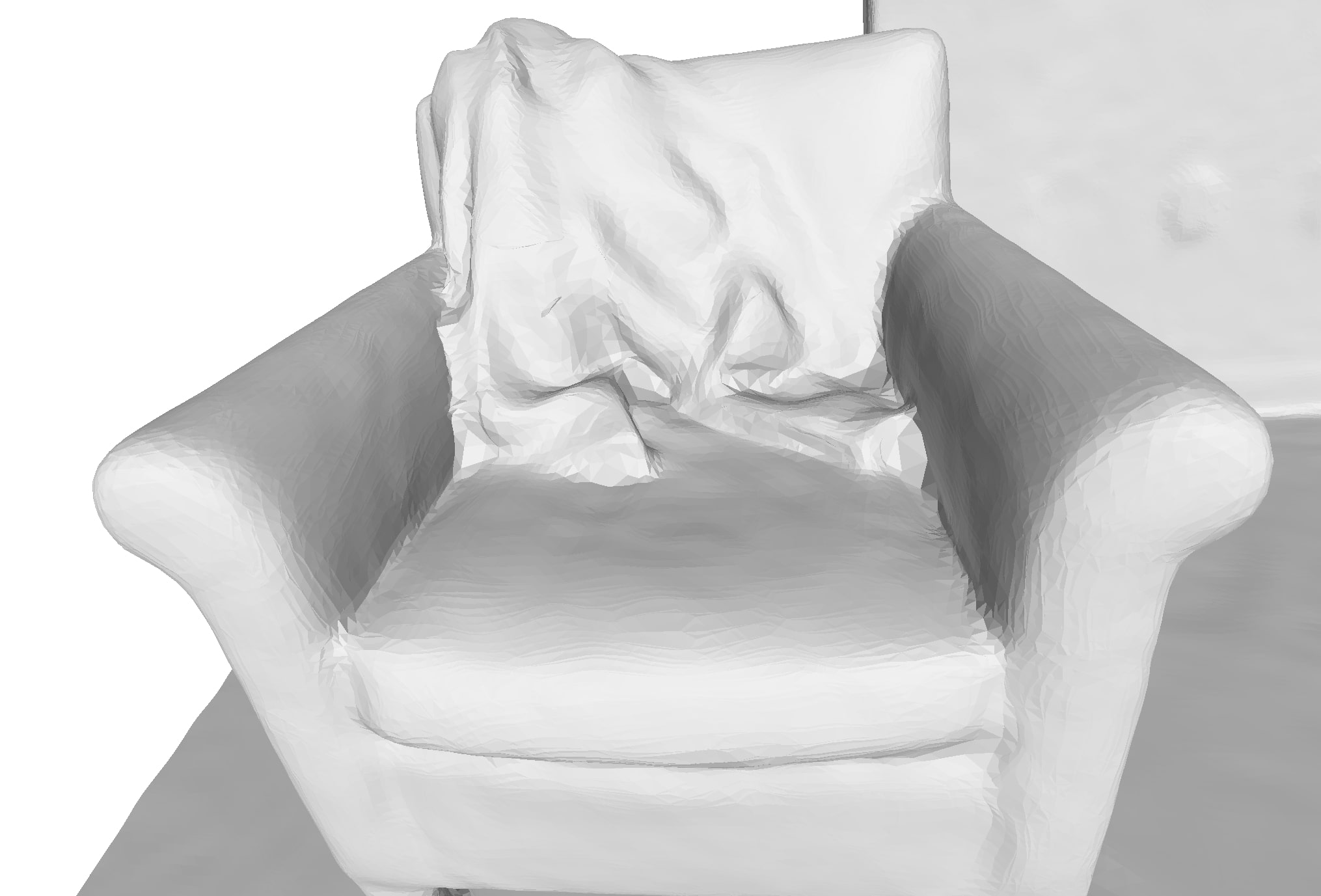} 
            \includegraphics[width=1\textwidth]{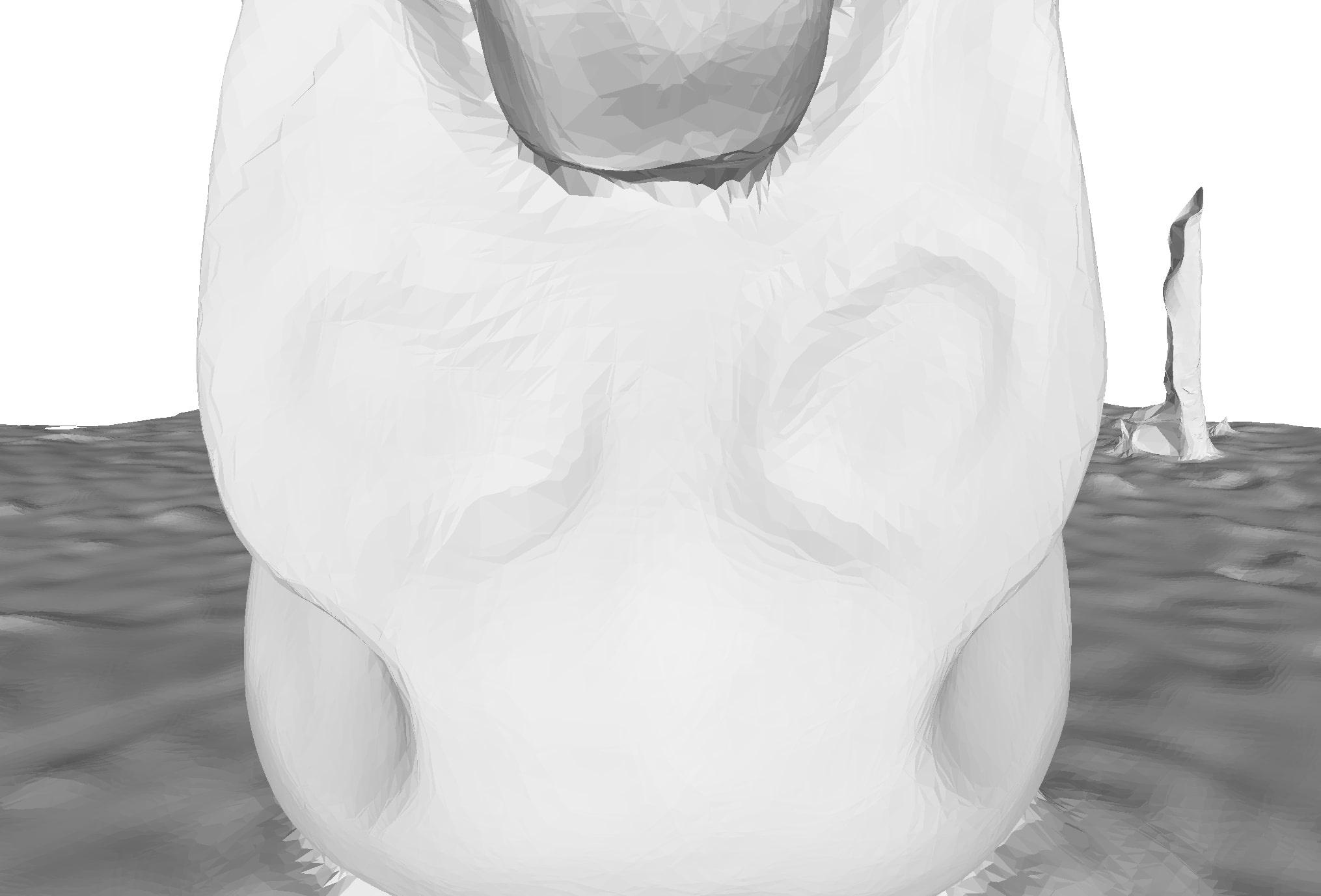} 
        \end{minipage}
    }
    \subfigure[GT]{
        \begin{minipage}[b]{0.18\textwidth}
		  \includegraphics[width=1\textwidth]{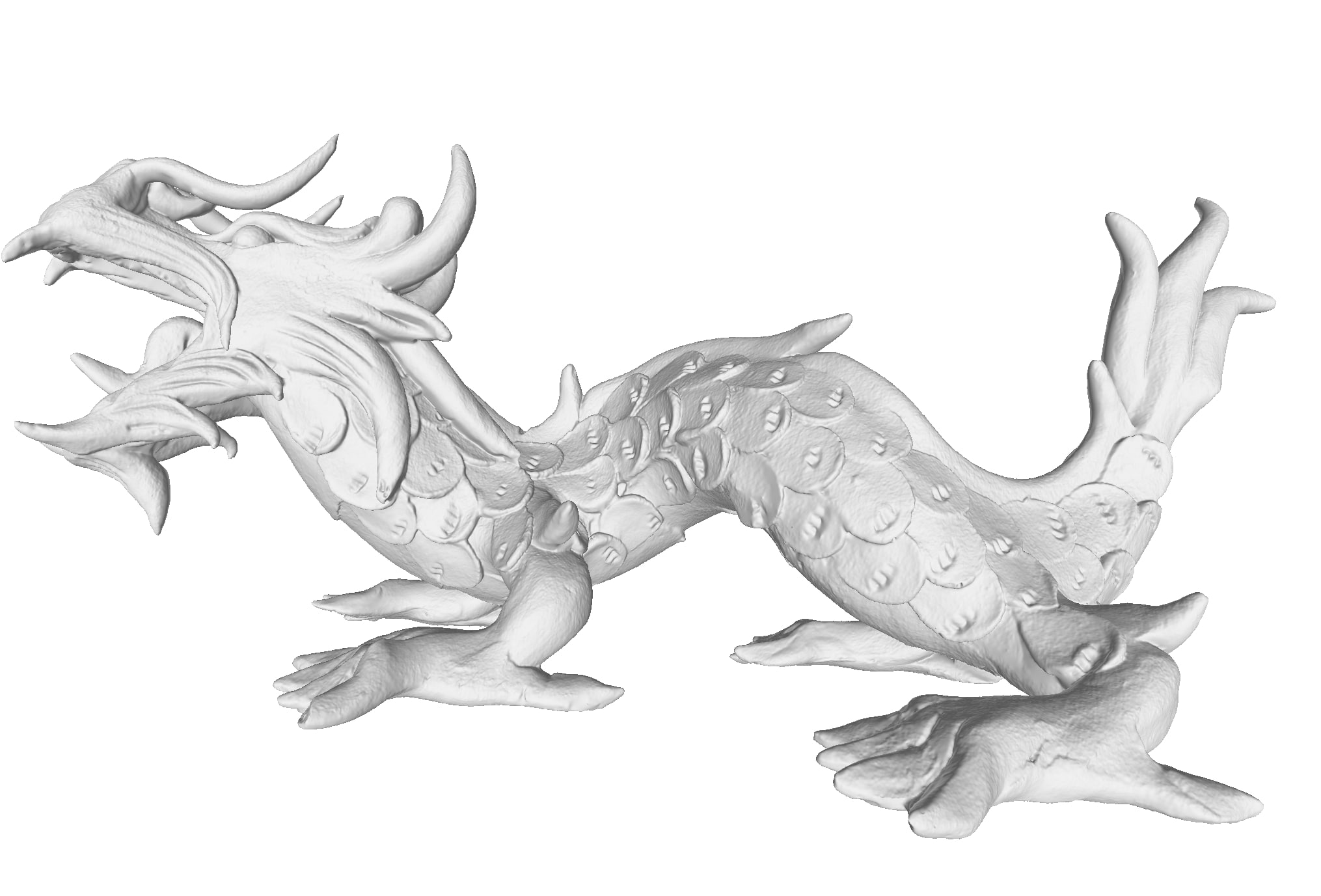} \\
		  \includegraphics[width=1\textwidth]{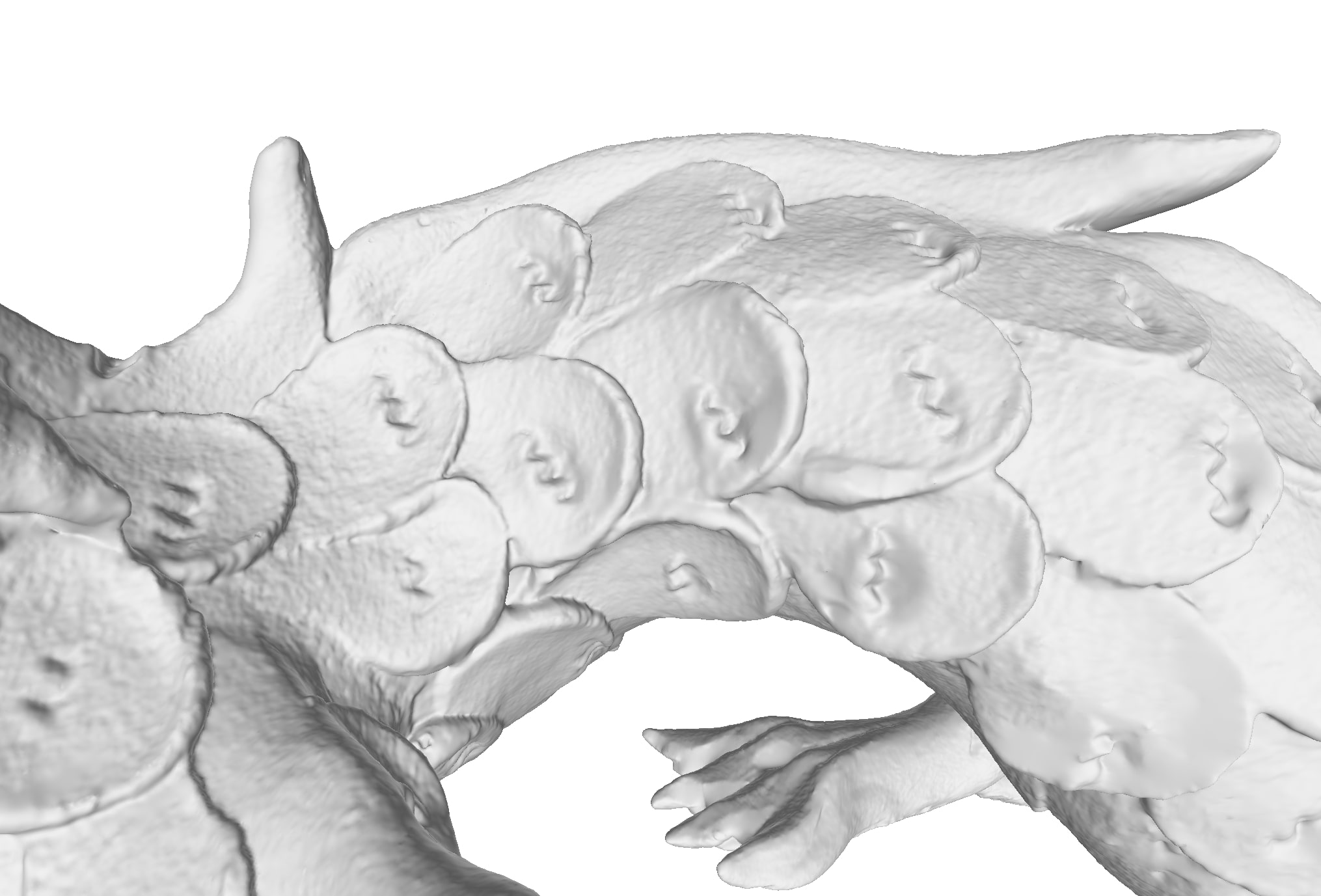} \\
    	\includegraphics[width=1\textwidth]{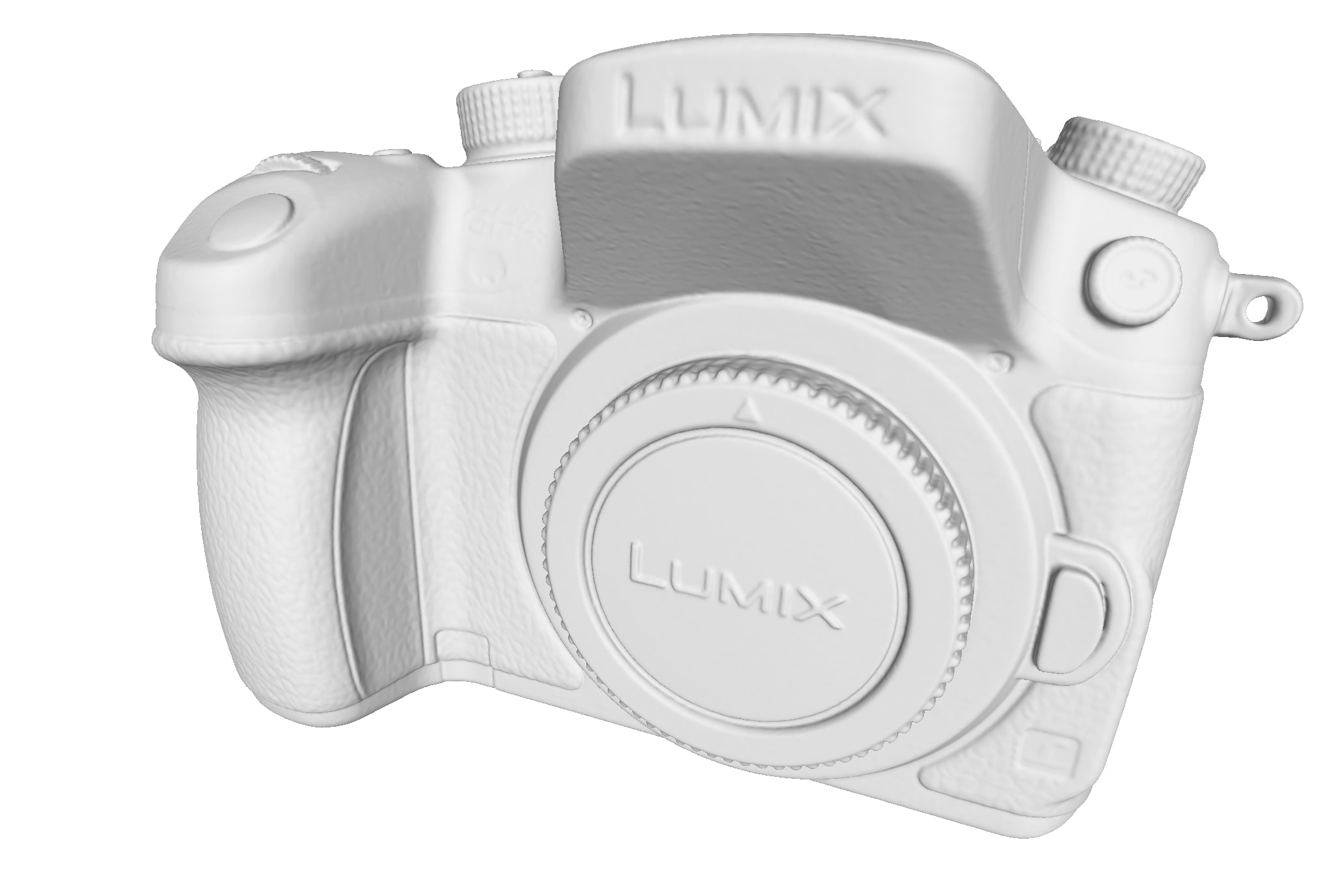} \\
            \includegraphics[width=1\textwidth]{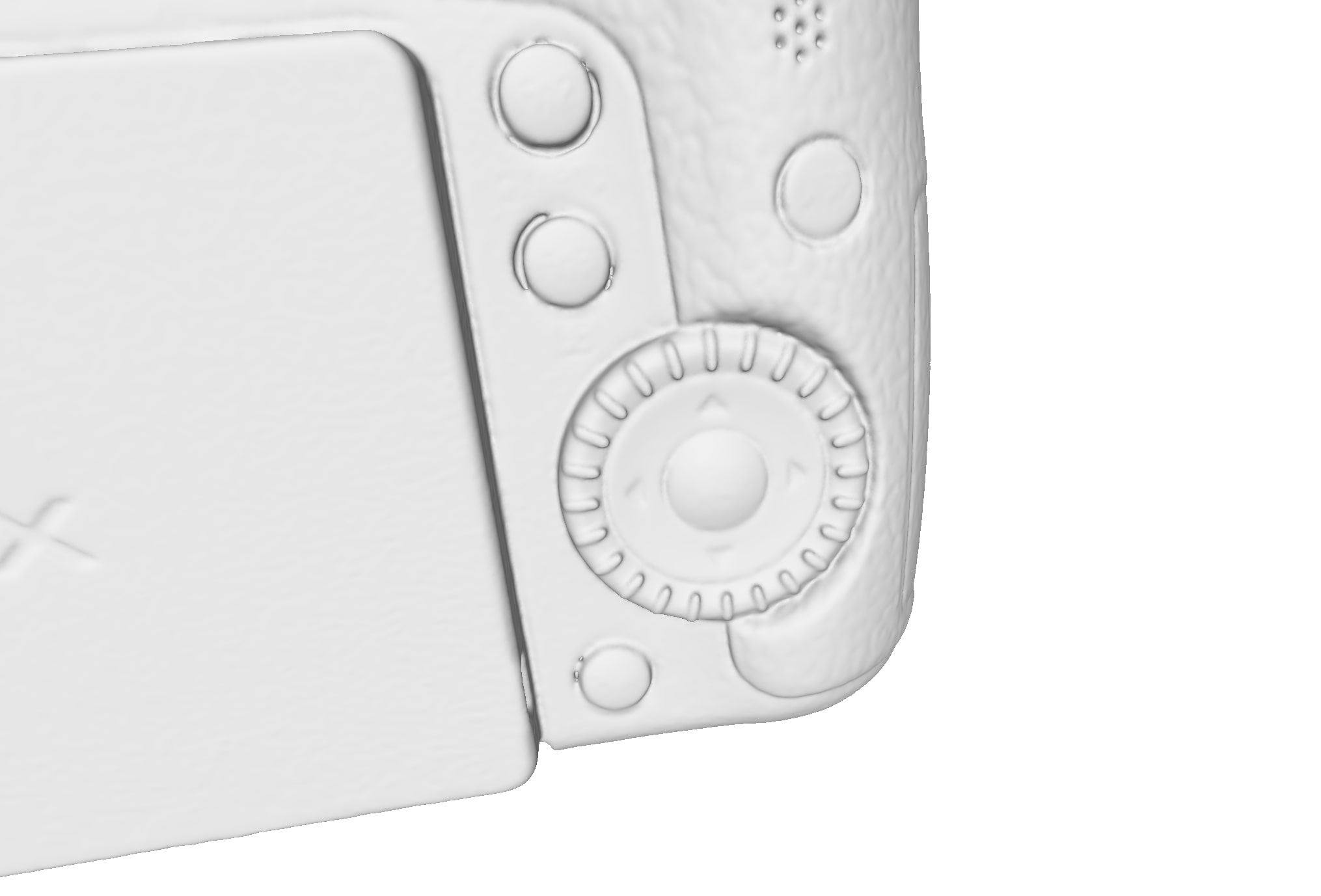} \\
    	\includegraphics[width=1\textwidth]{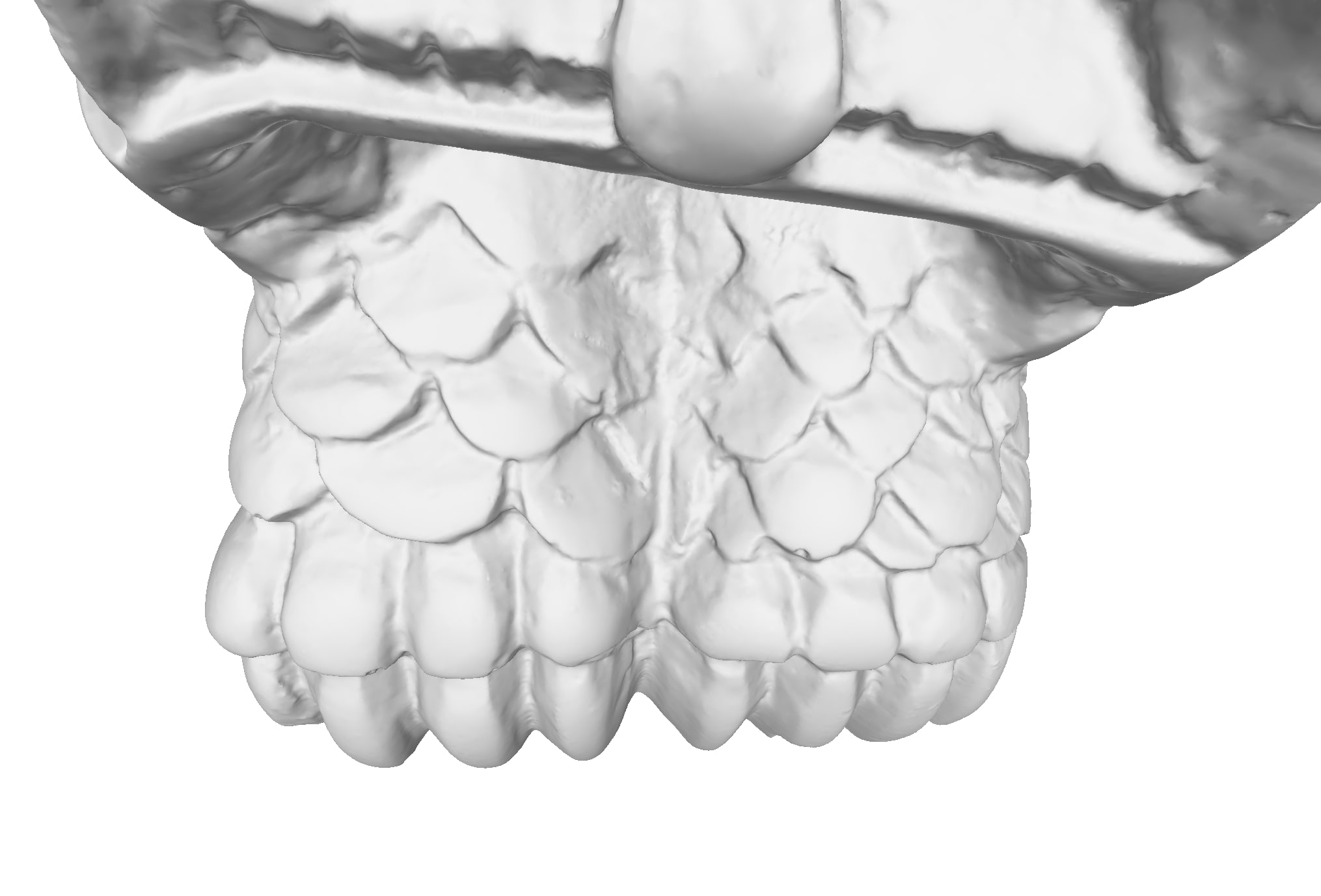} \\
            \includegraphics[width=1\textwidth]{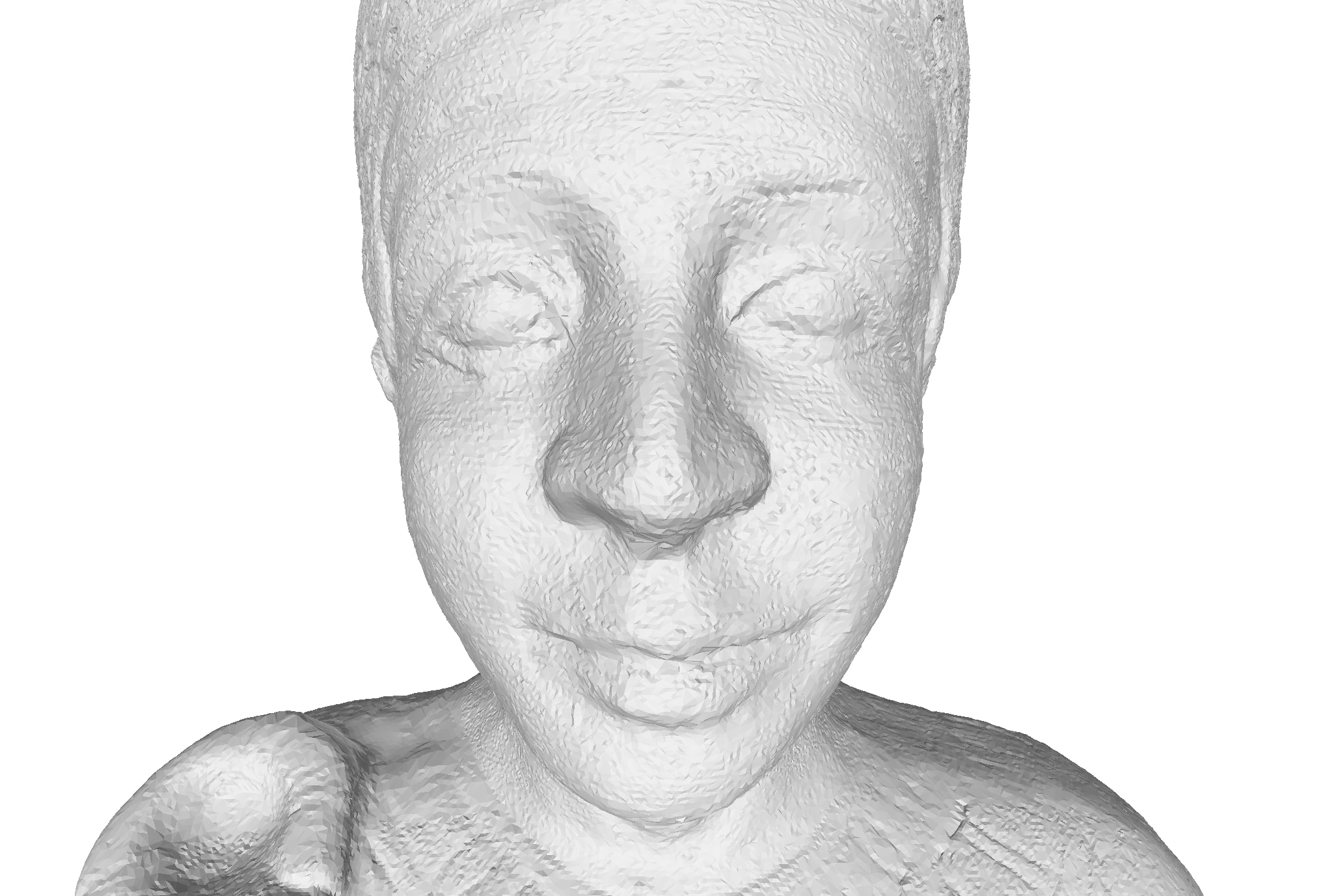} \\
    	\includegraphics[width=1\textwidth]{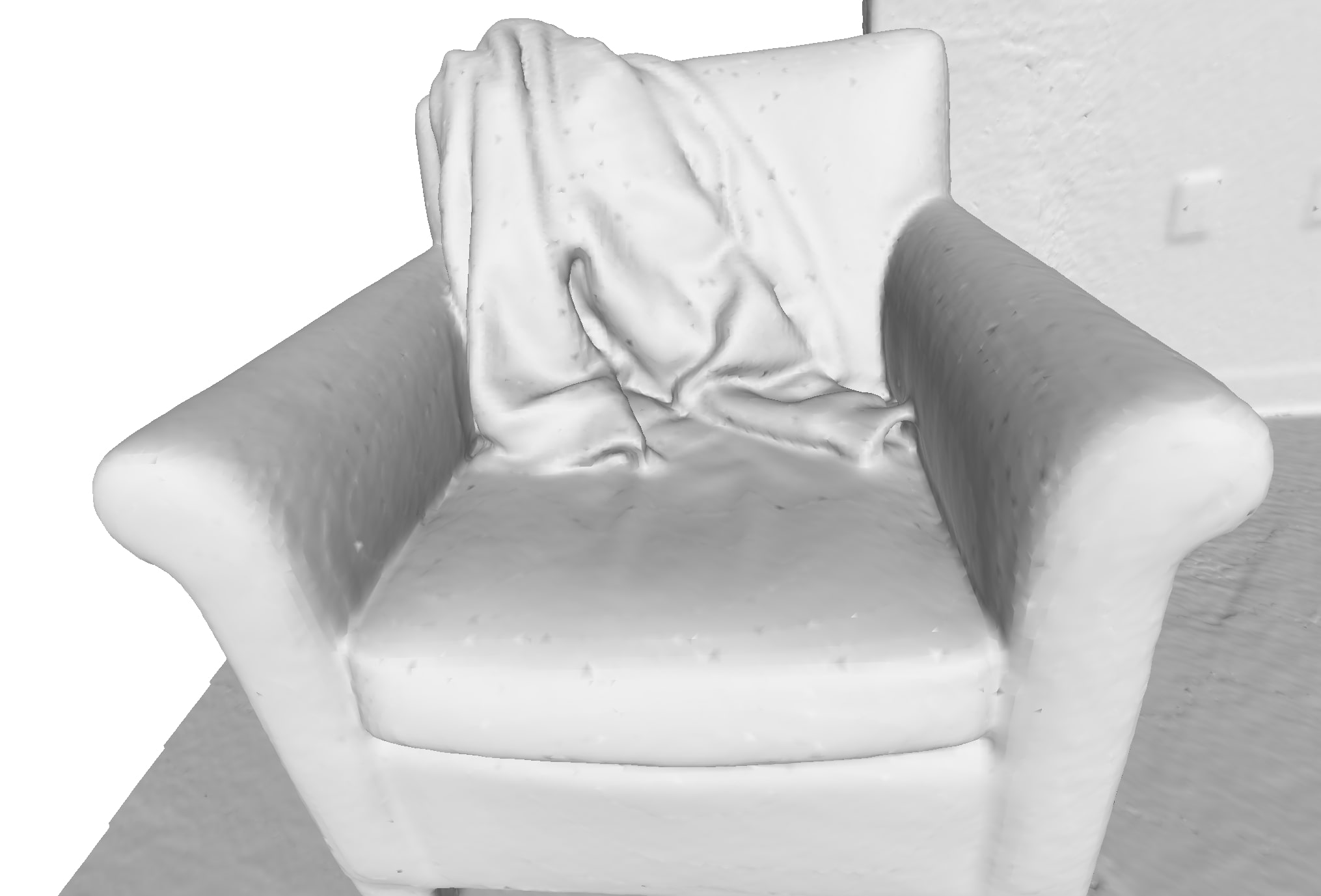} 
            \includegraphics[width=1\textwidth]{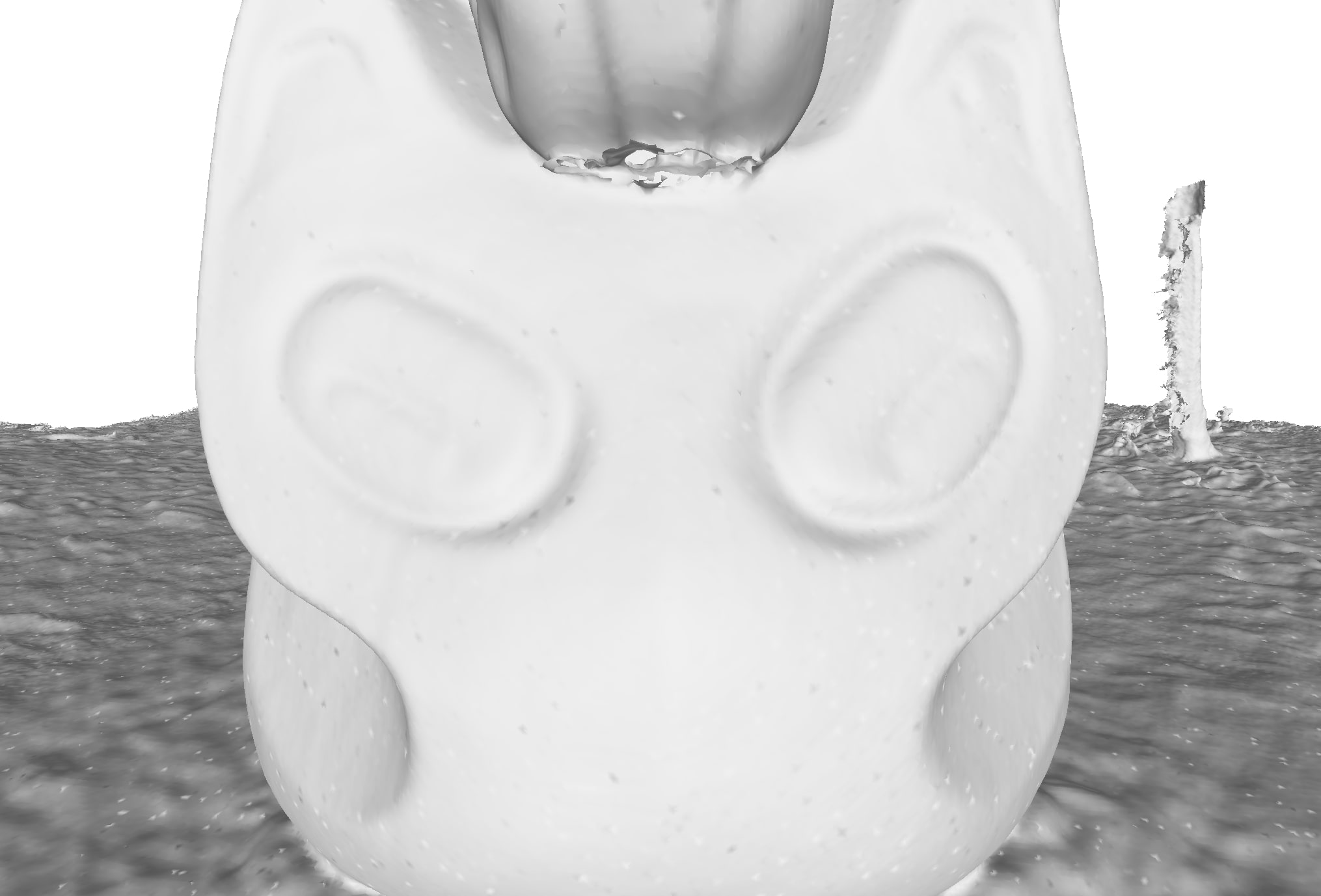} 
        \end{minipage}
    }
    \caption{More visual comparisons with DUDF, CAPUDF and LevelSetUDF on detailed models from the Stanford 3D Scan dataset and Stanford 3D Scene dataset. Our method learns more details.}
    \label{fig:3dScan-comparison}
\end{figure*}

\begin{figure*}[!htbp]
    \centering
    \subfigure[DUDF]{
        \begin{minipage}[b]{0.18\textwidth}
		  \includegraphics[width=1\textwidth]{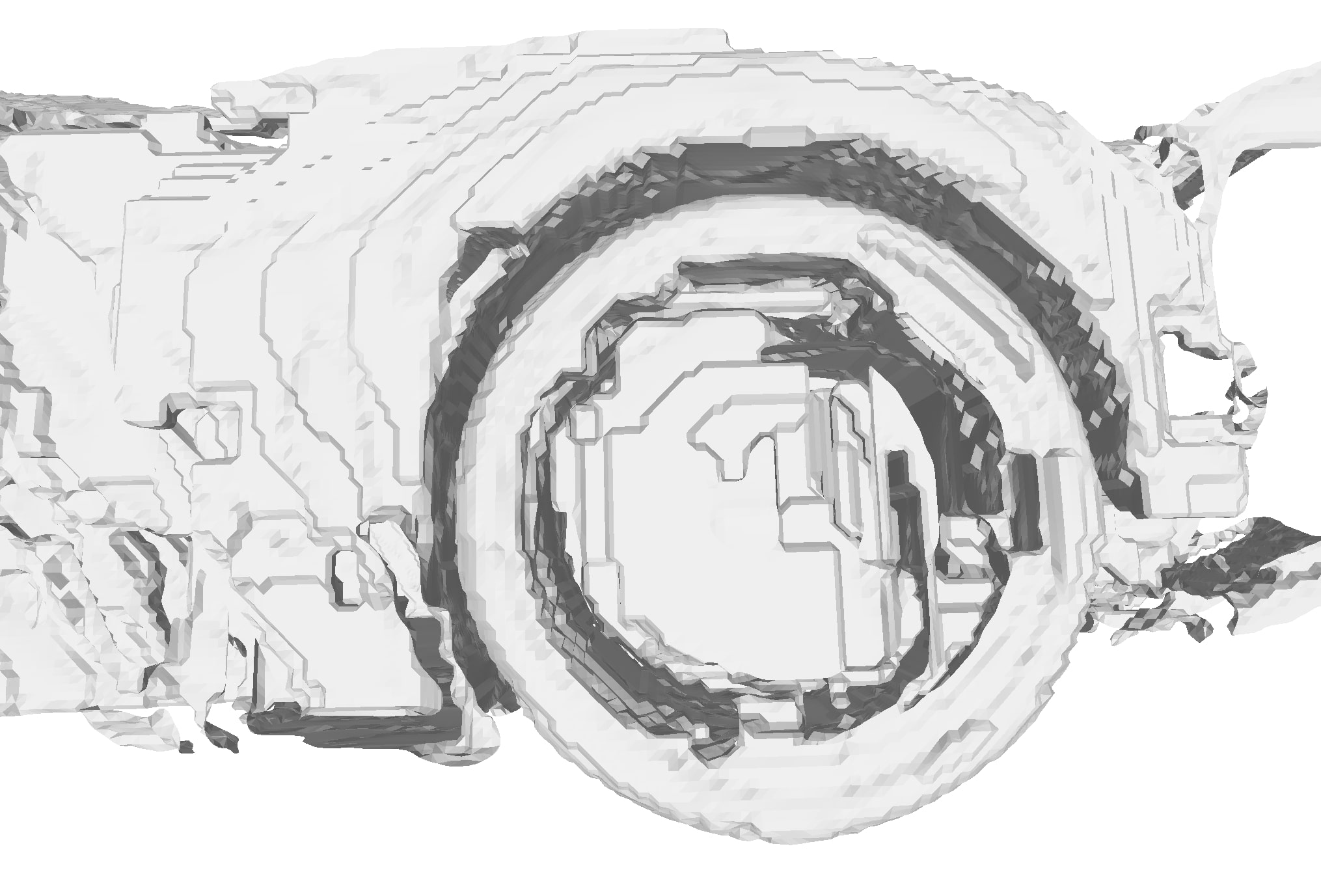} \\
		  \includegraphics[width=1\textwidth]{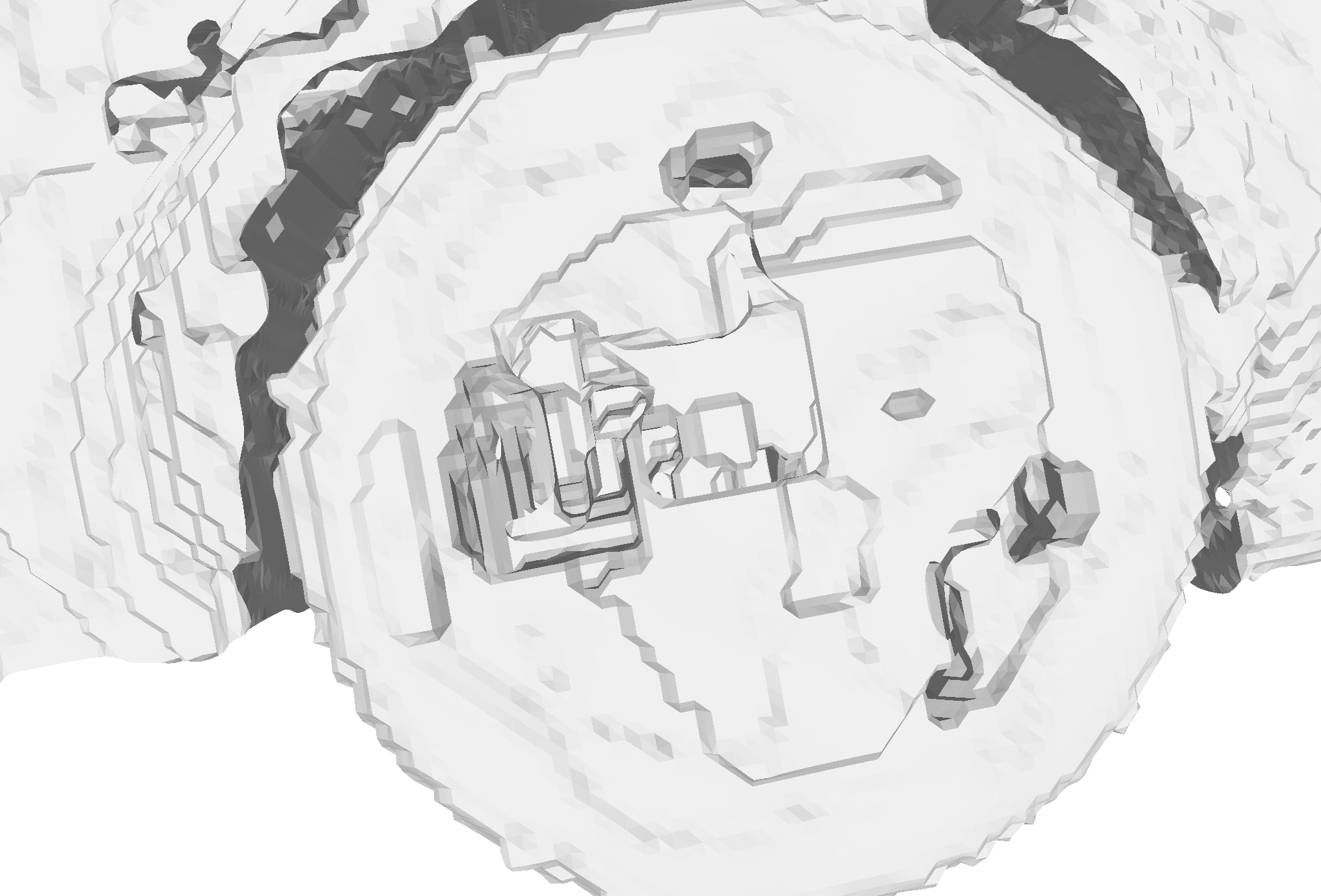} \\  
            \includegraphics[width=1\textwidth]{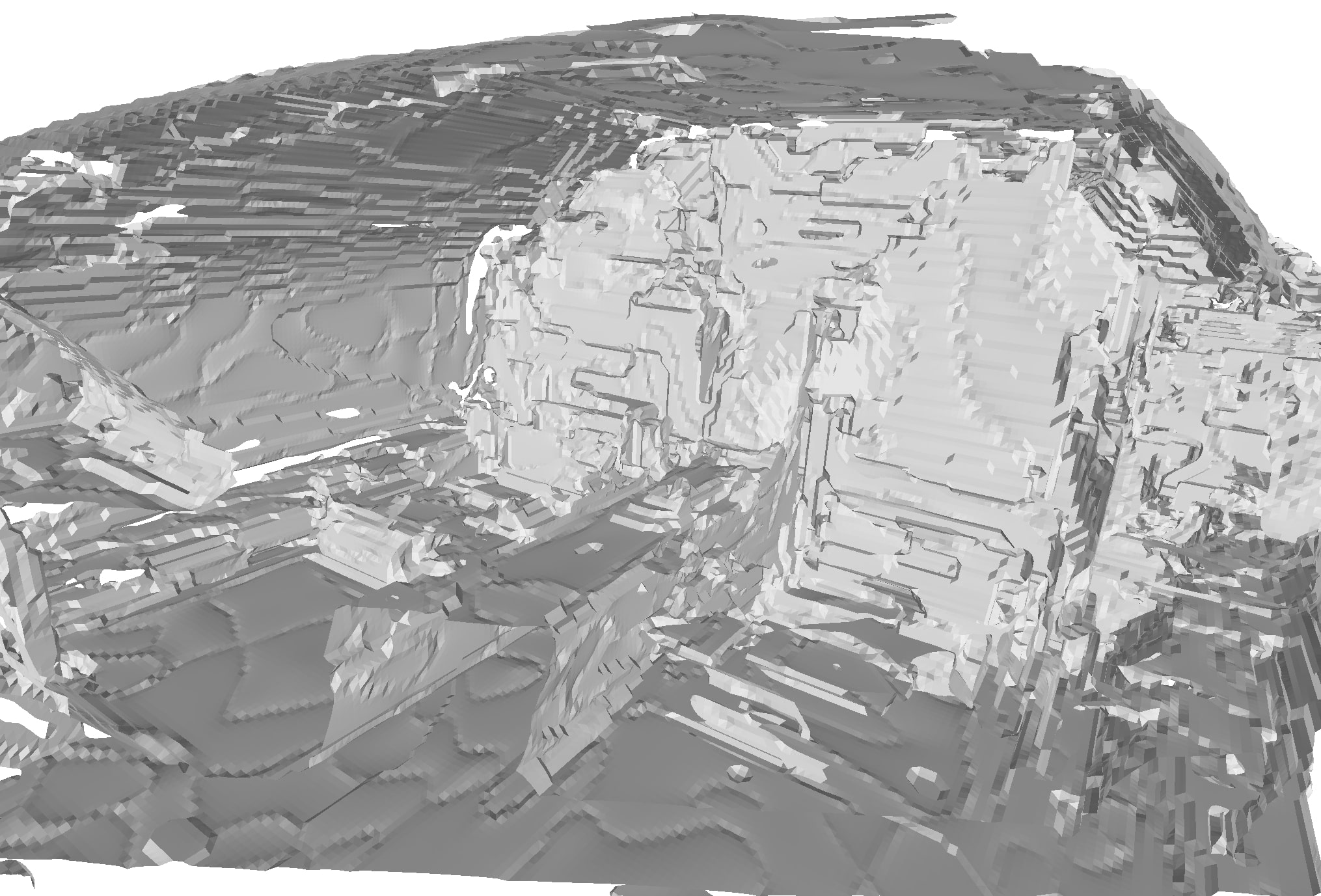} \\		  
            \includegraphics[width=1\textwidth]{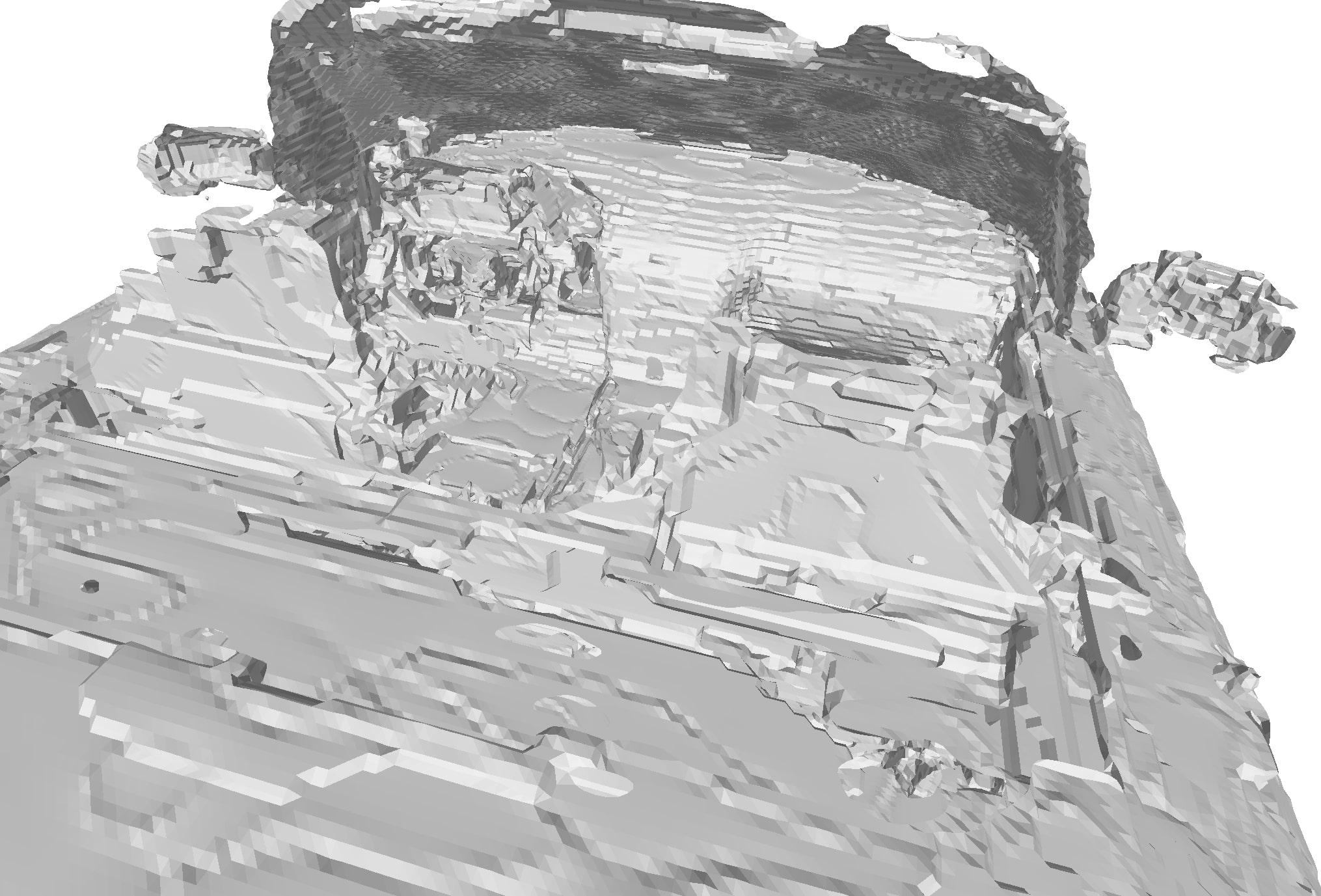} \\		  
            \includegraphics[width=1\textwidth]{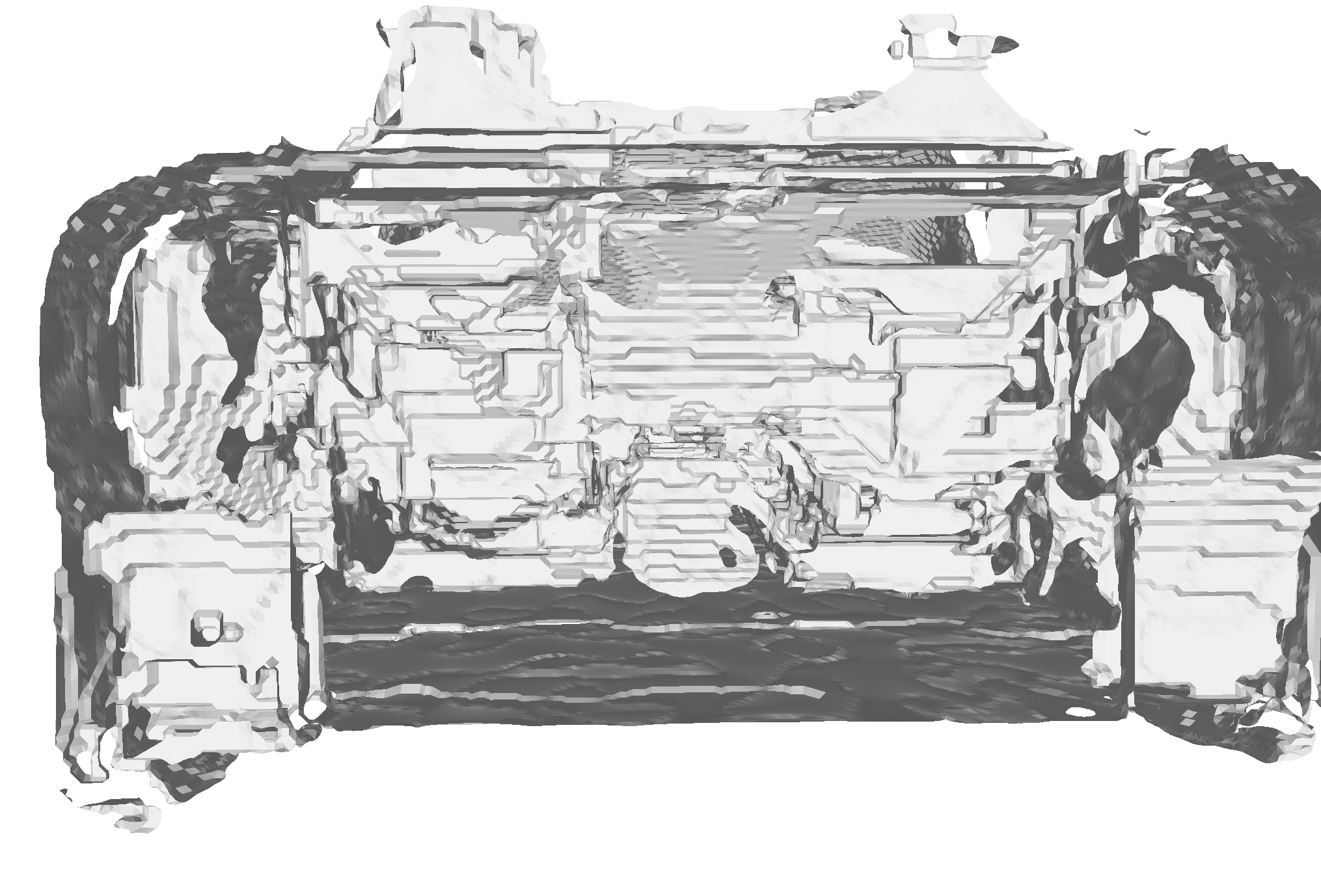} \\		    		  
            \includegraphics[width=1\textwidth]{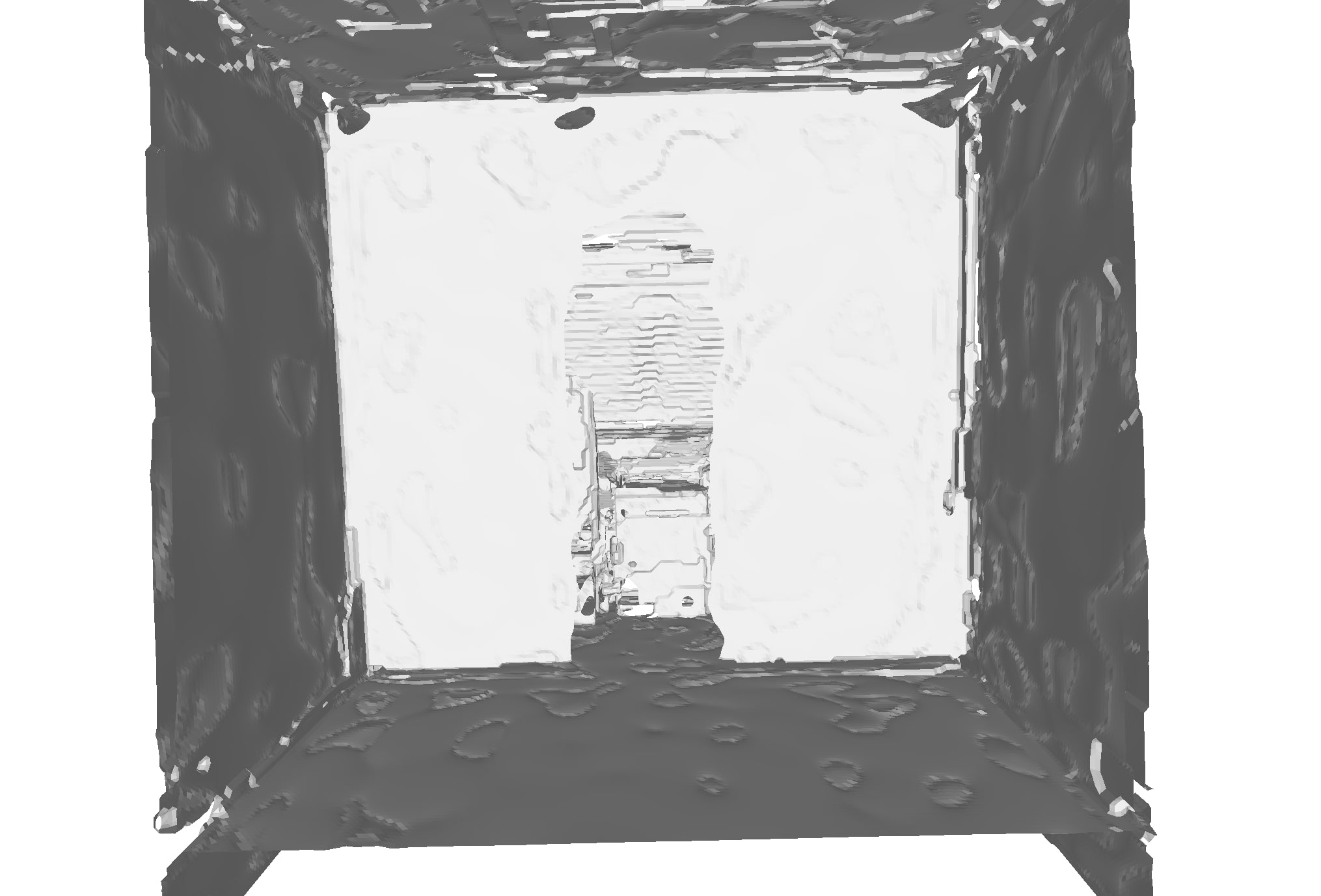} \\		 
            \includegraphics[width=1\textwidth]{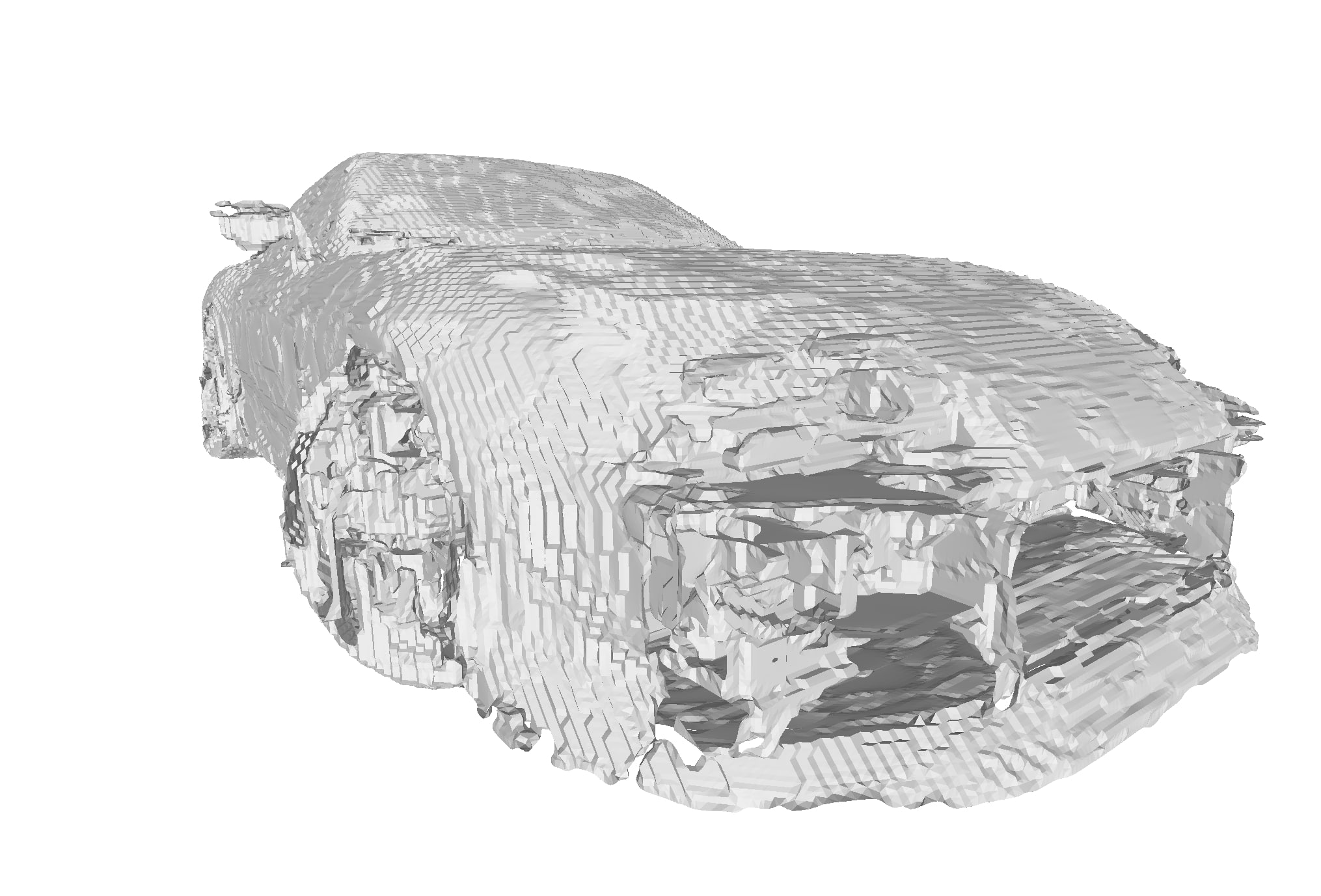} \\		  
            \includegraphics[width=1\textwidth]{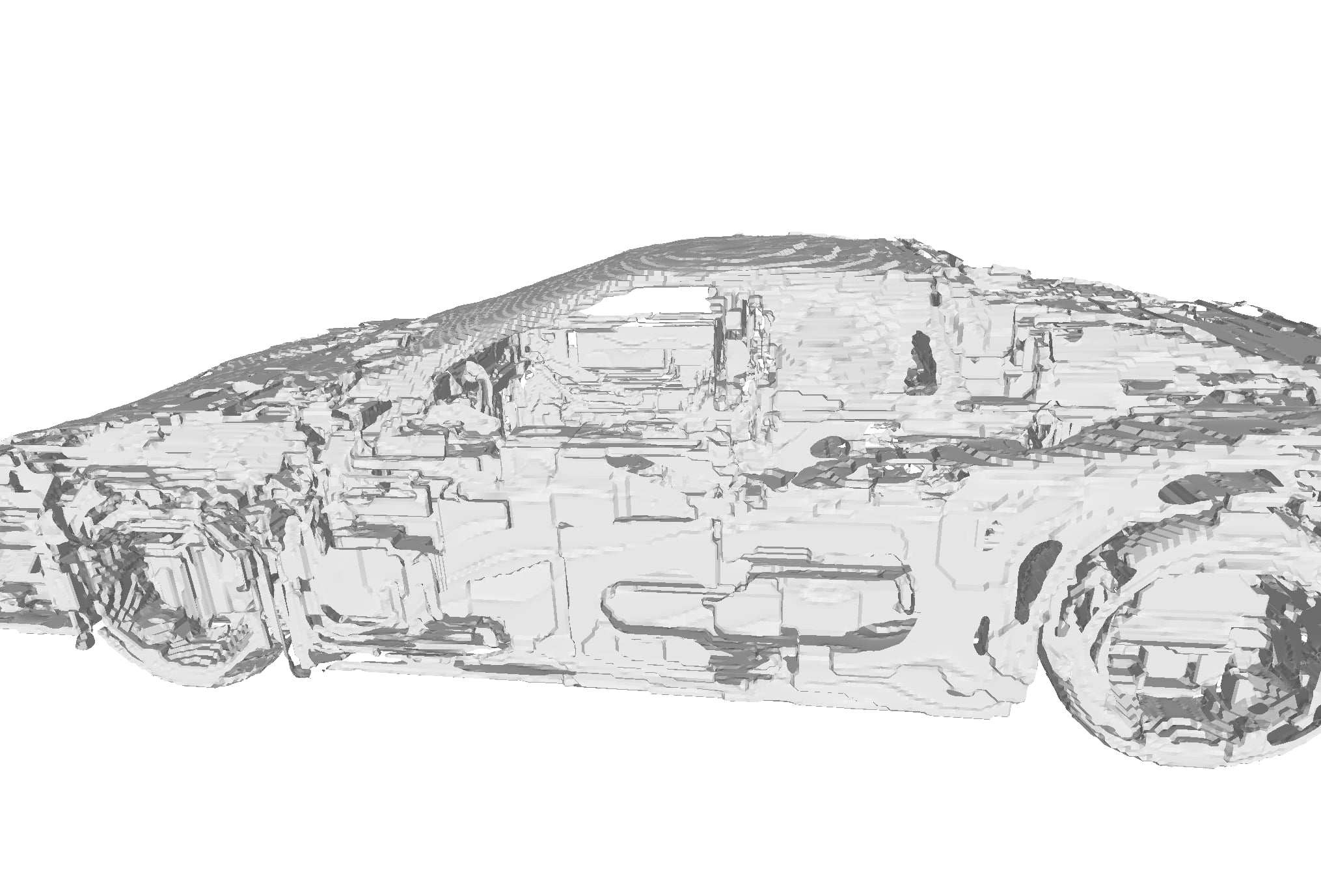} \\
            \includegraphics[width=1\textwidth]{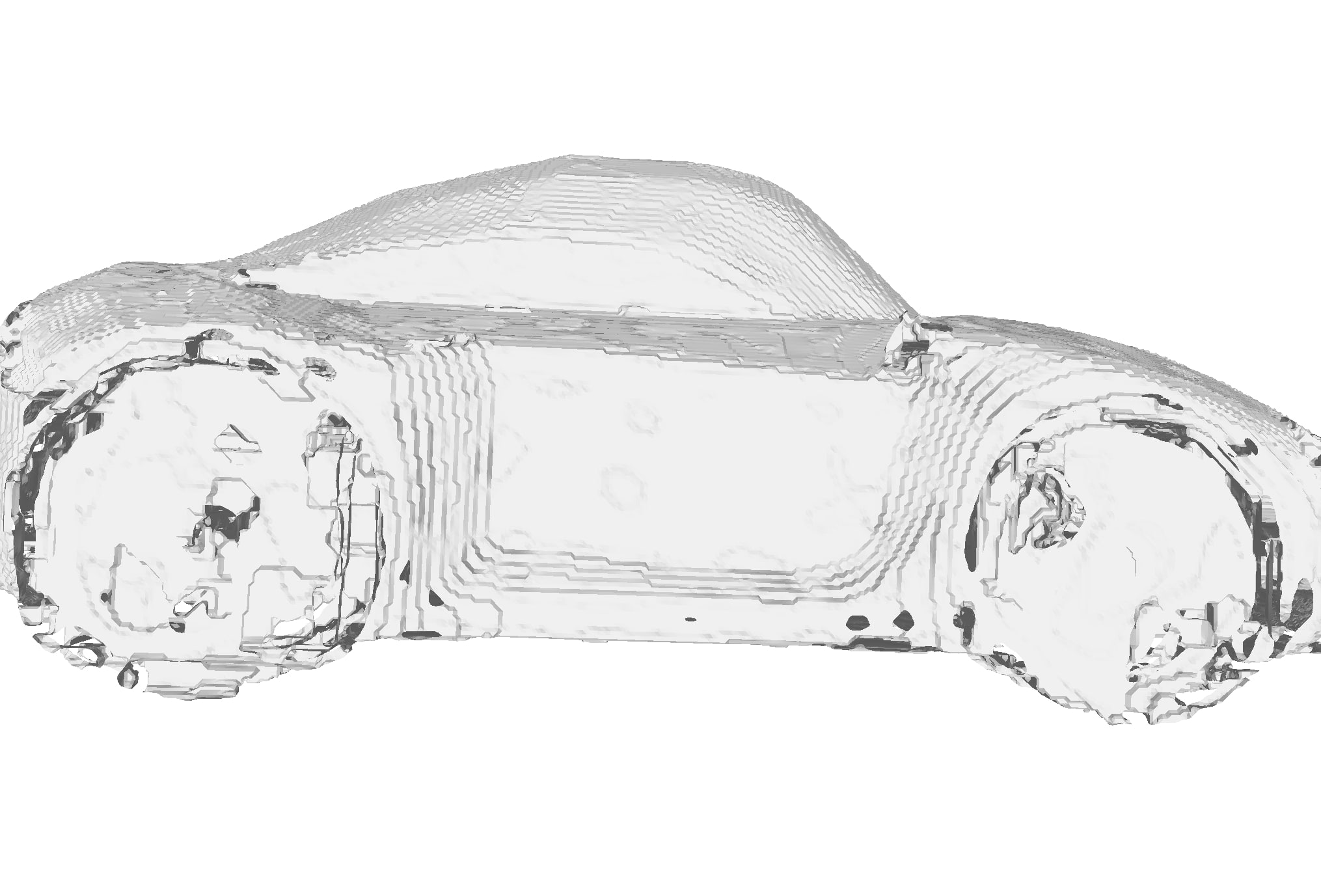}
        \end{minipage}
    }
    \subfigure[CAP-UDF]{
        \begin{minipage}[b]{0.18\textwidth}
		  \includegraphics[width=1\textwidth]{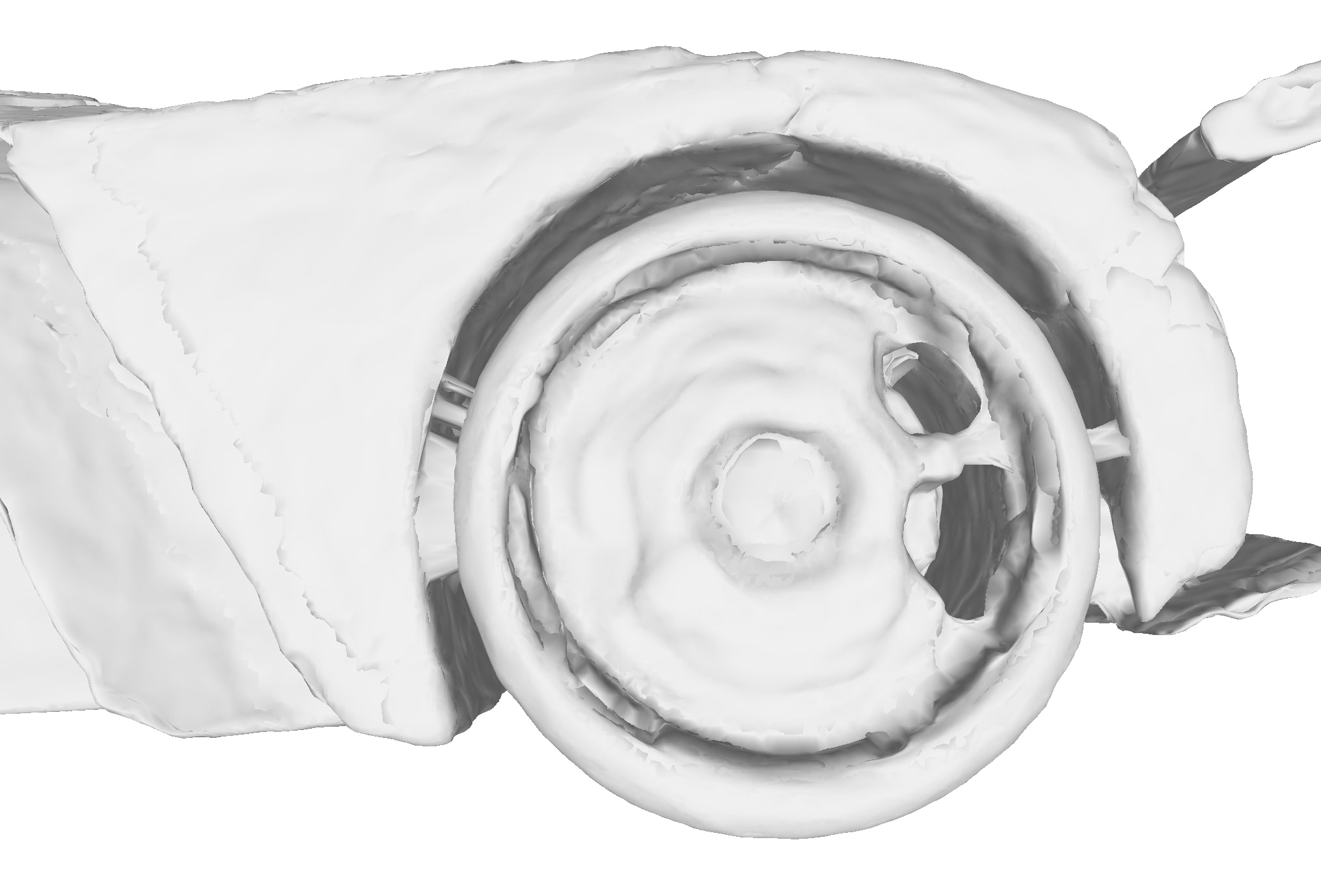} \\
		  \includegraphics[width=1\textwidth]{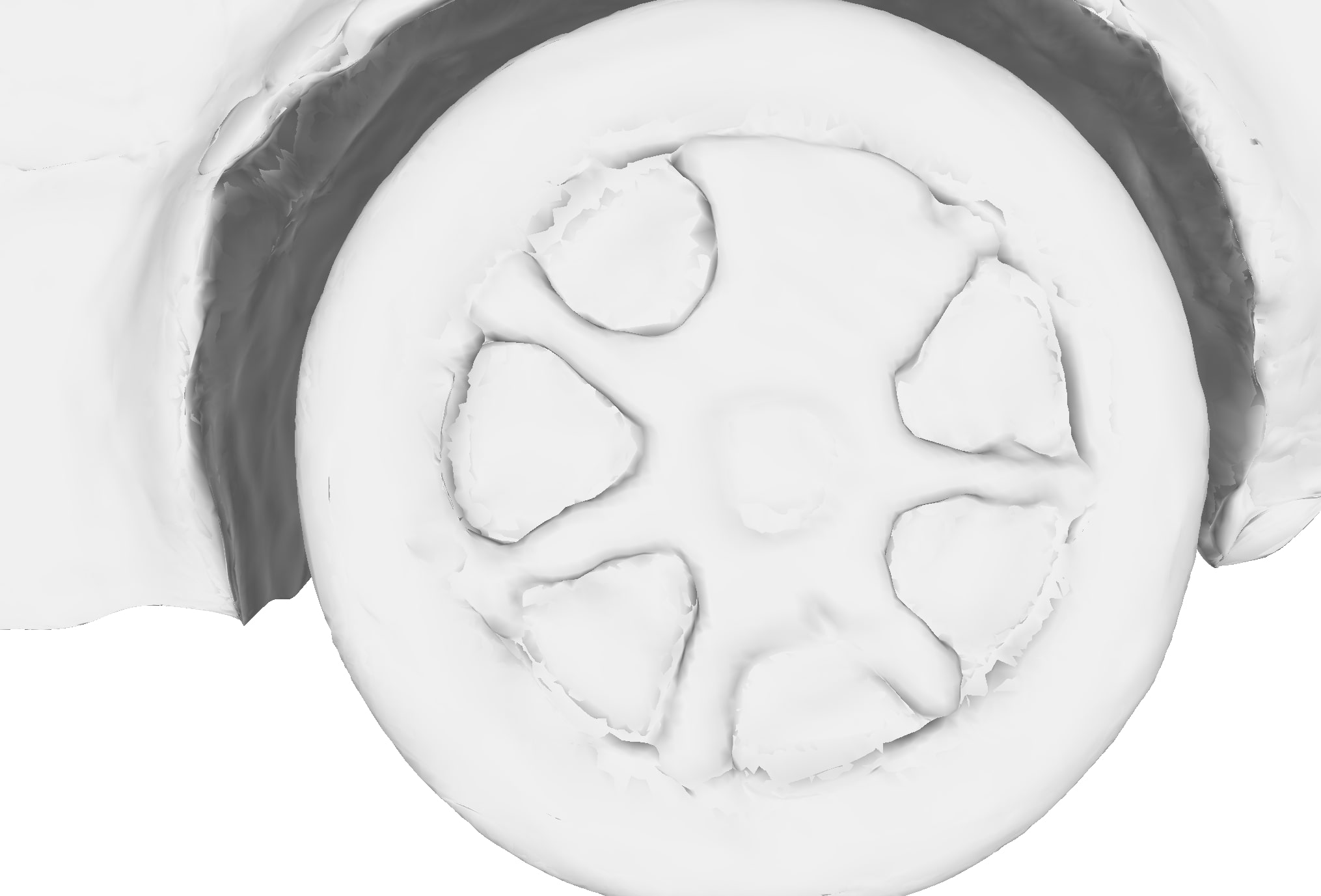} \\  
            \includegraphics[width=1\textwidth]{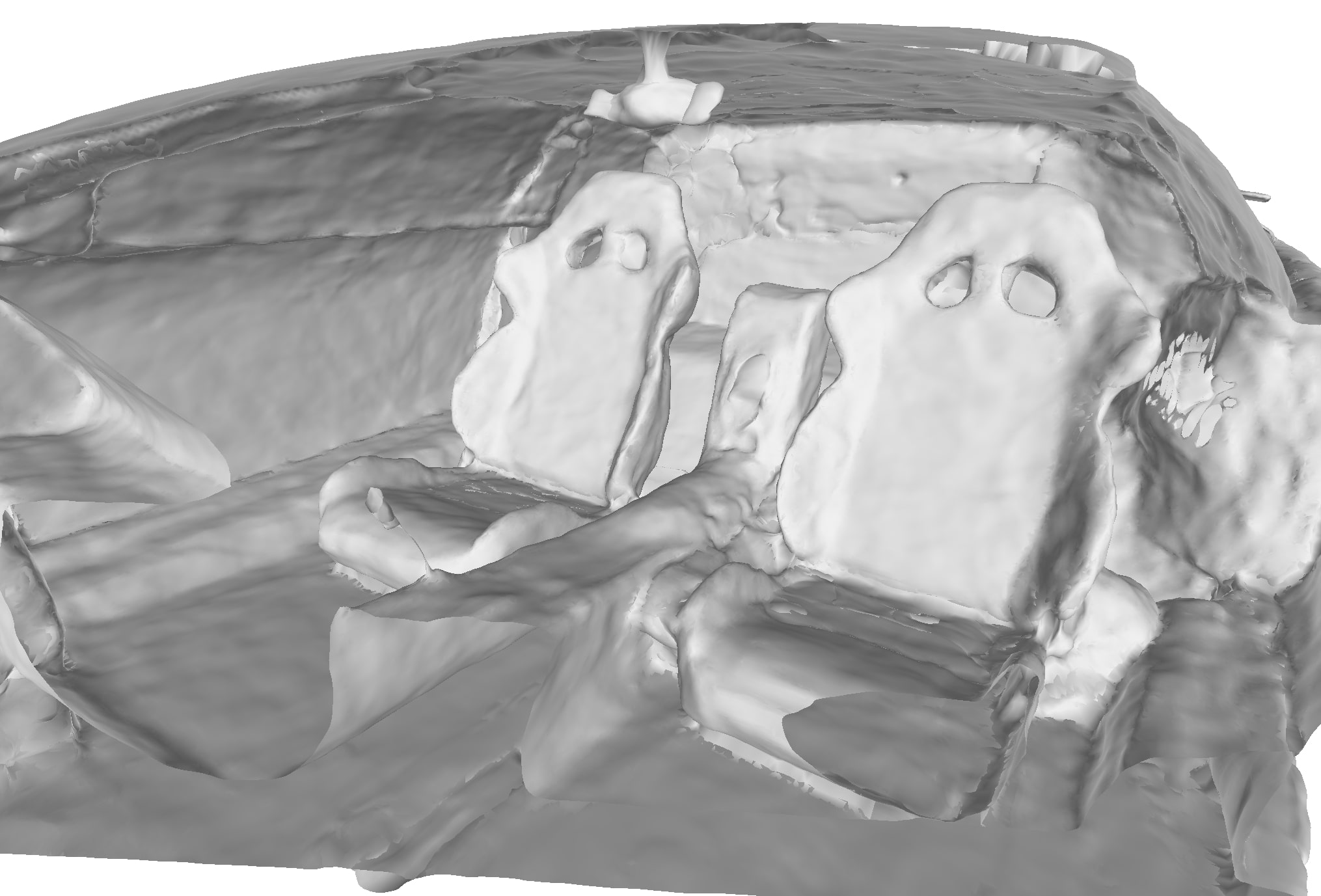} \\		  
            \includegraphics[width=1\textwidth]{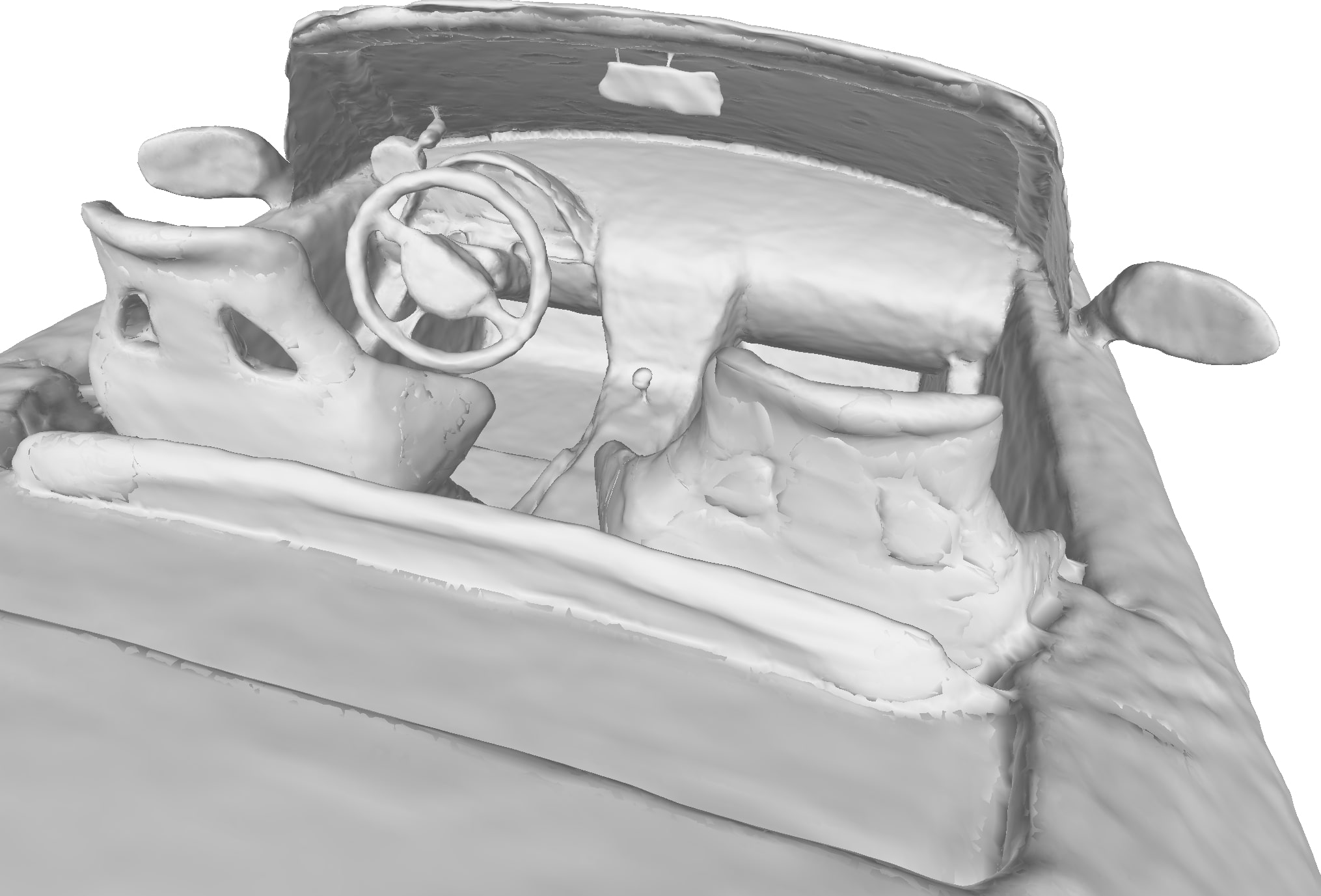} \\		  
            \includegraphics[width=1\textwidth]{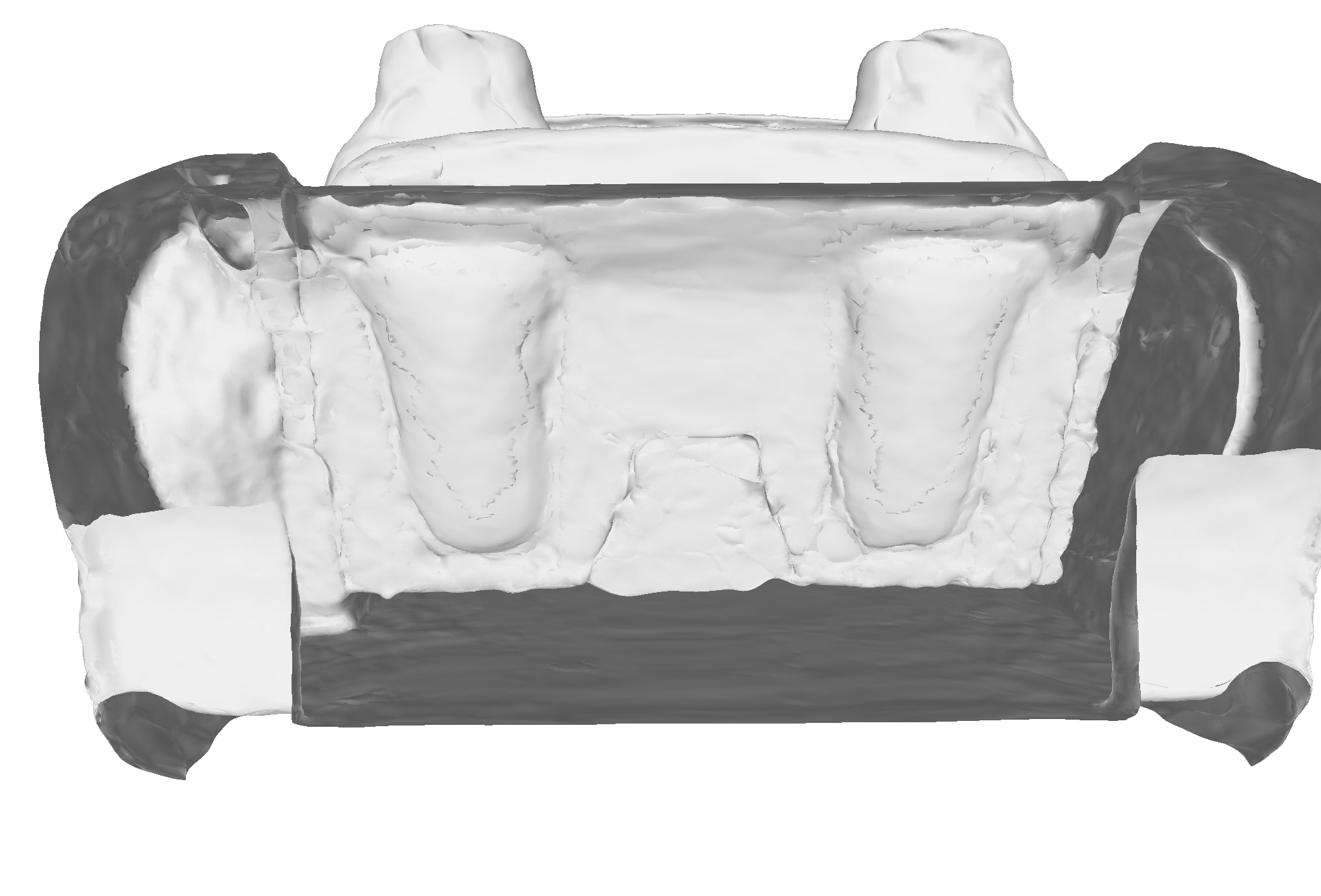} \\		    		  
            \includegraphics[width=1\textwidth]{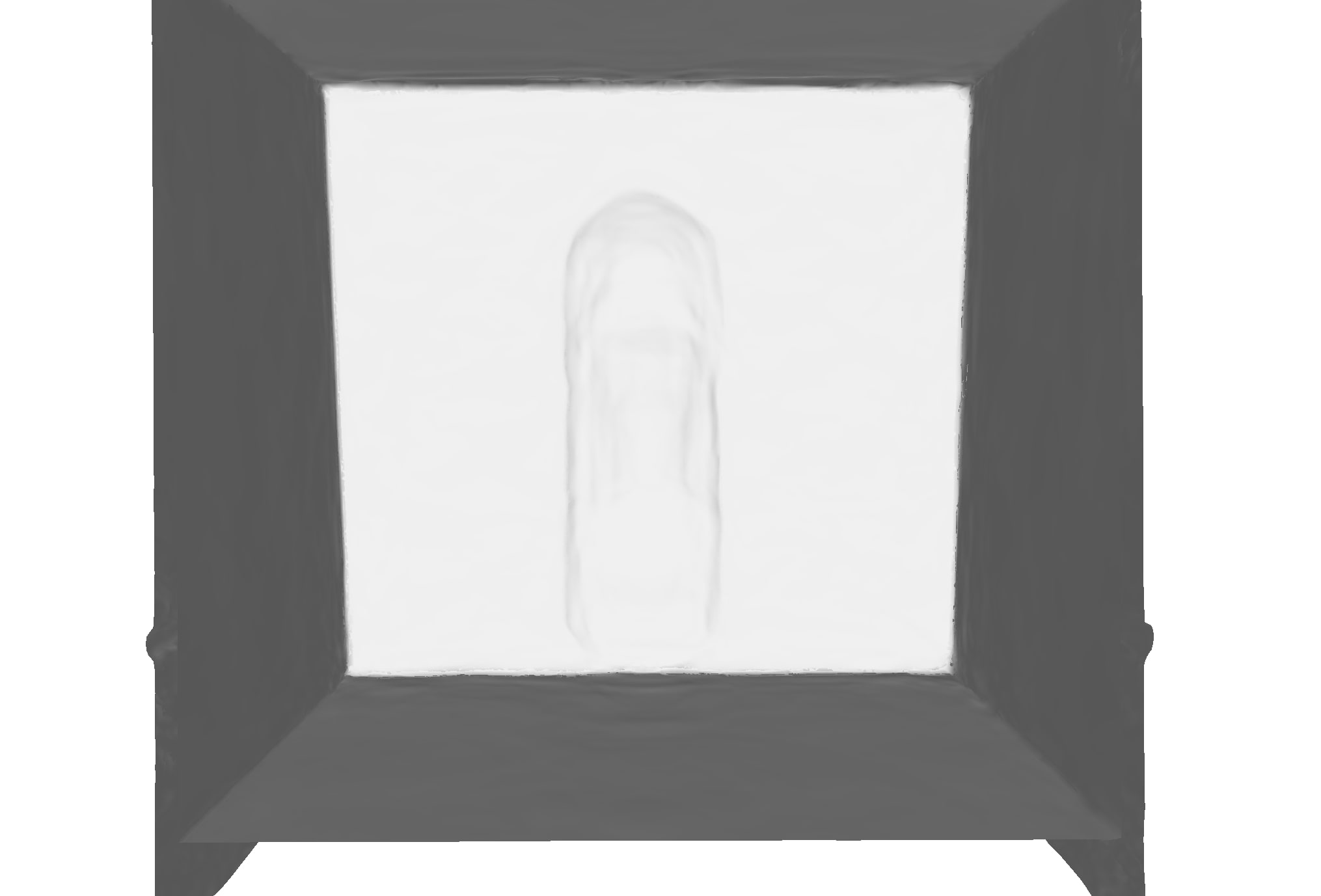} \\		 
            \includegraphics[width=1\textwidth]{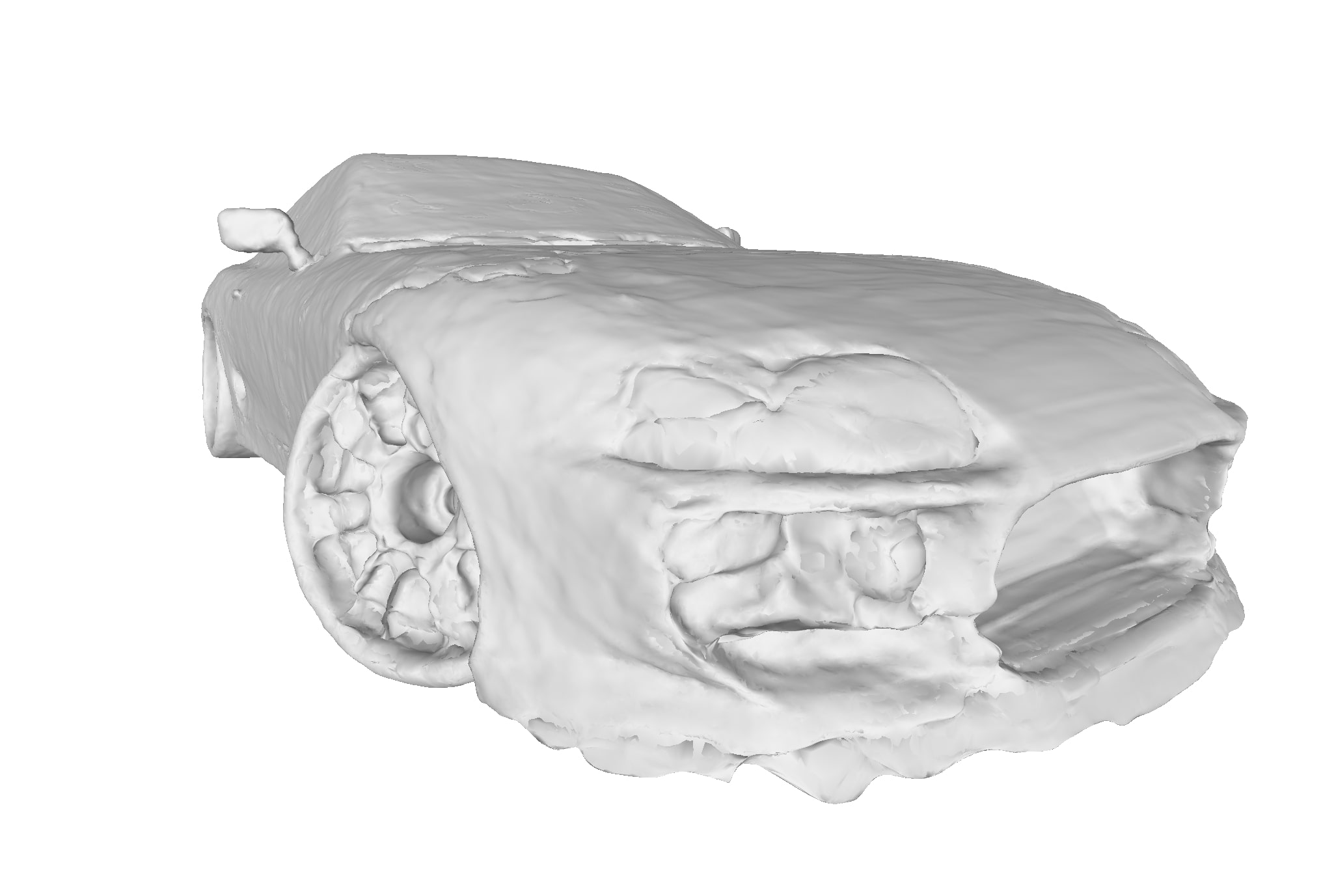} \\		  
            \includegraphics[width=1\textwidth]{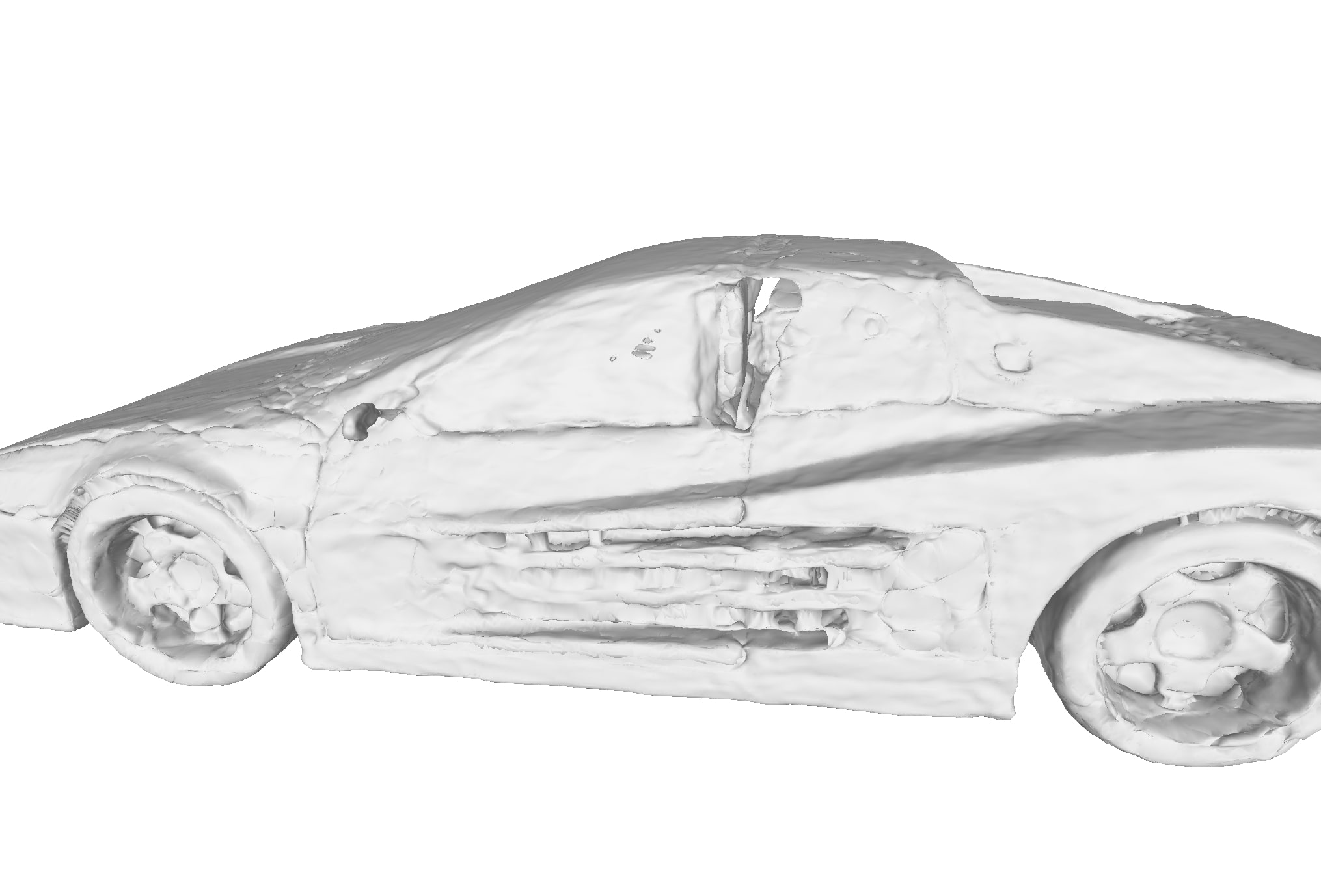} \\
            \includegraphics[width=1\textwidth]{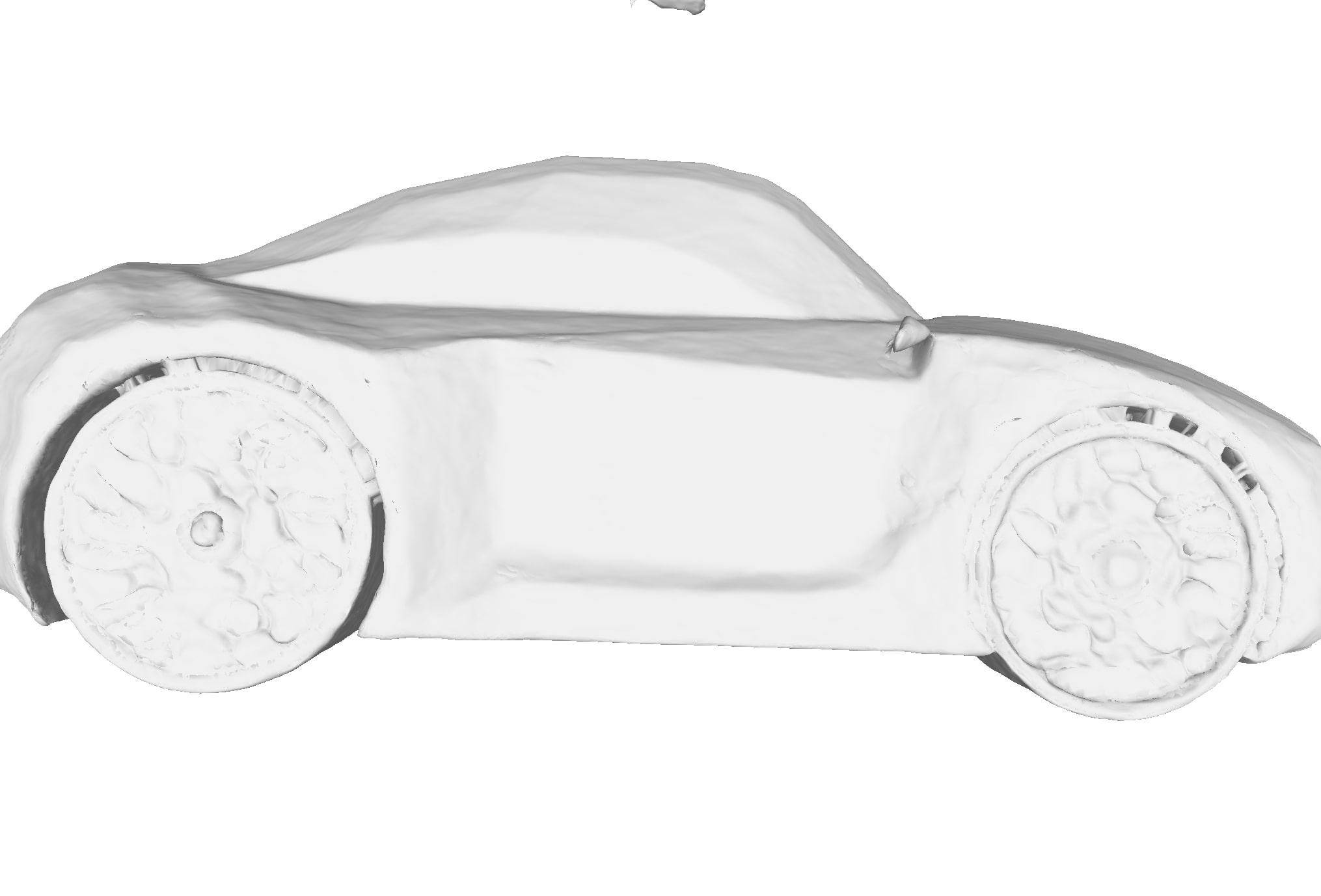}
        \end{minipage}
    }
    \subfigure[LevelSetUDF]{
        \begin{minipage}[b]{0.18\textwidth}
		  \includegraphics[width=1\textwidth]{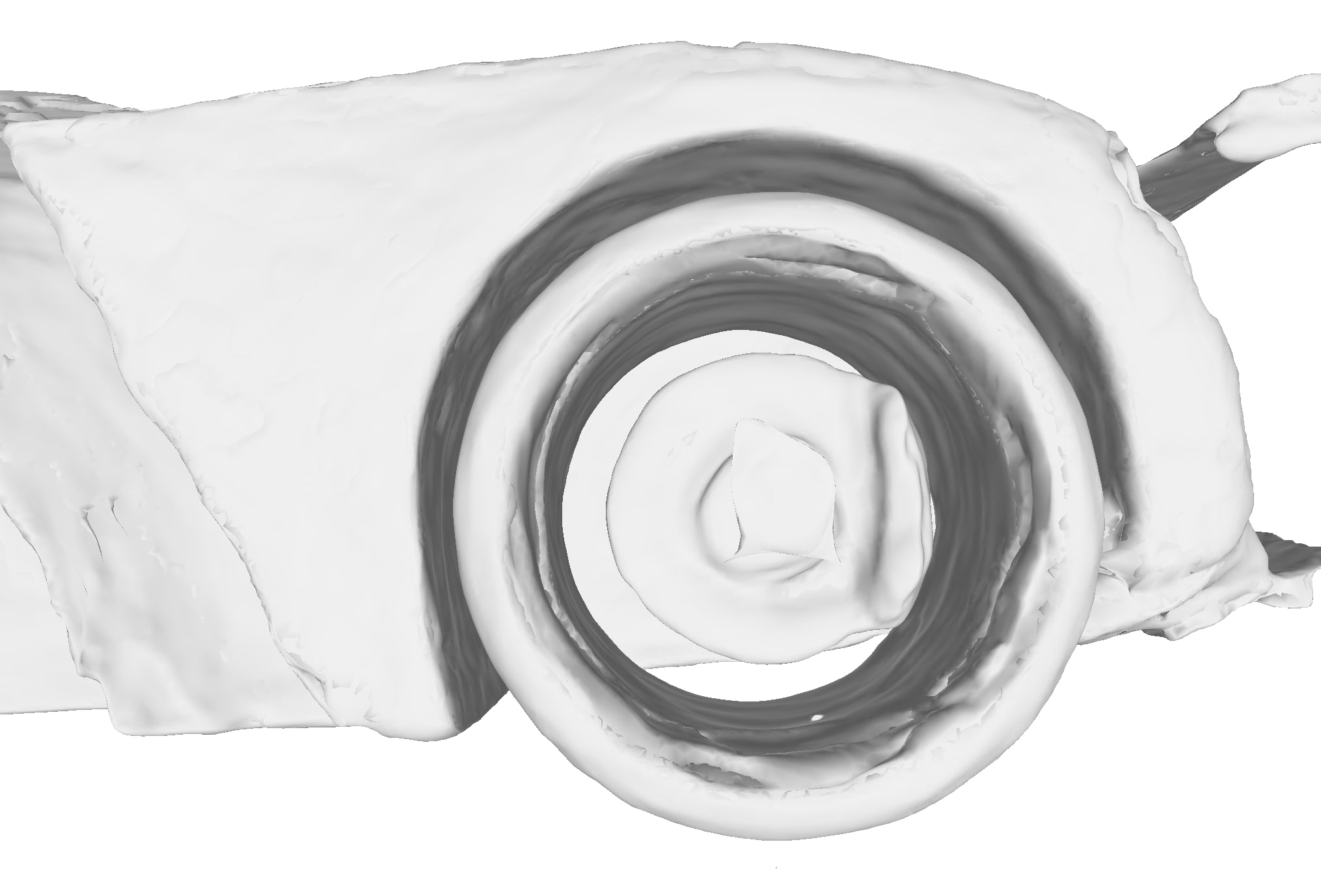} \\
		  \includegraphics[width=1\textwidth]{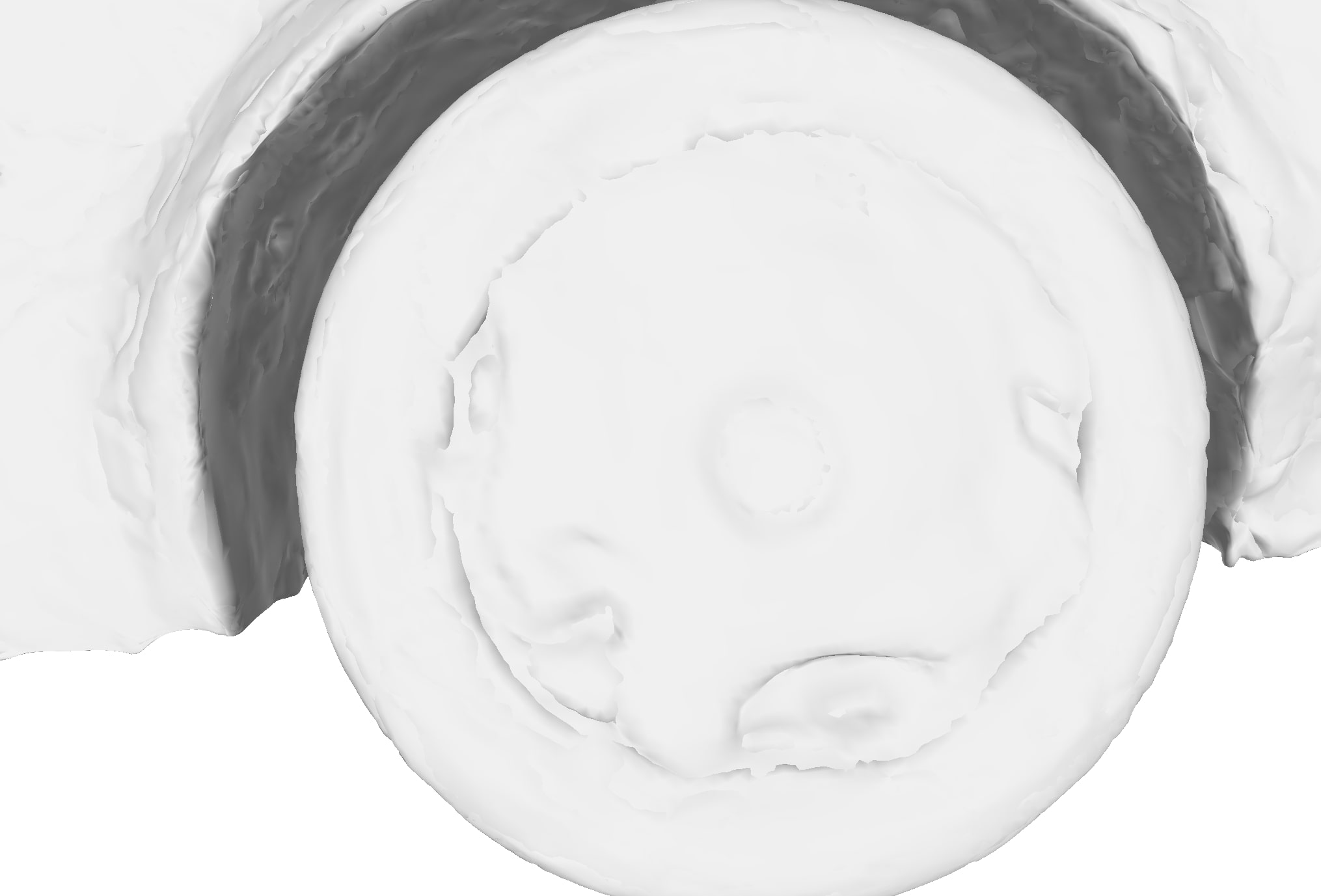} \\	  
            \includegraphics[width=1\textwidth]{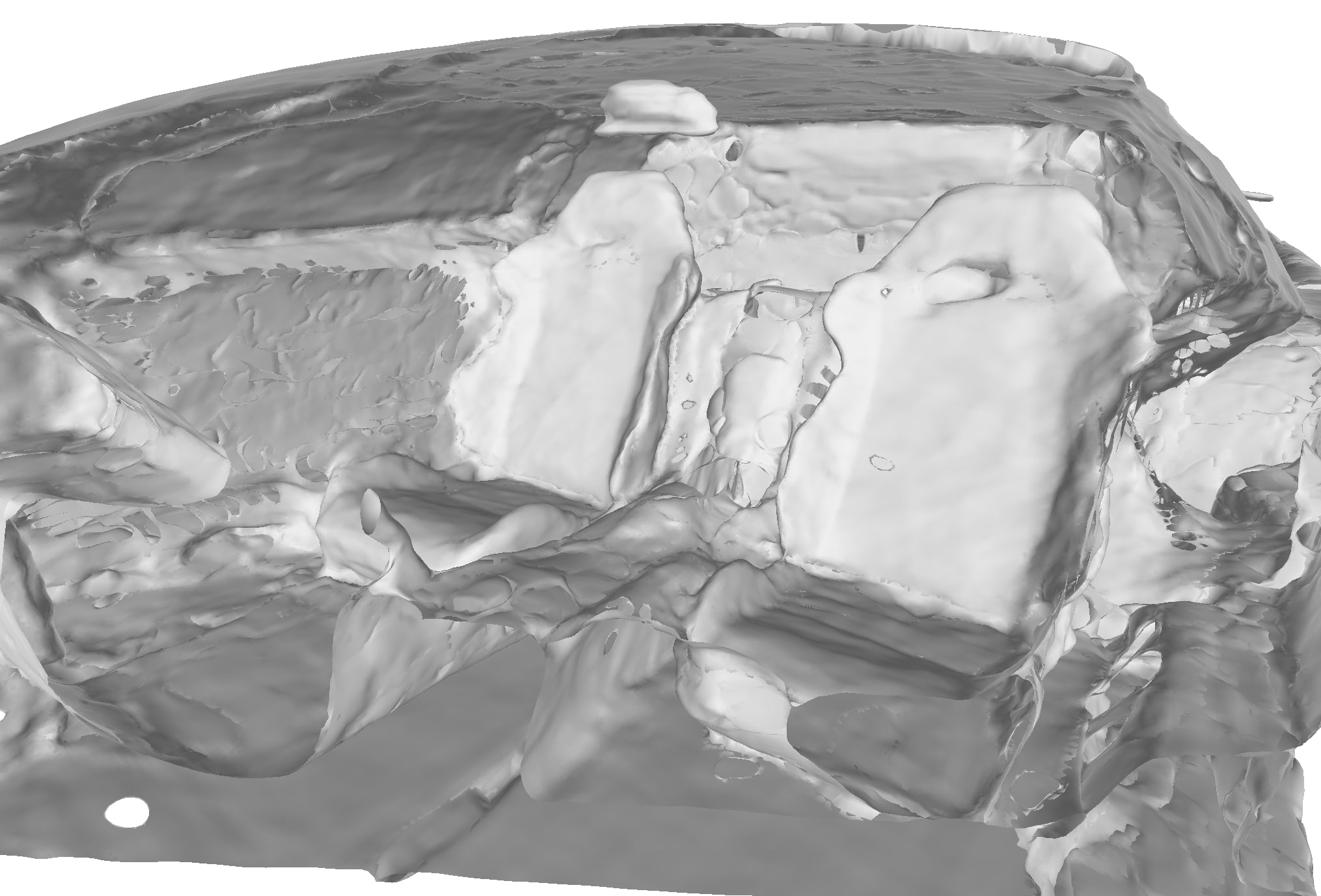} \\		  
            \includegraphics[width=1\textwidth]{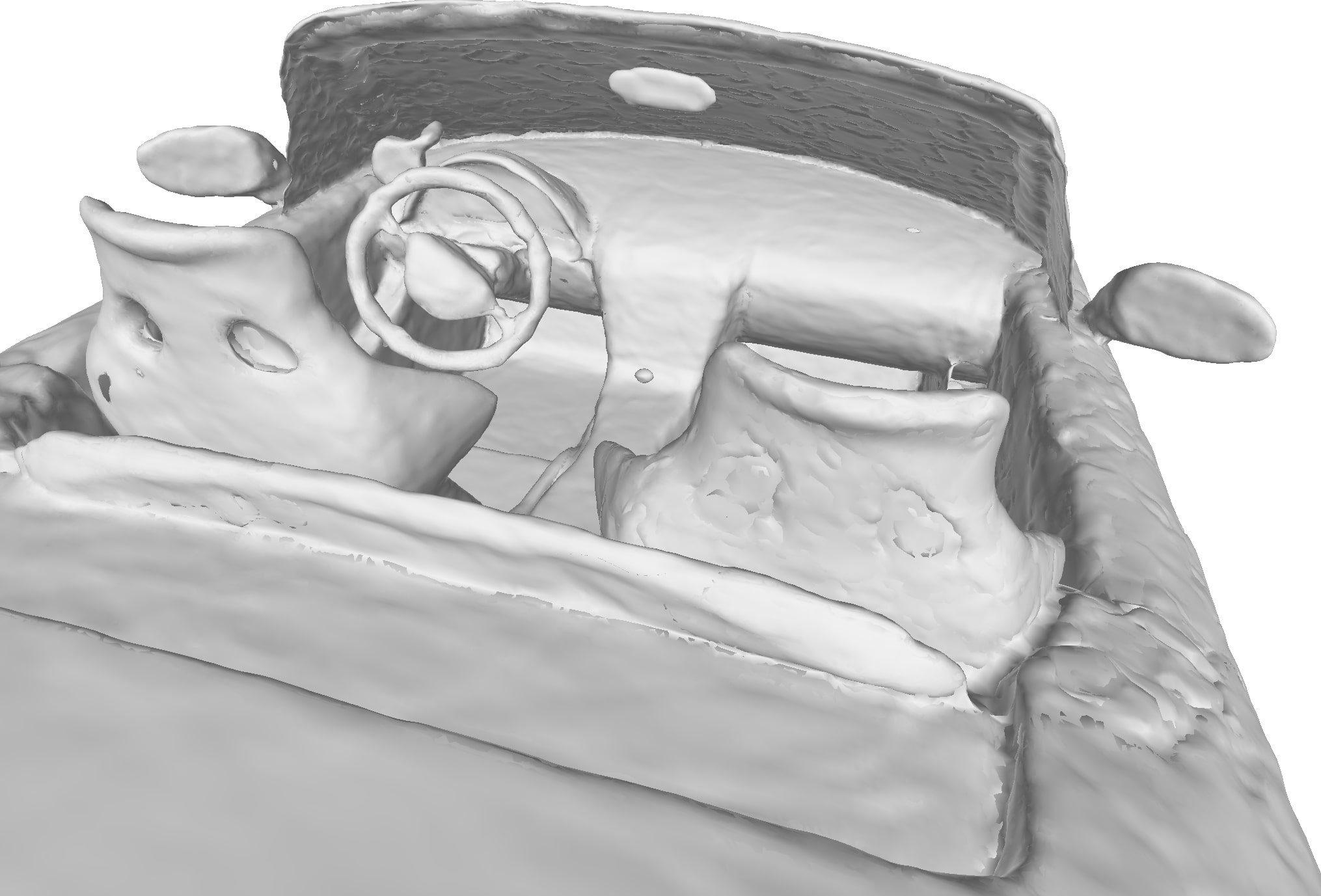} \\		  
            \includegraphics[width=1\textwidth]{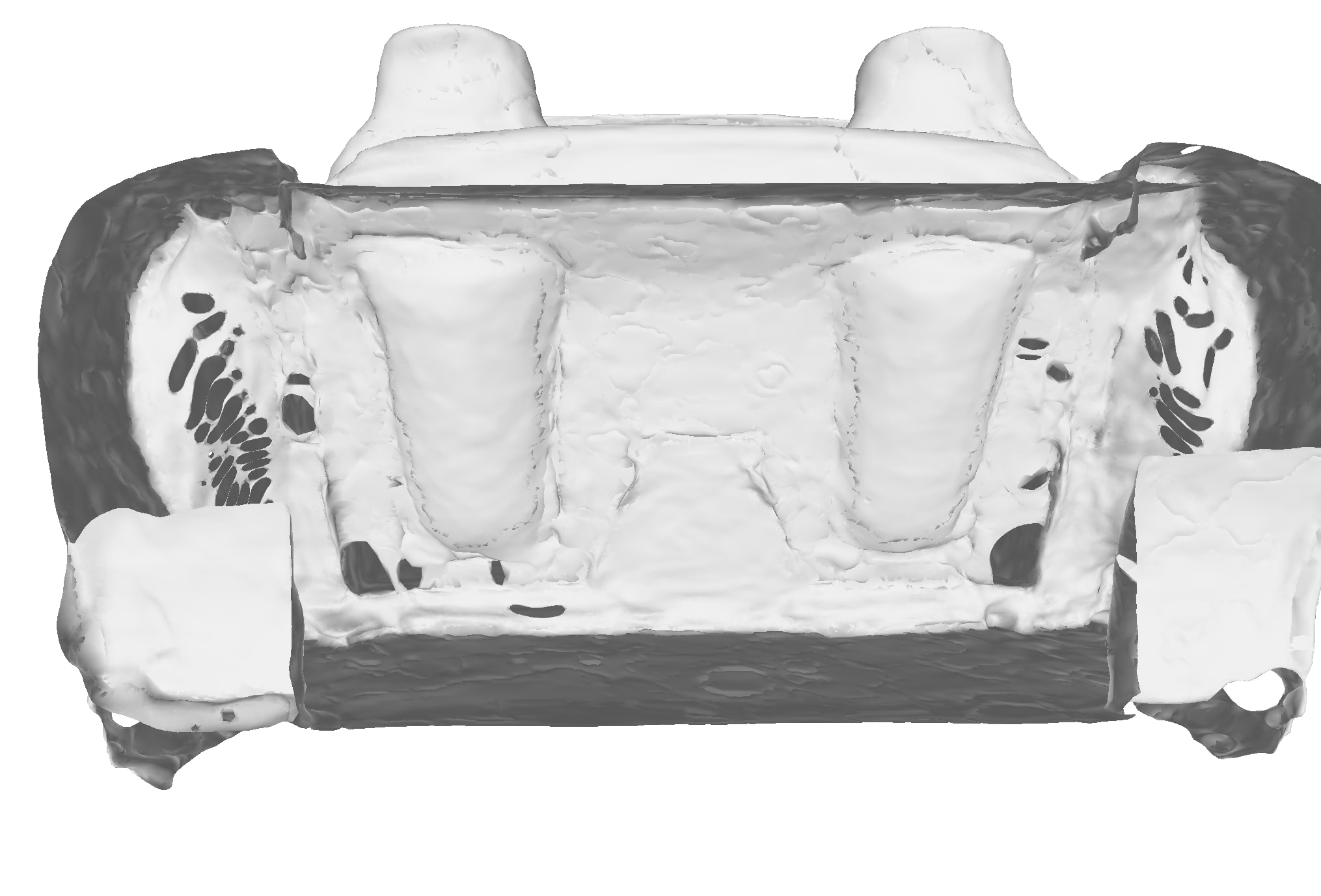} \\		  	  	  
            \includegraphics[width=1\textwidth]{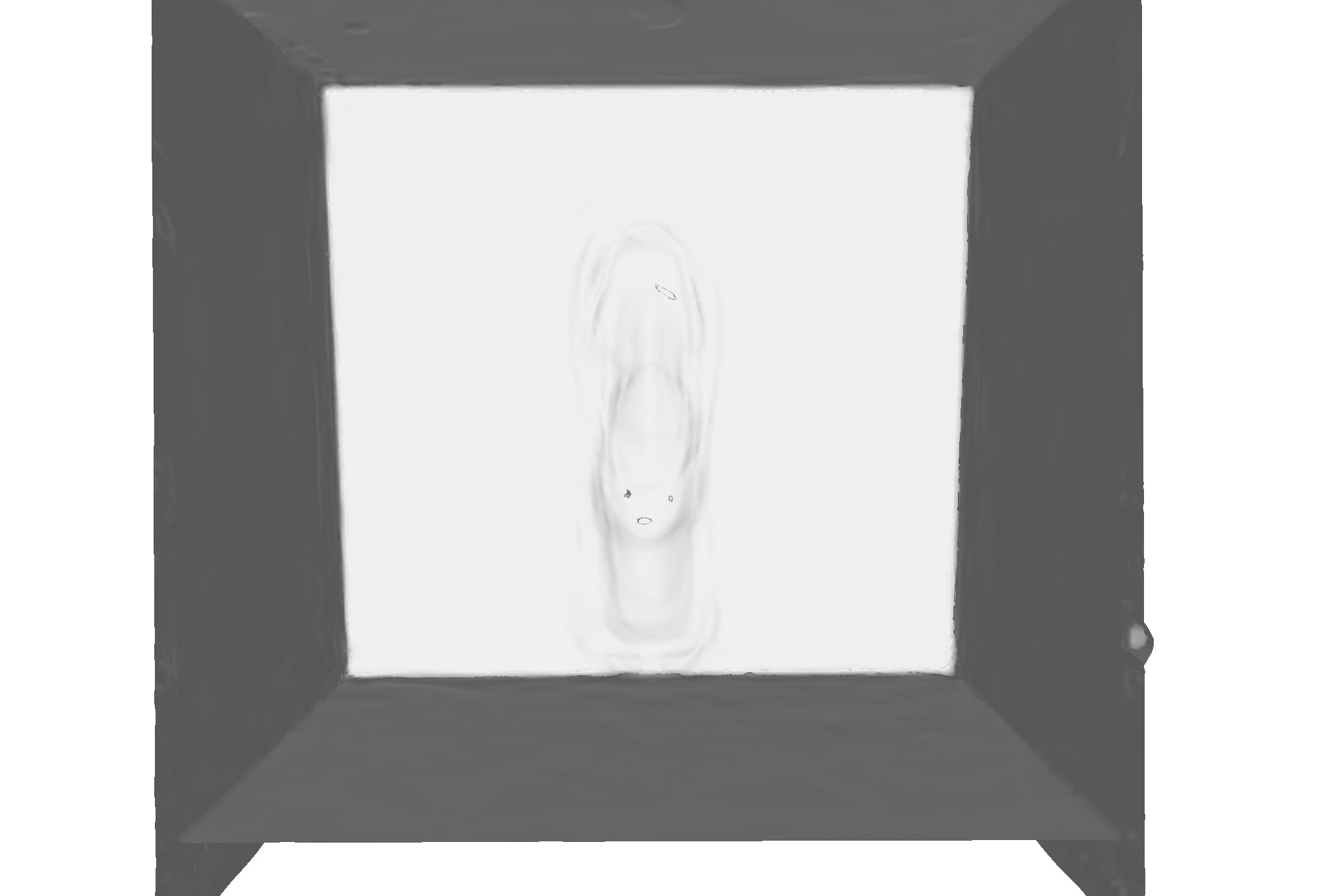} \\		 
            \includegraphics[width=1\textwidth]{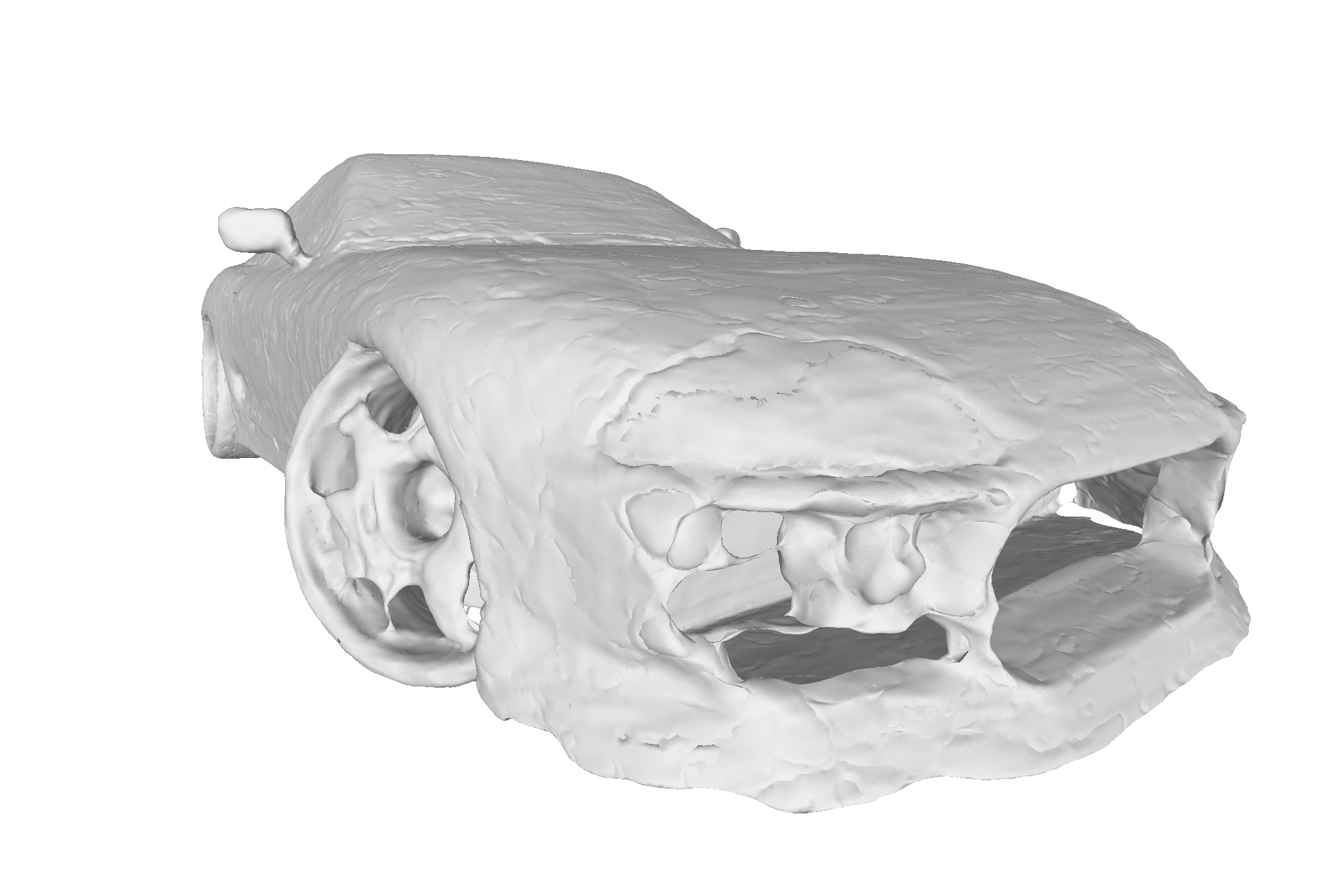} \\		  
            \includegraphics[width=1\textwidth]{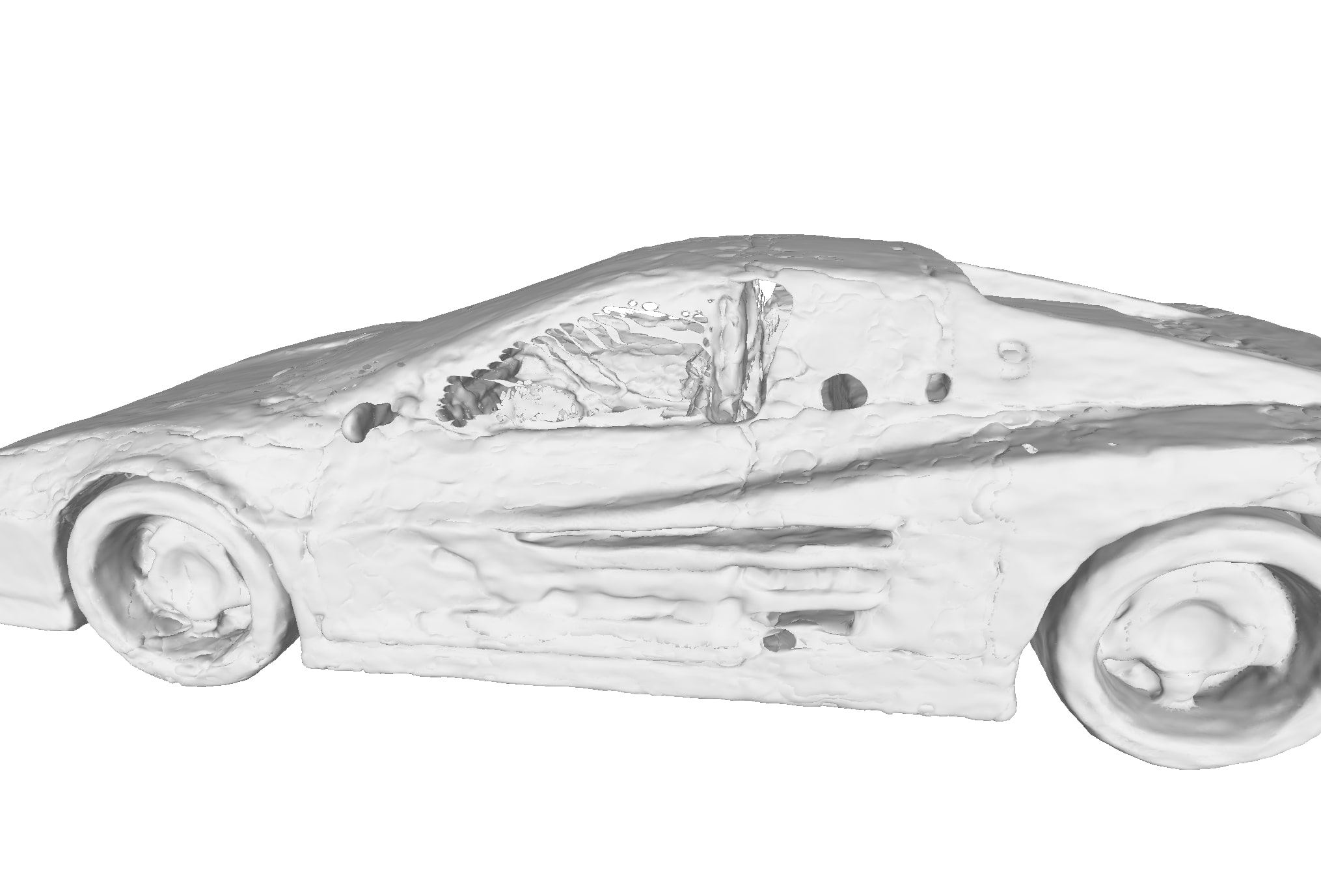} \\
            \includegraphics[width=1\textwidth]{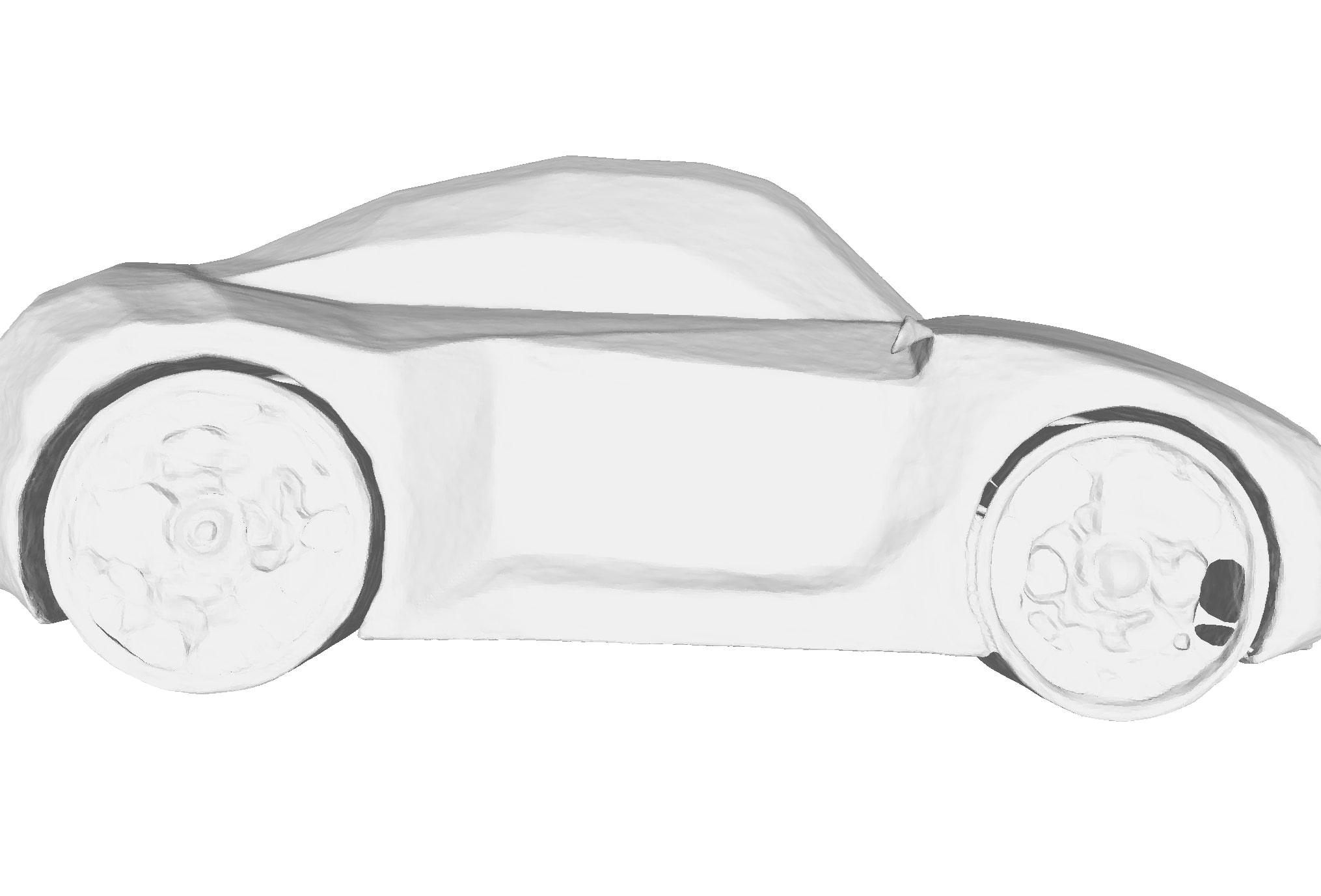}
        \end{minipage}
    }
    \subfigure[Ours]{
        \begin{minipage}[b]{0.18\textwidth}
		  \includegraphics[width=1\textwidth]{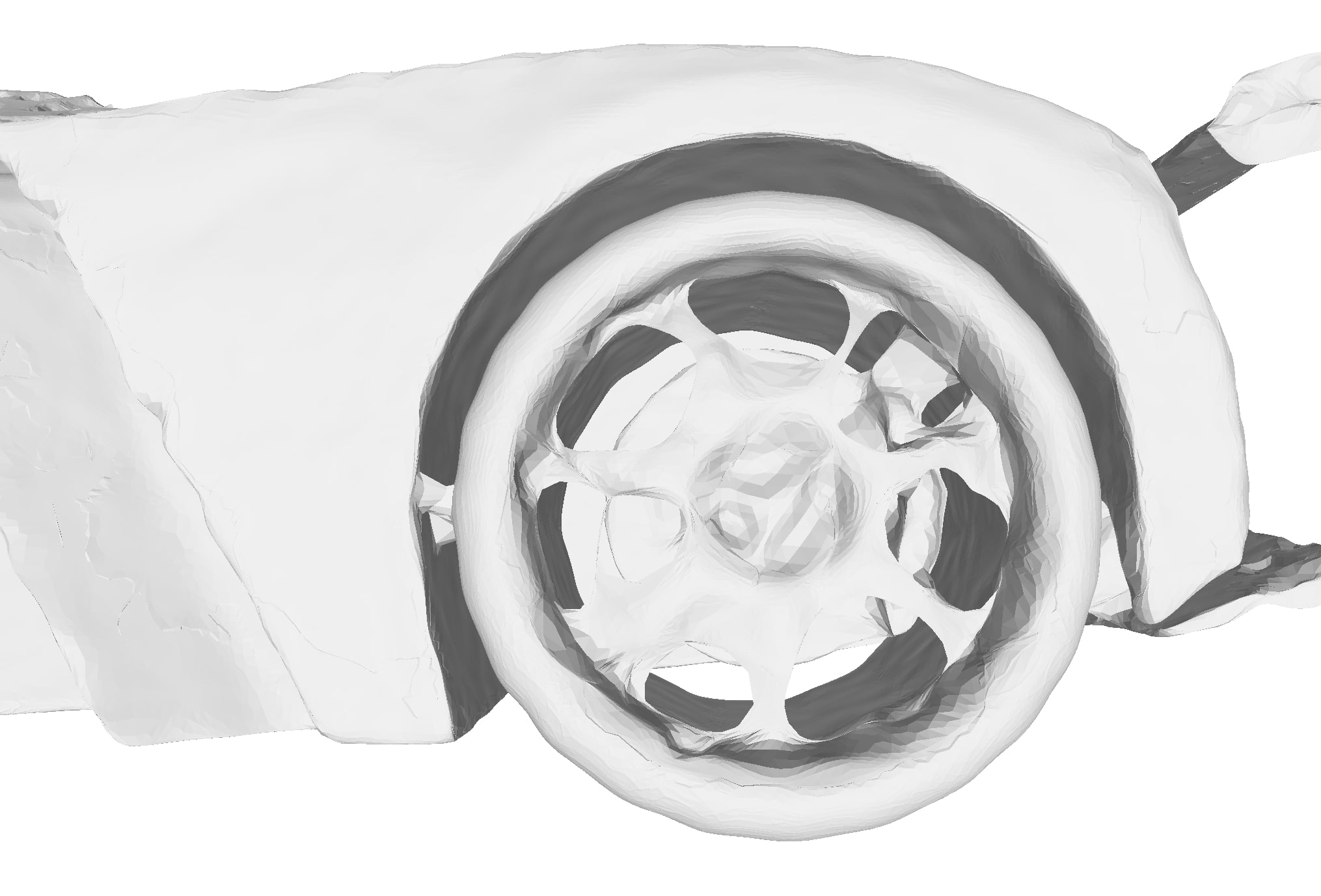} \\
		  \includegraphics[width=1\textwidth]{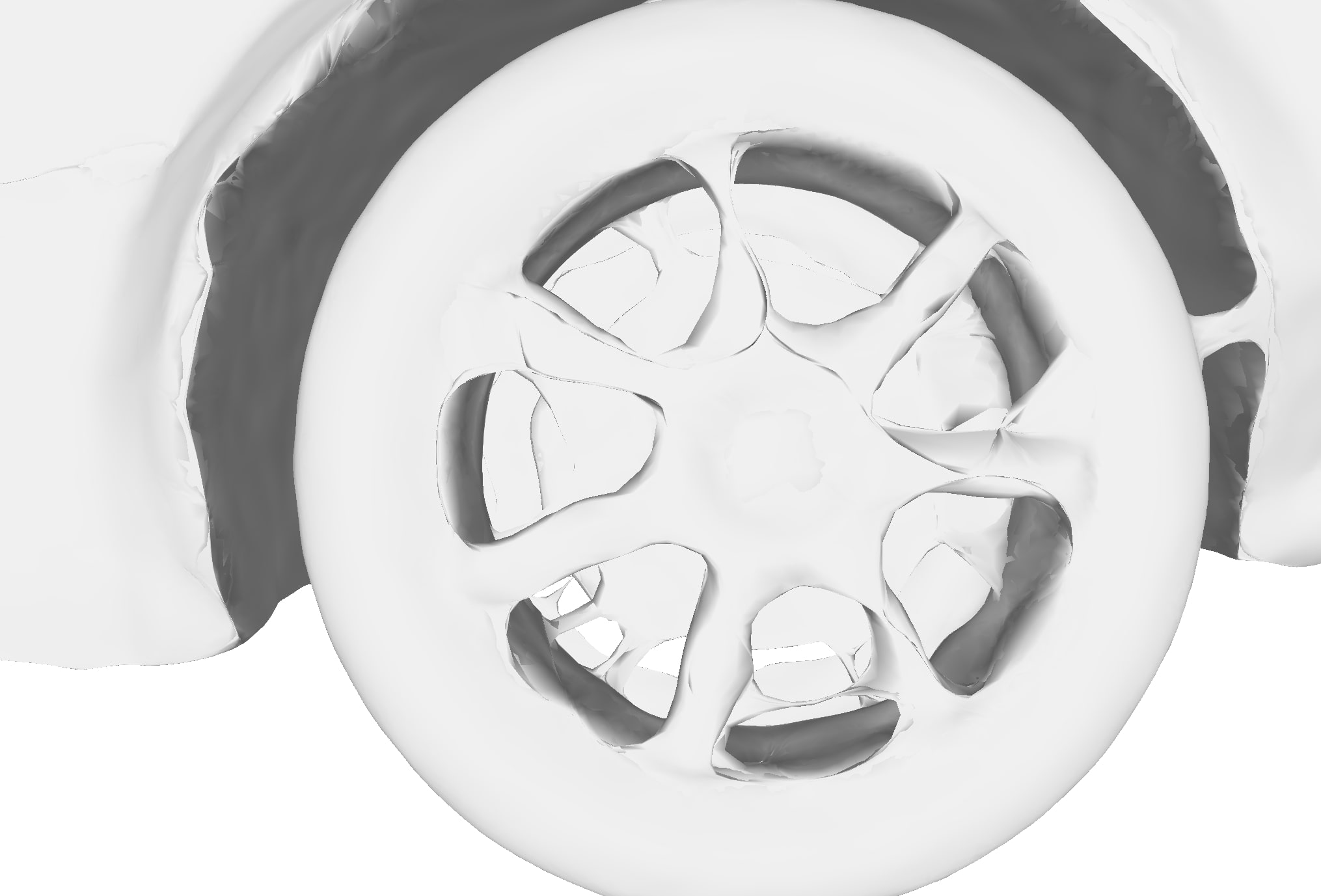} \\  
            \includegraphics[width=1\textwidth]{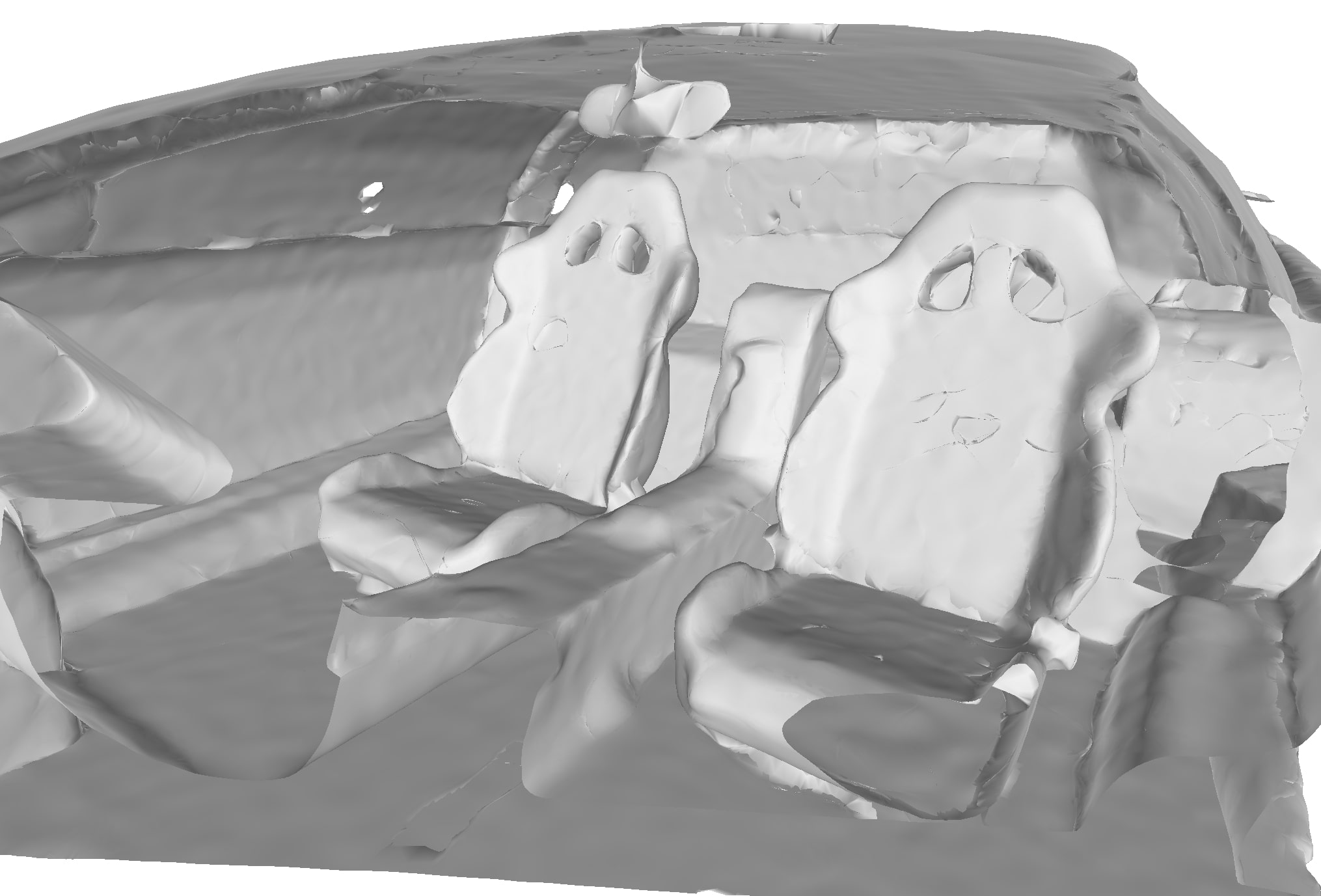} \\		  
            \includegraphics[width=1\textwidth]{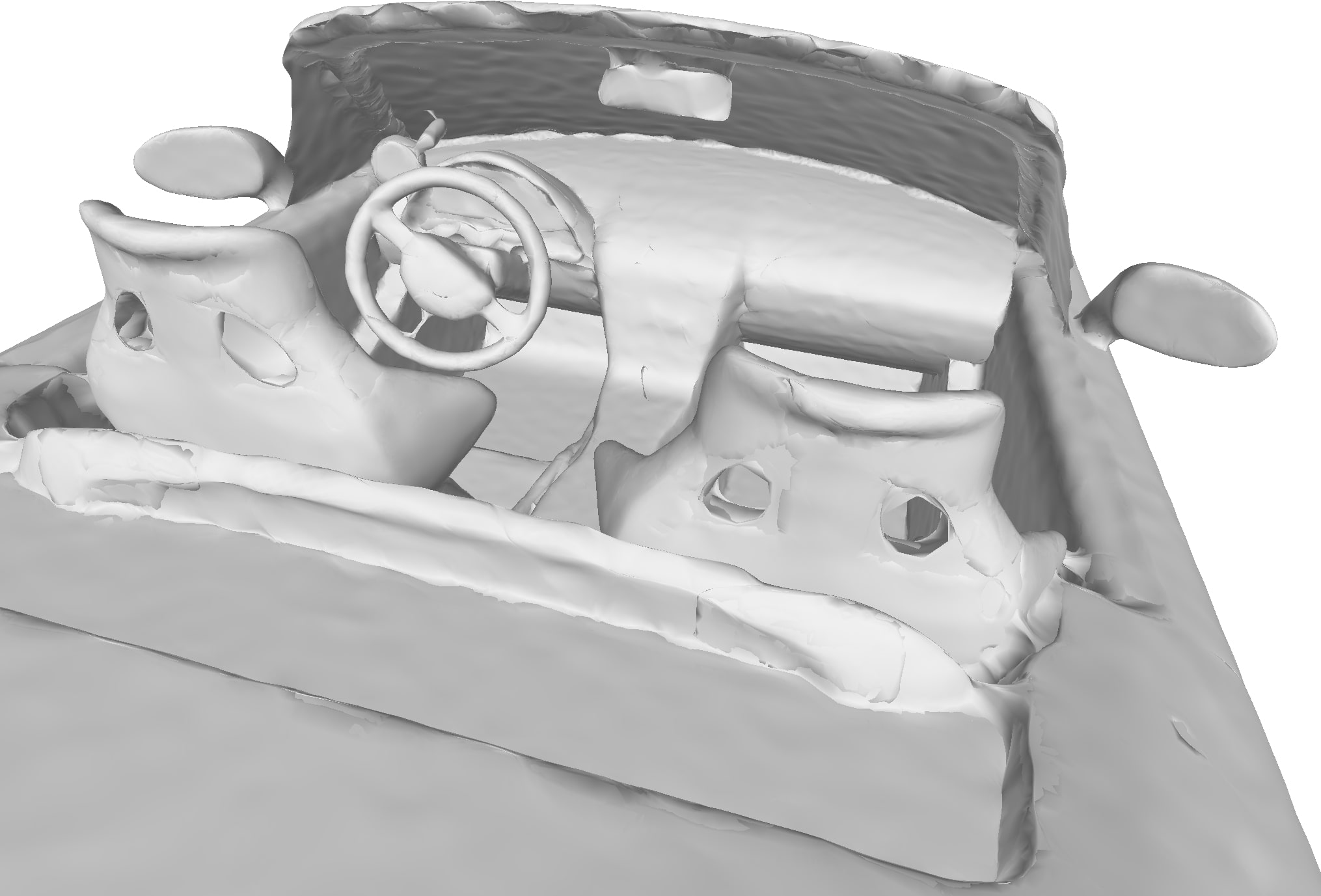} \\		  
            \includegraphics[width=1\textwidth]{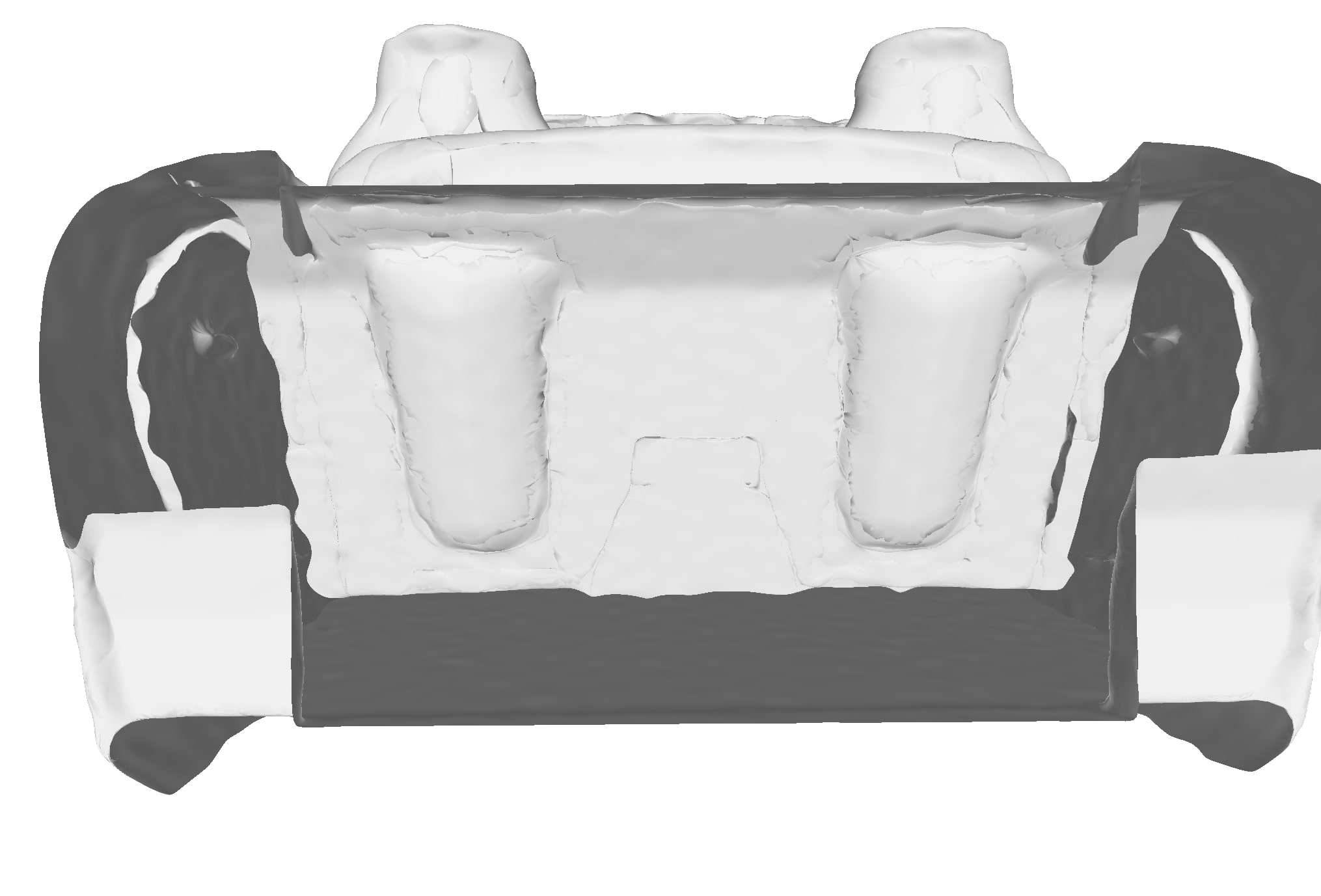} \\		  	  	  
            \includegraphics[width=1\textwidth]{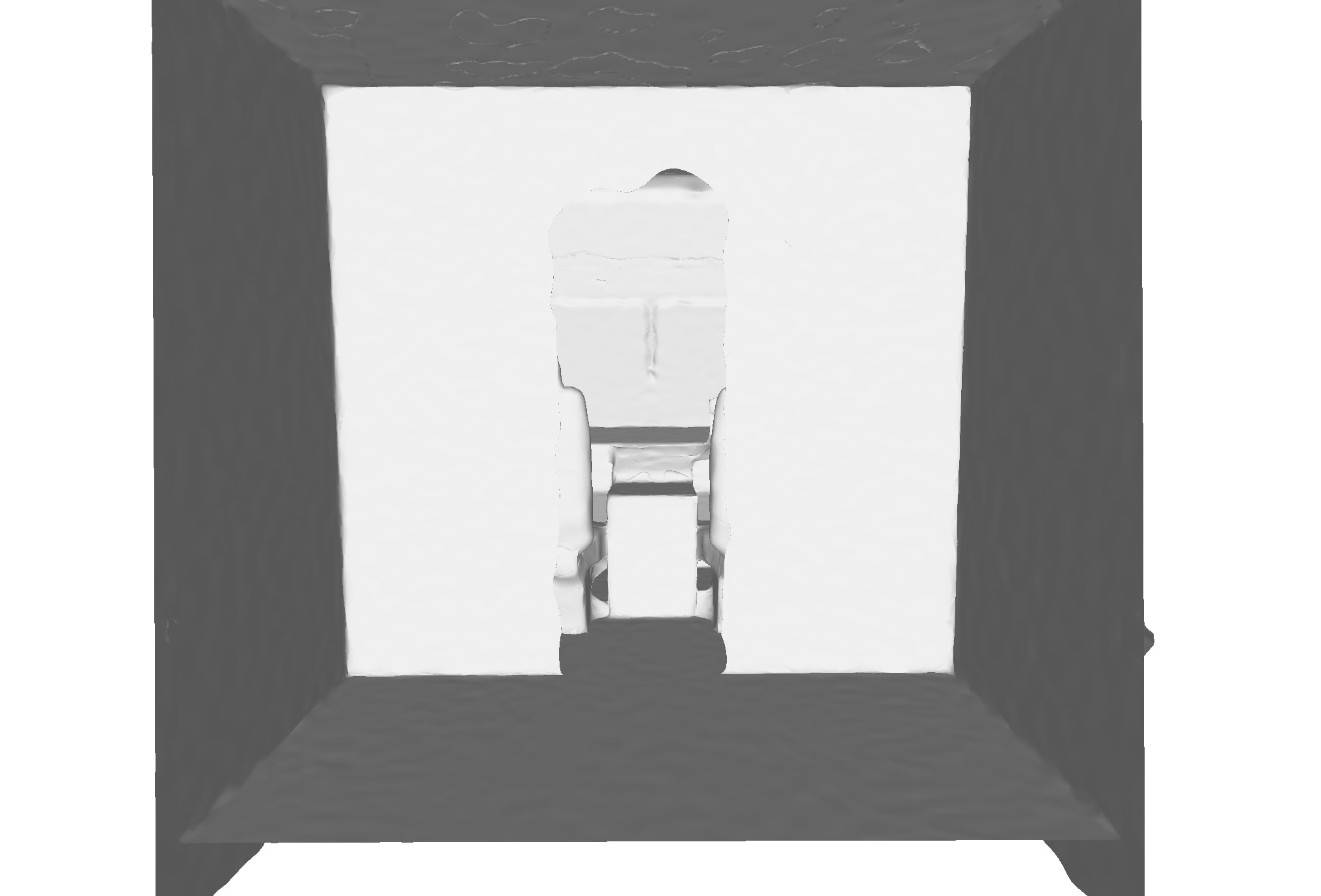} \\		 
            \includegraphics[width=1\textwidth]{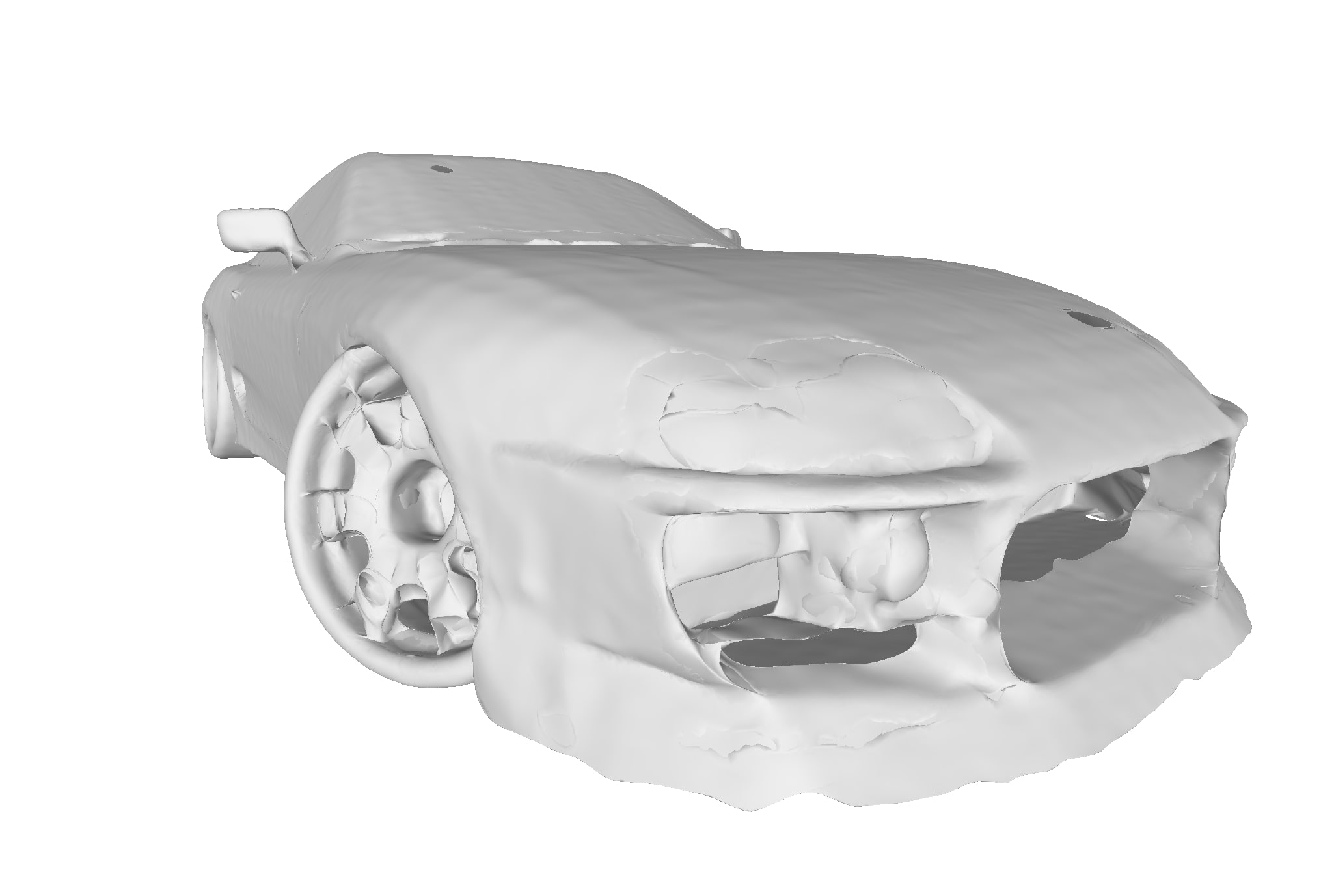} \\		  
            \includegraphics[width=1\textwidth]{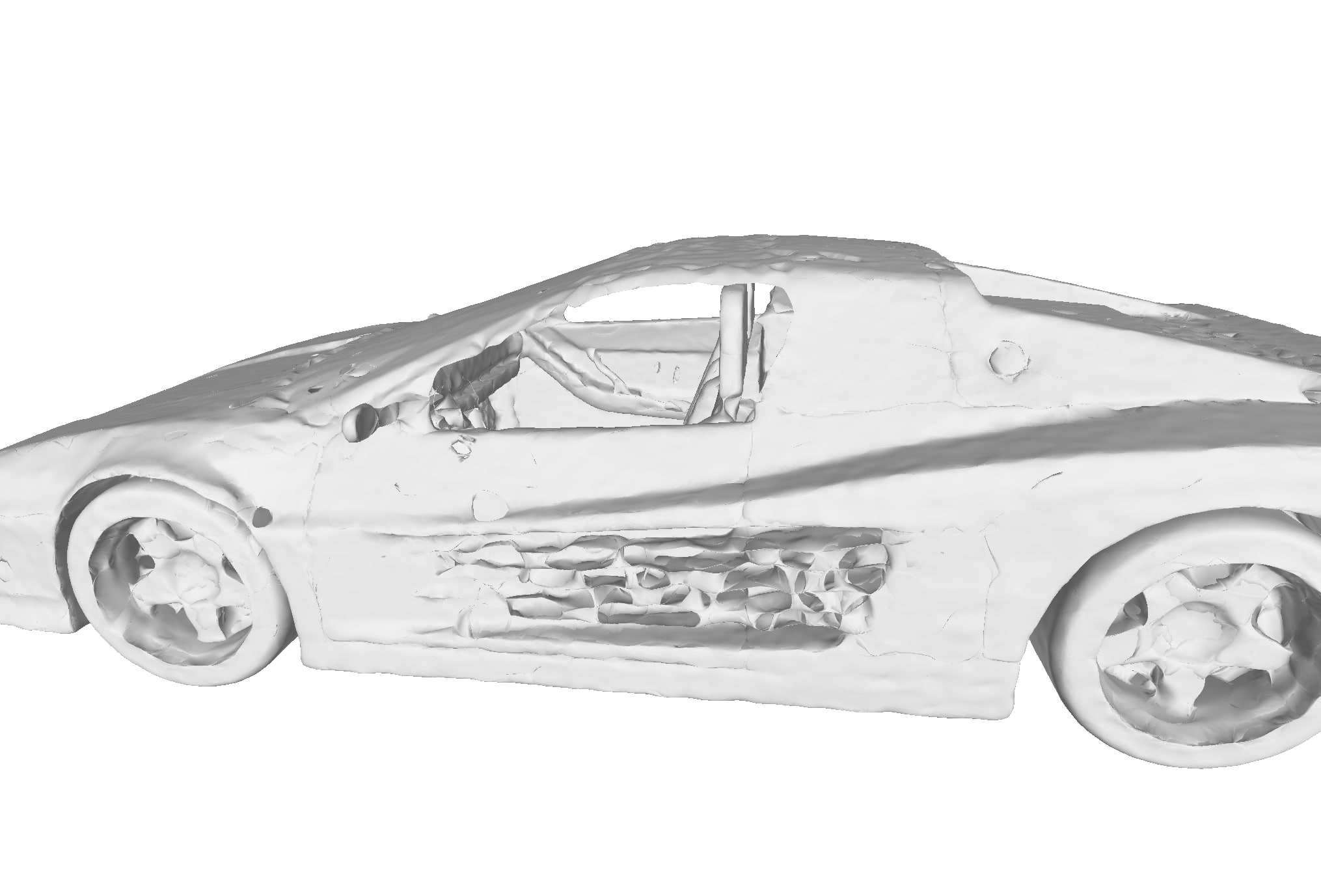} \\
            \includegraphics[width=1\textwidth]{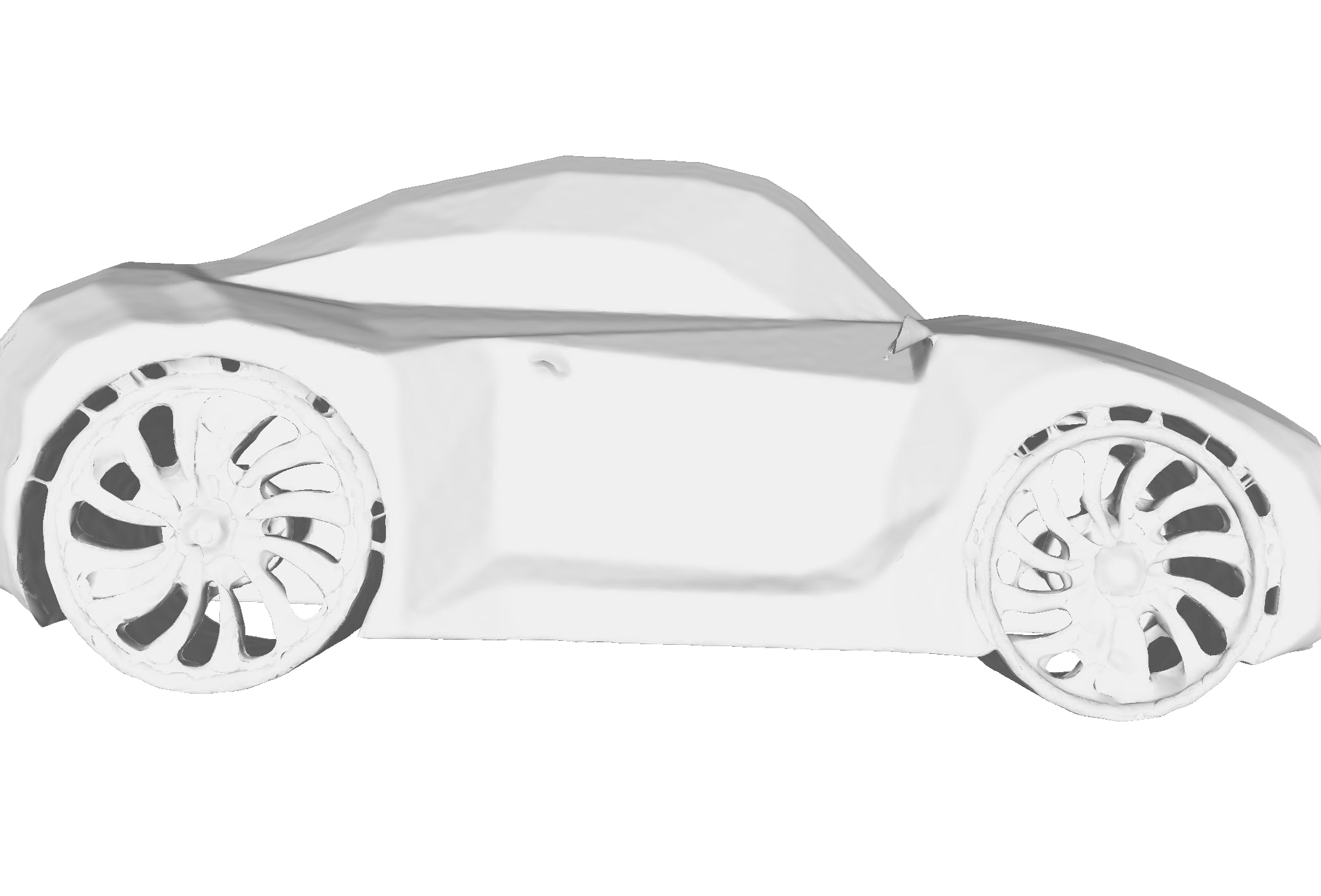}
        \end{minipage}
    }
    \subfigure[GT]{
        \begin{minipage}[b]{0.18\textwidth}
            \includegraphics[width=1\textwidth]{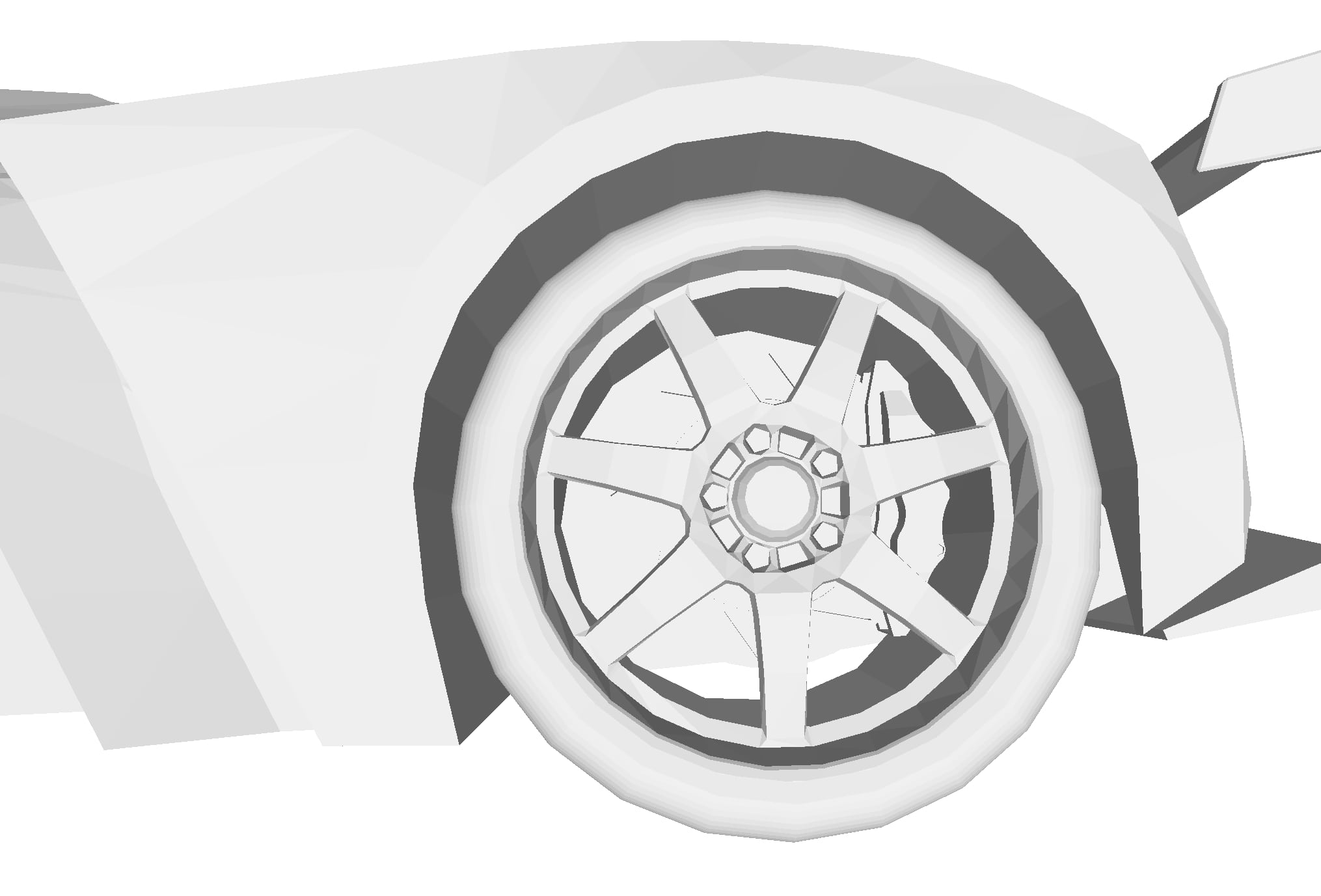} \\
		  \includegraphics[width=1\textwidth]{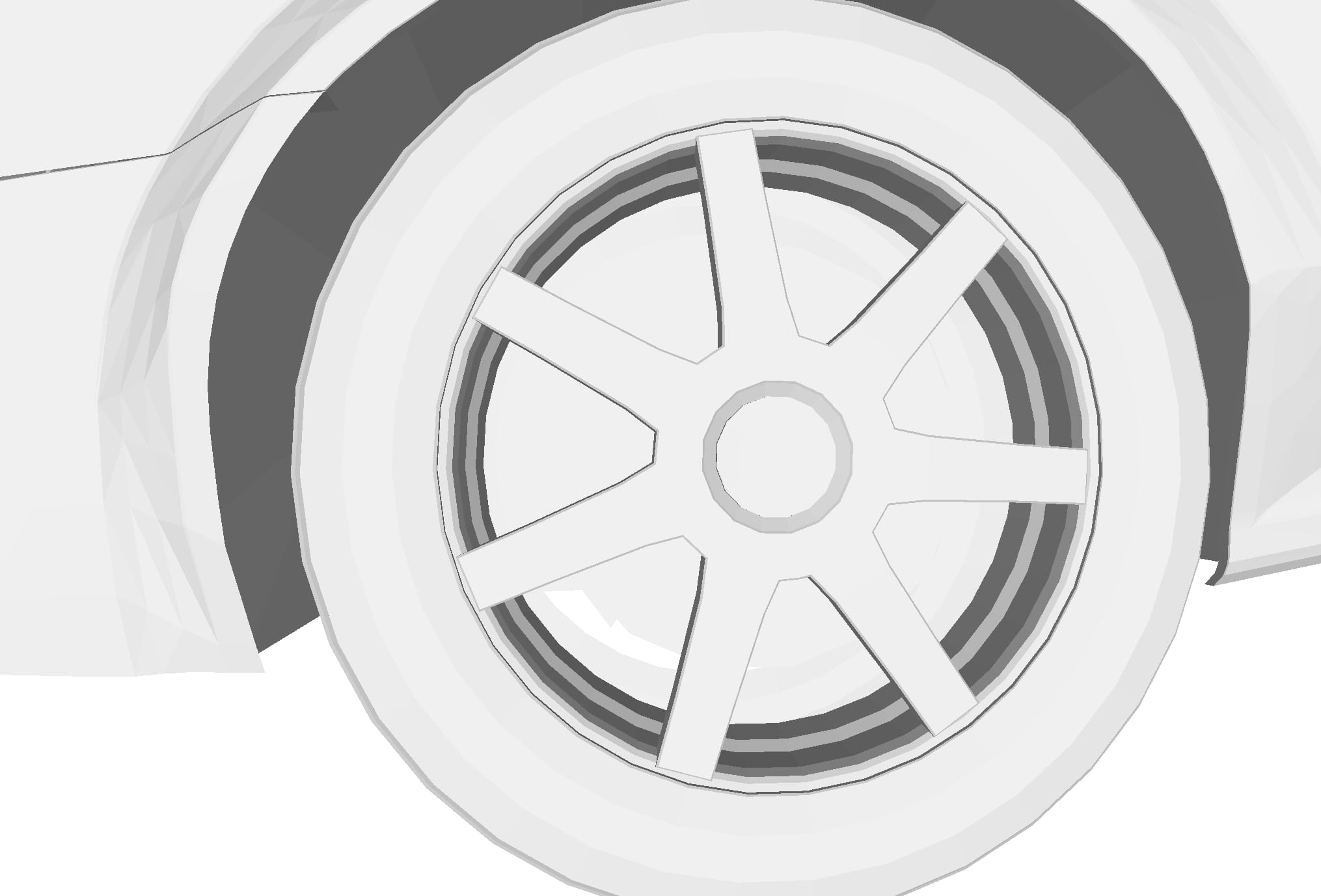} \\
            \includegraphics[width=1\textwidth]{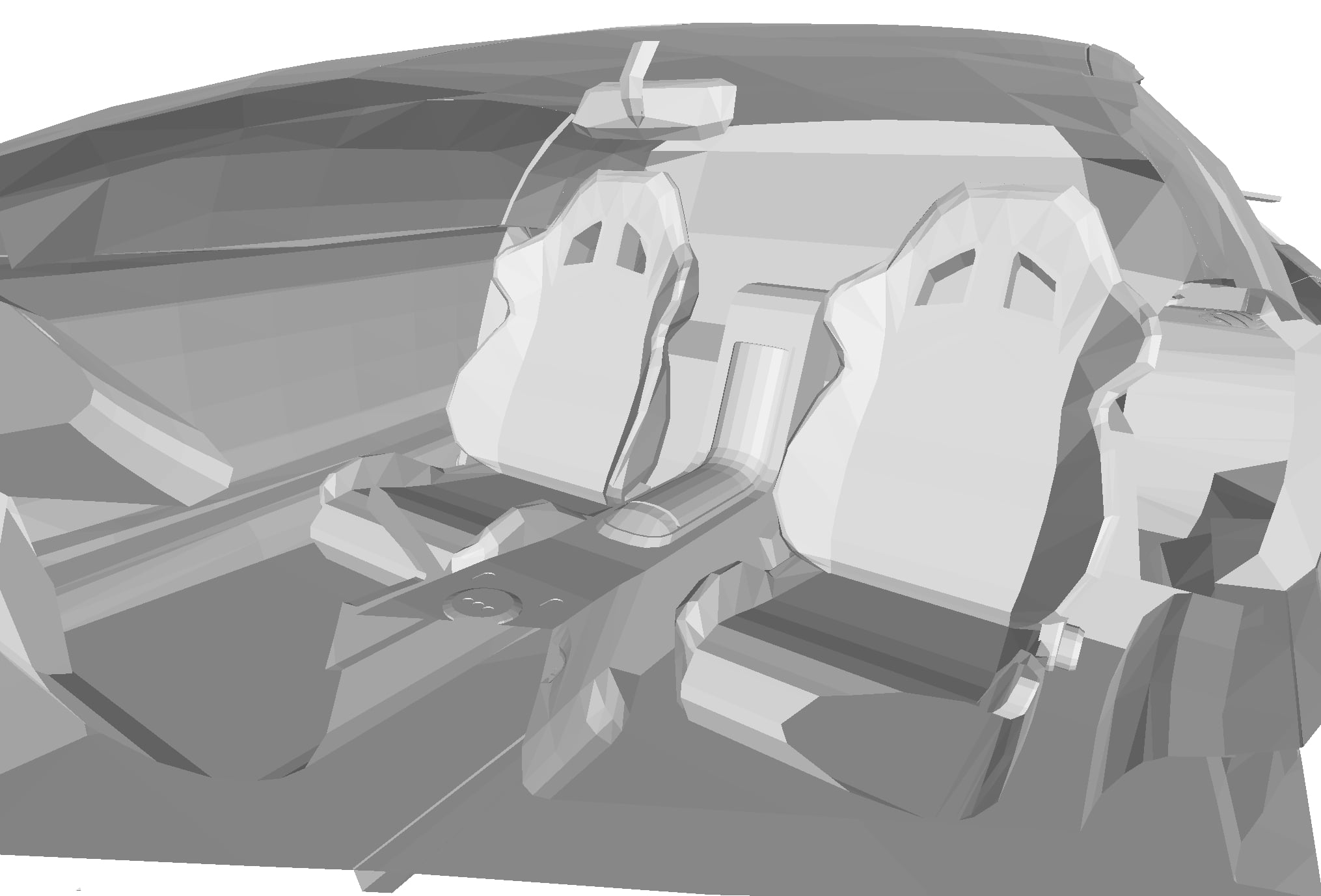} \\		  
            \includegraphics[width=1\textwidth]{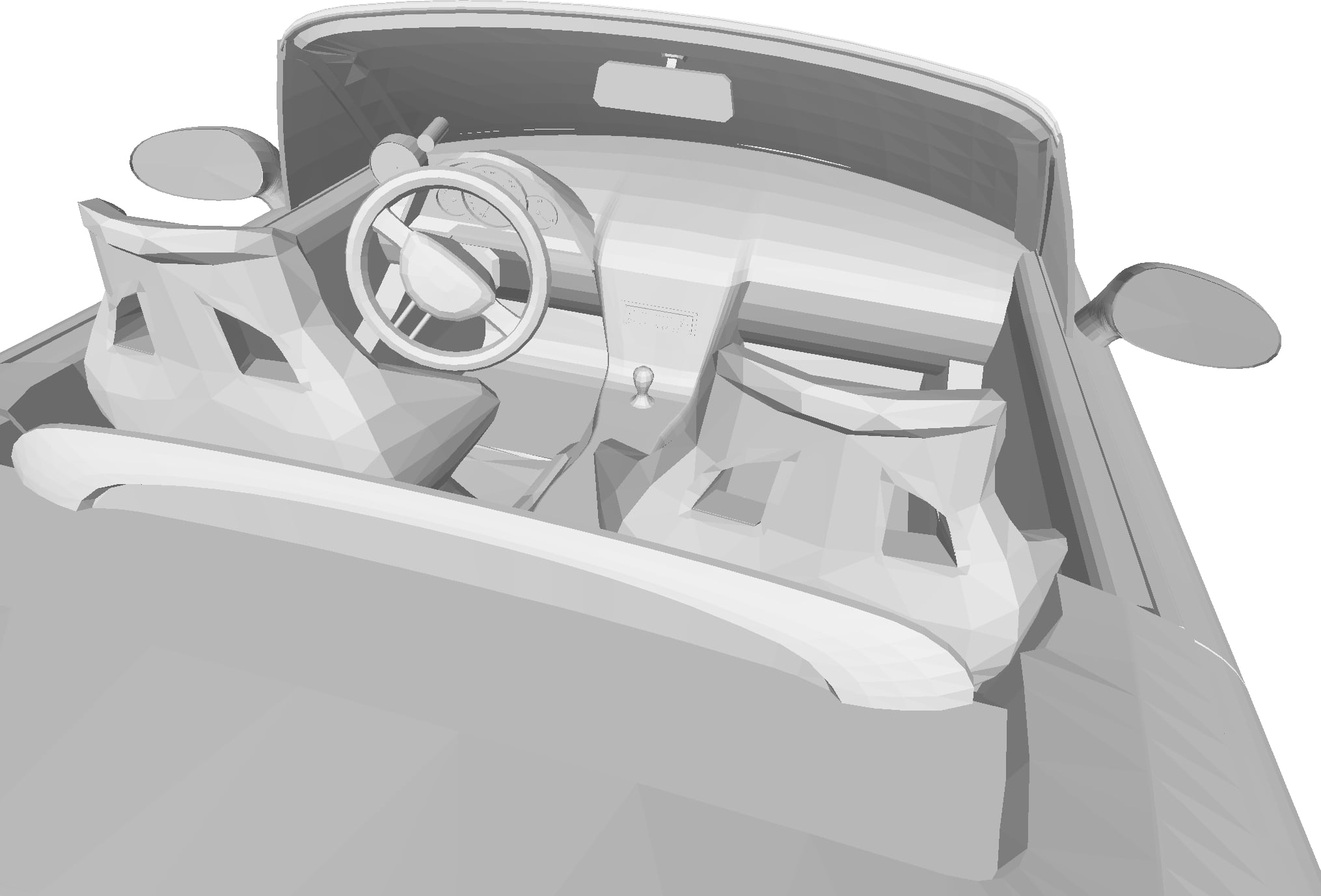} \\		  
            \includegraphics[width=1\textwidth]{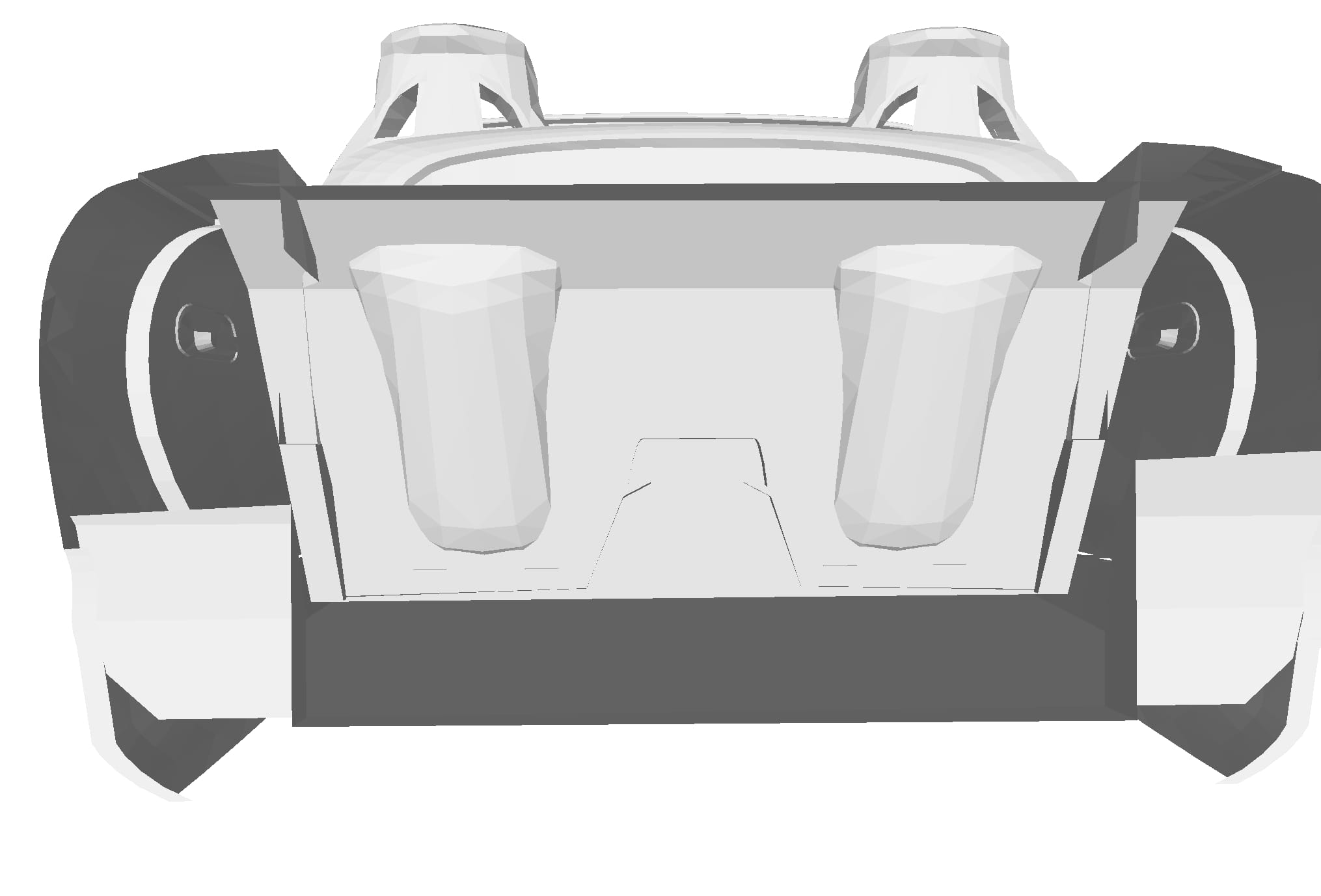} \\		  	  
            \includegraphics[width=1\textwidth]{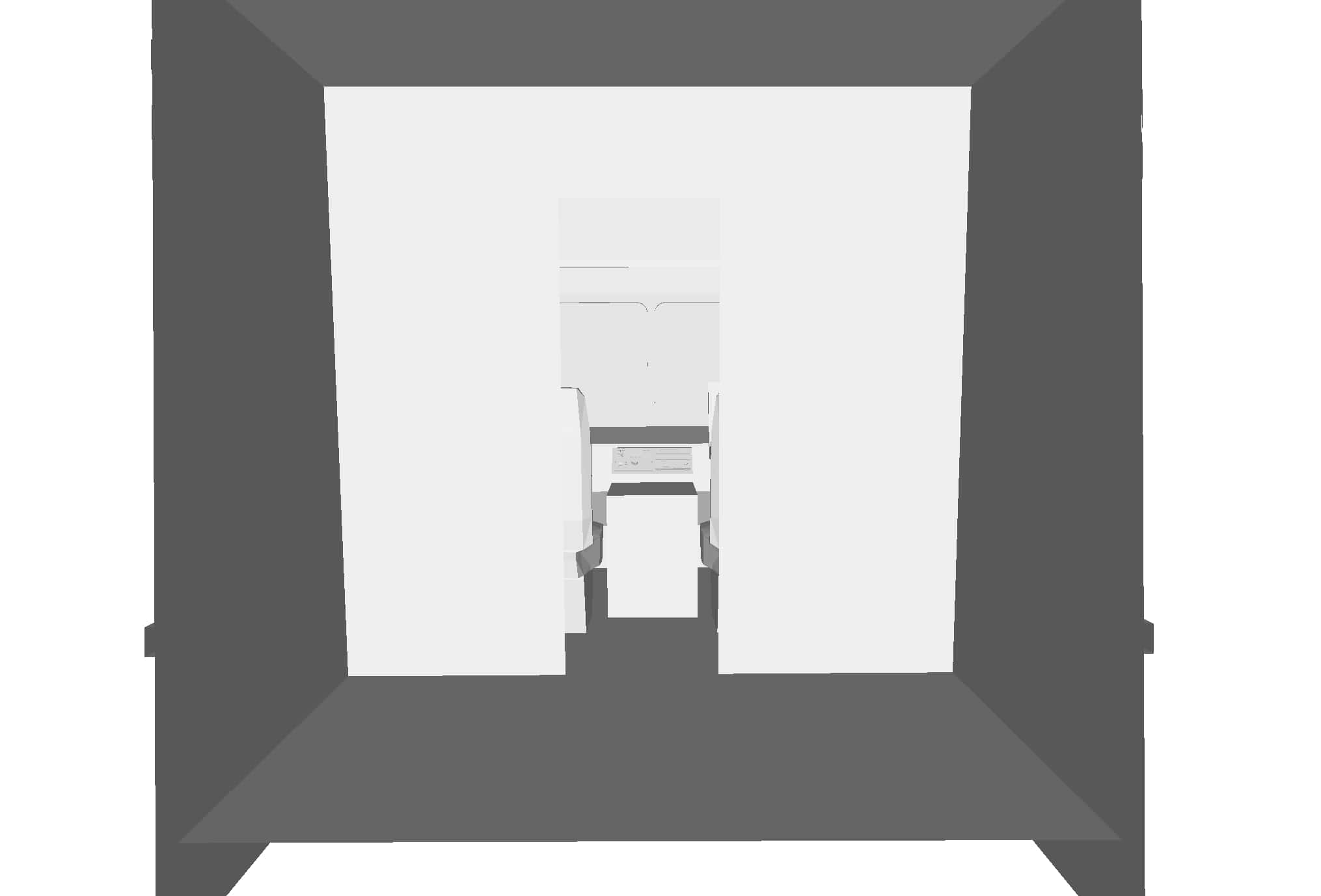} \\		 
            \includegraphics[width=1\textwidth]{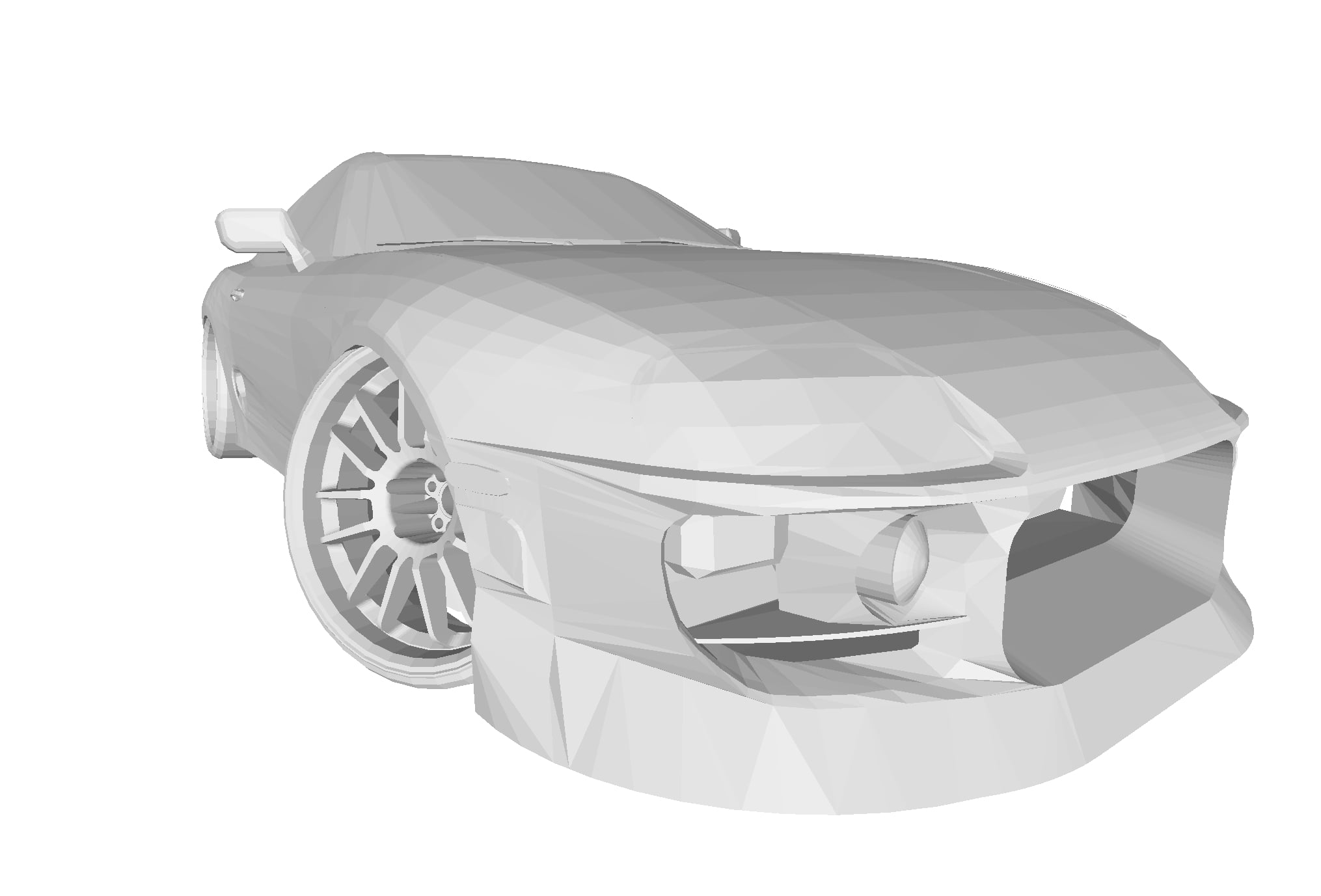} \\		  
            \includegraphics[width=1\textwidth]{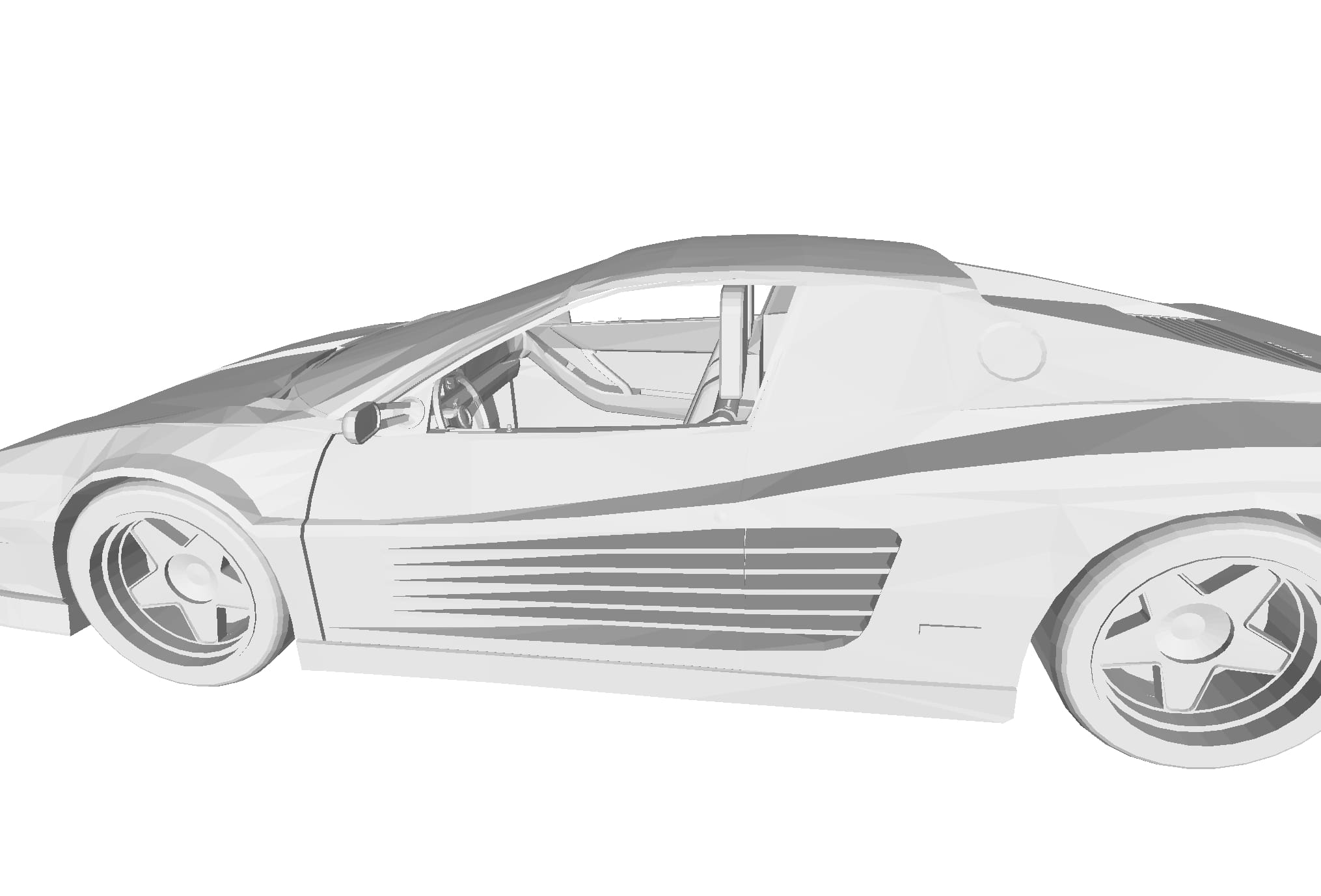} \\
            \includegraphics[width=1\textwidth]{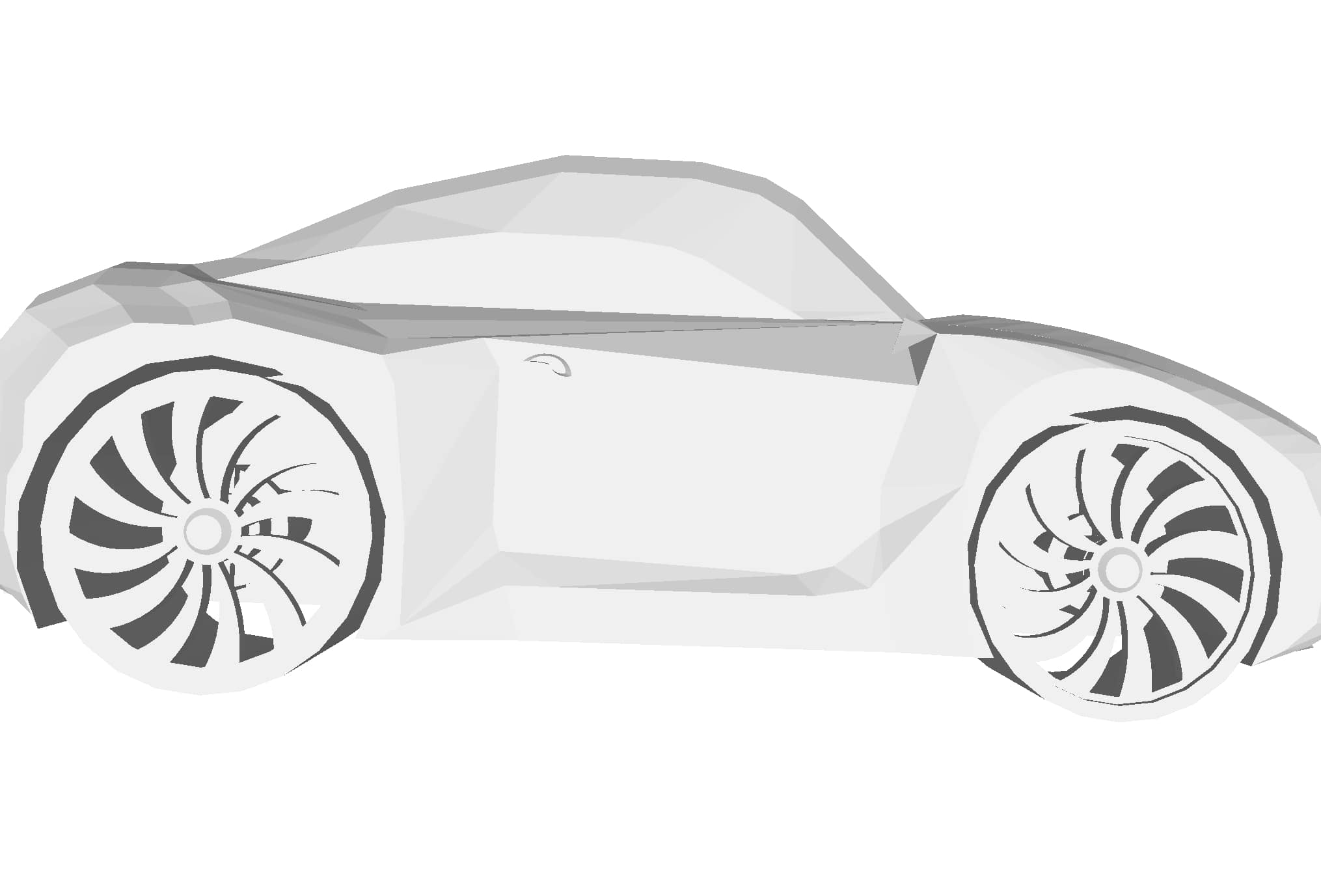}
        \end{minipage}
    }
    \caption{More visual comparisons with DUDF, CAPUDF and LevelSetUDF on car models from the ShapeNet-Cars dataset showcasing better reconstruction for complex internal structures.}
    \label{fig:shapenetCar-comparison}
\end{figure*}

We present additional comparisons with CAP-UDF~\cite{Zhou2022CAPUDF} and LevelSetUDF~\cite{Zhou2023levelset} by their original implementations as illustrated in Figure~\ref{fig:officail-comparison} and detailed in Table~\ref{tab:surf-comp-official}. The accuracy is lower than using DCUDF~\cite{Hou2023} for surface extraction. Our method produces higher quality results, by improved geometric detail and smoother shape boundaries. In Table~\ref{tab:zero-dev}, we show detailed results about the zero deviation from the zero level set. In Figure~\ref{fig:noise-comp}, we show visual comparisons for resilience to noisy input point clouds. In Figure~\ref{fig:3dScan-comparison} and \ref{fig:shapenetCar-comparison}, we show more results for models with details and models with complex topology and internal structures. The surfaces are all extracted by DCUDF except DUDF~\cite{fainstein2024dudf} by the original method. Our method outperforms DUDF, CAP-UDF and LevelSetUDF in terms of accuracy and topology.

\begin{table*}[!hb]
    \centering
    \setlength\tabcolsep{2pt}
    \resizebox{1.0\linewidth}{!}{
    \begin{tabular}{cccccccccccccc}
        \toprule
          & & \multicolumn{4}{c}{Stanford 3D Scene}& \multicolumn{4}{c}{Stanford 3D Scan}& \multicolumn{4}{c}{ShapeNet-Cars}\\
         \cmidrule(r){1-14}
         & & \multicolumn{2}{c}{Chamfer-L1 $(\downarrow)$} &\multicolumn{2}{c}{F-score $(\uparrow)$} & \multicolumn{2}{c}{Chamfer-L1 $(\downarrow)$} &\multicolumn{2}{c}{F-score $(\uparrow)$} & \multicolumn{2}{c}{Chamfer-L1 $(\downarrow)$} &\multicolumn{2}{c}{F-score $(\uparrow)$}\\
        \cmidrule(r){3-4}
        \cmidrule(r){5-6}
        \cmidrule(r){7-8}
        \cmidrule(r){9-10}        
        \cmidrule(r){11-12}
        \cmidrule(r){13-14}
         Method & Distance &  Mean & Median & $F1^{0.01}$& $F1^{0.005}$ &  Mean & Median & $F1^{0.01}$& $F1^{0.005}$ &  Mean & Median & $F1^{0.01}$& $F1^{0.005}$\\
        \midrule
         CAP-UDF & Unsigned  & 3.32 & 3.12& 99.36 & 84.98 &
         4.11 & 3.87& 99.12 & 69.24 &
         4.95 & 4.67& 95.49 & 55.9\\
         LeverSetUDF & Unsigned  & 3.16 & 2.93 & 99.32 & 85.90 &
         4.10 & 3.85& 99.13 & 69.42 &
         5.07 & 4.76& 94.98 & 54.30\\
        \midrule
         Ours & Unsigned & \textbf{3.09} &\textbf{2.85}& \textbf{99.41}&\textbf{86.38} &
         \textbf{4.08} &\textbf{3.83}& \textbf{99.14}&\textbf{69.59} & \textbf{4.91} &\textbf{4.58}& \textbf{95.53}&\textbf{56.98} \\
        \bottomrule
    \end{tabular}
    }
    \caption{
    Quantitative comparisons with the original results of CAP-UDF and LevelSetUDF. Chamfer distances are measured in the unit of $\times10^{-3}$. 
    }
    \label{tab:surf-comp-official}

\end{table*}

\begin{table*}[hb]
    \centering
    \resizebox{1.0\linewidth}{!}{
    \begin{tabular}{cccccccccccccc}
        \toprule
          & \multicolumn{5}{c}{Stanford 3D Scene}& \multicolumn{8}{c}{Stanford 3D Scan}\\
        \cmidrule(r){2-6}
        \cmidrule(r){7-14}

         Method & Burghers &  Copyroom & Lounge & Stonewall & Totempole & Asian dragon & Camera & Dragon & Dragon warrior &  Dragon wing & Statue ramesses & Thai statue & Vase lion\\
        \midrule
         DUDF        & 26.26 & 27.26 & 26.03 & 25.87 & 25.68 & 25.18 & 25.90 & 24.93 & 26.05 & 25.86 & 26.42 & 26.27 & 26.46\\
         CAP-UDF     & 3.94 & 6.64 & 4.81 & 2.66 & 2.19 & 3.39 & 0.26 & 0.92 & 2.52 & 2.38 & 1.54 & 2.31 & 0.74 \\
         LeverSetUDF & 1.98 & 3.63 & 3.06 & \textbf{0.83} & 1.12 & 1.24 & \textbf{0.31} & \textbf{0.49} & \textbf{0.80} & \textbf{0.72} & 1.90 & \textbf{1.50} & \textbf{0.45}\\
        \midrule
         Ours        & \textbf{1.43} & \textbf{1.61} & \textbf{1.48} & 0.92 & \textbf{0.99} & \textbf{1.02} & 0.94 & 1.48 & 2.22 & 1.92 & \textbf{0.53} & 1.69 & 1.26 \\
        \bottomrule
    \end{tabular}
    }
    \caption{Quantitative results for deviations from zero. We input vertices from reconstructed mesh into learned UDF, then calculate the average output value, which means average deviation from zero. Our method performs better in average, and is more stable than other methods.}
    \label{tab:zero-dev}
\end{table*}

\end{document}